\RequirePackage{silence} % :-\
\documentclass[ twoside,openright,titlepage,numbers=noenddot,headinclude,%1headlines,% letterpaper a4paper
                footinclude=true,cleardoublepage=empty,abstractoff, % <--- obsolete, remove (todo)
                BCOR=5mm,paper=a4,fontsize=11pt,%11pt,a4paper,%
                british,spanish,ngerman,american,%
                ]{scrreprt}

\PassOptionsToPackage{utf8}{inputenc}
  \usepackage{inputenc}

\PassOptionsToPackage{
  drafting=false,    % print version information on the bottom of the pages
  tocaligned=false, % the left column of the toc will be aligned (no indentation)
  dottedtoc=true,  % page numbers in ToC flushed right
  parts=true,       % use part division
  eulerchapternumbers=true, % use AMS Euler for chapter font (otherwise Palatino)
  linedheaders=true,       % chaper headers will have line above and beneath
  floatperchapter=true,     % numbering per chapter for all floats (i.e., Figure 1.1)
  listings=true,    % load listings package and setup LoL
  subfig=true,      % setup for preloaded subfig package
  eulermath=false,  % use awesome Euler fonts for mathematical formulae (only with pdfLaTeX)
  beramono=true,    % toggle a nice monospaced font (w/ bold)
  minionpro=false   % setup for minion pro font; use minion pro small caps as well (only with pdfLaTeX)
}{classicthesis}

\newcommand{\myTitle}{Hierarchical Representations for Spatio-Temporal Visual Attention Modeling and Understanding\xspace}
\newcommand{\mySubtitle}{...\xspace}

\newcommand{\myName}{Miguel Ángel Fernández Torres\xspace}
\newcommand{\myProf}{Dr. Iván González Díaz\xspace}
\newcommand{\myOtherProf}{Dr. Fernando Díaz de María\xspace}

\newcommand{\myFaculty}{Escuela Politécnica Superior\xspace}

\newcommand{\myUni}{Universidad Carlos III de Madrid\xspace}
\newcommand{\myLocation}{Leganés\xspace}
\newcommand{\myTime}{January 2019\xspace}
\newcommand{\myVersion}{version 1.0}

\newcounter{dummy} % necessary for correct hyperlinks (to index, bib, etc.)
\providecommand{\mLyX}{L\kern-.1667em\lower.25em\hbox{Y}\kern-.125emX\@}

\newcommand{\xv}{\mathbf{x}}

\newcommand{\cv}{\mathbf{c}}
\newcommand{\uv}{\mathbf{u}}

\DeclareOldFontCommand{\bf}{\normalfont\bfseries}{\mathbf}
    \usepackage{babel}

\usepackage{csquotes}
\PassOptionsToPackage{%
  backend=bibtex8,bibencoding=ascii,%
  language=auto,%
  style=numeric-comp,%
  sorting=none, %nyt, % name, year, title
  maxbibnames=10, % default: 3, et al.
  natbib=true % natbib compatibility mode (\citep and \citet still work)
}{biblatex}
    \usepackage{biblatex}

  \usepackage{amsmath}
  \usepackage{amssymb}
  \usepackage{nccmath}
  \usepackage{mathtools}
  \usepackage{multirow}
  
  \DeclareMathOperator*{\argmin}{arg\,min}

\PassOptionsToPackage{T1}{fontenc} % T2A for cyrillics
  \usepackage{fontenc}
\usepackage{textcomp} % fix warning with missing font shapes
\usepackage{scrhack} % fix warnings when using KOMA with listings package
\usepackage{xspace} % to get the spacing after macros right
\usepackage{mparhack} % get marginpar right
\PassOptionsToPackage{printonlyused,smaller}{acronym}
  \usepackage{acronym} % nice macros for handling all acronyms in the thesis
\usepackage{tabularx} % better tables
\usepackage{caption}
\usepackage{subfig}
\usepackage{listings}
\PassOptionsToPackage{hyperfootnotes=false,pdfpagelabels}{hyperref}
  \usepackage{hyperref}  % backref linktocpage pagebackref
  \usepackage{graphicx}

\hypersetup{%
  colorlinks=true, linktocpage=true, pdfstartpage=3, pdfstartview=FitV,%
  breaklinks=true, pdfpagemode=UseNone, pageanchor=true, pdfpagemode=UseOutlines,%
  plainpages=false, bookmarksnumbered, bookmarksopen=true, bookmarksopenlevel=1,%
  hypertexnames=true, pdfhighlight=/O,%nesting=true,%frenchlinks,%
  urlcolor=RoyalBlue, linkcolor=RoyalBlue, citecolor=webgreen, %pagecolor=RoyalBlue,%
  pdftitle={\myTitle},%
  pdfauthor={\textcopyright\ \myName, \myUni, \myFaculty},%
  pdfsubject={},%
  pdfkeywords={},%
  pdfcreator={pdfLaTeX},%
  pdfproducer={LaTeX with hyperref and classicthesis}%
}

\makeatletter
\@ifpackageloaded{babel}%
  {%
    \addto\extrasamerican{%
    }%
    \addto\extrasngerman{%
    }%
    }{\relax}
\makeatother

\usepackage{classicthesis}
\usepackage{pdflscape}
\usepackage{chngpage}
\usepackage{rotating}
\usepackage{dpfloat}
\usepackage{float}

\usepackage{adjustbox}
\usepackage[dvipsnames]{xcolor}
\usepackage{environ}
\usepackage{tikz}
\usetikzlibrary{calc,matrix,shapes.geometric, arrows}

\makeatletter
\let\matamp=&
\catcode`\&=13
\makeatletter
\def&{\iftikz@is@matrix
  \pgfmatrixnextcell
  \else
  \matamp
  \fi}
\makeatother

\newcounter{lines}
\def\endlr{\stepcounter{lines}\\}

\newcounter{vtml}
\newif\ifvtimelinetitle
\newif\ifvtimebottomline
\tikzset{description/.style={
  column 2/.append style={#1}
 },
 timeline color/.store in=\vtmlcolor,
 timeline color=red!80!black,
 timeline color st/.style={fill=\vtmlcolor,draw=black},
 use timeline header/.is if=vtimelinetitle,
 use timeline header=false,
 add bottom line/.is if=vtimebottomline,
 add bottom line=false,
 timeline title/.store in=\vtimelinetitle,
 timeline title={},
 line offset/.store in=\lineoffset,
 line offset=4pt,
}

\NewEnviron{vtimeline}[1][]{%
\setcounter{lines}{1}%
\stepcounter{vtml}%
\begin{tikzpicture}[column 1/.style={anchor=east},
 column 2/.style={anchor=west},
 text depth=0pt,text height=0.85ex,
 row sep=1ex,
 column sep=1em,
 #1
]
\tikzstyle{every node}=[font=\normalsize]
\matrix(vtimeline\thevtml)[matrix of nodes]{\BODY};
\pgfmathtruncatemacro\endmtx{\thelines-1}
\path[timeline color st] 
($(vtimeline\thevtml-1-1.north east)!0.5!(vtimeline\thevtml-1-2.north west)$)--
($(vtimeline\thevtml-\endmtx-1.south east)!0.5!(vtimeline\thevtml-\endmtx-2.south west)$);
\foreach \x in {1,...,\endmtx}{
 \node[circle,timeline color st, inner sep=0.15pt, draw=black, thick] 
 (vtimeline\thevtml-c-\x) at 
 ($(vtimeline\thevtml-\x-1.east)!0.5!(vtimeline\thevtml-\x-2.west)$){};
 \draw[timeline color st](vtimeline\thevtml-c-\x.west)--++(-3pt,0);
 }
 \ifvtimelinetitle%
  \draw[timeline color st]([yshift=\lineoffset]vtimeline\thevtml.north west)--
  ([yshift=\lineoffset]vtimeline\thevtml.north east);
  \node[anchor=west,yshift=16pt,font=\large]
   at (vtimeline\thevtml-1-1.north west) 
   {\textsc{Timeline \thevtml}: \textit{\vtimelinetitle}};
 \else%
  \relax%
 \fi%
 \ifvtimebottomline%
   \draw[timeline color st]([yshift=-\lineoffset]vtimeline\thevtml.south west)--
  ([yshift=-\lineoffset]vtimeline\thevtml.south east);
 \else%
   \relax%
 \fi%
\end{tikzpicture}
}

\tikzstyle{level0} = [rectangle, minimum width=5cm, minimum height=1cm, text centered, draw=black, fill=White]
\tikzstyle{level10} = [rectangle, minimum width=4cm, minimum height=1cm, text centered, draw=black, fill=NavyBlue!30]
\tikzstyle{level11} = [rectangle, minimum width=4cm, minimum height=1cm, text centered, draw=black, fill=YellowOrange!30]
\tikzstyle{level12} = [rectangle, minimum width=4cm, minimum height=1cm, text centered, draw=black, fill=Purple!30]
\tikzstyle{level20} = [rectangle, minimum width=2cm, minimum height=1cm, text centered, draw=black, fill=Red!30]
\tikzstyle{level21} = [rectangle, minimum width=2cm, minimum height=1cm, text centered, draw=black, fill=Gray!30]
\tikzstyle{level22} = [rectangle, minimum width=2cm, minimum height=1cm, text centered, draw=black, fill=Green!30]
\tikzstyle{level3} = [rectangle, minimum width=3cm, minimum height=1cm, text centered, draw=black, fill=White]
\tikzstyle{level4} = [rectangle, minimum width=2cm, minimum height=1cm, text width=4.5cm, draw=white, fill=White]
\tikzstyle{level41} = [rectangle, minimum width=2cm, minimum height=1cm, text width=3cm, draw=white, fill=White]
\tikzstyle{arrow} = [thick,->,>=stealth]

\usepackage{afterpage}

\usepackage{ragged2e}

\newlength\largefigure % to create a new length
\usepackage{diagbox}
\renewcommand{\chapterNumber}{
  \fontsize{70}{70}\usefont{T1}{doppio}{m}{n}}

\emergencystretch=\maxdimen
\begin{document}
\frenchspacing
\raggedbottom
\selectlanguage{british} % american ngerman
%\renewcommand*{\bibname}{new name}
%\setbibpreamble{}
\pagenumbering{roman}
\pagestyle{plain}
%********************************************************************
% Frontmatter
%*******************************************************
% \include{FrontBackmatter/DirtyTitlepage}
%*******************************************************
% Titlepage
%*******************************************************
\begin{titlepage}
    % if you want the titlepage to be centered, uncomment and fine-tune the line below (KOMA classes environment)
    \begin{addmargin}[-1cm]{-3cm}
    \begin{center}
    	%\includegraphics[width=12cm]{gfx/titlepage/uc3m.eps} \\ 		
        
        %\medskip
        \vspace*{5cm}
		
        \begingroup
            \color{MidnightBlue}\spacedallcaps{\myTitle} \\ \bigskip
        \endgroup
		\small{Presented by} \\
		\spacedlowsmallcaps{\myName} \\
		
		\vspace{5cm}
		
		%\small{Presented by} \\
        %\spacedlowsmallcaps{\myName} \\
        \medskip
        \medskip
        \small{in partial fulfillment of the requirements for the \\ Degree of Doctor in Multimedia and Communications} \\
        
        \vspace{1cm}
        
        %\spacedlowsmallcaps{\myDepartment} \\
        %\vspace{0.5cm}
        \spacedlowsmallcaps{\myUni}
        
        \vspace{3cm}
        
        \small{Advisors:} \\
        \spacedlowsmallcaps{\myProf} \\
        \spacedlowsmallcaps{\myOtherProf}

        \vspace{2cm}

        %\spacedlowsmallcaps{\myDepartment} \\

		\vspace{1cm}
		
        \myLocation, \myTime \\

    \end{center}
  \end{addmargin}
\end{titlepage}

\cleardoublepage\thispagestyle{empty}

\pdfbookmark[1]{Rights}{Rights}

\vspace*{\fill}

\justify
\noindent Some rights reserved. This work is licensed under the Creative Commons Attribution-NonCommercial-ShareAlike 4.0 International License. To view a copy of this license, visit \url{http://creativecommons.org/licenses/by-nc-sa/4.0/} or send a letter to Creative Commons, PO Box 1866, Mountain View, CA 94042, USA.

%\include{FrontBackmatter/Tribunal}
%\cleardoublepage\include{FrontBackmatter/Foreword}
\cleardoublepage%*******************************************************
% Dedication
%*******************************************************
\thispagestyle{empty}
%\phantomsection
%\refstepcounter{dummy}
\pdfbookmark[1]{Dedication}{Dedication}

\vspace*{3cm}

%\begin{center}
%    Stories are wild creatures, the monster said. \\
%    When you let them loose, who knows what havoc they might wreak? \\ \medskip
%    --- Patrick Ness, A Monster Calls
%\end{center}

\begin{center}
We experienced events in an order, \\ and perceived their relationship as cause and effect. \\ 
\medskip
\medskip
They experienced all events at once, and perceived a purpose underlying them all. A minimizing, maximizing purpose. \\
\medskip
\medskip
	--- Ted Chiang, Stories of Your Life and Others
\end{center}

\medskip

% \begin{center}
%     Dedicated to the loving memory of Rudolf Miede. \\ \smallskip
%     1939\,--\,2005
% \end{center}

\cleardoublepage%*******************************************************
% Acknowledgments
%*******************************************************
\pdfbookmark[1]{Agradecimientos}{acknowledgments}
%\bigskip

\begingroup
\let\clearpage\relax
\let\cleardoublepage\relax
\let\cleardoublepage\relax
\chapter*{Agradecimientos}
Ha llegado el momento. Quizás no sea el final que mi imaginación de niño soñó cuando comenzó esta tesis. Quizás no haya sido la aventura que esperábamos. 

\begin{center}
\emph{``Las historias son criaturas salvajes —dijo el monstruo—. Cuando las sueltas, ¿quién sabe los desastres que pueden causar?''}\\
\medskip
--- Patrick Ness, A Monster Calls
\end{center}

\noindent Pero, en el fondo, sí que ha sido la experiencia que necesitaba. Pasa el tiempo apenas sin darnos cuenta, y la vida nos va llevando por esos lugares que nos quedan por descubrir, que nos hacen falta para crecer. A veces el amor al que hay que renunciar a cambio de una lección aprendida es muy grande, pero vale la pena seguir dando pasos hacia delante cuando finalmente eres consciente del camino que te quedaba por recorrer. Dejando el pasado atrás, y pensando casi por primera vez en el presente más que en el futuro, me siento muy afortunado de haber podido vivir estos diez años de formación en la Universidad Carlos III de Madrid. Han dado para tanto, y todo tan bueno...

Creo que he aprendido a investigar. Y es aquí donde tengo que agradecer enormemente, en primer lugar, a Fernando, el haberme dado la oportunidad de formar parte del Grupo de Procesado Multimedia y descubrir mi pasión por la Visión Artificial. A Iván, porque su talento y dedicación han conseguido sacar lo mejor de mí estos años. Gracias a los dos por la confianza y paciencia que habéis tenido conmigo, por enseñarme a creer en mí.

A los compañeros de laboratorio, por su apoyo en los momentos que más lo necesitaba. A Fernando, por regalarme los mejores consejos y conversaciones, y por preocuparse de cada uno de nosotros. También a Tomás, Amaya, Álex, FerFer, Rubén y Carmen. A Luis, por todas las películas y conciertos compartidos, me siento muy afortunado de haber podido conocerte más allá de estas aulas. A Ascen, porque desde aquella vez que me ayudaste a preparar mi Erasmus en Viena no has dejado de apoyarme en todo lo que necesitaba, siempre es un gusto charlar contigo. Muy especialmente a mis dos compañeros de croquetones. Gracias a mi compañero sobre-saliente Antonio por los momentos más divertidos, por haber sido capaz de volver a sacar el niño que llevo dentro cuando más lo necesitaba, por enseñarme a reírle a la vida. También a Javi. Empezamos juntos esta aventura hace diez años, y me lo has puesto todo tan fácil siempre... gracias por aportar la calma cuando era día de tormenta. A Borja, Lorena y otros tantos compañeros de departamento junto a los que he crecido estos años. A Harold, por haberme prestado su ayuda para poder terminar esta tesis a tiempo.

He dado mis primeros pasos como profesor, y aquí también tengo que agradecer a muchos de los que he mencionado antes por ser un ejemplo a seguir, y por haber inculcado en mí esta vocación. No hay cosa que me haga más feliz cada semana que bajar a dar clase a un laboratorio y tratar de enseñar todo lo que puedo. Sí, sí la hay: Salir de una clase con esa sensación de haber compartido algo importante con los estudiantes que tengo a mi lado.

También he tenido oportunidad de viajar y conocer mundo. En Viena y West Lafayette he vivido la parte más impresionante de esta aventura. Gracias al Prof. Zygmunt Pizlo por acogerme en Purdue University y por lo mucho que ha aportado desde su perspectiva psicológica a esta tesis. Nunca olvidaré mis viajes a Chicago y Los Ángeles, los paseos por la noche junto a los rascacielos iluminados, todas las personas que fui encontrando por el camino. Tampoco a las que conocí en mi primera experiencia en un congreso en Bucharest. A Song y Souad, con las que he podido volver a encontrarme en Madrid no hace mucho.

Y entre investigación y docencia, también ha habido ratos para disfrutar. Y disfrutando hay veces que se ríe y otras que se llora, muchas de ellas de emoción, otras porque de los amigos también se aprende. Empezando por aquellos que he tenido la oportunidad de conocer en este lugar, o muy cerca, gracias a los compañeros de penúltimas y closing parties: Sergio, Unai, Adrián, Diego. A Cristina y Víctor, sois unos haters y lo sabéis. A mis comPICñeros preferidos: Gisela, Raquel y Rafa. A Alba, qué pena que ya no cojamos el tren juntos para perdernos de camino a casa.

Y siguiendo con los que me han acompañado fuera de la universidad. Gracias a mi gran compañero de cruces y cruzadas, y mejor amigo Pedro. ¿Cuál es el próximo destino? A Álvaro, que coge ahora el testigo de esta tesis y comienza la suya. Gracias por estar ahí siempre este último año, por recordarme lo importante que es quererte tal y como eres, eres el amigo que cualquiera querría tener. A Patri, por los muchos pasos que llevamos ya dados juntos, y otras tantas carreras que vamos a compartir pronto, por ser tan buena amiga. A Wilson, al que también le gusta Vetusta, tengo ganas de más conciertos juntos. A Morganne, por esas tardes recorriendo Madrid y hablando de cine.

Para terminar, a mi familia. A mis padres, por haber antepuesto sus intereses a los míos, porque me han dejado siempre hacer lo que más quería. A mi madre, porque hemos hecho esta tesis codo con codo. Gracias por tus abrazos pero, sobre todo, por ser mi ejemplo de perseverancia y compromiso en esta vida. A mi padre, con el que espero tener más tiempo a partir de ahora para compartir todas esas tecnologías que tanto le gustan. También a mi hermana, de la que admiro tanto su poca pereza, y a la que deseo que su esfuerzo tenga la recompensa que merece muy pronto, y podamos celebrar nuestros éxitos juntos. 

Seguramente me estoy dejando a muchas personas fuera de estas líneas, por falta de memoria o espacio, no por ello menos importantes. Gracias a todas ellas por haberse cruzado alguna vez por esta historia, por hacerla única y tan emocionante, por ser fundamentales en esta etapa de crecimiento personal y profesional que hoy cierro con los sentimientos un tanto encontrados. La tristeza por terminar algo tan grande, y las ganas y la ilusión de ver lo que está por llegar. Todavía queda mucho por descubrir. Sigamos caminando.

\medskip
\medskip
\medskip
\medskip

Miguel Ángel
\endgroup

\vfill

%*******************************************************
% Publications
%*******************************************************
\pdfbookmark[1]{Publications}{publications}
\chapter*{Published and submitted content}

%\graffito{This is just an early --~and currently ugly~-- test!}

Some parts of the following publications are included or extended in this thesis:

\begin{enumerate}
\item Fernández-Torres, M. Á., González-Díaz, I., \& Díaz-de-María, F. (2016, June). A probabilistic topic approach for context-aware visual attention modeling. In Content-Based Multimedia Indexing (CBMI), 2016 14th International Workshop on (pp. 1-6). IEEE.\\
\url{https://doi.org/10.1109/CBMI.2016.7500272}\\

Parts of this article are extended in Chapters \ref{ch:atom} and \ref{ch:atom_experiments} of the thesis. Whenever material from this source is included in this thesis, it is singled out with an explicit reference.

\item Fernández-Torres, M. Á., González-Díaz, I., \& Díaz-de-María, F. (submitted). Probabilistic Topic Model for Context-Driven Visual Attention Understanding. IEEE Transactions on Circuits and Systems for Video Technology. \\

Parts of this article are included or extended in Chapters \ref{ch:atom} and \ref{ch:atom_experiments} of the thesis. Whenever material from this source is included in this thesis, it is singled out with an explicit reference.
\end{enumerate}

%\noindent Put your publications from the thesis here. The packages \texttt{multibib} or \texttt{bibtopic} etc. can be used to handle multiple different bibliographies in your document.

%\begin{refsection}[ownpubs]
%    \small
%    \nocite{*} % is local to to the enclosing refsection
%    \printbibliography[heading=none]
%\end{refsection}
%
%\emph{Attention}: This requires a separate run of \texttt{bibtex} for your \texttt{refsection}, \eg, \texttt{ClassicThesis1-blx} for this file. You might also use \texttt{biber} as the backend for \texttt{biblatex}. See also \url{http://tex.stackexchange.com/questions/128196/problem-with-refsection}.

%*******************************************************
% Research
%*******************************************************
\pdfbookmark[1]{Research}{research}
\chapter*{Other research merits}

The following publications were part of my Ph.D. research, but the topics they cover are out of the scope of this thesis, so they are not included in it:

\begin{enumerate}
	\item Fernández-Martínez, F., Hernández-García, A., Fernández-Torres, M. A., González-Díaz, I., García-Faura, Á., \& de María, F. D. (2017). Exploiting visual saliency for assessing the impact of car commercials upon viewers. Multimedia Tools and Applications, 1-31.\\
	\url{https://doi.org/10.1007/s11042-017-5339-9}

	\item López-Labraca, J., Fernández-Torres, M. Á., González-Díaz, I., Díaz-de-María, F., \& Pizarro, Á. (2018). Enriched dermoscopic-structure-based cad system for melanoma diagnosis. Multimedia Tools and Applications, 77(10), 12171-12202. \\
	\url{https://doi.org/10.1007/s11042-017-4879-3}

	\item Martínez-Cortés, T., Fernández-Torres, M. Á., Jiménez-Moreno, A., González-Díaz, I., Díaz-de-María, F., Guzmán-De-Villoria, J. A., \& Fernández, P. (2014, October). A Bayesian model for brain tumor classification using clinical-based features. In Image Processing (ICIP), 2014 IEEE International Conference on (pp. 2779-2783). IEEE. \\
	\url{https://doi.org/10.1109/ICIP.2014.7025562}
\end{enumerate}
\cleardoublepage%*******************************************************
% Resumen extendido
%*******************************************************
\pdfbookmark[1]{Resumen extendido}{Resumen extendido}
\begingroup
\begin{otherlanguage}{spanish}
\let\clearpage\relax
\let\cleardoublepage\relax
\let\cleardoublepage\relax

\chapter*{Resumen extendido}
%Short summary of the contents in English\dots a great guide by
%Kent Beck how to write good abstracts can be found here:
%\begin{center}
%\url{https://plg.uwaterloo.ca/~migod/research/beckOOPSLA.html}
%\end{center}
En el siguiente resumen extendido se recogen los aspectos más relevantes de la presente Tesis doctoral. En primer lugar, se presenta la motivación del trabajo realizado. A continuación, se describen los principales objetivos y las contribuciones originales más destacadas. Finalmente, se resumen las conclusiones más relevantes, así como se mencionan posibles líneas futuras de investigación a partir del trabajo llevado a cabo.

\section*{Motivación de la tesis}
Dentro del marco de la Inteligencia Artificial, la Visión Artificial \cite{forsyth2011computer} es una disciplina científica que tiene como objetivo simular automaticamente las funciones del sistema visual humano, tratando de resolver tareas como la localización \cite{DeepSaliencyObject} y el reconocimiento \cite{krizhevsky2012imagenet} de objetos, la detección de eventos \cite{anomaliesDeLaCalle} o el seguimiento de objetos \cite{smeulders2014visual}.

A pesar de la gran variedad de sistemas automáticos que se han desarrollado para resolver estos problemas, algunos de ellos verdaderamente efectivos, la mayor parte requiere procesar grandes cantidades de información visual, lo cual influye negativamente en su eficiencia. A diferencia de estos sistemas, el sistema visual humano es capaz de seleccionar de manera casi inmediata los elementos más importantes para poder interactuar en un contexto dado, al mismo tiempo que se ve atraído por aquellos estímulos más sorprendentes o llamativos. Esto se debe a su función de atención visual, la cual puede entenderse como un proceso de optimización para la percepción y la cognición visual. Si fuéramos capaces de diseñar algoritmos para la interpretación de escenas que llevasen a cabo esta función, podríamos ayudar a usuarios y expertos en escenarios de aplicación complejos, en los que se requiere procesar una gran cantidad de información simultáneamente, tales como la conducción \cite{pradhan2005using}, la aeronáutica \cite{kilingaru2013monitoring} o la videovigilancia \cite{howard2011task}. Esto permitiría disminuir la probabilidad de que se produzcan errores humanos, así como agilizar los procesos de decisión de los expertos.

La atención visual se puede estudiar en dos dominios diferentes: espacial y temporal. Estos dan lugar a definir tres tipos de modelos computacionales: espacial, espacio-temporal y temporal \cite{Borji:2013:SVA:2412386.2412937}. La mayor parte de los modelos computacionales de atención visual existentes consideran la componente espacial para guiar el procesamiento de la información visual hacia las regiones más llamativas o interesantes de una escena. Además, la información que percibimos en el mundo es dinámica, por lo que es igual de importante modelar cómo cambia a lo largo del tiempo, lo que permite actualizar la atención espacial en función de las localizaciones seleccionadas previamente, así como seleccionar intervalos temporales de especial interés.

También es habitual distinguir entre dos familias de modelos de atención visual: modelos \emph{Bottom-Up}, basados en las características visuales de la escena; y modelos \emph{Top-Down}, los cuales tienen en cuenta un conocimiento a priori de la escena, o determinadas indicaciones para resolver una tarea  \cite{TreismanGelade80, wolfe1994guided}.

La motivación principal de esta tesis es, por tanto, el estudio y desarrollo de representaciones jerárquicas para el modelado y la interpretación de la atención visual espacio-temporal.

En particular, se proponen dos modelos computacionales de atención visual:

\begin{enumerate}
	\item En primer lugar, se presenta un modelo generativo probabilístico \emph{top-down} para el modelado y la interpretación de la atención visual en diferentes contextos. 

	\item En segundo lugar, se implementa una red profunda para el modelado de la atención visual. Esta arquitectura estima, en primer lugar, la atención visual espacio-temporal \emph{top-down}, para finalmente modelar la atención en el dominio temporal. Su diseño está orientado a su aplicación final en un escenario de videovigilancia.
\end{enumerate}

\section*{Objetivos y contribuciones originales de la tesis}
Tal y como se comenta en el apartado anterior, esta tesis puede dividirse en dos partes.

En la primera parte de la tesis, se introduce nuestra primera aproximación: un modelo generativo probabilístico para el modelado y la interpretación de la atención visual espacio-temporal. El modelo propuesto, al que se ha denominado \emph{visual ATtention Topic Model} o ATOM, es genérico, independiente del escenario de aplicación y está basado en las teorías psicológicas más destacadas sobre la atención visual \cite{TreismanGelade80,wolfe1994guided}. Además, considera la relación existente entre factores \emph{bottom-up} y \emph{top-down}.

Partiendo del conocido algoritmo \emph{Latent Dirichlet Allocation} o LDA propuesto por David Blei et al. \cite{LDA} para el análisis de corpus grandes de información textual, así como teniendo en cuenta dos de sus extensiones supervisadas \cite{sLDA,DBAyang}, nuestro modelo define la atención visual espacio-temporal \emph{top-down} como una combinación de subtareas latentes que, a su vez, se representan mediante combinaciones de características espacio-temporales de bajo, medio y alto nivel. 

En particular, esta primera aproximación da lugar a las siguientes contribuciones:

\begin{itemize}
	\item En primer lugar, se introduce un conjunto amplio de características de bajo nivel para el modelado de la atención visual, tales como el color, la intensidad, la orientación o el movimiento. A continuación, se procede a modelar características de medio y alto nivel, relacionadas con la estimación del movimiento de cámara en la escena y la detección de objetos.
	
	\item En segundo lugar, nuestro algoritmo incorpora un nivel intermedio formado por subtareas latentes. Este nivel permite acortar distancias entre la etapa de extracción de características y la de modelado de la atención visual, así como obtener representaciones de la atención más comprensibles y fáciles de interpretar.
	
	\item Además, nuestro modelo incorpora una variable categórica binaria que modela la atención visual en cada una de las localizaciones espaciales de una escena. Esta variable nos permite alinear automáticamente las subtareas determinadas por nuestro sistema con la información existente en aquellos lugares de la escena que atraen la atención de los usuarios.  
\end{itemize}

A continuación, se incluye en la tesis un análisis exhaustivo de este algoritmo. Para ello, el modelo ATOM se utiliza para estimar e interpretar la atención visual en diferentes contextos (exteriores, vídeojuegos, noticias, etc.), definidos dentro de dos amplias bases de datos de vídeo anotadas con fijaciones de los ojos de diferentes sujetos: CRCNS-ORIG \cite{Itti_Carmi09crcns} y DIEM \cite{mital2011clustering}. Esto permite ilustrar cómo nuestro modelo es capaz de aprender de manera efectiva representaciones jerárquicas de la atención visual adaptadas a diferentes escenarios. Además, los modelos obtenidos se utilizan para estimar la atención visual, comparando su eficiencia con la de otros modelos existentes en el estado del arte.

En la segunda parte de la tesis, se describe nuestra segunda aproximación: una red profunda que permite modelar la atención en el dominio temporal a partir de la atención visual espacio-temporal estimada. Nuestro algoritmo, al que se ha denominado \emph{Spatio-Temporal to Temporal visual ATtention NETwork} o ST-T-ATTEN, modela la atención a lo largo del tiempo considerando una variable basada en las fijaciones proporcionadas por diferentes sujetos desarrollando una misma tarea.

En primer lugar, se introduce la hipótesis fundamental del modelo, la cual establece que la atención en el dominio temporal puede estimarse midiendo la dispersión de la localización de las fijaciones facilitadas por diferentes usuarios. Además, se puede entender la dimensión temporal de la atención como un mecanismo de filtrado, el cual permite identificar intervalos temporales de especial importancia en secuencias de vídeo.

En particular, nuestra segunda aproximación da lugar a las siguientes contribuciones:

\begin{itemize}
	\item En primer lugar, se presentan tres arquitecturas para la extracción de características de alto nivel que permitan modelar la atención visual. Estas características, basadas en el color, el movimiento y los objetos presentes en la escena, servirán como entrada a nuestro sistema.
	
	\item En segundo lugar, se propone un \emph{ground-truth} temporal a nivel de frame basado en las fijaciones de diferentes sujetos. Este \emph{ground-truth} se obtiene atendiendo a la dispersión de las localizaciones fijadas. Además, permite validar la hipótesis fundamental de nuestro sistema, introducida anteriormente. Esta variable servirá para entrenar nuestros modelos para la estimación de la atención en el dominio temporal. 
	
	\item Finalmente, se puede distinguir dos etapas en nuestro modelo: 1) Una red para la estimación de la atención visual espacio-temporal, basada en una arquitectura de tipo codificador-decodificador convolucional, \emph{Convolutional Encoder Decoder} o CED \cite{huang2007unsupervised}; 2) Una red de tipo \emph{Long Short-Term Memory} o LSTM \cite{hochreiter1997long} para el modelado de la dimensión temporal de la atención.
\end{itemize}

Finalmente, los experimentos de la segunda parte de la tesis tienen como objetivo validar las diferentes configuraciones propuestas para cada una de las etapas del modelo ST-T-ATTEN. Para ello, se hará uso de la base de datos BOSS \cite{BOSS}, la cual contiene secuencias de vídeo grabadas en un contexto ferroviario, en las que tienen lugar diferentes eventos anómalos. Nuestro objetivo es determinar la configuración óptima para la arquitectura completa propuesta, así como motivar su uso como mecanismo de filtrado de información visual en un escenario de videovigilancia. 

\section*{Conclusiones}
A lo largo de la tesis se han propuesto dos algoritmos jerárquicos para modelar la atención visual en secuencias de vídeo. 

El primer algoritmo, presentado en la primera parte de la tesis y denominado ATOM, es un modelo generativo probabilístico para la estimación e interpretación de la atención visual espacio-temporal. La definición del sistema propuesto es genérica e independiente del escenario de aplicación. Además, el modelo se fundamenta en las teorías psicológicas más importantes acerca de la atención visual \cite{TreismanGelade80,wolfe1994guided}, las cuales han establecido que la atención visual no se basa directamente en la información proporcionada por los procesos visuales tempranos, sino en una representación contextual derivada de los mismos.

Utilizando como base el algoritmo LDA \cite{LDA} y dos de sus extensiones supervisadas \cite{sLDA,DBAyang}, ATOM define la atención visual espacio-temporal \emph{top-down} como una combinación de subtareas latentes que, a su vez, se representan mediante combinaciones de características espacio-temporales de bajo, medio y alto nivel. Por tanto, dado un frame en una secuencia de vídeo, el sistema recibe a su entrada un conjunto de mapas de características (color, intensidad, movimiento, etc.). A continuación, define un nivel de subtareas latentes entre la etapa de extracción de características y la de modelado de la atención visual. Finalmente, se utiliza una variable categórica binaria para alinear las subtareas definidas con la información derivada de las fijaciones de diferentes sujetos. Esta variable se genera a partir de un modelo de regresión logística aplicado sobre las subtareas, teniendo en cuenta la proporción en la que las mismas se dan en el frame.

El análisis llevado a cabo en la primera parte de la tesis demuestra la habilidad del modelo ATOM para aprender de manera efectiva, a partir de un conjunto amplio de características visuales, representaciones jerárquicas de la atención visual adaptadas específicamente a diferentes contextos (exteriores, vídeojuegos, deportes, noticias, etc.). Para ello, se ha hecho uso de dos amplias bases de datos de vídeo, CRCNS-ORIG \cite{Itti_Carmi09crcns} y DIEM \cite{mital2011clustering}. Por otro lado, los experimentos muestran la facilidad de comprensión de las representaciones de la atención visual obtenidas por nuestro modelo, gracias a su uso de características tradicionales, tales como el color o el movimiento. Además, se observa que la detección de objetos o elementos sencillos como el rostro de las personas o el texto en una pantalla, modelados a continuación a partir de distribuciones espaciales discretas, así como el uso de \emph{Redes Neuronales Convolucionales} o CNNs para obtener mapas de características basados en objetos, incrementa notablemente el rendimiento del sistema, permitiendo mejorar los resultados obtenidos por una gran variedad de métodos en el estado del arte a la hora de estimar la atención visual espacio-temporal.

En la segunda parte de la tesis, se describe nuestra segunda aproximación, denominada ST-T-ATTEN. Con este nuevo modelo se da un paso hacia adelante y se estima la atención en el dominio temporal, a partir de estimaciones de la atención visual espacio-temporal. La hipótesis fundamental de nuestro modelo establece que la atención en el dominio temporal puede estimarse midiendo la dispersión de la localización de las fijaciones proporcionadas por diferentes sujetos. En primer lugar, para demostrar esta hipótesis, se mide la correlación existente entre las secuencias definidas por el movimiento de los ojos de diferentes sujetos cuando sucede un evento importante o anómalo, dadas las secuencias de vídeo recogidas en la base de datos BOSS \cite{BOSS}. A pesar de que este nivel de atención temporal constituye un indicador muy útil para detectar eventos importantes en escenarios complejos y concurridos, la atención en el dominio temporal ha de ser considerada siempre como un mecanismo de filtrado que permite seleccionar intervalos de tiempo candidatos a contener eventos sospechosos y que, por tanto, reduce el procesado posterior que tendría que llevar a cabo un sistema de detección de anomalías. Teniendo en cuenta esta hipótesis, el algoritmo ST-T-ATTEN trata de modelar la atención en el dominio temporal a partir de estimaciones de la atención visual espacio-temporal.

Motivados por el reciente éxito de las CNNs para el aprendizaje de representaciones jerárquicas profundas, así como de las LSTMs para el modelado de series temporales, el algoritmo propuesto se compone de dos etapas. La primera etapa, definida como \emph{Spatio-Temporal visual ATtention NETwork} o ST-ATTEN, consiste en una red de tipo CED que recibe a su entrada tres mapas de características de alto nivel para modelar la atención visual, basados en el color, movimiento y los objetos presentes en la escena. Todos estos mapas se obtienen a partir de CNNs. A continuación, esta arquitectura de tipo codificador-decodificador convolucional permite tanto estimar mapas de atención visual espacio-temporal como obtener representaciones latentes de la atención visual. Además, se proponen dos configuraciones para este módulo del sistema, las cuales se diferencian en las capas inicial y final del codificador y decodificador, respectivamente. En la primera configuración, estas capas son convolucionales, mientras que en la segunda son convolucionales de tipo LSTM.

La segunda etapa de nuestro sistema, denominada \emph{Temporal ATtention NETwork} o T-ATTEN, es una arquitectura de tipo LSTM, la cual permite estimar, para cada frame en una secuencia de vídeo, la atención en el dominio temporal. También se distingue entre dos versiones de T-ATTEN, dependiendo de si éste recibe a su entrada el mapa de atención visual espacio-temporal a la salida del decodificador en ST-ATTEN o, en cambio, la representación latente generada por el codificador.

A continuación, se ha evaluado la arquitectura ST-T-ATTEN propuesta en el escenario de videovigilancia definido por la base de datos BOSS \cite{BOSS}, la cual incluye secuencias de vídeo que han sido grabadas en un contexto ferroviario, y que contienen diferentes tipos de eventos anómalos o sospechosos (varios abusos a mujeres, el robo de un teléfono móvil, una pelea entre pasajeros, etc.). El objetivo principal de los experimentos de la segunda parte de la tesis es evaluar las diferentes arquitecturas propuestas para nuestro modelo ST-T-ATTEN. A partir de estos experimentos, se ha determinado que la mejor configuración para nuestra arquitectura consiste, en primer lugar, en una etapa ST-ATTEN con capas convolucionales, la cual permite fusionar de manera efectiva la información proporcionada por los tres mapas de características a su entrada. Después, el módulo T-ATTEN ofrece similares prestaciones tanto si recibe a su entrada un mapa o una representación latente de la atención visual. 

Finalmente, se describen dos aplicaciones potenciales a nivel de usuario de nuestra propuesta. Por un lado, dado un escenario de videovigilancia, la atención estimada en el dominio temporal puede aplicarse para seleccionar en tiempo real las cámaras más importantes en un array de monitores, dirigiendo la atención de los operadores hacia aquellas cámaras que potencialmente muestran anomalías o eventos sospechosos. Por otro lado, esta atención temporal se puede aplicar también en tareas off-line que implican la visualización de una gran cantidad de horas de grabaciones de videovigilancia, reduciendo la cantidad de información que los operadores tienen que procesar. Por tanto, se puede concluir que, introduciendo algunas mejoras al sistema propuesto, éste podría ser capaz de proporcionar a los operadores una experiencia completa de la atención visual, identificando no únicamente las localizaciones más llamativas de la escena, sino también seleccionando intervalos temporales relevantes, de acuerdo con los eventos que han tenido lugar previamente en la escena, así como con los eventos que están sucediendo en otras cámaras al mismo tiempo. 
 
\section*{Líneas futuras de investigación}
Finalmente, en esta sección se identifican y comentan las líneas futuras de investigación más prometedoras en relación con el trabajo presentado en esta tesis.

Llegados a este punto, no cabe duda acerca de las enormes ventajas que tiene el modelado de la atención visual dentro del campo de la Inteligencia Artificial. Tampoco sobre las infinitas posibilidades que un concepto tan abstracto tiene para el procesado y la interpretación de un mundo en el que cada vez se maneja una mayor cantidad de datos. A pesar de la gran variedad de modelos computacionales de atención visual existentes en la literatura, todavía queda mucho camino por recorrer, no sólo para lograr un sistema que modele automáticamente esta función cognitiva, sino también para entender cómo el sistema visual humano lleva a cabo este proceso de optimización.

Teniendo en cuenta los dos paradigmas más populares en la actualidad para el aprendizaje de representaciones, los cuales se basan en el Aprendizaje Profundo, \emph{Deep Learning} o DL, y los Modelos Gráficos Probabilísticos, \emph{Probabilistic Graphical Models} o PGM, nuestras contribuciones han demostrado la importancia tanto de la tarea de \emph{percibir}, desempeñada por representaciones jerárquicas profundas, como de la habilidad de \emph{deducir}, característica de los modelos PGM, a la hora de modelar e interpretar la atención visual.

En primer lugar, es importante conseguir buenas representaciones del mundo que nos rodea para modelar la atención, y es ahí donde las redes profundas y, en particular, las CNNs, desempeñan un papel fundamental en la percepción automática. Además, dado que la atención visual lleva a cabo no una, sino varias tareas complejas, es fundamental poder interpretar cómo un modelo computacional hace uso de las representaciones jerárquicas proporcionadas por redes profundas. Esto se puede conseguir a partir de métodos probabilísticos que permitan definir relaciones entre las variables observadas. Esta dirección, definida recientemente como \emph{Bayesian Deep Learning} o BDL \cite{wang2016towards}, es la que queremos seguir en trabajos futuros, prestando una especial atención a la aplicación de BDL a los modelos probabilísticos de temas latentes \cite{LDA,sLDA,DBAyang}, los cuales constituyen la base de nuestra primera aproximación para la interpretación de la atención visual: ATOM. Si consiguiéramos definir subtareas no solamente en el espacio, sino también a lo largo del tiempo, podríamos establecer relaciones entre los conceptos reconocidos en una o varias secuencias de vídeo, tanto en la misma escena como en escenas diferentes.

En segundo lugar, se ha demostrado en la segunda parte de la tesis las importantes ventajas que tiene modelar la atención en el dominio temporal, la cual permite seleccionar intervalos temporales de especial importancia en secuencias de vídeo. Estos intervalos seleccionados ayudan a reducir, además, la carga computacional en posibles aplicaciones a nivel de usuario. Desde esta perspectiva, la atención visual apenas ha sido tratada en el estado del arte hasta la fecha, a pesar de su utilidad para el procesado y el análisis de grandes cantidades de información visual, en aplicaciones como la detección de anomalías.

Una línea de investigación interesante que no se ha tratado en esta tesis es la interpretación de las secuencias definidas por el movimiento de los ojos, lo cual facilitaría la implementación de sistemas de mayor comprensión y utilidad para estimar la variación de la atención visual a lo largo del tiempo. Para ello, creemos que el uso de métodos de aprendizaje por refuerzo o \emph{Reinforcement Learning} puede ser un camino prometedor a seguir \cite{mnih2014recurrent}.

Por último, estamos motivados a continuar estudiando el modelado tanto de la atención visual espacio-temporal como de la atención en el dominio temporal en secuencias de vídeo reproducidas al mismo tiempo, con el objetivo de ayudar a los expertos en escenarios complejos y concurridos. Para ello, en los próximos meses se procederá a anotar bases de datos de vídeo grandes, tales como VIRAT \cite{oh2011large} o UCF-Crime \cite{sultani2018real}, con fijaciones de diferentes sujetos, las cuales servirán para un análisis más completo de la arquitectura ST-T-ATTEN propuesta, así como para introducir posibles mejoras en la misma.
\end{otherlanguage}
\endgroup
%
%\vfill

\cleardoublepage%*******************************************************
% Abstract
%*******************************************************
\renewcommand{\abstractname}{Abstract}
\pdfbookmark[1]{Abstract}{Abstract}
\begingroup
\let\clearpage\relax
\let\cleardoublepage\relax
\let\cleardoublepage\relax

\chapter*{Abstract}
%Short summary of the contents in English\dots a great guide by
%Kent Beck how to write good abstracts can be found here:
%\begin{center}
%\url{https://plg.uwaterloo.ca/~migod/research/beckOOPSLA.html}
%\end{center}
This PhD. Thesis concerns the study and development of hierarchical representations for spatio-temporal visual attention modeling and understanding in video sequences. More specifically, we propose two computational models for visual attention. First, we present a generative probabilistic model for context-aware visual attention modeling and understanding. Secondly, we develop a deep network architecture for visual attention modeling, which first estimates top-down spatio-temporal visual attention, and ultimately serves for modeling attention in the temporal domain. 

The first part of the thesis introduces our first proposal: a generative probabilistic framework for spatio-temporal visual attention modeling and understanding. The model proposed is generic, independent of the application scenario and founded on the most outstanding psychological studies about attention. Moreover, it considers the existing concurrence between bottom-up and top-down factors. 

Drawing in the well-known Latent Dirichlet Allocation method for the analysis of large corpus of data, and some of its supervised extensions, our approach defines task- or context-driven visual attention in video as a mixture of latent sub-tasks, which are in turn represented as combinations of low-, mid- and high-level spatio-temporal features. Latent sub-tasks discovered are automatically aligned to the information drawn from human fixations by means of a categorical variable response, which is generated by a logistic regression model over the sub-task proportions. Therefore, our algorithm incorporates an intermediate level formed by latent sub-tasks, which bridges the gap between features and visual attention, and enables to obtain more comprehensible interpretations of attention guidance. 

The experiments related to our first approach demonstrate its ability to successfully learn hierarchical representations of visual attention, specifically adapted to diverse contexts, on the basis of a wide set of features. Besides, results show how our proposal significantly outperforms quite a few competent methods in the literature when estimating visual attention. 

The second part of the thesis presents our second proposal: a deep network architecture that takes a step further and goes from spatio-temporal visual attention prediction to attention estimation in the temporal domain. The model is fundamentally supported by the assumption that a measurement of task-driven visual attention in the temporal domain can be drawn from the dispersion of fixation locations recorded from several observers. Although this temporal level of attention constitutes a useful clue to detect important events in crowded and complex scenarios, attention in the temporal domain should be considered as an early filtering mechanism, which selects candidate time segments to contain suspicious events, and therefore reduces the later processing devoted to the anomaly detection. 

Based on this hypothesis, and inspired by the recent success of Convolutional Neural Networks for learning deep hierarchical representations and Long Short-Term Memory Units for time series forecasting, our approach is composed of two stages. On the one hand, the first stage consists of a Convolutional Encoder Decoder network that receives at its input three high-level feature maps for visual attention guidance: RGB-based, motion and objectness. Then, through an encoding-decoding architecture, the network concurrently estimates spatio-temporal visual attention maps and extracts latent representations of visual attention. On the other hand, the second stage involves an architecture based on Long Short-Term Memory Units that estimates, for each frame in a video sequence, a temporal attention response. We propose different configurations for both stages, in order to assess various architectures of our proposal.

Finally, the second approach proposed is evaluated in a video surveillance scenario, which contains video sequences recorded in a railway transport context, with different types of suspicious or anomalous events. In addition, we discuss two potential end-user applications for our proposal. On the one hand, given a surveillance scenario, the estimated temporal attention response could be applied to select in real-time the most outstanding screens from the monitoring array, thus driving operator's attention to scenes that potentially show anomalies or suspicious events. On the other hand, this response could be also applied in off-line tasks which imply reviewing many hours of surveillance recordings, reducing the information to be processed by the operator. 

%\vfill
%
%\begin{otherlanguage}{spanish}
%\pdfbookmark[1]{Resumen}{Resumen}
%\chapter*{Resumen}
%COMPLETAR
%
%\end{otherlanguage}
%
\endgroup
%
%\vfill

%*******************************************************
% Table of Contents
%*******************************************************
\pagestyle{scrheadings}
%\phantomsection
\refstepcounter{dummy}
\pdfbookmark[1]{\contentsname}{tableofcontents}
\setcounter{tocdepth}{2} % <-- 2 includes up to subsections in the ToC
\setcounter{secnumdepth}{3} % <-- 3 numbers up to subsubsections
\manualmark
\markboth{\spacedlowsmallcaps{\contentsname}}{\spacedlowsmallcaps{\contentsname}}
\tableofcontents
\automark[section]{chapter}
\renewcommand{\chaptermark}[1]{\markboth{\spacedlowsmallcaps{#1}}{\spacedlowsmallcaps{#1}}}
\renewcommand{\sectionmark}[1]{\markright{\thesection\enspace\spacedlowsmallcaps{#1}}}
%*******************************************************
% List of Figures and of the Tables
%*******************************************************
\clearpage
\begingroup
    \let\clearpage\relax
    \let\cleardoublepage\relax
    %*******************************************************
    % List of Figures
    %*******************************************************
    %\phantomsection
    \refstepcounter{dummy}
    \addcontentsline{toc}{chapter}{\listfigurename}
    \pdfbookmark[1]{\listfigurename}{lof}
    \listoffigures

    \vspace{8ex}
    \newpage

    %*******************************************************
    % List of Tables
    %*******************************************************
    %\phantomsection
    \refstepcounter{dummy}
    \addcontentsline{toc}{chapter}{\listtablename}
    \pdfbookmark[1]{\listtablename}{lot}
    \listoftables

    \vspace{8ex}
    \newpage

    %*******************************************************
    % List of Listings
    %*******************************************************
    %\phantomsection
    %\refstepcounter{dummy}
    %\addcontentsline{toc}{chapter}{\lstlistlistingname}
    %\pdfbookmark[1]{\lstlistlistingname}{lol}
    %\lstlistoflistings

    %\vspace{8ex}

    %*******************************************************
    % Acronyms
    %*******************************************************
    %\phantomsection
    \refstepcounter{dummy}
    \addcontentsline{toc}{chapter}{Acronyms}
    \pdfbookmark[1]{Acronyms}{acronyms}
    \markboth{\spacedlowsmallcaps{Acronyms}}{\spacedlowsmallcaps{Acronyms}}
    \chapter*{Acronyms}
    \begin{acronym}[UMLX]
    	\acro{AI}{Artificial Intelligence}
    	\acro{AIM}{Attention based on Information Maximization}
    	\acro{AT}{Attractive Topic}
    	\acro{ATOM}{visual Attention TOpic Model}
    	\acro{AUC}{Area Under \acs{ROC} Curve}
    	\acro{AWS}{Adaptive Whitening Saliency}
    	\acro{AWS-D}{Dynamic Adaptive Whitening Saliency}
    	\acro{BDL}{Bayesian Deep Learning}
    	\acro{BN}{Batch Normalization}
    	\acro{BoW}{Bag-of-Words}
    	\acro{BPTT}{Back-propagation Through Time}
    	\acro{BU}{Bottom-Up}
    	\acro{CC}{Correlation Coefficient}
    	\acro{CCTV}{Closed-Circuit TeleVision}
		\acro{CED}{Convolutional Encoder Decoder}    	
    	\acro{CNN}{Convolutional Neural Network}
    	\acro{CONV}{Convolutional}
    	\acro{CPU}{Central Processing Unit}
    	\acro{CRF}{Conditional Random Field}
    	\acro{C-A}{Context-Aware}
    	\acro{C-G}{Context-Generic}
    	\acro{DBA}{Dirichlet-Bernoulli Alignment}
    	\acro{DL}{Deep Learning}
    	\acro{DNN}{Deep Neural Network}
    	\acro{DoG}{Difference of Gaussians}
    	\acro{EDN}{Encoder-Decoder Network}
    	\acro{ELBO}{Evidence Lower Bound}
    	\acro{ELU}{Exponential Linear Unit}
    	\acro{EM}{Expectation-Maximization}
    	\acro{EMA}{Exponential Moving Average}
    	\acro{FC}{Fully-Connected}
    	\acro{FIT}{Feature Integration Theory}
    	\acro{FOA}{Focus of Attention}
    	\acro{FP}{False Positive}
    	\acro{GAN}{Generative Adversarial Network}
    	\acro{GBVS}{Graph-Based Visual Saliency}
    	\acro{GMM}{Gaussian Mixture Model}
    	\acro{GPU}{Graphics Processing Unit}
    	\acro{GRU}{Gated Recurrent Unit}
        \acro{GSM}{Guided Search Model}
        \acro{GT}{Ground-Truth}
        \acro{HMM}{Hidden Markov Model}
        \acro{HOG}{Histogram of Oriented Gradients}
        \acro{HVS}{Human Visual System}
        \acro{ICA}{Independent Component Analysis}
        \acro{ICL}{Incremental Coding Length}
        \acro{IID}{Independent and Identically Distributed}
        \acro{IOR}{Inhibition of Return}
        \acro{IFT}{Inverse Fourier Transform}
        \acro{ILSVRC}{ImageNet Large Scale Visual Recognition Competition}
        \acro{IT}{Inhibiting Topic}
        \acro{KL}{Kullback-Leibler divergence}
        \acro{KNN}{K-Nearest Neighbors}
        \acro{LDA}{Latent Dirichlet Allocation}
        \acro{LGN}{Lateral Geniculate Nucleus}
        \acro{LN}{Layer Normalization}
        \acro{LSK}{Local Steering Kernels}
        \acro{LSTM}{Long Short-Term Memory}
        \acro{LTM}{Latent Topic Model}
        \acro{MAE}{Mean Absolute Error}
        \acro{MBGD}{Mini-Batch Gradient Descent}
        \acro{ML}{Machine Learning}
        \acro{MLP}{Multi-layer Perceptron}
        \acro{MPSE}{Mean Pairwise Squared Error}
        \acro{MRF}{Markov Random Field}
        \acro{MSE}{Mean Squared Error}
        \acro{NLP}{Natural Language Processing}
        \acro{NN}{Neural Network}
        \acro{NSS}{Normalized Scanpath Saliency}
        \acro{NUS}{Non-Uniform Sampling}
        \acro{PCA}{Principal Component Analysis}
        \acro{PCC}{Pearson Correlation Coefficient}
        \acro{PDF}{Probability Density Function}
        \acro{PGM}{Probabilistic Graphical Models}
        \acro{PLSA}{Probabilistic Latent Semantic Analysis}
        \acro{POOL}{Pooling}
        \acro{PQFT}{Phase spectrum of Quaternion Fourier Transform}
        \acro{RAM}{Random Access Memory}
        \acro{RBM}{Restricted Boltzmann Machine}
        \acro{ReLU}{Rectified Linear Unit}
        \acro{RMSProp}{Root Mean Squared Propagation}
        \acro{RNN}{Recurrent Neural Network}
        \acro{ROC}{Receiver Operating Characteristic}
        \acro{ROI}{Region of Interest}
        \acro{SAM}{Saliency Attentive Model}
        \acro{sAUC}{Shuffled Area Under ROC Curve}
        \acro{SC}{Superior Colliculus}
        \acro{SDSR}{Saliency Detection by Self-Resemblance}
        \acro{SGD}{Stochastic Gradient Descent}
        \acro{SIFT}{Scale-Invariant Feature Transform}
        \acro{sLDA}{Supervised Latent Dirichlet Allocation}
        \acro{sNSS}{Shuffled Normalized Scanpath Saliency}
        \acro{SM}{Saliency Map}
        \acro{SSD}{Sum of Squared Differences}
        \acro{ST-ATTEN}{Spatio-Temporal visual ATtention NETwork}
        \acro{ST-T-ATTEN}{Spatio-Temporal to Temporal visual ATtention NETwork}
        \acro{SUN}{Saliency Using Natural Statistics}
        \acro{SVM}{Support Vector Machine}
        \acro{T-ATTEN}{Temporal ATtention NETwork}
        \acro{TD}{Top-Down}
        \acro{TP}{True Positive}
        \acro{VAM}{Visual Attention Map}
        \acro{V1}{Primary Visual Cortex}
        \acro{VIDYA}{Variable Index Dynamic Average}
        \acro{WMAP}{Weighted Maximum Phase Alignment}
        \acro{WTA}{Winner-Take-All}
        \acro{1D}{one dimensional}
        \acro{2D}{two dimensional}
        \acro{3D}{three dimensional}
    \end{acronym}

\endgroup

\cleardoublepage
%********************************************************************
% Mainmatter
%*******************************************************
\cleardoublepage
\pagestyle{scrheadings}
\pagenumbering{arabic}
%\setcounter{page}{90}
% use \cleardoublepage here to avoid problems with pdfbookmark
\cleardoublepage
%\part{Some Kind of Manual}\label{pt:manual}
%************************************************
\chapter{Introduction}\label{ch:introduction}
\chaptermark{Introduction}
%************************************************

This introductory chapter is organized as follows. First, we make a short introduction to visual attention in Section \ref{sec:visual_attention_intro}, enumerating its different types and possible applications. Secondly, in Section \ref{sec:hierarchical_representations_intro}, we discuss the use of hierarchical representations for visual attention, going from feature engineering to feature learning, through a brief description of the types of \acf{ML} covered by the systems presented in the thesis. Then, in Section \ref{sec:objectives}, we introduce the main focus of the thesis, which is the study and development of hierarchical representations for spatio-temporal visual attention modeling and understanding. Finally, Section \ref{sec:structure} summarizes the structure and contributions of this dissertation.
   
\section{Visual attention}\label{sec:visual_attention_intro}
``The world and its universe are, to anything or anyone with senses, incomprehensibly big data.'' (Mark Andrejevic, 2014) \cite{andrejevic2014big} 

We have been always surrounded by data. However, never before had we lived in such a data-driven world. Nowadays, unstoppable technological advances make possible to capture almost everything, anywhere and anytime, which has resulted in a massive amount of information that is necessary to filter and process. 

Within the framework of \acf{AI}, Computer Vision \cite{forsyth2011computer} emerged in the late 1960s with the objective of automatically simulating the \acf{HVS} functions. Drawing from the visual information captured in digital images and video sequences, this interdisciplinary field seeks to discover good representations of the real-world in order to carry out particular tasks such as object location \cite{DeepSaliencyObject} and recognition \cite{krizhevsky2012imagenet}, event detection \cite{anomaliesDeLaCalle} or visual tracking \cite{smeulders2014visual}.

In spite of the wide variety of systems that are continuously released and improved to solve these tasks, some of them truly effective, they still need to process large amounts of visual information for achieving high performances, which dramatically impacts on their efficiency. Human beings, however, inherently select the most important elements to interact in a context and, besides, are rapidly attracted by striking stimulus. And this is thanks to the visual attention function of the \acs{HVS}, which can be understood as an optimization process for visual cognition and perception. If we were able to design image-understanding algorithms that accomplish this operation, we could use them to reduce their computational cost. At the same time, we would help users and experts when dealing with applications and complex scenarios which require processing large amounts of information simultaneously, such as driving \cite{pradhan2005using}, aviation \cite{kilingaru2013monitoring} and video surveillance \cite{howard2011task}, reducing the probability of human errors and speeding up the decision making processes.

Visual attention can be readily identified in two different domains, spatial and temporal, which allow to define three types of computational models for visual attention: spatial, spatio-temporal and temporal \cite{Borji:2013:SVA:2412386.2412937}. Most of existing models consider a spatial component to guide information processing to conspicuous locations or areas of particular interest in a scene. Moreover, visual information in real world is dynamic, so it is equally important to model how it changes over time, in order to update spatial attention based on previously selected locations, which allows modeling visual attention in a spatio-temporal manner, as well as selecting time segments of special importance.   

It is also common to distinguish between two families of visual attention models: \emph{stimulus-driven} \acf{BU} models, which are based on visual features of the scene, and \emph{goal-driven} \acf{TD} approaches, which take into account prior knowledge or advanced indications \cite{TreismanGelade80, wolfe1994guided}. Eye movements play a major role in this second type of models, by providing information about which locations are essential for perception and how long they are fixated \cite{Yarbus1967, wolfe1994guided}. Although we live in a spatio-temporal reality, the majority of existing computational models for visual attention are \acs{BU} and have been built for still images. What is more, the few available \acs{TD} methods have been designed for well-determined scenarios, and are not applicable to other contexts. Finally, there is still room for models that take advantage of the demonstrated concurrence between \acs{BU} and \acs{TD} factors.

\section{Hierarchical representations for visual attention}\label{sec:hierarchical_representations_intro}
At present, most of the computer vision-based applications are addressed via feature-based algorithms, which often imply \acf{ML} and optimization methods. The performance of these applications is highly dependent on features or representations extracted from the visual information beforehand. Hence, features constitute themselves a major and prevailing area of research, which has rapidly evolved in the last few years, from traditional handcrafted feature engineering to high-level representation learning \cite{bengio2013representation}.

\subsection{Feature engineering}
Traditional \emph{feature engineering} involves transforming the domain knowledge of the data into features or properties common to all objects or items considered in a particular task. This process is difficult, time-consuming and requires expert knowledge \cite{ng2013machine}. Furthermore, the performance of a \acs{ML} model significantly depends on the quality and quantity of the features obtained. The better the features are, the simpler and more flexible the model needed will be. 

Attending to their semantic meaning, image features can be classified into three groups \cite{martinet2012mid}: low-, mid- and high-level descriptors. Low-level descriptors, such as color histograms, texture and shape features, capture either global or local visual properties, and can be directly extracted from the whole image or local regions, respectively. Mid-level features constitute an intermediate step between low and high level, and rely on a global analysis of low-level descriptors, in order to perform annotation or similarity matching tasks, among others. High-level features represent semantic concepts, interpretable by humans, such as faces, cars or any kind of objects, as well as simple categorizations (``urban vs. countryside'', ``indoor vs. outdoor'', etc.).

In the field of visual attention, a great effort has been made from multiple perspectives to determine which features better represent those conspicuous areas of the scene for observers \cite{wolfe2014approaches}. According to the most widely accepted psychological theories \cite{TreismanGelade80, wolfe1994guided}, there are three features which mainly attract human attention: intensity or luminance contrast, color and orientation. Then, some other attributes, such as motion, shape or faces, might be useful to develop a system that simulates the \acs{HVS}. Most computer vision researchers have modeled these properties separately, in order to develop computational mechanisms for predicting visual attention. However, only few works have tried to understand how they are combined to perform this function.

In contrast to feature engineering, \emph{feature or representation learning} \cite{bengio2009learning, bengio2013representation} encompasses those \acs{ML} techniques that automatically transform the data at the system input into abstract representations which allow an \acs{AI} to understand the world around us, improving its performance when solving a particular task. Indeed, representation learning often constitutes a preprocessing stage previous to a prediction problem. In the following subsections, we first briefly describe the types of \acs{ML} covered by the systems presented in this thesis. Afterwards, we introduce representation learning and discuss the issues that should be addressed by a good representation. 

\subsection{Machine Learning}\label{sec:machine_learning}
Two definitions of \acf{ML} are usually highlighted. First, the informal, traditional one stated by Arthur Samuel in 1959: \emph{``Machine learning is the field of study that gives computers the ability to learn without being explicitly programmed."} \cite{samuel1959some}. The more recent, by Tom Mitchell, establishes that \emph{``A computer program is said to learn from experience E with respect to some class of tasks T and performance measure P, if its performance at tasks in T, as measured by P, improves with experience E."} \cite{Mitchell:1997:ML:541177}.

\begin{itemize}
	\item \emph{Experiences E} are related to the nature of \acs{ML} methods, which define the way they process the information available in a database or collection or examples. Examples are features measured or extracted from the existing objects or events in the world. For instance, we can measure the value of the pixels of an image, but we can also extract its corresponding edge image. 
	
	According to the ``no free lunch'' theorem of David Wolpert and William Macready \cite{wolpert1997no}, there is no \acs{ML} algorithm universally better than any other, and the goal of \acs{ML} is thus to determine what is the way of experiencing that provides an \acs{AI} with the most relevant distributions to understand a particular real-world scenario.
	
	The two main different types of \acs{ML} are supervised and unsupervised learning. In this thesis, we will attend both to their generative and discriminative paradigms from either a probabilistic or a functional perspective, with the purpose of framing elaborated contributions to spatio-temporal visual attention, based on \acfp{LTM} (Chapter \ref{ch:atom}, section \ref{sec:latent_topic_models}) and \acfp{DNN} (Chapter \ref{ch:anomaly_detection}, section \ref{sec:deep_neural_networks}).
		
	A third less common, but also active, research area of \acs{ML} is known as \emph{reinforcement learning}, which consists on learning suitable actions to perform in order to maximize a reward function. Reinforcement learning has recently been employed in computer vision for object location \cite{caicedo2015active,buenoa2017hierarchical} and image classification \cite{mnih2014recurrent}. The authors of the latter reference also tested it to automatically play a simple game.
	
	\item \emph{Tasks T} determine how \acs{ML} algorithms process the examples in a database. Some examples of \acs{ML} tasks are classification, regression and density estimation. Tasks in computer vision such as image classification, retrieval and segmentation have been tackled both in supervised \cite{melanomas,ferfer2017exploiting,martinez2014bayesian} or unsupervised \cite{GONZALEZDIAZ20132437,zhang2008image} experiences. The paramount tasks we aim to solve in this thesis are spatio-temporal and temporal visual attention modeling: first, we interpret how spatio-temporal visual attention works in several contexts, and then we apply it in a video surveillance scenario for temporal modeling of attention.
	
	\item \emph{Performance measures P} evaluate how a \acs{ML} algorithm works, and are often tailored to the tasks carried out by the system. This evaluation is performed using a test set of data different from the one used for training the system. Metrics to assess spatio-temporal visual attention and temporal attention estimation are in the scope of Chapters \ref{ch:atom_experiments} and \ref{ch:surveillance_experiments}, which cover the experiments undertaken during the thesis.
\end{itemize}

The interested reader is referred to \cite{Bishop:2006:PRM:1162264, Murphy2013Machine, goodfellow2016deep} for further insight into the different methods and concepts in \acs{ML}. In an attempt to include all the existing \acs{ML} algorithms in a unique taxonomy, Goodfellow et al. propose in \cite{goodfellow2016deep} an easy recipe: combine a database, a function or probability distribution to approximate, an optimization procedure and a model.

\subsection*{Supervised learning}\label{sec:supervised_learning}
\begin{figure*}[!t]
\centering
	\includegraphics[trim=0cm 0cm 0cm 0cm, width=1\textwidth]{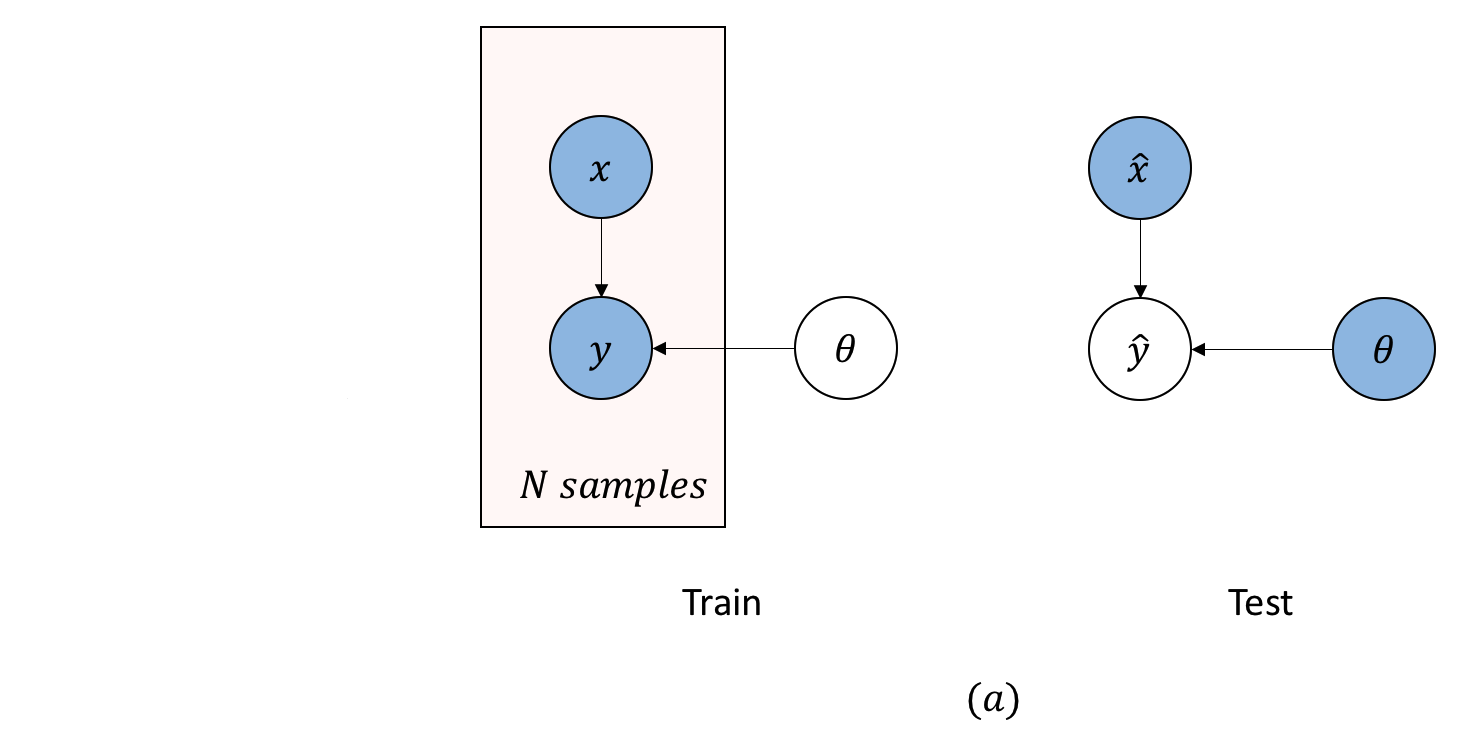}\\	
	\includegraphics[trim=0cm 0cm 0cm 0cm, width=1\textwidth]{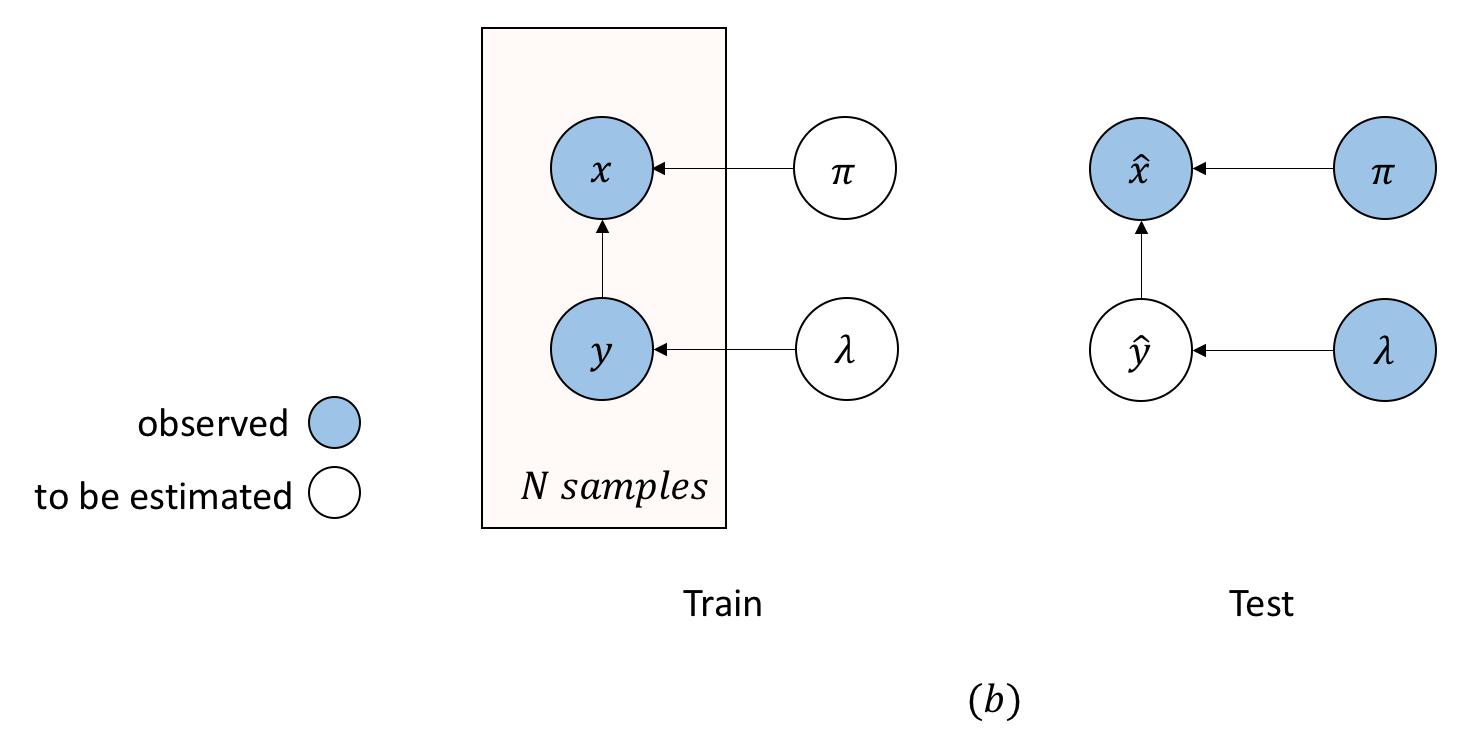}
\caption[Graphical representation of Bayesian (a) discriminative and (b) generative supervised models.]{Graphical representation of Bayesian (a) discriminative and (b) generative supervised models. Shaded nodes represent $N$ independent input-output example pairs $(x,y)$, and white nodes indicate parameter vectors to be estimated. Edges show the conditional dependence between variables.}
\label{31_supervised}
% \vspace{-0.4cm}
\end{figure*}

\emph{Predictive} or \emph{supervised learning} is the most common experience of \acs{ML}. Its goal is to predict the value of a \emph{response variable} vector or target $\hat{\mathbf{y}}$ given the value of a vector $\hat{\mathbf{x}}$ of input features, by means of a model learned from a \emph{training set} $\{(\mathbf{x}_{n},\mathbf{y}_{n})\}_{n=1}^{N}$ of $N$ input-output example pairs sampled from a true data distribution $\mathsf{D}$. This can be obtained via a discriminative or a generative model \cite{ng2002discriminative, bishop2007}. Figure \ref{31_supervised} shows the graphical representation of these two approaches. A graphical structure defines the conditional dependence between variables in a model \cite{koller2009graphical}.

While in a classification or recognition task $\mathbf{y}_{n}$ is a categorical variable from a finite set $\mathbf{y}_{n} \in \{1,...,C\}$, the problem is called regression if $\mathbf{y}_{n}$ involves one or more continuous variables. Considering that models in Figure \ref{31_supervised} are parametric, we will illustrate their differences by solving the following supervised classification problem.

Given the training set introduced before, let us denote individually $\mathbf{X}=\{\mathbf{x}_{1},...,\mathbf{x}_{N}\}$ as the set of $N$ input vectors and $\mathbf{Y}=\{\mathbf{y}_{1},...,\mathbf{y}_{N}\}$ as their corresponding classes, assumed independently sampled from the same distribution $\mathsf{D}$.

On the one hand, we can address \emph{discriminative approaches} from a deterministic or a probabilistic point of view:

\begin{itemize}
	\item From a \emph{deterministic} or functional perspective, the objective of supervised applications is to learn the mapping $f:\mathbf{X} \mapsto \mathbf{Y}$ between the input feature space $\mathbf{X}$ and the class space $\mathbf{Y}$. This mapping is defined by means of the optimal function $f^{*}$, from a set of parametrized functions $F$, that minimizes the expected value of a loss function $L$ given samples drawn from the true distribution $\mathsf{D}$:

\begin{gather}
f^{*} = \argmin_{f \in F} E_{(\mathbf{x},\mathbf{y}) \sim \mathsf{D}} \left[L\left(f(\mathbf{x}_{n}),\mathbf{y}_{n}\right)\right],
\end{gather} 		

\noindent where $E \left[ \cdot\right]$ stands for the expected value. The loss function $L$ computes the difference between the predicted label $\hat{\mathbf{y}}_{n} = f(\mathbf{x}_{n})$ and the true label $\mathbf{y}_{n}$ and is chosen according to the task performed, just as evaluation metrics.

Because it is not possible to access to all samples in the true data distribution $\mathsf{D}$, the problem is intractable and can be only optimized considering the available $N$ training samples, assuming that they are \acf{IID}, and expecting that they are representative of $\mathsf{D}$, so that it can be expressed as follows:

\begin{gather}
f^{*} = \argmin_{f \in F} \sum_{n=1}^{N} L\left(f(\mathbf{x}_{n}),\mathbf{y}_{n}\right).
\end{gather} 	
   
	\item Alternatively, from a \emph{probabilistic} point of view, $\mathbf{X}$ and $\mathbf{Y}$ are considered random variables and the objective is achieved by obtaining the conditional distribution $p(\mathbf{y}|\mathbf{x})$ of the target $\mathbf{y}$ given the observations $\mathbf{x}$. The aim is to determine the class $\hat{\mathbf{y}}$ associated with a new unseen input vector $\hat{\mathbf{x}}$, which implies evaluating the following conditional distribution:

\begin{gather}
p(\hat{\mathbf{y}}|\hat{\mathbf{x}},\mathbf{X},\mathbf{Y})
\end{gather} 

This distribution can be represented as $p(\hat{\mathbf{y}}|\hat{\mathbf{x}},\theta)$, where $\theta$ constitutes its corresponding set of parameters. Considering the $N$ independent training samples, the likelihood function is expressed as:

\begin{gather}
p(\mathbf{Y}|\mathbf{X},\theta) = \prod_{n=1}^{N} p(\mathbf{y}_{n}|\mathbf{x}_{n},\theta).
\end{gather} 

Assuming a prior $p(\theta)$, its product with the likelihood function provides a joint distribution $p(\theta,\mathbf{Y}|\mathbf{X})$ of the parameters $\theta$ and the classes $\mathbf{Y}$ given the observations $\mathbf{X}$. Then, we can obtain the posterior distribution of $\theta$ as:

\begin{gather}
p(\theta|\mathbf{X},\mathbf{Y}) = \frac{p(\theta,\mathbf{Y}|\mathbf{X})}{p(\mathbf{Y}|\mathbf{X})} = \frac{p(\theta)p(\mathbf{Y}|\mathbf{X},\theta)}{\int p(\theta)p(\mathbf{Y}|\mathbf{X},\theta)d\theta}
\end{gather}

Marginalizing the predictive distribution with respect to $\theta$ weighted by the posterior distribution, we are able to predict $\hat{\mathbf{y}}$ for unseen samples $\hat{\mathbf{x}}$:

\begin{gather}
p(\hat{\mathbf{y}}|\hat{\mathbf{x}},\mathbf{X},\mathbf{Y}) = \int p(\hat{\mathbf{y}}|\hat{\mathbf{x}},\theta)p(\theta|\mathbf{X},\mathbf{Y})d\theta.
\end{gather}
\end{itemize}

On the other hand, \emph{generative approaches} learn a probabilistic model of the joint distribution $p(\mathbf{x},\mathbf{y}|\theta)$ of the feature vector $\mathbf{x}$ and the class label $\mathbf{y}$, conditioned on a set of parameters $\theta=\{\lambda,\pi\}$. Given a prior probability for the classes $p(\mathbf{y}|\lambda)$ together with a class-conditional density for each class $p(\mathbf{x}|\mathbf{y},\pi)$, we can express $p(\mathbf{x},\mathbf{y}|\theta)$ as:

\begin{gather}
p(\mathbf{x},\mathbf{y}|\theta) = p(\mathbf{y}|\lambda)p(\mathbf{x}|\mathbf{y},\pi).
\end{gather}

\noindent Then, the joint distribution is obtained by drawing from the $N$ independent training samples as follows:

\begin{gather}
p(\mathbf{X},\mathbf{Y},\theta) = p(\theta)\prod_{n=1}^{N} p(\mathbf{x}_{n},\mathbf{y}_{n}|\theta).
\end{gather}

\noindent This distribution has to be maximized in order to determine the most probable value of $\theta$. According to Bayes' Rule:

\begin{gather}
p(\mathbf{X},\mathbf{Y},\theta)=p(\theta|\mathbf{X},\mathbf{Y})p(\mathbf{X},\mathbf{Y}). 
\end{gather}

\noindent Hence, maximizing $p(\mathbf{X},\mathbf{Y},\theta)$ is equivalent to maximize the posterior distribution $p(\theta|\mathbf{X},\mathbf{Y})$. The posterior distribution can be then used to evaluate $p(\hat{\mathbf{y}}|\hat{\mathbf{x}},\mathbf{X},\mathbf{Y})$ on new samples $\hat{\mathbf{x}}$, in order to make predictions $\hat{\mathbf{y}}$.

The main advantage of generative approaches with respect to discriminative ones is that the joint distribution $p(\mathbf{X},\mathbf{Y},\theta)$ models how the data has been generated, which allows to create new synthetic feature vectors $\hat{\mathbf{x}}$ that follow the same distribution than the the existing samples. In addition, this implies that these methods can benefit from the mixture of labeled and unlabeled data in semi-supervised frameworks. Well-known examples of generative methods are \acfp{GMM} \cite{Bishop:2006:PRM:1162264} and \acfp{HMM} \cite{baum1966statistical}.  

Despite this additional capability, traditional learning algorithms showed generative models limited performance to find optimal model parameters, which leads to the true distributions of the data. Hence, discriminative approaches often provide better generalization performances. \acfp{SVM} \cite{cortes1995support} and \acfp{NN} \cite{goodfellow2016deep} are examples of discriminative methods, as well as the widely used linear and logistic regression models \cite{Murphy2013Machine}, on which trending \acfp{DNN} are based, further explained in section \ref{sec:deep_neural_networks}.

\subsection*{Unsupervised learning}\label{sec:unsupervised_learning}
\begin{figure*}[!t]
\centering
	\includegraphics[trim=0cm 0cm 0cm 0cm, width=0.7\textwidth]{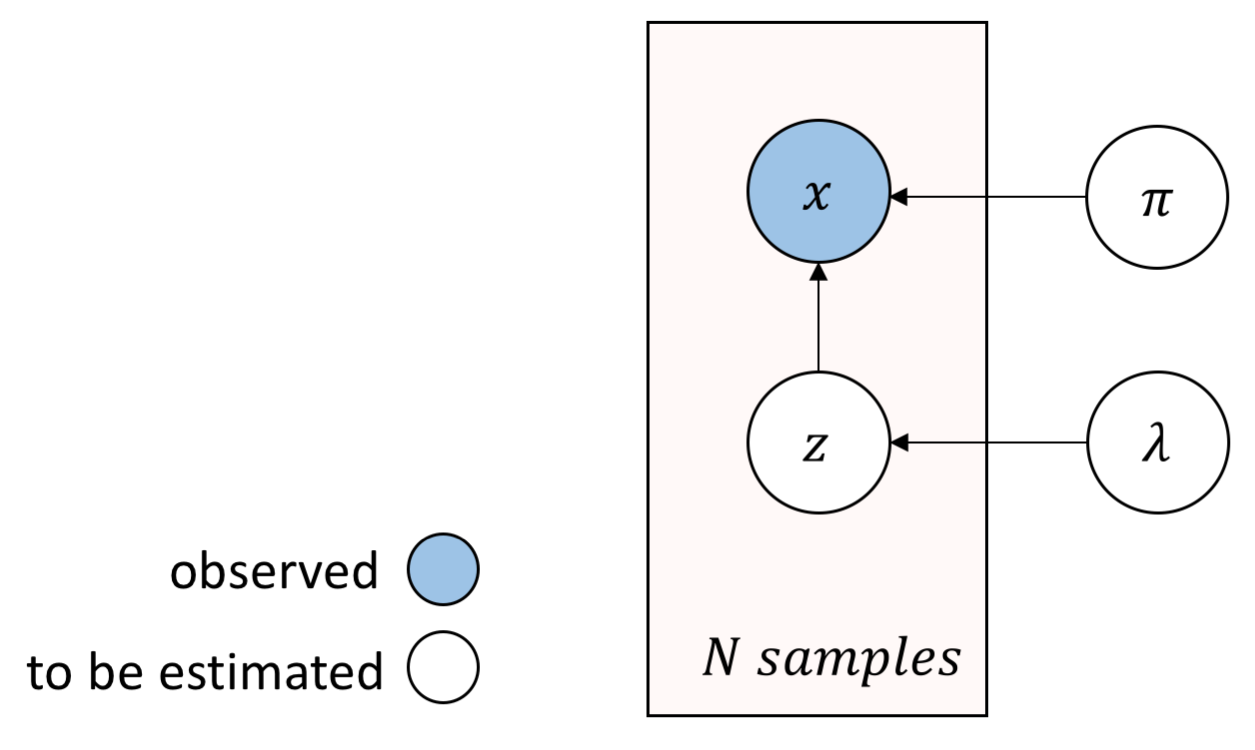}	
\caption[Graphical representation of Bayesian directed generative unsupervised models.]{Graphical representation of Bayesian directed generative unsupervised models. Shaded nodes represent $N$ independent inputs $x$, and white nodes indicate hidden variables $z$ and parameter vectors to be inferred. Edges show the conditional dependence between variables.}
\label{32_unsupervised}
% \vspace{-0.4cm}
\end{figure*}

The second main type of \acs{ML} is called \emph{descriptive} or \emph{unsupervised learning}. Also known as \emph{knowlegde discovery}, its objective is to find patterns of interest in the data, given a set of unlabeled inputs $\mathsf{D} = \{x_{n}\}_{n=1}^{N}$, by means of hidden variables. It should be noted that, despite the frequent use of this type of variables in unsupervised methods, they can be also arisen by supervised models, such as the \acfp{EDN} introduced in section \ref{sec:autoencoders}.

\emph{Latent} or \emph{hidden} variables $\bf z$ are representations of the data not directly observed but rather inferred from other variables that can be directly measured. They reduce the dimensionality of the observable data providing a model that explains and makes this information easier to understand. The underlying structures and relations established can be helpful for clustering data into groups, such as in the well-known \emph{K-means} \cite{Murphy2013Machine} algorithm, and also to reduce the dimensionality of high-dimensional vectors, as in \acs{PCA} \cite{Murphy2013Machine}. 

In this thesis, we will use the well-known \acf{LDA} \cite{LDA} directed generative model, which is detailed in section \ref{sec:lda}, for visual attention understanding. Directed generative models compute the distribution $p({\bf x}|\theta)$ of the data $\bf x$ given the parameters $\theta=\{\lambda,\pi\}$ as follows, by means of the prior $p(z|\lambda)$ of the latent variables $\bf z$ and the conditional distribution $p({\bf x}|z,\pi)$ that establishes the relationship between latent and observed variables:
	
	\begin{gather}
		p({\bf x}|\theta) = \sum_{z} p(z|\lambda)p({\bf x}|z,\pi).
	\end{gather} 

\noindent Figure \ref{32_unsupervised} shows a basic graphical representation for this type of models. \acfp{GAN} \cite{goodfellow2014generative} constitute another recent example of directed generative architecture.
%	
%	\item \emph{Undirected generative models}: In contrast, undirected models define directly the joint distribution $p({\bf x},{\bf z}|\theta)$ of the observed variables $\bf x$ and the hidden variables $\bf z$, so that the distribution of the data is:
%	
%	\begin{gather}
%		p({\bf x}|\theta) = \sum_{z} p({\bf x},z|\theta).
%	\end{gather} 
%	
%	\acfp{MRF} \cite{Murphy2013Machine} and \acfp{RBM} \cite{hinton2012practical} are famous examples of this type of unsupervised learning. 
%\end{itemize}  

\subsection{Representation learning}
One of the current key challenges of \acs{ML} is to model and understand complex abstract concepts such as attention or emotion. In the same way that human beings are able to efficiently process information, researchers pursue automatic methodologies capable of separating, given raw input data, useful from irrelevant information, relating it to basic interpretable features (e.g. color, shape), and representing it in a structured or hierarchical way.

According to the outstanding review of Yoshua Bengio et al. about representation learning \cite{bengio2013representation}, we should take account of the following aspects in order to achieve a good representation: 

\begin{itemize}
	\item A good representation is one that involves multiple explanatory factors of the observed input, which are useful to solve a particular supervised task. 
	\item A hierarchical organization of explanatory factors is always desirable, which describes the world around us by establishing relationships from less abstract concepts (e.g. movie, film director, actress, etc.), to more abstract ones (e.g. art, entertainment, saliency, etc.). 
	\item Semi-supervised frameworks are helpful to take advantage of the capability of complex unsupervised models, which provide latent representations of the world. They allow to maintain a connection between these hidden representations and our semantic concepts and categories, which contribute to a better understanding of how machines see our reality.
	\item Making associations between tasks facilitates solving applications for which we do not have enough information annotated or knowledge to interpret the scenarios they imply. Methodologies such as multi-task and transfer learning are in line with this objective \cite{pan2010survey}.
	\item We should keep in mind the existing correlation between nearby observations, which are often associated with the same semantic or categorical concepts, and change their representations similarly at different spatial and temporal scales.
	\item Finally, it is important to strive for the simplicity of factor dependencies, which is essential to reach efficient representations.
\end{itemize}

Nowadays, we can basically identify two paradigms for representation learning: \acf{DL} and \acf{PGM} \cite{wang2016towards}.

Inspired by the hierarchical architecture of the biological neural system, \acs{DL} methods can be understood as representation methods with multiple layers of representation \cite{bengio2009learning}. Starting with the raw input at the bottom layer, each layer is composed of simple non-linear units that transform its input into a new representation. This representation constitutes the input of a higher, slightly more abstract layer, being the output of the final top layer a lower-dimensional feature at a very high level.

On the other hand, \acs{PGM} learn a set of latent random variables, and make use of structures that define relationships between these variables, in an attempt to represent distributions over the observed data.

It is worth noting the great contribution of \acs{DL} representations to perception tasks such as seeing, proved by object recognition and tracking applications \cite{krizhevsky2012imagenet,wang2013learning}; hearing, performed by speech recognition or audio retrieval systems \cite{hinton2012deep, hamel2010learning}; or reading, carried out by sentiment analysis and machine translation methods \cite{zhang2018deep, sutskever2014sequence}. However, we are still far from a completely understanding of the representations derived from deep architectures. In contrast, \acs{PGM} have stood out by their ability of thinking and understanding, dealing better with uncertainty than \acs{DL}, at the expense of performing worse in perception tasks. The integration of both paradigms, which has been denoted as \acf{BDL}, seems to be the way forward for machine intelligence.

From traditional feature engineering techniques modeling the world around us, to widely adopted feature learning methods at present, we come to the following conclusion: We have been a lot of time trying to teach machines how to define surroundings in our language, by means of handcrafted features based on our experiences. However, machines are not like humans. They have their own language, probably the reason for the success of deep representation learning. We have reached a point where it is quite complex to argue about machine representations or semantics. Their comprehensive capacity sometimes seems to be beyond our scope. Now it is time to understand how machines learn from experiences of this world, shaped like multimedia content, such as audio, images or video sequences, which closely approximate our reality. Will machines perform in tasks like visual attention in a similar way than humans? Will be necessary to let machines choose first their own paths to solve these tasks, and then develop translation mechanisms to interpret them? Without doubt, we are at the beginning of a new promising and exciting era for \acs{AI}.

%\afterpage{
%\begin{landscape}
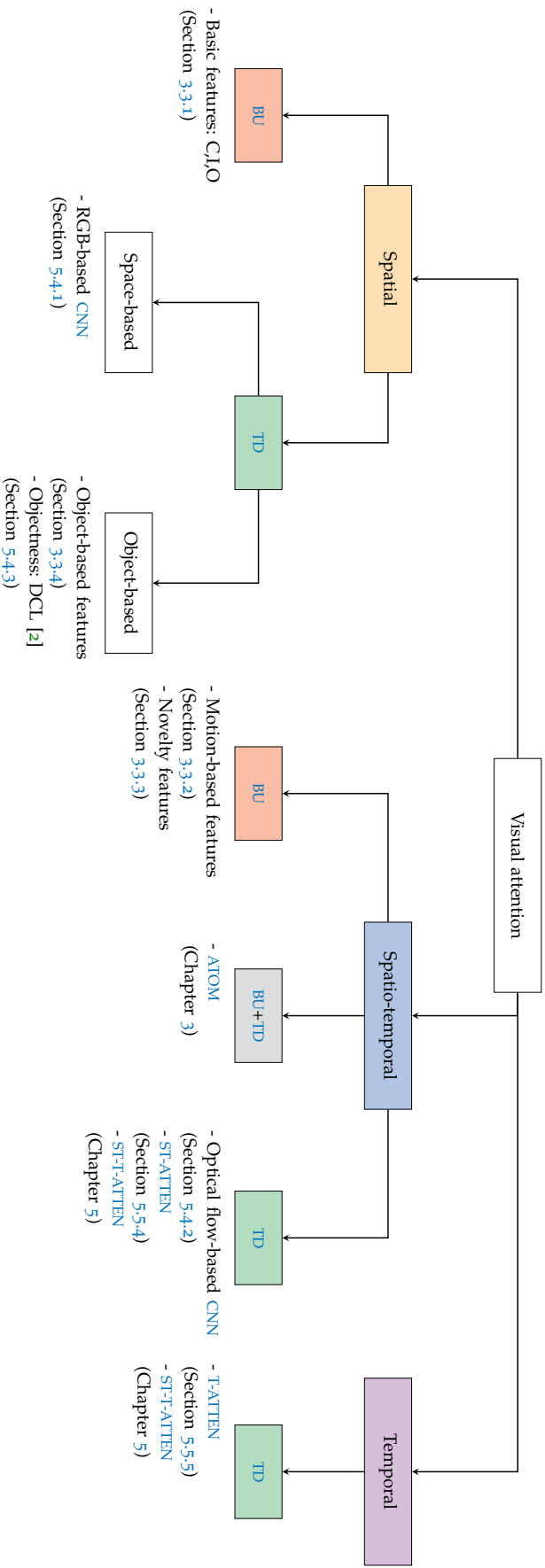
\begin{sidewaysfigure}
\centering
\scalebox{0.7}{\begin{tikzpicture}[node distance=2.75cm]
\node (spatio-temporal) [level10] {Spatio-temporal};
\node (visual_attention) [level0, above of=spatio-temporal, xshift=-3cm] {Visual attention};
\node (spatial) [level11, left of=spatio-temporal, xshift=-13cm] {Spatial};
\node (temporal) [level12, right of=spatio-temporal, xshift=7cm] {Temporal};
\node (sbu) [level20, below of=spatial, xshift=-3.5cm] {\acs{BU}};
\node (std) [level22, below of=spatial, xshift=3.5cm] {\acs{TD}};
\node (stbutd) [level21, below of=spatio-temporal] {\acs{BU}+\acs{TD}};
\node (stbu) [level20, left of=stbutd, xshift=-2cm] {\acs{BU}};
\node (sttd) [level22, right of=stbutd, xshift=2cm] {\acs{TD}};
\node (ttd) [level22, below of=temporal] {\acs{TD}};
\draw [arrow] (visual_attention) -| (spatial);
\draw [arrow] (visual_attention) -| (spatio-temporal);
\draw [arrow] (visual_attention) -| (temporal);
\draw [arrow] (spatial) -| (sbu);
\draw [arrow] (spatial) -| (std);
\draw [arrow] (spatio-temporal) -| (stbu);
\draw [arrow] (spatio-temporal) -- (stbutd);
\draw [arrow] (spatio-temporal) -| (sttd);
\draw [arrow] (temporal) -- (ttd);
\node (s-space-based) [level3, below of=std, xshift=-3cm] {Space-based};
\node (s-object-based) [level3, below of=std, xshift=3cm] {Object-based};
\draw [arrow] (std) -| (s-space-based);
\draw [arrow] (std) -| (s-object-based);
\node (harel) [level4, below of=sbu, yshift=1.5cm] {- Basic features: C,I,O \\ (Section \ref{sec:basic_features})};
\node (rgb_cnn) [level4, below of=s-space-based, yshift=1.5cm] {- RGB-based \acs{CNN} \\ (Section \ref{sec:cnn_rgb})};
\node (objectness) [level4, below of=s-object-based, yshift=1cm] {- Object-based features \\ (Section \ref{sec:object_features}) \\ - Objectness: DCL \cite{DeepSaliencyObject} \\ (Section \ref{sec:cnns_based_features6})};
\node (bu_motion) [level4, below of=stbu, yshift=1cm] {- Motion-based features \\ (Section \ref{sec:motion_features}) \\ - Novelty features \\ (Section \ref{sec:novelty_features})};
\node (atom) [level41, below of=stbutd, yshift=1.5cm] {- \acs{ATOM} \\ (Chapter \ref{ch:atom})};
\node (td_motion) [level4, below of=sttd, yshift=0.5cm] {- Optical flow-based \acs{CNN} \\ (Section \ref{sec:cnn_motion}) \\ - \acs{ST-ATTEN} \\ (Section \ref{sec:stva_autoencoder}) \\ - \acs{ST-T-ATTEN} \\ (Chapter \ref{ch:anomaly_detection})};
\node (td_lstm) [level4, below of=ttd, yshift=1cm] {- \acs{T-ATTEN} \\ (Section \ref{sec:tva_modeling}) \\ - \acs{ST-T-ATTEN} \\ (Chapter \ref{ch:anomaly_detection})};
\end{tikzpicture}}
\caption[Visual attention features and representation models covered by the different systems presented in the thesis, classified according to the spatial, spatio-temporal or temporal dimension in video sequences where they are modeled.]{Visual attention features and representation models covered by the different systems presented in the thesis, classified according to the spatial, spatio-temporal or temporal dimension in video sequences where they are modeled. Sections and chapters where they are described are indicated next to each item.}\label{fig:13_visual_attention}
\end{sidewaysfigure}
%\end{landscape}
%}

% \clearpage
\section{Goals and context of the thesis}\label{sec:objectives}
In this section, we discuss the main focus of this thesis, which concerns the study and development of hierarchical representations for spatio-temporal visual attention modeling and understanding. 

Specifically, the thesis makes the following two main contributions towards our goals:

\begin{enumerate}

	\item We introduce a hierarchical generative probabilistic model for context-aware visual attention modeling and understanding. 
	
	Our first approach, which we have called \acf{ATOM}, models visual attention in the spatio-temporal domain by considering the existing concurrence between \acs{BU} and \acs{TD} factors.
	
	\item We develop a deep network architecture for visual attention modeling, which is oriented to be applied in a video surveillance scenario.
	
	We have called our second proposal as \acf{ST-T-ATTEN}. It first estimates \acs{TD} spatio-temporal visual attention, which ultimately serves for modeling visual attention in the temporal domain.
	
\end{enumerate} 

Our particular contributions associated with both systems are mentioned in the next section, which also summarizes the main content of each chapter of the thesis. 

In order to contextualize our contributions with respect to the existing types of visual attention models (Section \ref{sec:visual_attention_intro}) and the different methodologies for visual information representation (Section \ref{sec:hierarchical_representations_intro}), we include two diagrams. In both diagrams, next to each item, sections or chapters where features and representation models are explained are indicated.

On the one hand, Figure \ref{fig:13_visual_attention} outlines the  features and representations for visual attention guidance covered throughout the thesis. They are classified according to the dimension (spatial, spatio-temporal or temporal) that they model in video sequences, and constitute a wide and complete framework to give context to our proposals. Reading from left to right, the diagram goes from spatial through spatio-temporal to temporal representations, which are classified according to the two main families of visual attention models introduced in section \ref{sec:visual_attention_intro}: \acs{BU} and \acs{TD} implementations. In addition, for the case of \acs{TD} spatial methods, the diagram differentiates between space-based features, which rely on the information drawn from eye fixations, and object-based features, related to salient objects in the scene. 

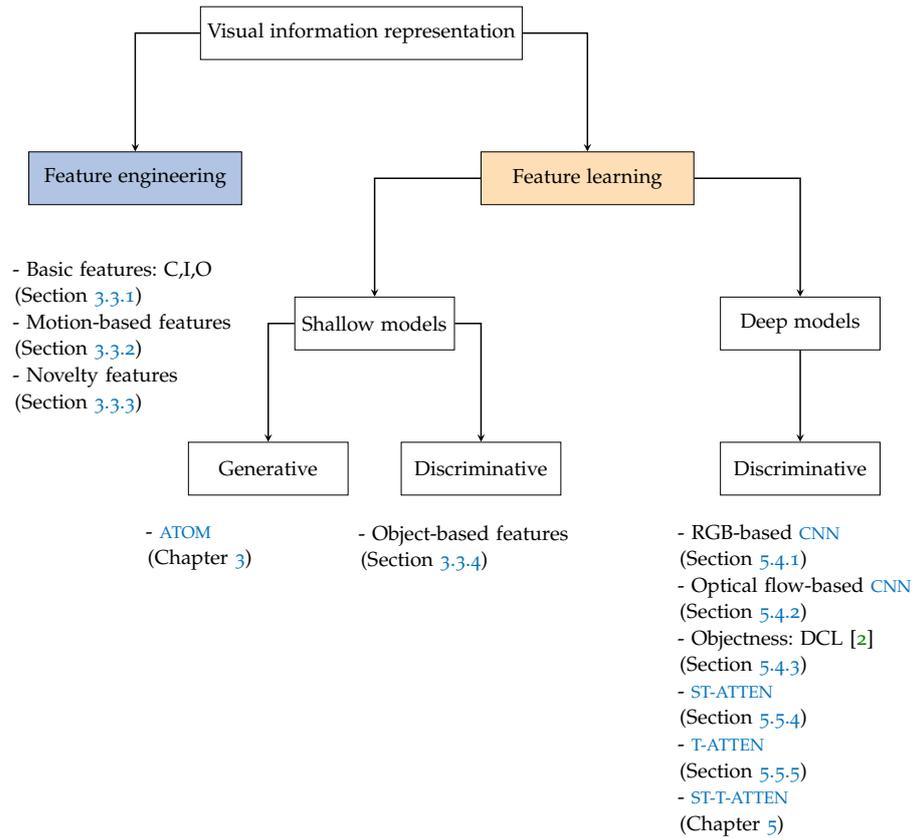
\begin{figure*}[!t]
\centering
\scalebox{0.7}{\begin{tikzpicture}[node distance=2.75cm]
\node (data_representations) [level0] {Visual information representation};
\node (feature) [level10, below of=data_representations, xshift=-4.25cm] {Feature engineering};
\node (representation) [level11, below of=data_representations, xshift=4.25cm] {Feature learning};
\node (shallow) [level3, below of=representation, xshift=-4cm] {Shallow models};
\node (deep) [level3, below of=representation, xshift=4cm] {Deep models};
\node (shallow_generative) [level3, below of=shallow, xshift=-2cm] {Generative};
\node (shallow_discriminative) [level3, below of=shallow, xshift=2cm] {Discriminative};
\node (deep_discriminative) [level3, below of=deep] {Discriminative};
\draw [arrow] (data_representations) -| (feature);
\draw [arrow] (data_representations) -| (representation);
\draw [arrow] (representation) -| (shallow);
\draw [arrow] (representation) -| (deep);
\draw [arrow] (shallow) -| (shallow_generative);
\draw [arrow] (shallow) -| (shallow_discriminative);
\draw [arrow] (deep) -- (deep_discriminative);
\node (feature_engineering) [level4, below of=feature, yshift=-0.25cm]{- Basic features: C,I,O \\ (Section \ref{sec:basic_features}) \\ - Motion-based features \\ (Section \ref{sec:motion_features}) \\ - Novelty features \\ (Section \ref{sec:novelty_features})};
\node (shallow_generative_models) [level4, below of=shallow_generative, yshift=1.25cm]{- \acs{ATOM} \\ (Chapter \ref{ch:atom})};
\node (shallow_discriminative_models) [level4, below of=shallow_discriminative, yshift=1.25cm]{- Object-based features \\ (Section \ref{sec:object_features})};
\node (deep_discriminative_models) [level4, below of=deep_discriminative, yshift=-1.25cm]{- RGB-based \acs{CNN} \\ (Section \ref{sec:cnn_rgb}) \\ - Optical flow-based \acs{CNN} \\ (Section \ref{sec:cnn_motion}) \\ - Objectness: DCL \cite{DeepSaliencyObject} \\ (Section \ref{sec:cnns_based_features6}) \\ - \acs{ST-ATTEN} \\ (Section \ref{sec:stva_autoencoder}) \\ - \acs{T-ATTEN} \\ (Section \ref{sec:tva_modeling}) \\ - \acs{ST-T-ATTEN} \\ (Chapter \ref{ch:anomaly_detection})};
\end{tikzpicture}}
\caption[Visual attention features and representation models covered by the different systems presented in the thesis, classified according to the types of processes that they involve.]{Visual attention features and representation models covered by the different systems presented in the thesis, classified according to the types of processes that they involve. Sections and chapters where they are explained are indicated next to each item.}\label{fig:12_hierarchical_representations}
\end{figure*}

On the other hand, features and representation models for visual attention guidance are depicted in Figure \ref{fig:12_hierarchical_representations}, according to the feature engineering or feature learning processes that they involve. Within the context of feature learning, we distinguish between shallow models, which are those with one or few levels of representation, and deep models with multiple layers of representation, representing the new Computer Vision paradigm. Furthermore, the diagram also reflects the difference between generative and discriminative methods. 

% understand how features for visual attention guidance are combined to perform this function, and it is in this direction in which we want to make our contributions in the first part of the thesis. 

\section{Structure of the thesis and contributions}\label{sec:structure}
In this section, we present the structure of the dissertation, introducing our main scientific contributions in the corresponding chapters. 

Chapter \ref{ch:visual_attention} makes a review of the most relevant and recent related work in perception and visual attention from a multidisciplinary perspective. 

\begin{itemize}
\item First, we review visual attention from a neurophysiological perspective, mainly describing the mechanisms of the eye and the brain for visual selection and representation. 

\item Then, we introduce the most outstanding psychological theories of visual attention, as well as some noticeable studies on the role of eye movements.

\item Finally, we summarize some of the existing computational models of visual attention, attending to either Bayesian models or deep learning-based approaches close to the contributions of the thesis.
\end{itemize}

Then, we have developed two computational systems for visual attention, which constitute the main contributions of this thesis.

Chapter \ref{ch:atom} introduces our first proposal: a generative probabilistic framework for spatio-temporal visual attention modeling and understanding. 

We first briefly discuss the related work in computational visual attention modeling. The model proposed, which we have called \acf{ATOM}, is generic, independent of the application scenario and founded on the most outstanding psychological studies about attention. Drawing in the well-known \acf{LDA} \cite{LDA} method for the analysis of large corpus of data and some of its supervised extensions \cite{sLDA,DBAyang}, our approach defines task- or context-driven visual attention in video as a mixture of latent sub-tasks, which are in turn represented as combinations of low-, mid- and high-level spatio-temporal features. 

In particular, we make the following contributions in this chapter:

\begin{itemize}
	\item We introduce feature engineering for visual attention guidance, providing a wide set of handcrafted features, which are later used in our experiments. Starting from basic and novelty spatio-temporal low-level features, such as color, intesity, orientation or motion, we move on to describe and model some mid- and high-level features related to camera motion estimation and object detection.
	
	\item Then, our algorithm incorporates an intermediate level formed by latent sub-tasks, which bridges the gap between features and visual attention, and enables to obtain more comprehensible interpretations of attention guidance.

	\item Moreover, we generate a categorical binary response for each spatial location to model visual attention. This allows to automatically align the sub-tasks discovered to a binary response by means of a logistic regression, which fully corresponds to the definition of human fixations.
\end{itemize}

Chapter \ref{ch:atom_experiments} provides an in-depth analysis of \acs{ATOM}. For that purpose, our model is used for context-driven visual attention modeling and understanding in two large-scale video databases annotated with eye fixations: CRCNS-ORIG \cite{Itti_Carmi09crcns} and DIEM \cite{mital2011clustering}. We illustrate how our approach successfully learns hierarchical guiding representations adapted to several contexts. Moreover, we analyze the models obtained, as well as perform a comparison with quite a few \emph{state-of-the-art} methods.

Chapter \ref{ch:anomaly_detection} describes our second proposal: a deep network architecture that goes from spatio-temporal visual attention prediction to attention estimation in the temporal domain. The system proposed, which we have named \acf{ST-T-ATTEN}, models visual attention over time as a fixation-based response. 

First, we review the most relevant and recent works in visual attention estimation applying deep learning-based architectures. Then, we introduce the fundamental hypothesis of the second part of the thesis: attention in the temporal domain can be predicted using the dispersion of gaze locations recorded from several subjects.

Indeed, visual attention in the temporal domain can be understood as a filtering mechanism, which allows to select time segments of special importance in video sequences. Hence, it could be used to prevent human errors and speed up decision making processes in real applications which require watching large amounts of visual information, such as the task of video surveillance.

We make the following particular contributions in this chapter:

\begin{itemize}
	\item We describe three feature learning architectures for visual attention guidance, which provide input feature maps to our system: RGB-based spatial, optical flow-based and objectness-based networks. 
	\item We propose a frame-level fixation-based temporal ground-truth, which is computed attending to the dispersion at fixation spatial locations from several subjects. Furthermore, we validate the fundamental hypothesis introduced above. We will use this variable to train our models to estimate attention in the temporal domain.
	\item Our proposed \acs{ST-T-ATTEN} is built on the combination of two modules: 1) A \acf{ST-ATTEN} for spatio-temporal visual attention estimation, which consists on a \acf{CED} \cite{huang2007unsupervised} network; 2) A \acf{T-ATTEN} for modeling visual attention in the temporal domain, based on \acf{LSTM} \cite{hochreiter1997long} units, widely used for time series forecasting.
\end{itemize}

Chapter \ref{ch:surveillance_experiments} describes the experiments conducted to validate the different configurations proposed for the \acs{ST-T-ATTEN} modules. We make use of the BOSS \cite{BOSS} database, which contains videos recorded in a railway transport context with different anomalous events, with the aim of determining the optimal configuration for the whole \acs{ST-T-ATTEN} proposed, as well as motivating its use as an information filtering mechanism in a video surveillance application.

Finally, in Chapter \ref{ch:conclusions}, we summarize the conclusions drawn from the main contributions of the thesis, which serve to outline future lines of research.

%*****************************************
%*****************************************
%*****************************************
%*****************************************
%*****************************************

%\ctparttext{You can put some informational part preamble text here.
%Illo principalmente su nos. Non message \emph{occidental} angloromanic
%da. Debitas effortio simplificate sia se, auxiliar summarios da que,
%se avantiate publicationes via. Pan in terra summarios, capital
%interlingua se que. Al via multo esser specimen, campo responder que
%da. Le usate medical addresses pro, europa origine sanctificate nos se.}
%\part{The Showcase}\label{pt:showcase}
%*****************************************
\chapter{A multidisciplinary perspective on visual attention}\label{ch:visual_attention}
%*****************************************

\section{Introduction}\label{sec:introduction_2}
%SIMONE FRINTOP
%Computational Visual Attention Systems and their Cognitive Foundations: A Survey
%- World
%- Eye: Information capture. Fixations mechanism.
%- Brain: Attention is the process by which the brain controls and tunes information processing.
%- Image
%https://www.ics.uci.edu/~majumder/vispercep/vispercepold.htm
%PIZLO Course
%Palmer, 1999
During the 1970s, scientists from several disciplines began to show a great interest in understanding how optical images could be processed to extract useful information about the environment \cite{Palmer}. This resulted in the emergence of vision science. 

Vision science \cite{Palmer} is defined as an interdisciplinary branch from cognitive science, which is devoted to the study of visual mental states and processes from different, compatible and complementary perspectives. All of them talk about vision in the common language of computation, by accepting that it may take place not only in living organisms, through eyes and brains, but also when information from cameras is processed in ad-hoc programmed computer systems.  

Over the past few decades, psychologists have tried to explain visual perception through a vast amount of theories and models. Moreover, neurophysiologists have made experiments to monitor neuron activity. Furthermore, computational neuroscientists have built neural network architectures to simulate how these neurons represent and react to visual stimuli. Drawing on these findings, computer vision scientists have sought to develop computational models and algorithms which automatically address the cognitive functions involved in attention. 
  
Indeed, a great world full of visible information is opened to us, and the \acf{HVS} has the paramount responsability of dealing with attentive processes. Due to the limited capacity of the brain to process such a big amount of sensory input, attention involves the inherent search operations that reformulate and optimize generic perception and cognition problems so that they become tractable \cite{TsotsosBook}. Eye movements allow acquiring and tracking visual stimuli, unconsciously highlighting the most conspicuous \cite{BruceT05} \cite{Itti_Baldi06nips} areas in a particular context, or willingly selecting those that aid to solve a particular task \cite{sprague2003eye}. 

This thesis presents a framework for visual attention estimation and understanding from a computational view, not only applying existing computer vision techniques, but also contemplating psychological arguments. This chapter makes a review of the most relevant and recent related work in perception and visual attention, bearing in mind all the perspectives differentiated above. The purpose of this \emph{state-of-the-art} is thus to provide a broad overview of the visual attention research by identifying the basis of our work.
 
\section*{Chapter overview}
Starting from the structure and the processes involved in the eyes and the brain, visual attention is reviewed from a neurophysiological perspective in Section \ref{sec:neuro_va}. We discuss the difference between \emph{overt attention}, which implies eye movements and fixations, and \emph{covert attention}, which is more related to the mechanisms of the brain for visual selection and representation. Then, Section \ref{sec:psy_va} introduces the most outstanding psychological theories of visual attention, which allude to \emph{early representation} features that guide the attention of observers. Moreover, we also cover some noticeable studies on the the role of eye movements. Finally, we summarize some of the existing computational models of visual attention in Section \ref{sec:computational_va}, mainly attending to those approaches close to the contributions of the thesis.

\section{Neurophysiological basis of visual attention}\label{sec:neuro_va}
This section describes the structure of the \acs{HVS}, attending to the regions of the eye and the nervous system that take part in the process of visual perception. According to the Professor Stephen E. Palmer \cite{Palmer}, \emph{``visual perception is an information extraction process that involves the acquisition of knowledge about objects and events in the environment''}. This information comes from the light that is emitted or reflected by objects. 

It should be noted that \acs{HVS} has an extraordinary ability to select only the necessary information in order to interact with a given scenario, being able to infer the rest with sufficient accuracy. Hence, given an image, vision implies a heuristic process to infer the most likely environmental condition that could have produced it. 
  
\subsection{Human Visual System: eye and brain}\label{sec:neuro_va1}
Both eyes and brain are essential for visual perception. The complete eye-brain system must perform adequately to obtain trustworthy visual data. 

\begin{figure*}[!t]
\centering
    \includegraphics[width=1\textwidth]{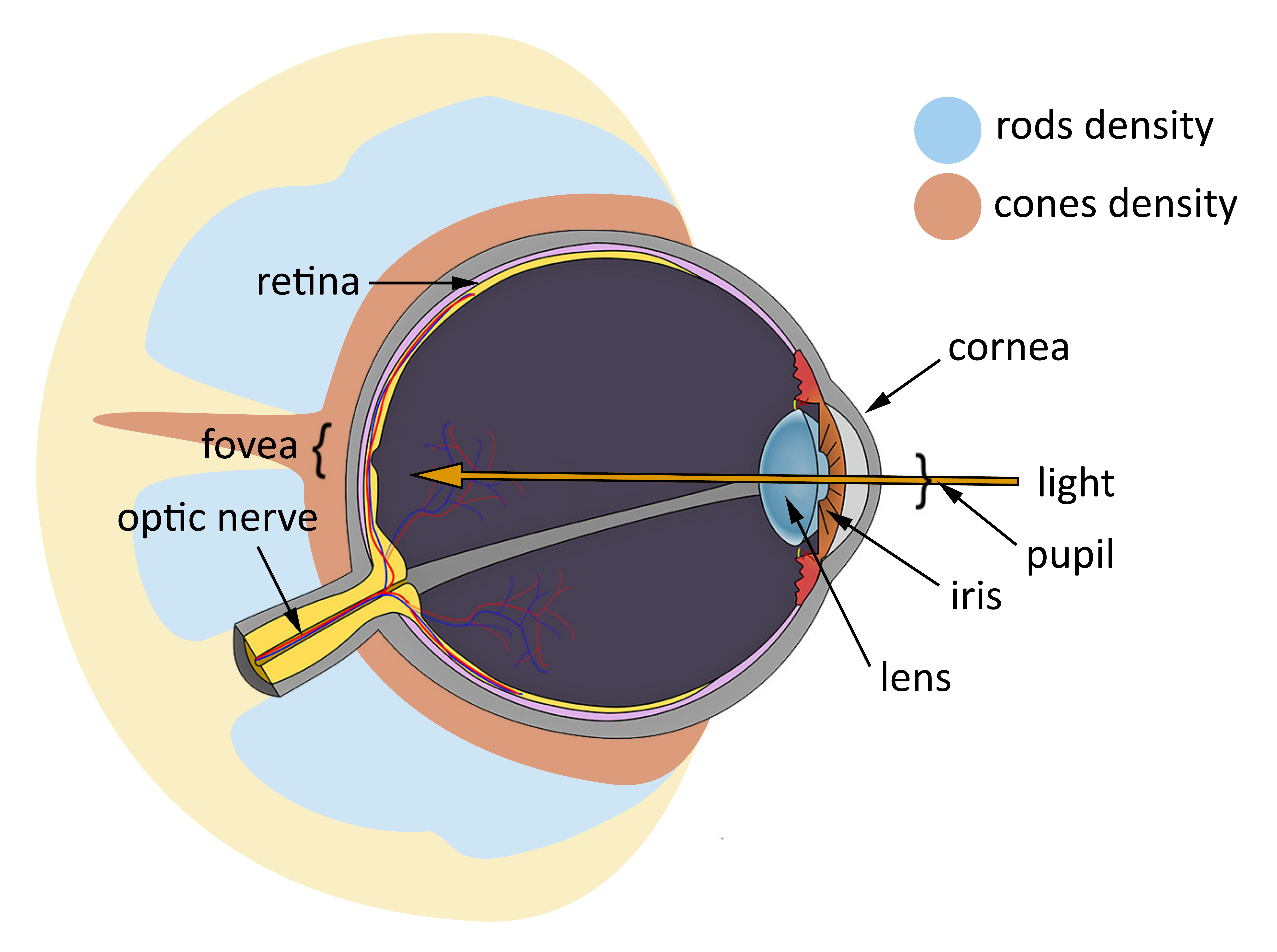} %\quad
    \caption[Diagram of the human eye.]{Diagram of the human eye. Rods and cones densities are drawn around the retina in blue and red, respectively. Adapted from Wikimedia Commons \cite{human_eye_diagram}.}
    \label{fig:hvs_eye}
\end{figure*}

\subsubsection*{The eye} 
The structure of the human eye is shown in Figure \ref{fig:hvs_eye}. Humans have two approximately spherical eyes. They are situated at two holes in the skull called \emph{eye sockets}, which are placed at about the horizontal midline of the head. Eyes are moved by six small and strong \emph{extraocular muscles}, which are responsible for eye movements, allowing to scan different regions of the visual field. They are monitored by several nuclei in the brain stem, via the oculomotor neurons.

Several parts of the eye carry out optical functions. First, eyes collect the light that enters through the \emph{cornea}. The light crosses an opening in the \emph{iris} called \emph{pupil}, behind which the \emph{lens} is located. Finally, incoming light projects an image onto the \emph{retina}, a curved surface at the back of the eye. The retina is composed of more than 100 million light-sensitive cells, known as \emph{photoreceptors}, which transform light into neural activity. There are two types of photoreceptors: rods and cones. As can be seen in Figure \ref{fig:hvs_eye}, rods, which are longer and more numerous (about 120 million), are located everywhere in the retina except at its center. They are highly sensitive to light, so they allow us to see at low light levels or \emph{scotopic conditions}. In contrast, most of the cones, which are shorter and fewer (8 million), are clustered in the \emph{fovea}, situated at the center of the retina. They are much less sensitive to light, used under normal lighting or \emph{photopic conditions}, and also in all experiences of color.

Light comprises photons, many small units of energy that produce electrical changes in photoreceptors, and the information travels via the \emph{optic nerves} to the visual centers in the brain. 

\subsubsection*{The brain}\label{sec:brain}
\begin{figure*}[!t]
\centering
	\includegraphics[trim=0cm 0cm 0cm 0cm, width=0.9\textwidth]{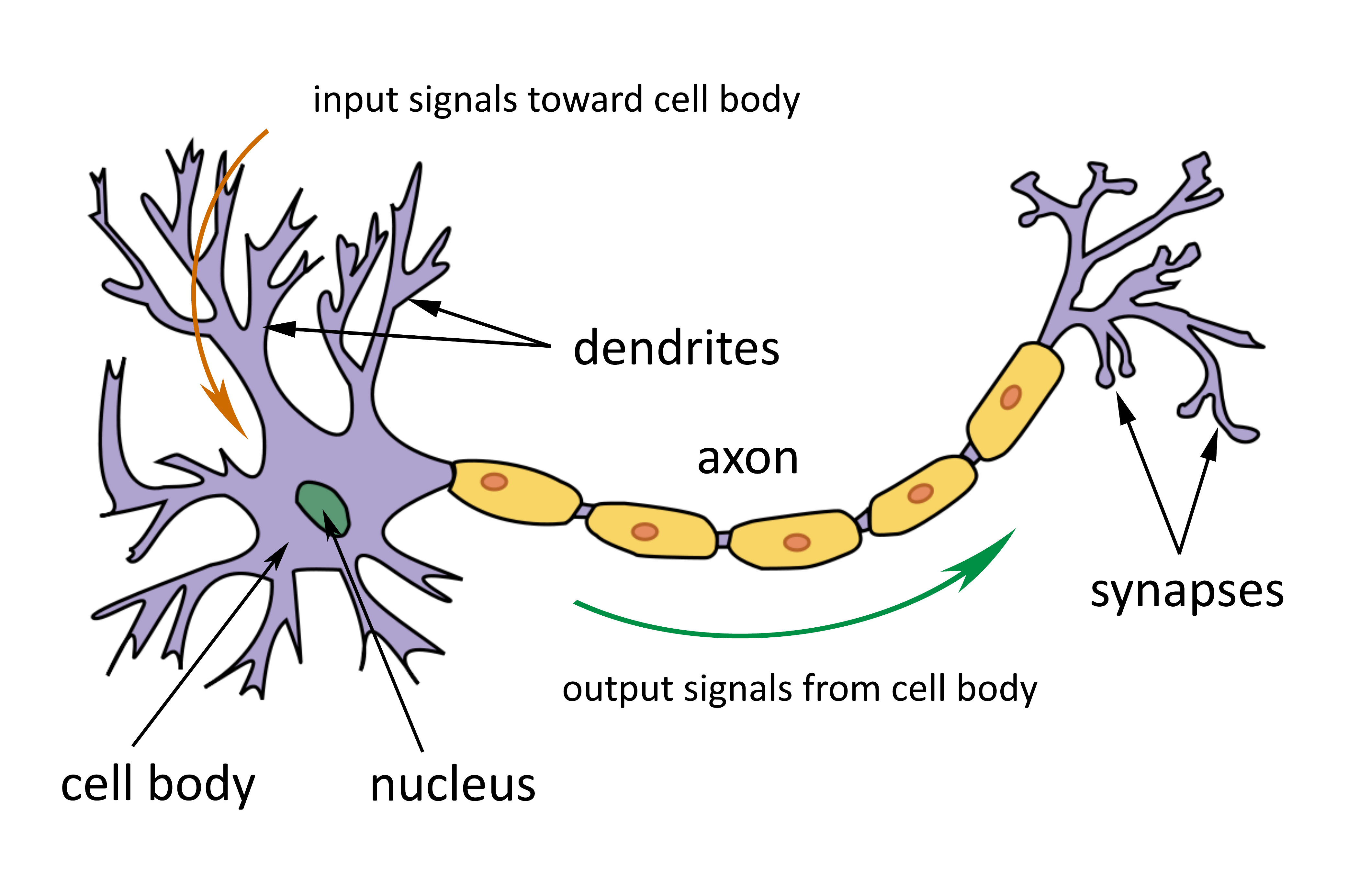}	
\caption[Schematic diagram of a biological neuron.]{Schematic diagram of a biological neuron. Adapted from \cite{Palmer,cs231n}.}
\label{fig:221_neuron}
% \vspace{-0.4cm}
\end{figure*}
From the fovea in each eye, the optic nerves cross over to the opposite side of the brain, leading the information from the left half of the visual field to the right side of the brain and vice versa. The brain processes this information, in order to make it useful for observers.

\emph{Neurons} constitute the basic computational cells of the brain. The human brain is composed of around 100 billion neurons. As shown in Figure \ref{fig:221_neuron}, a neuron first receives electrical signals coming from other neurons across the \emph{dendrites}. Within the \emph{cell body}, where the \emph{nucleus} is located, these inputs are converted into a series of output spikes that are propagated through its \emph{axon} to other neurons. The \emph{firing rate} of the neuron determines the frequency of the spikes. Finally, \emph{synapses} connect the axon to the dendrites of the following neurons.   

There are two pathways on each half of the brain. One of them arrives to the \emph{\acf{SC}}, which seems to be involved in the control of eye movements by processing information related to the location of objects in the world; the second and larger pathway goes to the \emph{occipital} or \emph{\acf{V1}}.

\begin{figure*}[!t]
\centering
	\includegraphics[trim=0cm 0cm 0cm 0cm, width=1\textwidth]{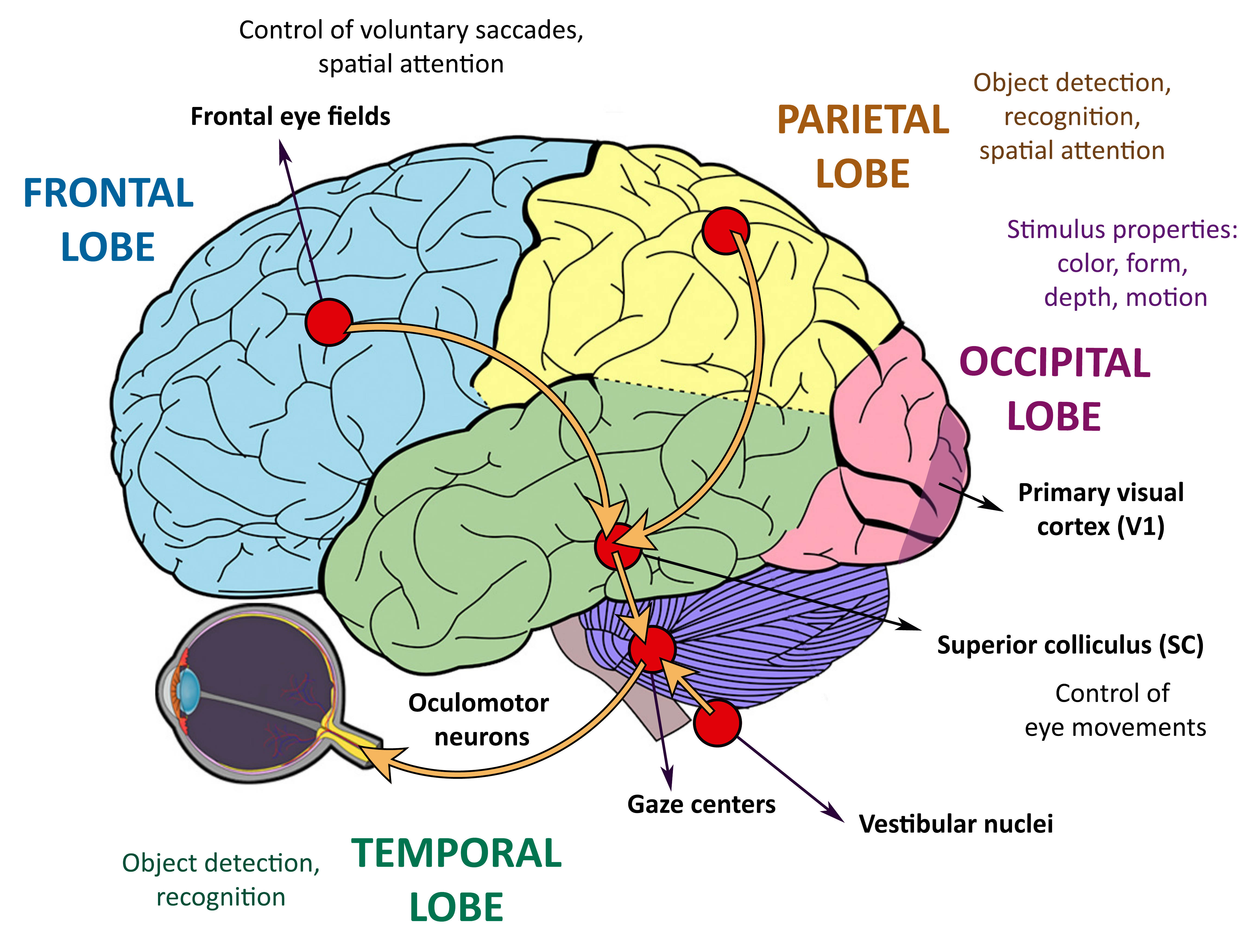}	
\caption[Diagram of the human brain.]{Diagram of the human brain. Arrows indicate the connection between the eye and the principal areas in the brain involved in the visual attention process: frontal eye fields and posterior parietal cortex, which guide spatial attention, and the \acf{SC}, which controls both eye movements and covert shifts of attention. Eye and brain diagrams taken from Wikimedia Commons \cite{human_eye_diagram, brain_diagram}.}
\label{fig:221_brain}
% \vspace{-0.4cm}
\end{figure*}

Nowadays, we have an evidence about the function of the \emph{occipital}, \emph{parietal} and \emph{temporal lobes} of the visual cortex, which are identified in Figure \ref{fig:221_brain}. The cortical processing begins at \acs{V1} cells, where spatial \emph{receptive fields} respond to visual stimuli, so that a mapping of the information from the retina is produced. According to the scale-space theory of computer vision \cite{lindeberg2013computational}, receptive fields encode simple visual patterns of light, such as oriented edges or color blotches, which constitute the first stages of visual processing, by reflecting the symmetry properties of the world that surrounds us. Moreover, a reduced set of operations over these simple patterns allow to obtain a wide variety of complex underlying representations for visual perception. \acfp{CNN} \cite{lecun1989backpropagation} emerged in an attempt to reproduce the function of receptive fields, recently demonstrating an astonishing performance in a lot of applications. They will be described in section \ref{sec:convolutional_neural_networks}, as the basis of our contributions for visual attention estimation in the temporal domain in Chapter \ref{ch:anomaly_detection}.

The information from \acs{V1} is then projected to other parts of the occipital lobe, and also to areas of the parietal and temporal ones. Some studies \cite{zeki1978functional, livingstone1988segregation} suggest that these different regions involve small maps where several properties derived from the retinal stimulation are coded in parallel, such as color, form, depth and motion. Regarding the parietal and temporal cortex, they seem to be responsible for the identification and location of objects, respectively. 

\subsection{Visual attention}\label{sec:neuro_va2}
%YARBUS
%Eye movements (See journal article), visual attention
Visual perception is inherently selective \cite{Palmer}. We are able either to globally process the information in a scene, or to focus our vision more on some particular objects.
We sometimes even attend locally to their specific parts or properties, depending on their importance in the activity we are performing. In addition, we choose quite automatically where to fixate our vision next. For instance, although there may be a lot of appliances in a kitchen, we immediately head towards the fridge if we want to drink something, without looking at the toaster or the dishwasher. %In the same way, if we are going to cook a pizza, we barely have to think about going from the fridge, where we will first take it, to the oven. 

All these strategies for selecting and processing information in the visual field are related to attention. Two different acts of visual attention can be distinguished. First, attention is called \emph{overt} if it is external and observable by others, implying eye movements to fixate from one object to another. Different fixations of a context contain useful information which is shaped like visual images on the retina. Then, part of the fixated information is selected by \emph{covert} attention to be fully processed. Covert selections are, conversely, internal and unobservable by others. They do not imply eye movements, but allow to shift our gaze to peripheral Regions Of Interest (\acsp{ROI}), chosen from the information processed.

\subsubsection*{Overt attention: eye movements and fixations} 
Visual attention can be thus described, on the one hand, as a temporal process that involves a sequence of eye fixations preceded by different types of eye movements. This results in a series of instantaneous spatial locations of the visual axis called \emph{gaze points}.

There are four basic types of \emph{eye movements} \cite{purves2001neuroscience}. If we look at still images or static objects, we mainly perform saccadic movements to scan over them. \emph{Saccades} are very quick (20-40 ms) voluntarily or involuntarily ballistic jumps between two points of fixation. During a saccade, both eyes drift in the same direction. Moreover, the trajectory of a saccade cannot be changed when the eyes are in motion.

In real-life situations, where either the viewer or the objects are moving, three more types of eye movements can be found:

\begin{itemize}

\item \emph{Smooth pursuit movements}, which are slow in comparison with saccades, are used to track the position of moving objects. The ability of the \acs{HVS} to take clear images from tracked objects depends on how fast they move, being less accurate at speeds higher than 30 degrees per second, when subjects start to use saccades to follow objects. 

\item \emph{Vergence movements} allow \acs{HVS} to fixate objects located at different depths. In this type of movement, eyes move in opposite directions and have an angle of convergence that depends on the distance of the target from the observer. Eyes seeing nearby objects strongly converge.

\item \emph{Vestibular movements} are controlled by the vestibular system in the inner ear, and contribute to keep the target fixed on the fovea when the head changes its position and orientation (ego-motion). In this situation, eyes compensate the ego-motion by moving in the opposite direction of the head, normally at its same speed.

\end{itemize}

When the eyes stop examining the scene, \emph{fixations} take place and the \acs{HVS} takes comprehensive information about what is being looked at. Although it is not often mentioned, fixations are not directly measurable, but composed of minute microsaccades, tremor and drift movements that focus the eyes on the target, generating multiple gaze point samples. They have a particular duration, usually between 50-600 ms, and can reveal meaningful information about attention and understanding. Given a specific context, the time to first fixation in a conspicuous location or target is short, while a long fixation duration may suggest a greater effort to make sense of a stimulus, or an appealing one.

All these movements have disparate neural mechanisms, which are spread in different areas of the brain, as shown in Figure \ref{fig:221_brain}. First, frontal eye fields in the frontal cortex control the voluntary saccades. On the other hand, both smooth pursuit and vergence movements require visual feedback, so they are monitored by means of information from the motion channels in visual cortex and binocular disparity channels in occipital cortex, respectively. Finally, vestibular movements result from disturbances in the fluid of the semicircular canals of the inner ear. These are monitored by the vestibular system, which connects to the oculomotor neurons to provide them with the correct eye velocity signal. The \acs{SC} is also involved in the control of eye movements, as was mentioned above. 

\subsubsection*{Covert attention: Relation with overt attention}
On the other hand, \emph{visual attention} concerns a set of complex covert processes that aid an observer to select and gather the most outstanding information within the visual field, with the aim of successfully solving a cognitive problem in a particular environment.

Covert attention is usually directed at the \acs{ROI} fixated by the eyes. Professor Stephen E. Palmer \cite{Palmer} makes an interesting metaphor in order to explain the relationship between eye movements and covert attention: \emph{"Attention is like an internal eye that can be moved around to sample the visual field much as the eye can be moved around to sample the visual world."} What is more, there is evidence that eye movements usually follow attentional movements. Thus, covert attention, viewed as the primary function of visual selection, controls overt saccades, which play an important but supporting role, driving them to the appropriate locations and enhancing the perception of events happening there. 

Therefore, it is known that there exist a strong correlation between the areas in the brain that controls eye movements and covert attention: \acs{SC} controls both eye movements and covert shifts of attention, while frontal eye fields and the posterior parietal cortex are responsible for guiding spatial attention. Arrows in Figure \ref{fig:221_brain} indicate the connection between the eye and the principal areas in the brain involved in the visual attention process.

Finally, many studies agree in saying that this frontoparietal network may involve an attentional \emph{priority map} \cite{bisley2011neural}, which represents items in the visual world according to their importance in a particular situation. We will discuss more about the utility of this representation of visual attention in the next section, now from a psychophysical perspective.

\section{Psychophysical theories of visual attention}\label{sec:psy_va}
%TREISMAN AND GELADE, WOLFE...
%Psychological Basis of Visual Attention
%(See journal article)
In this section, we gather the statements of the most outstanding psychological theories of visual attention, which establish relevant features for the perception of objects and support the majority of existing computational attention systems \cite{frintrop2010computational}. 

Two fundamental theories have been the most influential: the \acf{FIT} \cite{TreismanGelade80} and the \acf{GSM} \cite{wolfe1994guided}. Based on the foundations of these studies, we can differentiate between two main families of visual attention models: \acf{BU} or \emph{stimulus-driven} and \acf{TD} or \emph{task-oriented}. In addition, it is also worth mentioning the importance of eye movements in scene perception, as explained in the famous classic study of Yarbus \cite{Yarbus1967}. 

In addition, we refer for the first time to several aspects that have been considered in the design of the probabilistic model for visual attention understanding presented in Chapter \ref{ch:atom}.

\begin{figure*}[!t]
\centering
	\includegraphics[trim=0cm 0cm 0cm 0cm, width=1\textwidth]{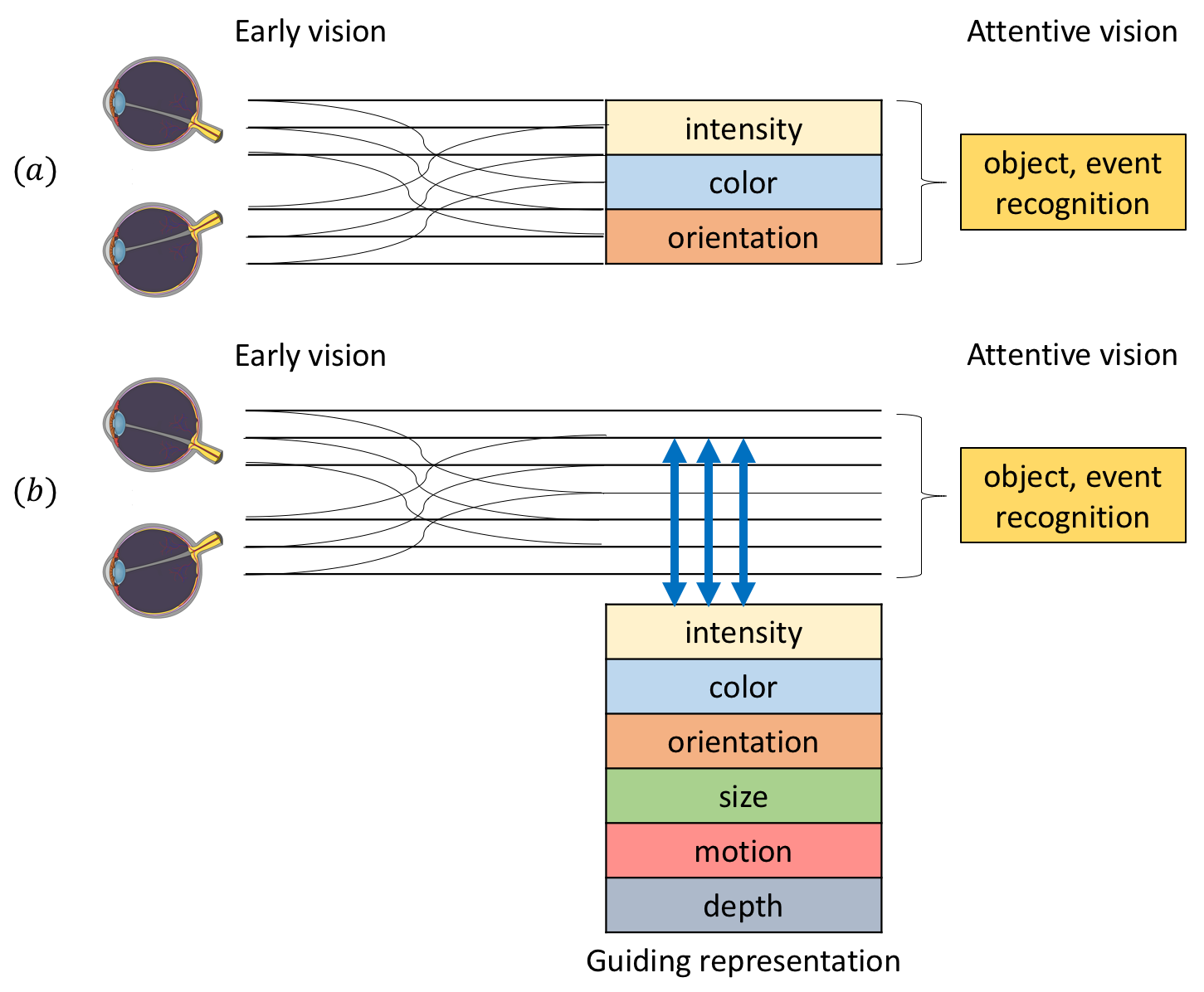}	
\caption[Diagram representations of (a) the \acf{FIT} and (b) the \acf{GSM} \cite{wolfe1994guided}.]{Diagram representations of (a) the \acf{FIT} \cite{TreismanGelade80}, where visual attention is modeled as a combination of intensity, color and orientation pre-attentive features, and (b) the \acf{GSM} \cite{wolfe1994guided}, which introduces ``guiding representation'' as a control mechanism before object recognition, and also mentions new features such as size, motion or depth. Adapted from \cite{WolfeHorowitz04}. Eye diagrams taken from Wikimedia Commons \cite{human_eye_diagram}.}
\label{fig:23_psycho_theories}
% \vspace{-0.4cm}
\end{figure*}

\subsection{The Feature Integration Theory}
Treisman and Gelade introduced the \acf{FIT} \cite{TreismanGelade80} in 1980, which states that several features or attributes are identified early, automatically and in parallel across the visual field, while objects are registered separately as a conjunction of these features at a later serial stage. The model is depicted in Figure \ref{fig:23_psycho_theories}(a). A master map of location results from this combination of attributes, which indicates where the objects are, prior to their recognition.

In light of this theory, Treisman discussed the difference between feature and conjunction search processes, when looking for a target in a scenario full of distractors. In a \emph{feature} or parallel search process, we seek through distractors that differ from the target by a unique feature. In contrast, Treisman asserted that a \emph{conjunction} or serial search assumes a more complex and time-consuming task, since the distractors have one or more features in common with the target. The more attributes aid to discern the target from the distractors, the easier the search for a target is. Nevertheless, if we know the features of the target in advance, conjunction search performance can be improved by inhibiting the features which are exclusive from distractors.

% In the same way that other behavioral analysis [COMPLETECITE], \acs{FIT} mentions three basic features: intensity or luminance contrast, color and orientation.

\subsection{The Guided Search Model}
The standard \acs{FIT} defines parallel and serial search as autonomous processes that cannot share information between them. By contrast, Wolfe's \acf{GSM} 2.0 \cite{wolfe1994guided} from 1994 claims that the \acs{FIT} attentive serial search has to be guided by useful information in the preattentive parallel processes, which divided their corresponding set of stimulus into distractors and candidate targets. 

\acs{GSM} thus supports that attention can be guided towards specific targets by modulating gains associated with low-level features. Indeed, visual search is a continuous process and, consequently, the information from the parallel processes to the serial process can be updated over time.

Subsequents works by Wolfe \cite{WolfeHorowitz04,Wolfe2007guided} 
present the idea of ``guiding representation'' or guidance as a control device located to one side of the main pathway from early vision to object recognition, as shown in Figure \ref{fig:23_psycho_theories}(b). It controls the access to the attentional bottleneck, so the guidance is abstracted from the main pathway despite of not being part of the pathway itself. Thus, the way we see stimulus in the world is different from the representations upon which guidance is founded. Rather than altering the stimulus such as filters would do, this module guides attention as a CCTV operator working at a public building (e.g. a train station or a university) would do. Based on an abstract representation of some notions (e.g. threat, suspicious object), the operator selects some parts of the scenario that receive more attention than others. 

Two ways of guidance are possible, \acf{BU} and \acf{TD}, which correspond to the two main types of computational visual attention systems. \acs{BU} attention is fast, involuntary and mainly based on characteristics of the visual scene (\emph{stimulus-driven}) such as color, orientation, motion or depth. By contrast, \acs{TD} attention is slow, voluntary and determined by cognitive phenomena like knowledge, expectations or advanced indications (\emph{goal-driven}). 

Guidance can be ultimately represented as an activation map. An activation map drawn only by \acs{BU} signals is a \acf{SM} \cite{koch_shifts}. Instead, \acs{TD} guidance provides a priority or \acf{VAM} that sorts by relevance the items existing in the visual field for selective attention and recognition \cite{wolfe2014approaches}. 

It is also worth noting the concept of scene guidance incorporated in the version 4.0 of \acs{GSM} \cite{wolfe2011visual}, which results in a non-selective pathway in the visual search process: Observers are able to determine very rapidly the global properties or `gist' of a given scene (e.g. a highway with intense traffic) before they selectively attend to the most conspicuous objects (e.g. a damaged car). 
% Completar esta parte si finalmente se incorpora a alguna parte de nuestro modelo!!!

Hence, guidance is not based directly on the information provided by early visual processes but on a coarse and contextual representation derived from them. This interpretation of visual attention supports the main assumption of our model in Chapter \ref{ch:atom} and opens the door to the inclusion of an intermediate layer mapping the low level stimuli to an intermediate representation. 
% Dejo sin explicar re-entrant process (Wolfe, Oxford Attention, Pag. 23)
% y busqueda de objetos como proceso de difusion (Pag. 36)

\subsection{What are the attributes that guide attention?}
A significant number of authors have contributed to determine a wide set of features that might drive attention, as outlined in the excellent review by Wolfe \cite[Chapter 2]{wolfe2014approaches}. In this survey, features are grouped depending on their consensual reliability, which is determined taking into account both the number of studies that support them and the convergence of their demonstrations. Some of these features are more effective than others, not only because of what they measure but also depending on the quality of the visual support where they are computed. 

First, Treisman and Gelade's \acs{FIT} \cite{TreismanGelade80} mentioned three basic features: intensity or luminance contrast, color and orientation. These are considered undoubted attributes, so they constitute the basis of almost all the existing models that explain visual attention.

Additionally, Wolfe's \acf{GSM} 2.0 \cite{wolfe1994guided} enumerated other attributes that humans can perceive efficiently and thus could be also considered salient in a scene: motion, scale or size, shape and depth. According to McLeod et al. \cite{mcleod1988visual}, motion speed and direction could be represented separately. Moreover, it is also unclear if shape should be defined as a whole or, alternatively, as a family of simpler attributes such as curvature, line termination or closure. Then, depth aid to modulate features like size. Near objects stand out from far ones, which seem to be smaller. Wolfe later extended this list \cite{WolfeHorowitz04}, raising doubtful or complicated cases such as novelty, faces or other semantic categories (e.g. `car' , `dog').  

Based on psychologists intuition, we have selected some of these attributes as input to our approach for visual attention understanding, with the purpose of appraising their utility in diverse contexts. Chapter \ref{ch:atom} introduces this set of features, as well as the image processing techniques used for their extraction.

%\begin{table}
%    \myfloatalign
%    \begin{tabularx}{\textwidth}{lll} \toprule
%        \tableheadline{Undoubted attributes} & \tableheadline{Probable attributes}
%        & \tableheadline{Complicated attributes} \\ \midrule
%        Intensity & Shape & Novelty \\
%        Color & Depth & Faces \\
%        Orientation &  & Text \\
%        Motion & & Semantic category (e.g. 'car', 'dog') \\
%        Size & & \\
%        \bottomrule
%    \end{tabularx}
%    \caption[]{Examples of attributes that might guide visual attention. Features are grouped according to their consensual realiability, which has been determined taking into account both the number of studies that support them and the convergence of their demonstrations. Table adapted from \cite{WolfeHorowitz04} \cite{wolfe2014approaches}}  \label{tab:guiding_atributes_examples}
%\end{table}

\subsection{Eye movements}
The role of eye movements in scene perception had already been studied before the introduction of the perception theories referred above. According to the revision of previous research on high-level scene perception made by J. M. Henderson and A. Hollingworth \cite{henderson1999high}, one can figure out what are the procedures that control \emph{where} and \emph{how long} each fixation point tends to remain centered at a particular location to provide a complete understanding of scene. 

Yarbus classic study of 1967 \cite{Yarbus1967} showed that, although first few fixations in a scene seem to be controlled by global characteristics, positions of later fixations are not random but landed on regions that are useful or essential for perception. Eyes are either driven by \acs{TD} factors that direct fixations toward task-driven informative locations (e.g. cooking, driving) or led to low-level image discontinuities called salient regions (e.g. bright regions, edges). The time the eyes remain in a given region also depends on its visual and semantic properties. 

The experience of a complete and integrated visual world is thus based on an abstract representation that covers general information about the scene combined with perceptual information arisen from fixations. By examining eye movements, it could be possible to infer the underlying factors affecting fixations or task at hand, even to interpret observer thoughts \cite{defendingYarbus}. 

This is the purpose of our model in Chapter \ref{ch:atom}, which introduces an intermediate level between feature extraction and visual attention computation stages based on the information drawn from fixations. This level consists of latent sub-tasks that can be used to determine why some locations are more conspicuous than others. Thus, rather than directly learn a predictor of human attention over low-level visual features, our method provides a hierarchical interpretation of visual attention, advantageous for further comprehensive analysis.

The experiments conducted by Yarbus have motivated researchers to assess the possibilities of eye tracking for assistance in real applications such as industry control \cite{sharma2016eye}, health-care \cite{hermens2013eye} and video surveillance \cite{howard2011task}. Experts in these applications have to process a large amount of visual information at the same time, which implies a high cognitive effort that might be reduced by modeling fixations behavior along time. Indeed, it has been observed that there is a strong correlation between fixation patterns of different viewers performing the same task, specially during an important or suspicious event \cite{howard2013suspiciousness,sharma2016eye}.

The latter is the ultimate objective of our system in Chapter \ref{ch:anomaly_detection}, which is trained to model a temporal attention response arisen from fixations dispersion across viewers. 

\section{Computational modeling of visual attention}\label{sec:computational_va}
%%Evolution of computational visual attention models
%%(See article with scheme + State-of-the-art in visual attention)
\begin{figure*}
\begin{vtimeline}[description={text width=10cm}, 
 row sep=1ex, timeline color=Red] 
 % use timeline header]
 % timeline title={The title}]
1985 & Koch and Ullman \cite{koch_shifts} \endlr
1998 & Itti et al. \cite{Itti:1998:MSV:297843.297870} \endlr
\end{vtimeline}

\begin{vtimeline}[description={text width=10cm}, 
 row sep=1ex, timeline color=Green]
2003 & Sprage and Ballard \cite{sprague2003eye} \endlr
\end{vtimeline}

\begin{vtimeline}[description={text width=10cm}, 
 row sep=1ex, timeline color=YellowOrange]
\ \ \ \ \ \ \ \ & Torralba \cite{Torralba:03} \endlr
\end{vtimeline}

\begin{vtimeline}[description={text width=10cm}, 
 row sep=1ex, timeline color=Green]
2005 & Bruce and Tsotsos (\acs{AIM}) \cite{BruceT05} \endlr
\end{vtimeline}

\begin{vtimeline}[description={text width=10cm}, 
 row sep=1ex, timeline color=Gray]
\ \ \ \ \ \ \ \ & Navalpakkam and Itti \cite{1641004} \endlr
\end{vtimeline}

\begin{vtimeline}[description={text width=10cm}, 
 row sep=1ex, timeline color=YellowOrange]
\ \ \ \ \ \ \ \ & Itti and Baldi \cite{Itti_Baldi06nips} \endlr
\end{vtimeline}

\begin{vtimeline}[description={text width=10cm}, 
 row sep=1ex, timeline color=Red] 
2006 & Harel et al. (\acs{GBVS}) \cite{Harel07graph-basedvisual} \endlr
2007 & Hou and Zhang \cite{HouZhang} \endlr
\end{vtimeline}

\begin{vtimeline}[description={text width=10cm}, 
 row sep=1ex, timeline color=Gray]
\ \ \ \ \ \ \ \ & Peters and Itti \cite{peters2007beyond} \endlr
\end{vtimeline}

\begin{vtimeline}[description={text width=10cm}, 
 row sep=1ex, timeline color=Purple]
2008 & Zhang et al. (\acs{SUN}) \cite{doi:10.1167/8.7.32} \endlr
\end{vtimeline}

\begin{vtimeline}[description={text width=10cm}, 
 row sep=1ex, timeline color=Red] 
2009 & Hou and Zhang (\acs{ICL}) \cite{hou2009dynamic} \endlr
\ \ \ \ \ \ \ \ & Seo and Milanfar (\acs{SDSR}) \cite{SeoMilanfar} \endlr
\end{vtimeline}

\begin{vtimeline}[description={text width=10cm}, 
 row sep=1ex, timeline color=Green]
\ \ \ \ \ \ \ \ & Judd et al. \cite{judd2009learning} \endlr
\end{vtimeline} 

\begin{vtimeline}[description={text width=10cm}, 
 row sep=1ex, timeline color=Purple]
\ \ \ \ \ \ \ \ & Zhang et al. (Dynamic \acs{SUN}) \cite{zhang2009sunday} \endlr
\end{vtimeline}

\begin{vtimeline}[description={text width=10cm}, 
 row sep=1ex, timeline color=Red] 
2010 & Rahtu et al. (ESA-D) \cite{Rahtu2010} \endlr
\ \ \ \ \ \ \ \ & Guo and Zhang (\acs{PQFT}) \cite{5223506} \endlr
\ \ \ \ \ \ \ \ & Goferman et al. \cite{Goferman} \endlr
\end{vtimeline}

\begin{vtimeline}[description={text width=10cm}, 
 row sep=1ex, timeline color=Green]
\ \ \ \ \ \ \ \ & Elazary and Itti \cite{Elazary20101338} \endlr
\end{vtimeline}

\begin{vtimeline}[description={text width=10cm}, 
 row sep=1ex, timeline color=Red] 
2011 & López-García et al. (\acs{WMAP}) \cite{lopez2011scene} \endlr
\end{vtimeline}

\begin{vtimeline}[description={text width=10cm}, 
 row sep=1ex, timeline color=Purple]
\ \ \ \ \ \ \ \ & Li et al. \cite{5733396} \endlr
\end{vtimeline}

\begin{vtimeline}[description={text width=10cm}, 
 row sep=1ex, timeline color=Red] 
2012 & García-Díaz et al. (\acs{AWS}) \cite{GallegaSaliency2012} \endlr
\end{vtimeline}

\begin{vtimeline}[description={text width=10cm}, 
 row sep=1ex, timeline color=NavyBlue]
2014 & Vig et al. \cite{vig2014deep} \endlr
\ \ \ \ \ \ \ \ & Kümmerer et al. (DeepGaze I) \cite{kummerer2014deepgaze1} \endlr
2015 & Huang et al. (SALICON) \cite{huang2015salicon} \endlr
2016 & Kümmerer et al. (DeepGaze II) \cite{kummerer2016deepgaze2} \endlr
\end{vtimeline}

\begin{vtimeline}[description={text width=10cm}, 
 row sep=1ex, timeline color=Sepia]
\ \ \ \ \ \ \ \ & Li et al. (DCL) \cite{DeepSaliencyObject} \endlr
\end{vtimeline}

\begin{vtimeline}[description={text width=10cm}, 
 row sep=1ex, timeline color=Red]
2017 & Leborán et al.(\acs{AWS-D}) \cite{GallegaSaliency} \endlr
\end{vtimeline}

\begin{vtimeline}[description={text width=10cm}, 
 row sep=1ex, timeline color=NavyBlue]
\ \ \ \ \ \ \ \ & Kruthiventi et al. (DeepFix) \cite{kruthiventi2017deepfix} \endlr
\end{vtimeline}

\begin{vtimeline}[description={text width=10cm}, 
 row sep=1ex, timeline color=Rhodamine]
\ \ \ \ \ \ \ \ & Jiang et al. \cite{ledov} \endlr
2018 & Bak et al. \cite{bak2018spatio} \endlr
\ \ \ \ \ \ \ \ & Wang et al. \cite{dhf1k} \endlr
\ \ \ \ \ \ \ \ & Cornia et al. (\acs{SAM}) \cite{lstm2018cornia} \endlr
\end{vtimeline}

\begin{vtimeline}[description={text width=10cm}, 
 row sep=1ex, timeline color=Sepia]
\ \ \ \ \ \ \ \ & Wang et al. \cite{wang2018video} \endlr
\end{vtimeline}

\caption*{\tikz\draw[fill=Red] (0,0) circle (.5ex); \acs{BU} models\hspace{5mm} \tikz\draw[fill=Green] (0,0) circle (.5ex); \acs{TD} models\hspace{5mm} \tikz\draw[fill=Gray] (0,0) circle (.5ex); \acs{BU}+\acs{TD} models\hspace{5mm} \tikz\draw[fill=YellowOrange] (0,0) circle (.5ex); Bayesian \acs{BU} models\\ \tikz\draw[fill=Purple] (0,0) circle (.5ex); Bayesian \acs{TD} models\hspace{5mm} \tikz\draw[fill=NavyBlue] (0,0) circle (.5ex); Deep static models\hspace{5mm} \tikz\draw[fill=Rhodamine] (0,0) circle (.5ex); Deep dynamic models\hspace{5mm}\\ \tikz\draw[fill=Sepia] (0,0) circle (.5ex); Deep salient object detection models \hspace{5mm}}
\caption{Chronological timeline of the visual attention models in the \emph{state-of-the-art} reviewed in this thesis.}\label{fig:saliency_timeline}
\end{figure*}

So far, we have reviewed how vision and visual attention work in humans. Similarly, vision in camera systems depends on the interaction among light, surfaces that reflect light, and a visual system that can detect light. Furthermore, both human eyes and cameras share certain physical similarities, and the same optical function: they gather light reflected from objects in order to obtain a sharply-focused image. 

However, while humans acquire knowledge about their surroundings, being able to respond to a given situation, cameras have no perceptual capabilities, so they can not interpret their recordings at all. As cameras are not able to deal with attention processes, computer vision researchers have developed automatic systems which efficiently determine the most appealing regions from images or videos, by means of \acfp{VAM}.

This section provides a review of some of the existing visual attention algorithms in the \emph{state-of-the-art}, referring both to \acs{BU} and \acs{TD} methods. We will specially focus on those that are closely related to our approaches, and also on promising deep learning-based architectures. All the models explained in the following paragraphs are summarized in Figure \ref{fig:saliency_timeline} in chronological order. For further information, a exceptional detailed survey on visual attention modeling is presented by Borji et al. \cite{Borji:2013:SVA:2412386.2412937}.

\subsection{Bottom-up versus top-down approaches}\label{sec:computational_bu_td} 
On the basis of Treisman and Gelade's \acs{FIT} \cite{TreismanGelade80}, Koch and Ullman \cite{koch_shifts} presented in 1985 a design to combine early vision features, and defined the concept of \acs{SM}, which was subsequently mentioned in Wolfe's \acs{GSM} as a mechanism to model local visual attention driven by the set of visual stimuli in the scene. 

The first implementation and verification of a \acs{BU} model, performed by Itti et al. \cite{Itti:1998:MSV:297843.297870}, and incorporating color, intensity and orientation features, would nevertheless come more than ten years later. After that, Harel et al. \cite{Harel07graph-basedvisual} proposed a saliency algorithm based on graphs, which extracted the same features at different scales. These two representations are the most frequently employed in the literature due to their good performance in a variety of applications \cite{yubing2011spatiotemporal, wang2013airport, ferfer2017exploiting}.

The great majority of visual attention models developed are \acs{BU} approaches, as the ones explained below. It is also surprising the lack of use and modeling of spatio-temporal and high-level features to address visual attention in real scenarios or videos, although almost half of the methods found in the literature are dynamic ones.

% Bottom-up
% - Characteristics of the visual scene
% - Salient object detection
\begin{itemize}
\item \emph{Itti et al., 1998} \cite{Itti:1998:MSV:297843.297870}: Related to \acs{FIT} \cite{TreismanGelade80} and in accordance with the biological architecture proposed by Koch and Ullman \cite{koch_shifts}, it was the first existing implementation of a spatial saliency model. It decomposes visual input into a set of topographic feature maps of color, intensity and orientation. The three conspicuity maps are normalized and summed constituting the \acs{SM}, whose maximum determines the most salient image region. 

Additionally, the authors propose a later stage to determine the order in which fixations may occur based on the \acs{SM} obtained. To this end, the \acs{SM} is modeled as a \acf{2D} layer of leaky integrate-and-fire neurons which feeds into a biologically-plausible \acf{WTA} neural network. Each \acs{SM} neuron excites its corresponding \acs{WTA} neuron, until the most salient ("the winner") fires. This causes the \acf{FOA} to be shifted to the location of the winner neuron. After this, all neurons are reset, and an \acf{IOR} mechanism is activated, which either allows the next most salient area to become the following winner or prevents the \acs{FOA} from shortly reaching an attended region. 

\item \emph{Harel et al., 2006} \cite{Harel07graph-basedvisual}: This spatial \acs{BU} saliency algorithm based on graphs extracts the same multi-scale features than Itti et. al \cite{Itti:1998:MSV:297843.297870}. \acf{GBVS} then computes a fully connected graph over all feature map grid locations, by assigning weights between each two nodes that are proportional to the similarity of feature values and their spatial distance. The obtained graphs proceed as Markov chains that define an equivalence relation either between nodes and states or edge weights and transition probabilities. Their associated equilibrium distribution results in activation maps, whose combination gives rise eventually to the \acs{SM}.

\item \emph{Hou and Zhang, 2007} \cite{HouZhang}: Hou and Zhang first proposed a spectral residual saliency model for static images. On the basis that statistical singularities in the spectrum may determine the anomalous regions of a given image, they derive its amplitude and phase. Then, they compute the log-spectrum of the down-sampled image. After that, the spectral residual is obtained by multiplying by a local average filter, and subtracting the result from the original version. Finally, the \acs{SM} is built in the spatial domain by using the \acf{IFT} and squaring the value of each spatial location. 

\item \emph{Hou and Zhang, 2009} \cite{hou2009dynamic}: Their next approach to saliency, called \acf{ICL}, is a principle that tries to maximize the overall entropy gain of a given set of sample visual features, now both in dynamic and static settings. Salient cues are those unexpected feature values that produce an entropy gain in the perception state. Therefore, features with large coding length increments will allow to reach attention selectivity. 
 
\item \emph{Seo and Milanfar, 2009} \cite{SeoMilanfar}: Given an image or video, the \acf{SDSR} spatio-temporal framework computes local regression kernel descriptors. Each pixel or voxel in the \acs{SM} indicates the statistical likelihood of saliency of a feature matrix given its surrounding feature matrices. According to the authors, the use of \acf{LSK} as features instead of conventional filter responses captures the underlying local structure of the data, even in the presence of significant distortions. % Se puede ampliar la explicacion, ver paper Master

\item \emph{Rahtu et al., 2010} \cite{Rahtu2010}: This spatio-temporal method, named ESA-D, proposes a combination of a statistical saliency measure based on contrasts in illumination, color and motion, together with a \acf{CRF} model for salient object detection. The motivation of the authors to determine the most appealing areas by minimizing an energy function derived from a \acs{CRF} is that saliency estimation objective is usually to achieve an object-level segmentation instead of a pixel-level one. 

\item \emph{Guo and Zhang, 2010} \cite{5223506}: The authors proposed the \acf{PQFT} method, which carries out a quaternion representation of video frames by means of intensity, color and motion features. %Moreover, they also implement a foveation approach based on a wavelet multiresolution representation, with the purpose of improving video compression efficiency.  

\item \emph{Goferman et al., 2010} \cite{Goferman}: This context-aware saliency model takes into account four psychological principles of human attention: 1) local low-level features, such as color and contrast; 2) global factors, which stand out features that differ from the norm; 3) visual organization rules related to forms and 4) a high-level face detector.    

\item \emph{López-García et al., 2011} \cite{lopez2011scene}: The authors proposed the \acf{WMAP} measure as a spatial visual attention estimator with the purpose of significantly accelerating a scene recognition task, preserving its performance. The approach considers both efficient coding, in order to reduce the redundancy of the input data, and the detection of important attributes of the image via local edge phase and energy.

\item \emph{García-Díaz et al., 2012} \cite{GallegaSaliency2012}: The \acf{AWS} model provides a measure of saliency by considering the local energy variability in the $Lab$ color space. First, a Gabor bank of filters is applied to the luminance channel $L$, extracting several multioriented multiresolution features. Additionally, a multiscale decomposition of the $a$ and $b$ color components is computed. Finally, a \acf{PCA} is performed over all these low-level representations in order to decorrelate them, obtaining the local measure of variability that underlies the \acs{SM}. 

\item \emph{Leborán et al., 2017} \cite{GallegaSaliency}: This approach, called \acf{AWS-D}, is an extension of the \acs{AWS} \cite{GallegaSaliency2012} explained above that computes either static or dynamic saliency maps. The hypothesis of the authors was that saliency has a strong relationship with the variability of the local energy measured over a statistically decorrelated and normalized space. Thus, in order to estimate saliency, the model looks for the space-time points with maximum variability in the distribution of the local energy across spatio-temporal scales and orientations. In contrast to other spatio-temporal methods, it does not rely on information derived from a explicit background model or the estimation of the optical flow to compute motion features. 

\end{itemize}

% Top-down
% - + eye movements 
In contrast, \acs{TD} architectures are still scarce and very often tailored to well-defined scenarios. In such cases, the evaluation of the whole scheme is performed regardless of the capability of the guidance tool. What is more, most \acs{TD} approaches guide attention towards specific targets by modulating gains associated with low-level stimuli.

\begin{itemize}
\item \emph{Sprague and Ballard, 2003} \cite{sprague2003eye}: They presented a reinforcement learning method that combines action selection and visual perception in a sidewalk navigation task. 

\item \emph{Bruce and Tsotsos, 2005} \cite{BruceT05}: The \acf{AIM} model proposed computes a \acs{VAM} based on the Shannon's self-information measure. At each image region, saliency is the information that the region conveys with respect to its surroundings. The information of the visual feature is inversely proportional to the likelihood of observing it. Consequently, to calculate this feature, the \acf{PDF} has to be estimated. Moreover, in order to reduce the dimensionality of the problem, an \acf{ICA} is performed. Thus, given a local image region, the probability of observing the $RGB$ values is estimated via the product of the likelihood of the components associated with that region.      

\item \emph{Judd et al., 2009} \cite{judd2009learning}: The model is based on a linear \acf{SVM}, taking some image features and human fixations to define salient locations. 

\item \emph{Elazary and Itti, 2010} \cite{Elazary20101338}: They proposed a more flexible \acs{TD} model that can concurrently select the best features to guide attention and adjust the width of feature detectors. 

\end{itemize}

Finally, although suggested by the prevalent studies about attention \cite{Yarbus1967, wolfe1994guided}, just a few works proposed hybrid models incorporating \acs{BU} and \acs{TD} factors. 

\begin{itemize}
\item \emph{Navalpakkam and Itti, 2005} \cite{1641004}: The model optimizes the integration of \acs{BU} cues for target detection by maximizing the signal-to-noise ratio of the target versus background. 

\item \emph{Peters and Itti, 2007} \cite{peters2007beyond}: This model computes a task-dependent map based on the scene gist and the eye fixations gathered from a video game scenario. \acs{BU} and \acs{TD} integration is simply conducted as a multiplication of both components. 
\end{itemize}

\subsection{Bayesian models}\label{computational_bayesian} 
Since our model proposed in Chapter \ref{ch:atom} is based on a Bayesian formulation of visual attention, this section briefly introduces some Bayesian or probabilistic approaches found in the \emph{state-of-the-art}. 

On the one hand, probabilistic \acs{BU} algorithms make use of Bayes' rule to combine the features observed with prior constrains:

\begin{itemize}
\item \emph{Torralba, 2003} \cite{Torralba:03}: The author presented a Bayesian approach for visual search tasks. \acs{BU} saliency stands on a global feature based on the scene gist, which provides a shortcut to detect the presence or absence of objects in an image before exploring it.

\item \emph{Itti and Baldi, 2005} \cite{Itti_Baldi06nips}: They proposed a probabilistic framework for modeling saliency as ``surprise'' by computing the \acs{KL} divergence between the posterior and prior beliefs about image features, either in space or time domains. 
\end{itemize}

On the other hand, Bayesian \acs{TD} models are characterized by their capacity to learn from data, taking advantage of data statistics to model the underlying attention process and allowing to obtain interpretable relationships between data and visual fixations:

\begin{itemize}
\item \emph{Zhang et al., 2008} \cite{doi:10.1167/8.7.32}: This framework understands saliency as the point-wise mutual information between \acs{BU} local features and \acs{TD} search target features. The model, known as \acf{SUN}, tries to reproduce the visual experience acquired by an organism. To achieve this, it defines visual saliency as the probability of a searched target at every point in the visual field given the features observed. Using Bayes' rule, and assuming that feature and location are independent, the self-information is taken as definition of \acs{BU} saliency: the rarer a feature is, the more it will attract our attention. Given this definition, features are calculated as responses of biologically plausible linear filters, such as \acf{DoG} and Gabor filters, as same as in \cite{Itti:1998:MSV:297843.297870}, and also as the responses to filters learned from natural images, using \acs{ICA}. 

After this first approach, the model was extended to temporally dynamic scenes in \cite{zhang2009sunday}, characterizing the video statistics around each pixel using a bank of spatio-temporal filters with separable space-time components.

\item \emph{Li et al., 2011} \cite{5733396}: They provided a multi-task learning approximation for visual attention in video, where different ranking functions for fusing \acs{BU} and \acs{TD} maps are learned depending on the scene content.  
\end{itemize}

%% DEEP LEARNING
\subsection{Deep Neural Networks}\label{sec:computational_cnn} 
\acfp{CNN} \cite{lecun1989backpropagation, simard2003best}, the current dominant paradigm for many supervised tasks in computer vision, have also been applied to model visual attention achieving promising results, especially in the still image domain. Furthermore, either for \acsp{SM} refinement based on attention modules or visual attention estimation in video, researchers have recently drawn on recurrent \acf{LSTM}-based networks \cite{hochreiter1997long}.

\acsp{DNN} involve training end-to-end models according to a loss function, using a database of images or videos annotated with \acs{GT} fixations. They unify feature extraction, fusion and saliency prediction into a single structure. This usually improves the system performance at the expense of making the analysis of these stages more challenging, mainly due to the abstract nature of representations at the deepest layers of these strategies.

Among the first attempts to rely on deep learning for static saliency estimation, the following ones deserve our analysis:

\begin{itemize}
\item \emph{Vig et al., 2014} \cite{vig2014deep}: It constituted one of the first approaches that makes use of \acs{CONV} layers for saliency prediction, training a softmax classifier on top of them, so that \acsp{SM} are formulated as generalized Bernoulli distributions.

\item \emph{Kümmerer et al., 2014} \cite{kummerer2014deepgaze1}, \emph{2016} \cite{kummerer2016deepgaze2}: First, in 2014, the authors presented Deep Gaze I, which builds \acsp{SM} by using the object recognition model of Krizhevsky et al. \cite{krizhevsky2012imagenet} and a prior distribution to model the central fixation bias \cite{kummerer2014deepgaze1}. Further on, in 2016, Deep Gaze II \cite{kummerer2016deepgaze2} applies transfer learning to saliency prediction by fine-tuning a few layers on top of the features from a VGG network \cite{vgg2014}, also for object recognition.
 
\item \emph{Huang et al., 2015} \cite{huang2015salicon}: The SALICON fine- and coarse-scale model evaluates the use of four commonly-known differentiable saliency metrics as the objective function of a simple \acs{CNN} architecture, providing image \acsp{SM} which integrate information at different scales.  Furthermore, the authors introduced a large-scale image dataset \cite{jiang2015salicon} for training new models, annotated by means of a mouse-tracking procedure, which seemed to correlate well with human fixations.

\item \emph{Kruthiventi et al., 2017} \cite{kruthiventi2017deepfix}: The authors presented DeepFix, a fully \acs{CNN} built on top of a VGG network \cite{vgg2014} for hierarchically modeling the \acs{BU} mechanism of visual attention. The network captures semantics at multiple scales and information derived from the global context, and also models center-bias effect in human attention. 

\item \emph{Cornia et al., 2018} \cite{lstm2018cornia}: \acs{SAM} model is able to predict accurate \acsp{SM} by incorporating a neural attentive mechanism based on convolutional \acsp{LSTM} \cite{xingjian2015convolutional}. Given a \acs{SM}, the method refines it by iteratively focusing on the most prominent regions, and also considering the center bias existing in human fixations by learning a set of prior Gaussian maps. 
\end{itemize}

Despite the outstanding performance achieved by these approaches, they still miss some key elements \cite{DBLP:conf/eccv/BylinskiiRBOTD16}, mostly related to mis-detections of people, actions and text, and the relative importance assigned to them when they take place simultaneously. 

It should also be pointed out that only a few works have drawn on deep learning to tackle the estimation of visual attention in videos:

\begin{itemize}
\item \emph{Jiang et al., 2017} \cite{ledov}: Together with a large-scale eye-tracking database of generic videos, the authors proposed a \acs{CNN} to learn spatio-temporal features based on object motion, and also a two-layer convolutional \acs{LSTM} network to smooth the transition between \acsp{SM} of consecutive frames.

\item \emph{Bak et al., 2018} \cite{bak2018spatio}: The authors studied the use of dynamic models for saliency prediction in videos, providing several single and two-stream \acsp{CNN} and evaluating different fusion mechanisms to combine spatial and temporal information. They demonstrated the importance of considering inherent motion information, by training models on estimated optical flow. % \cite{DBLP:journals/corr/BakEE16}

\item \emph{Wang et al., 2018} \cite{dhf1k}: Similarly to \cite{ledov}, the authors shared a new large-scale database of videos with fixations, organized by their categories, and then presented a \acs{CNN}-\acs{LSTM} architecture with an attention mechanism. The \acs{CNN} encodes the static saliency information, which allows the \acs{LSTM} to learn temporal saliency representations for successive video frames.
\end{itemize}

Finally, \acsp{CNN} have also been applied to a particular type of saliency, closely related to the object detection task, both in images or videos. These models provide a map of objectness, measuring the probability that each image location belongs to an object. Nevertheless, these networks are often trained on databases with object segmentation masks as \acs{GT} instead of fixations.

\begin{itemize}
\item \emph{Li et al., 2016} \cite{DeepSaliencyObject}: The authors presented an end-to-end deep contrast network for salient object detection. The network consists of a pixel-level \acs{FC} stream, which generates a \acs{SM} with pixel-level accuracy, and a segment-wise spatial pooling stream, which improves the modeling of saliency discontinuities along object boundaries. Moreover, on top of these two streams, a \acs{CRF} model can be applied to improve the spatial coherence and contour localization. 
 
\item \emph{Wang et al., 2018} \cite{wang2018video}: They provided an efficient framework for object detection in videos that captures spatial and temporal saliency information via a short-term analysis consisting of learning from adjacent frame pairs, without the need to compute optical flow.
\end{itemize}

The reader is also referred to a recently released survey by Ali Borji about deep learning-based models for saliency prediction in images and videos \cite{borji2018saliency}, where the author also discusses emerging applications of these architectures and which aspects should be considered in order to improve them.

\subsection{Applications}\label{sec:computational_applications}
Nowadays, computer vision techniques have to deal with millions and millions of available data, just as the \acs{HVS} does. This is probably the prime reason why the effort in developing computational systems to accomplish the task carried out by visual attention has increased during the last few years. 

We can highlight two main purposes for visual attention modeling \cite{borji2018saliency}:

\begin{itemize}
\item The first, related to our contributions in Chapters \ref{ch:atom} and \ref{ch:atom_experiments}, is to understand \emph{how} visual attention works in humans, trying to describe the behavioral and neural processes involved. 

\item The second is to predict \emph{where} people look, in order to address traditional image and video applications, such as object \cite{ivanNUS,IEEEVideoObjectRecognition} and action \cite{IEEEVideoActionRecognition, WangCNNAction} recognition, video summarization \cite{Ma:2002:UAM:641007.641116}, patient diagnosis \cite{Chaabouni2017} or image quality assessment \cite{IEEEVideoImageQuality}, in broader and more complex scenarios, while providing efficient solutions and better performances. In line with the second objective, our contributions in Chapters \ref{ch:anomaly_detection} and \ref{ch:surveillance_experiments} pursue to facilitate the task of a \acs{CCTV} operator in a video surveillance scenario, by means of a deep architecture that models attention in the temporal domain. 
\end{itemize}

%\subsubsection*{Video surveillance}
%In line with the second objective, our contributions in Chapters \ref{ch:anomaly_detection} and \ref{ch:surveillance_experiments} pursue to facilitate the task of a \acs{CCTV} operator in a video surveillance scenario, by means of a deep architecture that models attention in the temporal domain. 
%
%A video surveillance scenario involves... COMPLETE

%Anomaly detection
%-Souad application
%-Ivan application (egocentric video)
%-studying car driving-related attentional mechanisms \cite{DBLP:journals/corr/PalazziSCAC16} and recognizing human activities \cite{WangCNNAction}.
%-surveillance,automatic target detection,navigational aids, robotic control

%\subsection{Models for eyes and gaze}\label{sec:eye_tracking_va}
%Eye tracking devices

%*****************************************
%*****************************************
%*****************************************
%*****************************************
%*****************************************

%\addtocontents{toc}{\protect\clearpage} % <--- just debug stuff, ignore
%\include{Chapters/Chapter03}
%************************************************
\chapter{A generative probabilistic model for spatio-temporal visual attention}\label{ch:atom} 
\chaptermark{A generative probabilistic model for visual attention}
%************************************************

\section{Introduction}\label{sec:introduction_4}
Modern computer vision techniques have to deal with vast amounts of visual data, which requires a computational effort that has often to be accomplished in challenging scenarios. The interest in solving these image and video applications efficiently has led researchers to develop visual attention-based methods to expertly drive the corresponding processing to conspicuous regions that either depend on the context or are based on specific requirements of the task. 

In this chapter, we propose a generative hierarchical probabilistic framework for spatio-temporal visual attention understanding and prediction in video. Our model is independent of the application scenario, and founded on the most outstanding psychological studies about attention and eye movements, which support that \emph{guidance} is not based directly on the information provided by early visual processes but on a contextual representation arisen from them. 

Drawing from the well-known \acf{LDA} \cite{LDA} method for the analysis of large corpus of data, and inspired by some of its supervised extensions \cite{sLDA,DBAyang}, our approach defines task- or context-driven visual attention as a mixture of latent sub-tasks, which are in turn modeled as a combination of specific distributions associated with low-, mid- and high-level spatio-temporal features. Learning from fixations gathered from human observers, we incorporate an intermediate level between feature extraction and visual attention estimation that enables to obtain guiding representations. 

\section*{Chapter overview}
The chapter is organized as follows. First, in Section \ref{sec:related_work4}, we review the most relevant and recent related work in perception and spatio-temporal visual attention, justify our claims, and present our main contributions. Then, Section \ref{sec:handcrafted_features} presents a broad set of example features which may potentially guide the attention of observers, and will be tested as input for our experiments in Chapter \ref{ch:atom_experiments}. Next, probabilistic \acfp{LTM} are introduced in Section \ref{sec:latent_topic_models}, putting special emphasis on \acs{LDA} and its supervised extensions, on the basis of which our approach is developed. Finally, Section \ref{sec:sub_atom} describes in detail our generative probabilistic framework for spatio-temporal visual attention understanding and prediction.

\section{Related work and main contributions}\label{sec:related_work4}
As it was discussed in Section \ref{sec:computational_bu_td}, most of the visual attention models developed thus far are \acf{BU} approaches \cite{Harel07graph-basedvisual, hou2009dynamic, Goferman, GallegaSaliency}, whereas \acf{TD} architectures are mostly tailored to scenarios where it is not critical to achieve a good estimation of visual attention to solve a particular task \cite{sprague2003eye, peters2007beyond}. Only a few works tackle the confluence between \acs{BU} and \acs{TD} factors, and there is still a lot of research to be done in real scenarios or videos. 

Probabilistic models have an undeniable potential: they are able either to estimate attention or to understand its process \cite{Torralba:03, doi:10.1167/8.7.32, 5733396}. However, their expressive ability is often limited to very simple single-layered fusion schemes built on top of features.

To overcome these shortcomings, we propose a general data-driven hierarchical probabilistic architecture to estimate visual attention in videos, which can be applied to different scenarios and tasks by simply learning from human fixations. 

Our \acs{LTM}-based design was first described in \cite{cbmi2016miguel}. It introduced an intermediate level formed by latent sub-tasks, which bridges the gap between features and visual attention, and enables to obtain more comprehensive interpretations of attention guidance. These representations provide additional information about how features are combined both in attracting and inhibiting spatial locations. Then, \acs{TD} visual attention is modeled as a linear regression over a set of learned intermediate sub-tasks rather than over the features themselves. Depending on the context, distinct features could draw visual attention. For instance, motion features are useful to follow players and track objects in outdoor scenes, while color, faces or text are more relevant in TV recordings. However, the fundamental basis of the system is, indeed, generic and independent from the application scenario. 

In a recent work \cite{csvtmiguel}, we updated this design, making two substantial contributions: 

\begin{enumerate}
\item We generated a categorical binary response for each spatial location to model visual attention, in contrast to the continuous variable used in our previous approach \cite{cbmi2016miguel}. The new system now allows to automatically align the sub-tasks discovered to a binary response by means of a logistic regressor, which fully corresponds to the definition of human fixations.

\item We extended the initial set of basic and novelty spatio-temporal low-level features presented in \cite{cbmi2016miguel}, including and modeling some new mid- and high-level features related to camera motion estimation and object detection, and taking advantage of powerful paradigms such as \acfp{CNN}. To do the latter, we make use of the features derived from a recently released deep contrast network for salient object detection with pixel-level accuracy \cite{DeepSaliencyObject}.
\end{enumerate}

For the sake of simplicity, we will only describe the current extended version of the model in this thesis \cite{csvtmiguel}, providing the results on visual attention estimation of the first approach as part of the comparison with the state-of-the-art methods carried out in Section \ref{sec:results5}.  

\section{Feature engineering for visual attention guidance}\label{sec:handcrafted_features}
According to the most leading psychology theories for computational attention systems \cite{TreismanGelade80} \cite{wolfe1994guided}, different simple features are early and pre-attentively processed in parallel to guide visual search in the human brain (see Section \ref{sec:psy_va}). 

Selective visual attention is built on what it is called the \emph{early representation} \cite{koch_shifts}, a set of conspicuity maps related with some \emph{elementary features} such as color, orientation or motion. These topographical maps do not only surround physical attributes, but also may be explained as relational aspects of these physical characteristics. We may even guide our attention by focusing on mid- and high-level features such as symmetry, faces or text.

In the experiments described in Chapter \ref{ch:atom_experiments}, a wide set of features has been considered. For the sake of completeness, we briefly describe the features in the following subsections.

%\afterpage{
%\begin{landscape}
\begin{figure*}
\centering
	\includegraphics[trim=0cm 0cm 0cm 0cm, width=0.9\textwidth]{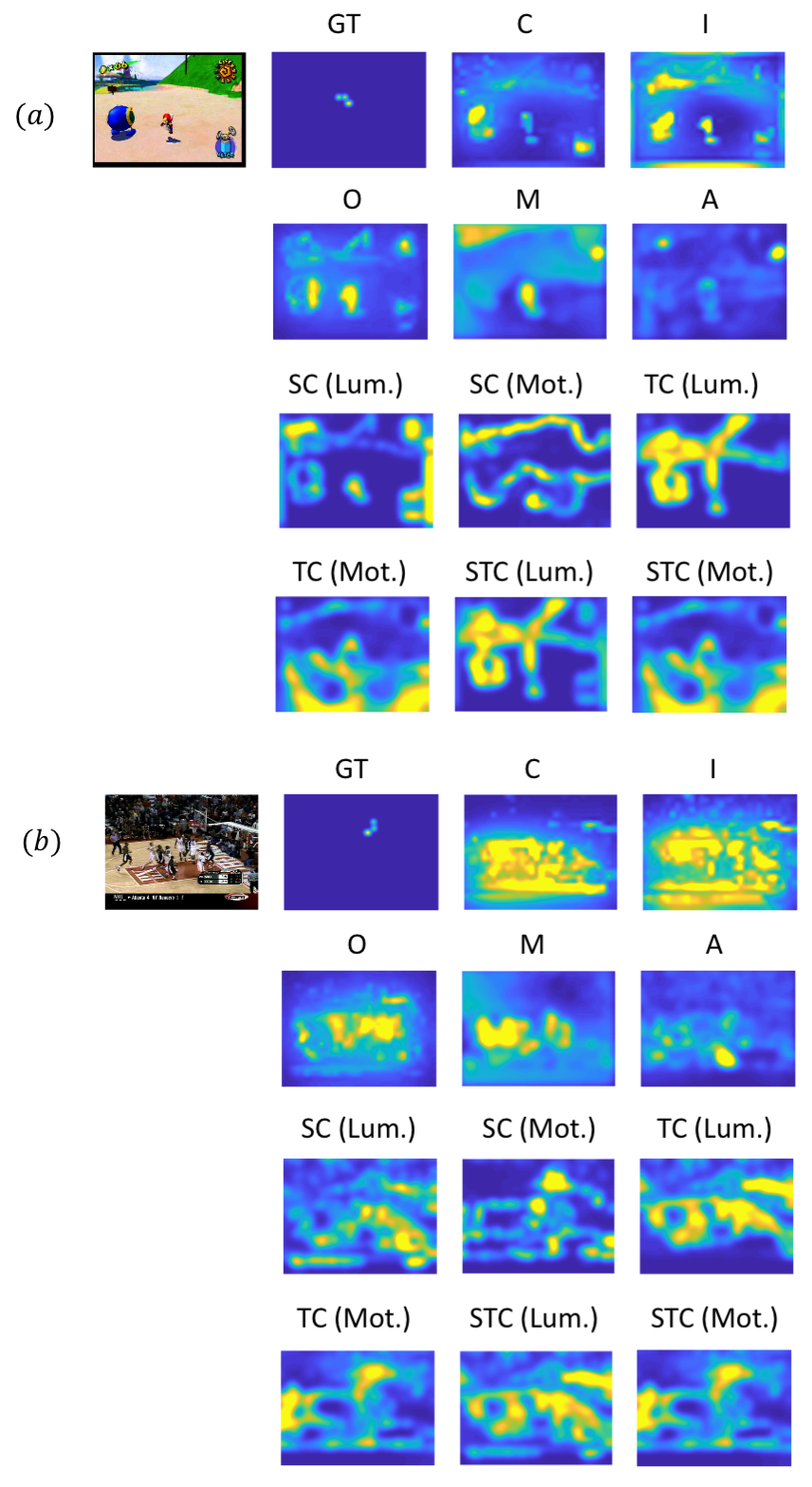}
\caption[Basic, motion-based and novelty feature maps computed for two example frames taken from Videogames (a) and Sports (b) categories from CRCNS-ORIG \cite{Itti_Carmi09crcns} database.]{Basic, motion-based and novelty feature maps computed for two example frames taken from Videogames (a) and Sports (b) categories from CRCNS-ORIG \cite{Itti_Carmi09crcns} database. Basic features are color (C), intensity (I) and orientation (O), extracted by using the \acs{BU} approach for saliency estimation of Harel et al. \cite{Harel07graph-basedvisual}. Motion-based features are velocity or motion magnitude (M) and acceleration (A). Novelty features are spatial coherency (SC (Lum.), SC (Mot.)), temporal coherency (TC (Lum.), TC (Mot.)) and spatio-temporal coherency (STC (Lum.), STC (Mot.)), computed either over the pixel intensity values $\mathcal{I}$ or the motion phase $\theta_{M}$.}
\label{fig:44_gaussian_features}
% \vspace{-0.4cm}
\end{figure*} 
%\end{landscape}
%}

\subsection{Basic features: color, intensity and orientation}\label{sec:basic_features}
% Siglas para figuras, se indican en la figura directamente! (simplificar listado de acrónimos, en el texto nunca se las llamará con una letra)
As stated in \acs{FIT} \cite{TreismanGelade80}, the majority of computational models of attention consider three early visual features: \emph{intensity} or \emph{luminance contrast} (I), \emph{color} (C) and \emph{orientation} (O). In this section, we briefly explain and compare how they are represented in the famous model of Itti et al. \cite{Itti:1998:MSV:297843.297870} and its update based on graphs \cite{Harel07graph-basedvisual}, which have been introduced in Section \ref{sec:computational_bu_td}. Due to their easy interpretation, effectiveness and prevalence almost up to the advent of \acsp{CNN}, we decided to make use of the activation maps from \cite{Harel07graph-basedvisual} in our experiments on visual attention understanding.

\subsubsection*{Feature maps extraction}
On the basis of the red ($r$), green ($g$) and blue ($b$) components of a given still image, five channels are obtained, which constitute the basis of the subsequent feature and activation maps. First, an intensity channel $I$, computed as the linear combination of the three components:

\begin{gather}
I = \frac{r+g+b}{3} 
\end{gather}

Then, $r$, $g$ and $b$ channels are normalized by $I$ to separate hue from intensity, and four additional color channels (red, green, blue and yellow) are calculated:

\begin{gather}
R = r - \frac{g+b}{2} \\
G = g - \frac{r+b}{2} \\
B = b - \frac{r+g}{2} \\
Y = \frac{r+g}{2} - \frac{|r-g|}{2} -b
\end{gather}

A multi-scale process is performed for feature extraction. For that purpose, five Gaussian pyramids ($I(\sigma)$, $R(\sigma)$, $G(\sigma)$, $B(\sigma)$ and $Y(\sigma)$) are generated by consecutively low-pass filtering and sub-sampling each channel. In total, each pyramid is composed of nine spatial scales $\sigma \in [0, 8]$. 

It should be noted that Harel et al. \cite{Harel07graph-basedvisual} method proposes the $DKL$ color space \cite{derrington1984chromatic} as a better alternative to the $RGBY$ model. This color space is composed by three axis. The first one represents luminance changes independently from variations in chromaticity, while along the others chromaticity varies without changing the excitation of blue-sensitive or red- and green-sensitive cones, respectively.

Finally, $I$ is convoluted by several oriented Gabor filters $O(\sigma,\theta)$ at different scales $\sigma$ and with multiple orientations $\theta \in \{ 0^{\circ}, 45^{\circ}, 90^{\circ}, 135^{\circ} \}$ in order to extract orientation-based maps.

Once the feature maps have been obtained, an activation map associated with each early visual feature is computed. To this end, the pioneer Itti et al. \cite{Itti:1998:MSV:297843.297870} model proposed a center-surround approach, while Harel et al. \cite{Harel07graph-basedvisual} presented a graph-based mechanism. Both are explained in the following paragraphs.

\subsubsection*{Center-surround activation maps formation}
Activation maps in \cite{Itti:1998:MSV:297843.297870} arise from the center-surround difference ($\ominus$) between ``center'' fine scales $c$ and ``surround'' coarser scales $s$. This operation, which tries to simulate the receptive field structure of neurons in the \acs{HVS} \cite{Palmer, lindeberg2013computational}, allows to detect prominent locations with respect to their surround. To this effect, it involves an interpolation to the finer scale and a point-by-point subtraction.

First, several intensity contrast maps, which correspond to neurons receptivity to dark spots on bright surrounds and vice versa, are obtained as follows:

\begin{gather}
\mathcal{I}(c,s) = |I(c) \ominus I(s)|, 
\end{gather}

\noindent where $\ominus$ denotes the across-scale difference between two maps. Secondly, different color maps are concerned with red/green ($\mathcal{RG}(c,s)$) and blue/yellow ($\mathcal{BY}(c,s)$) double opponencies, also perceivable in human visual cortex:

\begin{gather}
\mathcal{RG}(c,s) = |(R(c) - G(c)) \ominus (G(s)-R(s))| \\
\mathcal{RG}(c,s) = |(B(c) - Y(c)) \ominus (Y(s)-B(s))| 
\end{gather}

Local orientation information is also provided by orientation-selective neurons in primary visual cortex. Following the same process, orientation feature maps are computed to encode the local orientation contrast between center and surround scales:

\begin{gather}
\mathcal{O}(c,s,\theta) = |O(c,\theta) \ominus O(s,\theta)| 
\end{gather}

In accordance with the hypothesis that similar maps associated with a particular feature compete for saliency, while different features contribute independently to it, three separate activation or conspicuity maps are built for intensity contrast, color and orientation. Feature maps obtained have different dynamic ranges, so a normalization operation $\mathcal{N} (.)$ is applied to them before their combination. This operator not only removes amplitude differences between maps, but also stands out in each map those activation spots whose difference from the average is large. Finally, activation maps are computed by performing an across-scale addition between maps $\oplus$, reducing each feature map before to the fourth spatial scale considered:

\begin{gather}
\mathcal{\overline I} = \bigoplus\limits_{c} \bigoplus\limits_{s} \mathcal{N}(\mathcal{I}(c,s)) \\
\mathcal{\overline C} = \bigoplus\limits_{c} \bigoplus\limits_{s} [ \mathcal{N}(\mathcal{RG}(c,s)) + (\mathcal{BY}(c,s)) ] \\
\mathcal{\overline O} = \sum\limits_{\theta} \mathcal{N} \left(\bigoplus\limits_{c} \bigoplus\limits_{s} \mathcal{N}(\mathcal{O}(c,s,\theta))\right)
\end{gather}

\subsubsection*{Graph-based activation maps formation}
Alternatively, the Markovian approach presented in \cite{Harel07graph-basedvisual} tries to imitate the communication between neurons in the visual cortex when processing areas of a scene. Given a feature map $M$ at a particular scale, it establishes a fully-connected directed graph $G_{A}$ by connecting every location $(i,j)$ in $M(\sigma)$ with all other locations $(p,q)$. The dissimilarity $d((i,j)||(p,q))$ of $M(i,j)$ and $M(p,q)$ is defined as:

\begin{gather}
d((i,j)||(p,q)) \triangleq \left|log \frac{M(i,j)}{M(p,q)}\right|
\end{gather} 

Thus, a weight is assigned to the directed edge from location $(i,j)$ to location $(p,q)$:

\begin{gather}
w_{1}((i,j)||(p,q)) \triangleq d((i,j)||(p,q)) \cdot F(i-p,j-q)
\end{gather}

\noindent where $F$ is related to their spatial distance:

\begin{gather}
F(a,b) \triangleq exp\left(-\frac{a^{2}+b^{2}}{2\gamma^{2}}\right),
\end{gather}

\noindent being $\gamma$ a free parameter of the algorithm. A Markov chain is then defined on $G_{A}$ by normalizing the edge weights of each node to sum to 1, and drawing on the correspondence between locations and states, and edge weights and transition probabilities. The equilibrium distribution of this chain highlights those regions that have high dissimilarity with their surrounding, resulting in the expected conspicuousness map. Finally, the activation map associated with each early visual feature is obtained from the combination of all the normalized conspicuousness maps at different scales, according to a procedure similar to the one applied in \cite{Itti:1998:MSV:297843.297870}, which has been explained above. Figure \ref{fig:44_gaussian_features} includes examples of these feature maps for two frames in CRCNS-ORIG \cite{Itti_Carmi09crcns} database.

\subsection{Motion-based features}\label{sec:motion_features}
Motion is undoubtedly another feature that attracts our gaze. It was introduced to model visual attention for the first time in \cite{itti2003realistic}. Given two images, this neurobiological approach considers a motion map in terms of the difference between their corresponding Gabor orientation pyramids, capturing a wide range of object speeds. 

The motion-based features used as input for the method presented in this chapter of the thesis draw instead on the optical flow technique proposed in \cite{cliuphdthesis} for dense motion estimation. Moreover, a parametric motion model is obtained to estimate camera motion, which also serves to detect moving objects. Both are described below, together with the feature maps used in our attention model: \emph{velocity}, \emph{acceleration} and \emph{camera motion}.

\subsubsection*{Optical flow estimation} 
Optical flow \cite{szeliski2010computer} computes an independent estimate of motion $\mathbf{v_{n}}$ at each spatial location $\xv_{n} = (x_{n}, y_{n})$, which can be tackled by minimizing the \acf{SSD} between the intensity or brightness of corresponding pixels in two consecutive frames $I_{t-1}$ and $I_{t}$ in a video:

\begin{gather}\label{eq:ssd}
E_{SSD}(\{\mathbf{v_{n}}\}) = \sum_{n} [I_{t}(\xv_{n}+\mathbf{v_{n}})-I_{t-1}(\xv_{n})]^2,
\end{gather}

\noindent where $\{\mathbf{v_{n}}\}$ denotes the whole vector field. In order to efficiently optimize this cost function, an image pyramid is usually built and motion is estimated hierarchically from coarse to fine scales, as first suggested by Lucas and Kanade in \cite{lucaskanade1981}. The solution to this function is underconstrained, since we have two variables $\mathbf{v_{n}} = (u_{n}, v_{n})$ to determine and just one equation per pixel. For each pair of consecutive frames $I_{t-1}$ and $I_{t}$, the patch-based typical approach to this problem involves a local summation over overlapping regions $\xv_{n}$ and $(\xv_{n}+\mathbf{v_{n}} + \Delta \mathbf{v_{n}})$, as well as performing gradient descent on Eq. \ref{eq:ssd} using a Taylor series expansion of the displaced image function:

\begin{equation}
\begin{split}\label{eq:ssd_gd}
E_{SSD}(\{\mathbf{v_{n}} + \Delta \mathbf{v_{n}}\}) &= \sum_{n} [I_{t}(\xv_{n}+\mathbf{v_{n}} + \Delta \mathbf{v_{n}}) -I_{t-1}(\xv_{n})]^2 \\
&\approx \sum_{n} [I_{t}(\xv_{n}+ \mathbf{v_{n}})+ \mathbf{J_{t}}(\xv_{n}+ \mathbf{v_{n}})-I_{t-1}(\xv_{n})]^2 \\
&= \sum_{n} [\mathbf{J_{t}}(\xv_{n}+\mathbf{v_{n}}) \Delta \mathbf{v_{n}} + e_{n}]^2,
\end{split}
\end{equation}

\noindent where

\begin{gather} 
\mathbf{J_{t}}(\xv_{n}+ \mathbf{v_{n}}) = \nabla I_{t}(\xv_{n}+\mathbf{v_{n}}) = \left( \frac{dI_{t}}{dx}, \frac{dI_{t}}{dy} \right) (\xv_{n} + \mathbf{v_{n}})
\end{gather}

\noindent is the image gradient or Jacobian at $(\xv_{n}+\mathbf{v_{n}})$ and $e_{n} = I_{t}(\xv_{n} + \mathbf{v_{n}})-I_{t-1}(\xv_{n})$ is the temporal derivative or brightness change between images.

Horn and Schunck \cite{horn1981determining} later proposed a regularization-based variational framework to minimize Eq. \ref{eq:ssd} simultaneously over all flow vectors $\mathbf{v_{n}}$, which is known as the linearized optical flow constraint:

\begin{gather}
E_{SSD} = \iint \left[(I_{x}u+I_{y}v+I_{t})^2 + \alpha(||\nabla u||^2 + ||\nabla v||^2)\right] dxdy,
\end{gather}

\noindent denoting $(I_{x},I_{y})=\mathbf{J_{t}}(\xv_{n}+ \mathbf{v_{n}})$ and $I_{t}=e_{n}$ spatial and temporal derivatives, respectively. In addition, $\alpha$ is a regularization constant to be determined.

% Brox04highaccuracy is the GPU IMPLEMENTATION!!!
Using as baseline the algorithms in \cite{lucaskanade, Brox04highaccuracy}, and also including symmetric flow computation, Liu et al. \cite{cliuphdthesis} presented a layer-wise optical flow estimation method. Layered motion framework arises from the observation that motion in a scene is often associated with few objects at different depths, so that pixels motion can be estimated more accurately if they are grouped into suitable objects or layers. According to this approach, the optical flow constraint has three terms. First, a data term, which matches the two consecutive frames $I_{t-1}$ and $I_{t}$:

\begin{gather}
E^{(t)}_{data} = \int g*M_{t}(x, y) |I_{t}(x + u_{t}, y + v_{t}) - I_{t-1}(x, y)| \ dxdy,
\end{gather}

\noindent being $g$ a Gaussian filter, $M_{t}$ the visible layer mask that indicates which layers' pixels are not occluded, and $\mathbf{v}_{t}=(u_{t},v_{t})$ the flow field from $I_{t}$ to $I_{t-1}$. Second, a smoothness term, defined as:

\begin{gather}
E^{(t)}_{smooth} = \int (|\nabla u_{t}|^2 + |\nabla v_{t}|^2)^\eta \ dxdy,
\end{gather}

\noindent where $\eta$ constant varies between 0.5 and 1. Finally, symmetric matching is achieved by means of the following term:

\begin{equation}
\begin{split}\label{eq:of_sym}
E^{(t)}_{sym} = \int |u_{t}(x,y) + u_{t-1}(x+u_{t},y+v_{t})| + \\
 |v_{t}(x,y) + v_{t-1}(x+u_{t},y+v_{t})| \ dxdy.
\end{split}
\end{equation}

Note that $E^{(t-1)}$ terms are defined in a similar way. Thus, the optimization function consists of the sum of these terms:

\begin{gather}
E(\mathbf{v_{t}},\mathbf{v_{t-1}}) = \sum_{i=t-1}^{t} E^{(i)}_{data} + \alpha E^{(i)}_{smooth} + \beta E^{(i)}_{sym},
\end{gather}

\noindent being $\alpha$ and $\beta$ two additional regularization constants. Flow computation is performed from coarse to fine image pyramid levels, updating the visible layer mask $M_{t}$ after each scale. Given the flow estimated at the current scale, if $M_{t-1}(\xv+\mathbf{v_{t}})=0$ or the symmetry term in Eq. \ref{eq:of_sym} at this location is beyond a threshold, $M_{t}(\xv)=0$ for the next finer scale. 

\subsubsection*{Motion parameterization}
\begin{figure*}[t]
\centering
	\includegraphics[trim=0cm 0cm 0cm 0cm, width=1\textwidth]{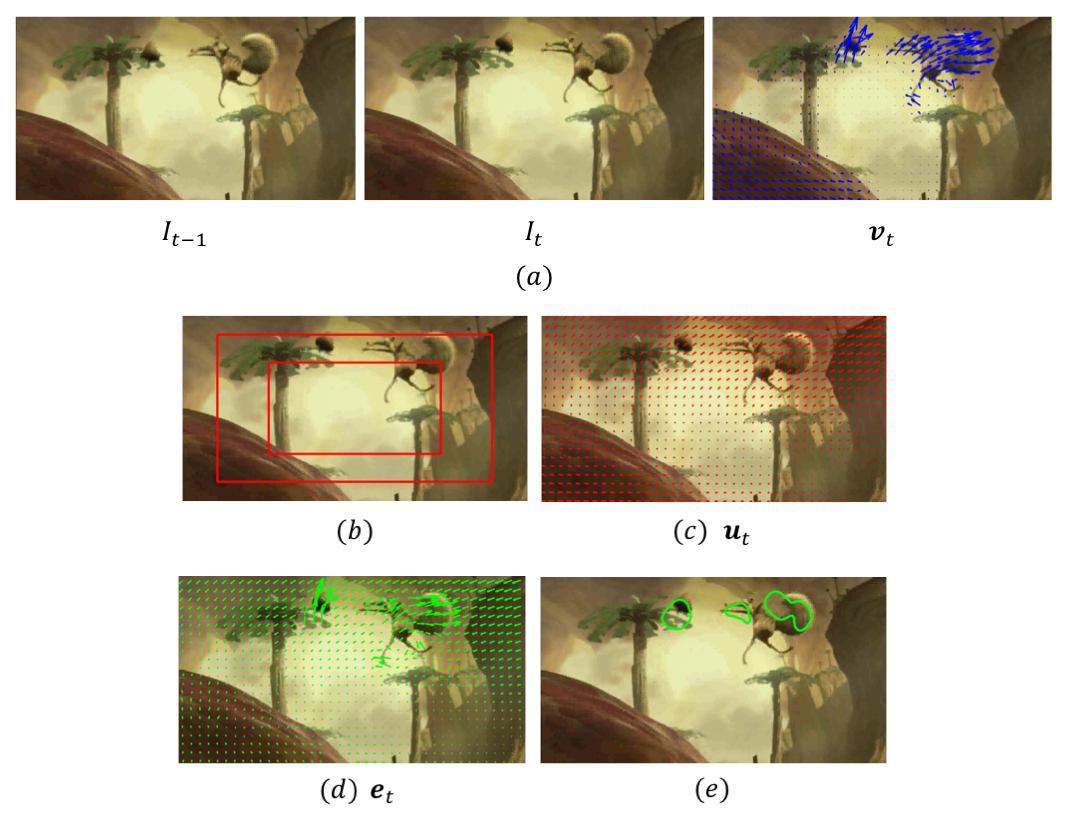}
\caption[Motion parameterization in two example frames taken from a commercials video in the DIEM \cite{mital2011clustering} database.]{Motion parameterization in two example frames taken from a Commercials video in the DIEM \cite{mital2011clustering} database. (a) Frames $I_{t-1}$ and $I_{t}$, together with the computed optical flow vector field ${\bf v}_{t}$. (b) Camera motion modeling. Assuming that moving objects are centered on the image and since optical flow estimation cloase to the edges is less accurate, we only consider spatial locations within the red inner ring to obtain the parametric camera motion model. (c) Camera motion modeled as a translation vector field ${\bf u}_{t}$. (d) Residual motion vector field ${\bf e}_{t}$. (e) Moving objects correspond to the green regions indicated in the image, with residual energy ${\bf e}^{2}_{t}$ higher than an empirically determined threshold $\xi$.}\label{fig:332_optical_flow}
% \vspace{-0.4cm}
\end{figure*}

From the vector field $\mathbf{v_{n}}$ computed, the correspondence between the spatial locations $\mathbf{x}_{n}$ and $\mathbf{x}'_{n}$ in two consecutive frames can be defined as:

\begin{gather}
\mathbf{x}'_{n} = \mathbf{x}_{n} + \mathbf{v_{n}}
\end{gather}

If $N_{t}$ is the total number of pixels of each frame, and we express each spatial location $n \in N_{t}$ in homogeneous coordinates, so that $\overline{\xv}_{n} = (x_{n},y_{n},1)^T$, we are able to represent both frames as two $Nx3$ matrices $\mathbf{X}$ and $\mathbf{X}'$. Then, a parametric model for camera motion can be obtained as:

\begin{gather}
\mathbf{X}' = \mathbf{X}\mathbf{H} \Rightarrow \mathbf{v} = \mathbf{X}(\mathbf{H}-\mathbf{I}) \Rightarrow \mathbf{v} = \mathbf{X}\mathbf{P} 
\end{gather}

\noindent being $\mathbf{H}$ and $\mathbf{P}$ the $3\times3$ matrices that define the geometric transformation between frames and the parametric camera motion model, respectively. According to an affine motion model \cite{hartley2003multiple}, $\mathbf{P}$ is defined as:

\begin{gather}
\mathbf{P} = \begin{bmatrix}
\mathbf{A} & \mathbf{u} \\
\mathbf{0}^T & 1  
\end{bmatrix}
\end{gather}

\noindent where $\mathbf{A}$ is a $2 \times 2$ non-singular matrix, $\mathbf{u}=(u_{x}, u_{y})$ a translation \acs{2D}-vector and $\mathbf{0}$ a null \acs{2D}-vector. The transformation has six degrees of freedom, which correspond to the six elements in $\mathbf{A}$ and $\mathbf{u}$, and is estimated as:

\begin{gather}
\mathbf{P} = \mathbf{X}^{+} \mathbf{u}
\end{gather}

\noindent being $\mathbf{X}^{+}$ the pseudoinverse of the matrix $\mathbf{X}$ of homogeneous coordinates. Assuming that moving objects tend to be centered on the image and since optical flow estimation close to the edges is less accurate, we only consider spatial locations within an inner ring for camera motion modeling, as shown in Figure \ref{fig:332_optical_flow}(b). 

Finally, if we want to detect moving objects on the scene, we use $\mathbf{P}$ to compute the energy $e^2_{n}$ of the residual motion for each pixel $\mathbf{x_{n}}$:

\begin{gather}
e^2_{n} = ||\mathbf{v_{n}}-\mathbf{P}\mathbf{x_{n}}||^2 \geq \xi
\end{gather}

Those pixels with residual energy higher than an empirically determined threshold $\xi$ will correspond to moving objects, as can be seen in the example in Figure \ref{fig:332_optical_flow}(e).

\begin{figure*}[!t]
\centering
	\includegraphics[trim=0cm 0cm 0cm 0cm, width=1\textwidth]{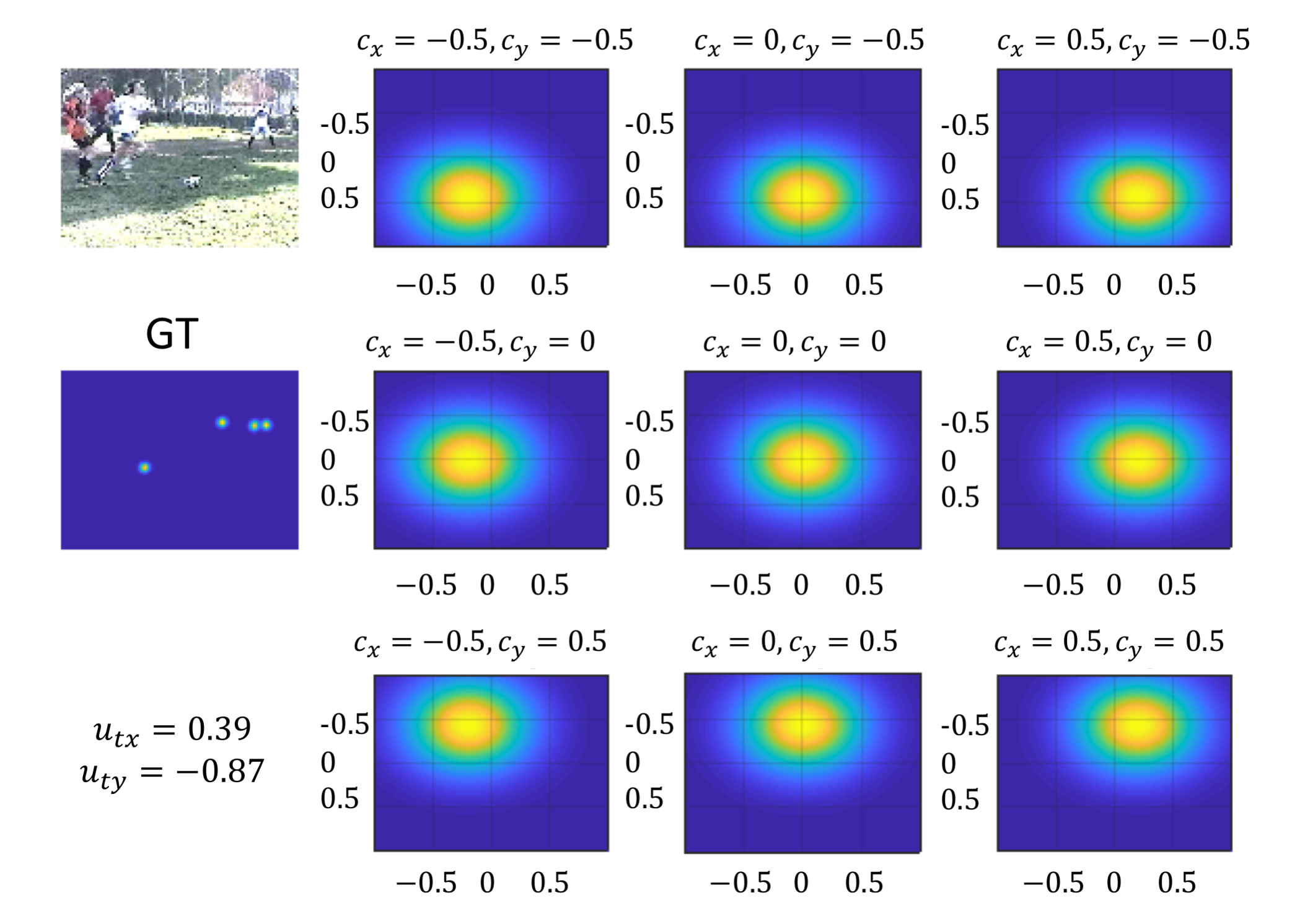}
\caption[Camera motion modeling in an Outdoor frame taken from the CRCNS-ORIG \cite{Itti_Carmi09crcns} database.]{Camera motion modeling in an Outdoor frame taken from the CRCNS-ORIG \cite{Itti_Carmi09crcns} database. Visual attention based on camera motion is modeled by means of a 2D Gaussian distribution $N(\cv_{z_n} \odot \uv, \Sigma_{z_n})$ over the spatial coordinates, where $\uv$ is the translation vector modeling the camera motion. Figure illustrates, given a sample value of $\uv$, the distribution learned for different $\cv = (c_{x}, c_{y})$ values, being $\cv$ the vector of parameters that establishes a relation between the camera motion and the predicted position of the attention. Variance is set to $\Sigma^{2}=diag(0.25)$ in order to cover a sufficiently wide area in the scene.}
\label{fig:442_camera_motion}
% \vspace{-0.4cm}
\end{figure*}

\subsubsection*{Feature extraction} 
Once we have computed optical flow and subtracted camera motion from motion vectors to detect moving objects, we compute two maps based on them. First, velocity or motion magnitude (M), which is calculated using the Euclidean or L2 norm as follows:

\begin{gather}
\mathcal{M}_{t} = ||{\bf v}_{t}||
\end{gather}

Then, acceleration (A), which is its absolute derivative:

\begin{gather}
\mathcal{A}_{t} = ||{\bf v}_{t}-{\bf v}_{t-1}||
\end{gather}

Examples of these two maps are shown in Figure \ref{fig:44_gaussian_features}.

Finally, camera motion may also influence viewers regarding a video. Indeed, as seeing in previous studies \cite{zygcameramotion2008}, observers tend to follow the camera motion direction to draw their attention to the new information and objects that emerge in the camera view.

As was defined above, $\xv_{n}=(x_{n},y_{n})$ is a \acs{2D} vector with spatial coordinates $x$ and $y$ associated with the spatial location $n$. Hence, the visual attention based on camera motion is modeled by means of a \acs{2D} Normal distribution over the spatial coordinates:

\begin{gather}
\mathcal{CM}_{t} \sim N(\cv \odot \uv_{t}, \Sigma), 
\end{gather}
 
\noindent where $\uv=(u_{x},u_{y})$ is the vector modeling the camera motion as a simple translation whose values are computed from a parametric affine motion model, as described above; $\odot$ stands for the Hadamard product between vectors, and $\cv=(c_{x},c_{y})$ is the vector of parameters that establishes a relation between the camera motion and the predicted position of the attention, and is learned during the training process. Figure \ref{fig:442_camera_motion} illustrates, given a sample value of $\uv$, examples of the distribution for several values of the vector $\cv$. Intuitively, higher absolute values of $\cv$ point to camera motion as an important feature for visual attention. If $\cv$ and $\uv$ have the same positive or negative sign, camera motion constitutes an attracting feature, and the \acs{2D} Gaussian distribution is shifted in its direction. In contrast, if they have the opposite sign, camera motion inhibits attention, and the distribution is shifted in the opposite direction. The second parameter $\Sigma$, which controls the spatial extent of the Gaussian distribution, has been empirically set to $\Sigma=diag(0.25)$ in order to cover a sufficiently wide area in the scene.

\subsection{Novelty features}\label{sec:novelty_features}
Those regions of the scene that continually change may also attract the attention of observers. In order to highlight them, novelty can be modeled by using the so-called \emph{coherence-based features}, which analyze the distribution of pixel values along space and time in order to detect areas where dispersion is large. To do this, we rely on the work done in \cite{6783970}, extracting \emph{spatial}, \emph{temporal} and \emph{spatio-temporal} coherence maps. In the following definitions, let us consider $f_{n}$ as the value of a given feature map at the location $n$ with spatial coordinates $\xv_{n}=(x_{n},y_{n})$, over which a coherence-based feature value is computed.

\begin{itemize}

\item Spatial Coherency (SC) identifies regions that belong to a well defined object, or where motion is stable, highlighting most changing regions, which can be more surprising and salient. For each pixel or spatial location $n$, it is calculated as the variance with respect to the mean $\mu_{n}$ of its neighbor values in a window of size $N \times N$, with $N=5$:

\begin{gather}
\mathcal{SC}_{n} = \frac{1}{N^2} \sum_{m \in R_{n}} (f_{m}-\mu_{n})^2
\end{gather}

\noindent where $R_{n}$ stands for the $N \times N$ neighborhood centered in the spatial location $n$.

\item Temporal Coherency (TC) is useful to distinguish between regions with small random motion (e.g. leaves falling from trees) and those with regular motion. Given a pixel $n$ in a current frame $t$, it is the variance with respect to the mean $\mu_{tn}$ of its value across the $T=7$ preceding frames, as the effect of motion in one frame on the scan path of the eye usually lasts for no more than $5-7$ frames \cite{abrams2003motion}:

\begin{gather}
\mathcal{TC}_{tn} = \frac{1}{T} \sum_{i \in [t-T+1,t]} (f_{in}-\mu_{tn})^2 
\end{gather}

\item Spatio-Temporal Coherency (STC) combines both previous measures and it is calculated for each pixel $n$ as the variance  with respect to the mean $\mu_{tn}$ of the set of pixel values within a $N \times N$ neighborhood, with $N=5$, across the $T=7$ preceding frames:

\begin{gather}
\mathcal{STC}_{tn} = \frac{1}{TN^2} \sum_{i \in [t-T+1,t]} \sum_{m \in R_{n}} (f_{im}-\mu_{tn})^2
\end{gather}

\end{itemize}

In total, 6 maps are computed: three over the pixel intensity values $I$ and three over the motion phase $\theta_{\mathcal{M}} = arctan \frac{v}{u}$. Examples of these maps are gathered in Figure \ref{fig:44_gaussian_features}.  

\subsection{Object-based features}\label{sec:object_features}
Despite the questionable conclusions of some psychologists \cite{WolfeHorowitz04,wolfe2014approaches} about the inclusion and modeling of faces and other semantic categories as attributes that guide attention, they have been considered in some computational approaches \cite{xu2014predicting, baluch2015mining}. Moreover, a recent analysis of Bylinskii et al. \cite{DBLP:conf/eccv/BylinskiiRBOTD16} gathers a list of under-predicted regions when estimating saliency in images, which mainly consists of parts of objects or subjects. 

Hence, we have considered detectors for some general-purpose objects that tend to attract visual attention, in order to appraise their utility in some contexts. In particular, cascade object detectors based on the Viola-Jones algorithm \cite{viola2001rapid} are used to detect people's \emph{frontal} (F) and \emph{profile faces} (PF), \emph{upper bodies} (B) and \emph{pedestrians} (P) and a detector working on the Harris corner response \cite{harris1988combined} is used to detect \emph{text} (T). Both methods are briefly explained here below. Many detectors for other visual concepts may also be included in our model without effort.

\subsubsection*{Cascade object detectors} 
Cascade classifiers \cite{viola2001rapid} are trained to detect objects with invariant aspect ratio at different scales. Therefore, in order to use them to locate an object whose appearance changes significantly when varying its orientation, such as in a face, it would be necessary to train a single detector for each of its views, as it is carried out in our approach by considering either frontal or profile faces. These kind of detectors are outstanding for being extremely fast, at the same time they can achieve high detection rates. 

A cascade detector involves several stages. Each of them constitutes an ensemble of weak learners, which are simple classifiers trained using AdaBoost \cite{freund1997decision}. Given a candidate window $x$ in an image, a weak classifier $h_{t}(x)$ is a threshold function that depends on a feature value $f_{t}(x)$. It can be formulated as follows:

\begin{gather}
h_{t}(x) = \begin{cases}
		-s_{t} &\quad\text{if } f_{t}(x) < \theta_{t} \\
		s_{t} &\quad\text{otherwise}
	\end{cases}
\end{gather} 

Simple features $f_{t}$ used, which are founded on the Haar filters introduced in \cite{papageorgiou1998general}, are calculated as the difference between the sum of the pixels within two sub-regions in $x$. The threshold $\theta_{t}$ and the polarity $s_{t} \in \pm 1$ are determined in the learning phase of the detector, on the basis of positive and negative samples of the object class for which it is trained. At each stage of the detector, a final strong decision is made as the weighted linear combination of the $T$ decisions made by all weak learners, being the weights $\alpha_{t}$ inversely proportional to their corresponding training errors:

\begin{gather}
h(x) = sgn \left( \sum_{t=1}^{T} \alpha_{t} h_{t}(x) \right)
\end{gather}

Under the assumption that the majority of regions of an image covered by a sliding window do not contain an object of interest, early stages of the cascade are designed to rapidly select the most promising sub-windows with a low false negative rate. Indeed, a cascade can be understood as a degenerate decision tree where, if a sub-window is rejected at any stage, no further processing is performed, dramatically decreasing the number of sub-windows to be evaluated. The complexity of the strong classifiers gradually increases until reaching the end of the cascade, with the purpose of achieving a final high detection rate. 

\begin{figure*}[t]
\centering
	\includegraphics[trim=0cm 0cm 0cm 0cm, width=1\textwidth]{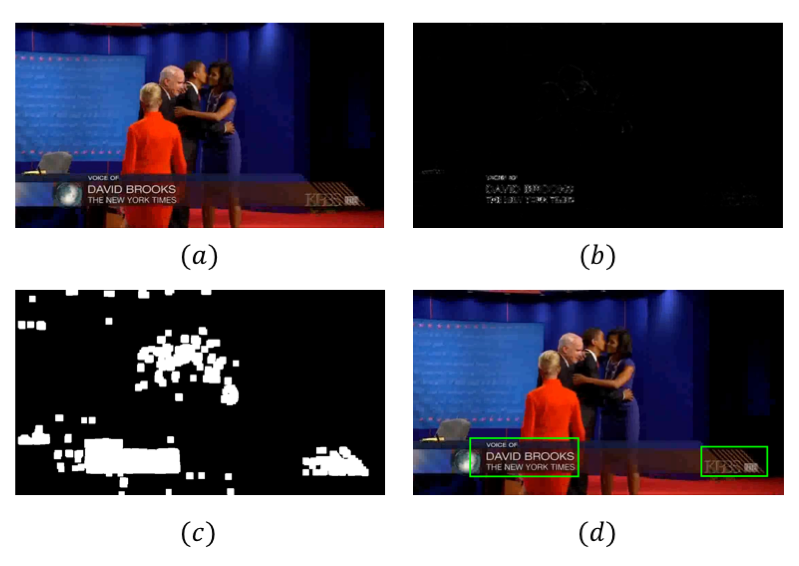}
\caption[Harris response for text detection in an example frame taken from a TV news video in the DIEM \cite{mital2011clustering} database.]{Harris response for text detection in an example frame taken from a TV News video in the DIEM \cite{mital2011clustering} database. (a) Example frame. (b) Absolute value of the Harris response computed in order to detect corners, which often correspond to text areas. (c) Binary mask obtained after applying to the response a non-maximum suppression operation consisting of a dilation. (d) Regions in the binary mask are filtered, selecting those which are horizontally or vertically aligned and occupy less than one third of the area of the whole image, which correspond to text bounding boxes.}\label{fig:334_text_detection}
% \vspace{-0.4cm}
\end{figure*}

\begin{figure}[!ht]
\centering
	\includegraphics[trim=0cm 0cm 0cm 0cm, width=0.95\textwidth]{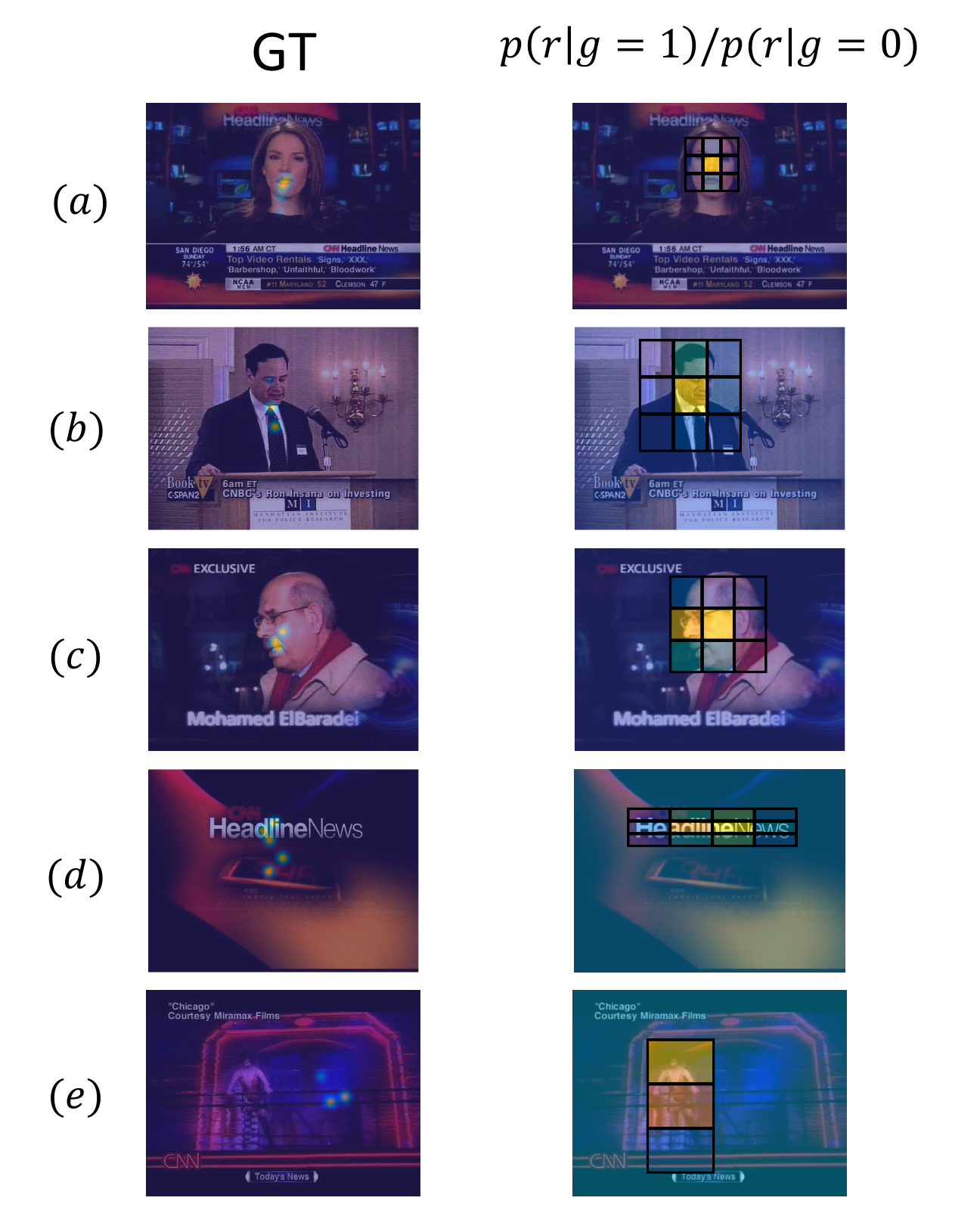}
\caption[Object-based feature maps computed for example frames taken from TVNews (a, b, c, d) and TalkShows (e) categories from CRCNS-ORIG \cite{Itti_Carmi09crcns} database.]{Object-based feature maps computed for example frames taken from TVNews (a, b, c, d) and TalkShows (e) categories from CRCNS-ORIG \cite{Itti_Carmi09crcns} database. (Left) Human fixations do not cover the whole object, but concentrate on particular areas/parts of the objects. (Right) Consequently, and based on the detected bounding box, we have divided the image into a set of subregions $r={0,1,...,R}$. Some of them ($r>0$) divide the object into several cells (9 for frontal (F) and profile faces (PF), and upper bodies (B); 3 for pedestrians (P) and 12 for text (T)). Moreover, an additional subregion $r=0$ is considered for the background, covering the rest of the image. Overlay heat maps highlight subregions where probabilities of each object for fixated points ($p(r|g=1)$, being $g \in \{0,1\}$ the ground truth variable indicating if the spatial location attracts or inhibits the attention) are substantially higher than those for non-fixated points ($p(r|g=0)$). Although the prior probability of objects is fairly lower than the probability of background in the database, it can be seen that objects are quite attractive for observers, due to the significant probability of internal cells given fixated locations.}
\label{fig:444_object_detector}
% \vspace{-0.4cm}
\end{figure} 

\subsubsection*{Harris response for text detection}
A simple text detector is proposed based on the commonly used Harris detector \cite{harris1988combined}, which locates corners in an image. Interest points in whose local neighborhood two dominant and different edge directions arise constitute corners, which often correspond to text areas. Corners are invariant to translation, rotation and illumination. 

In order to detect corners in a grayscale image $I$, the following second-moment matrix $M$ is computed, which is derived from its horizontal $I_{x}$ and vertical $I_{y}$ directional gradients:

\begin{gather}
M = \begin{bmatrix}
I_{x}I_{x} & I_{x}I_{y} \\
I_{x}I_{y} & I_{y}I_{y}  
\end{bmatrix}
\end{gather}

\noindent Then, the Harris response $HR$ is calculated as follows:

\begin{gather}
HR = det(M) -k \ tr(M)^2,
\end{gather}

\noindent where $k$ is a empirically determined constant. A non-maximum suppression operation consisting of a dilation is performed to find the local maxima in $HR$, which results in candidate text regions. Finally, we measure the area, orientation and eccentricity of the candidate regions, in order to obtain one or several bounding boxes. We assume that texts are usually horizontally or vertically aligned, and occupy less than one third of the area of the whole image. The complete process is illustrated in Figure \ref{fig:334_text_detection}.

\subsubsection*{Feature modeling} 
The output bounding boxes from the detectors described above are used to generate high level spatial feature maps. Visual attention usually points to particular locations within the objects, so this fact has to be considered when modeling these features. Since the size of the detected bounding boxes is often large, if we use a \acs{2D} Gaussian centered in the bounding box that contains a particular object, for instance, we are notably emphasizing the center of the object with respect to its surroundings. However, attention may be generally fixed at some elements of the object and not only at its center, such as in the case of faces or pedestrians, where subjects often look at the eyes or upper body part, respectively. Rather than directly considering the detected boxes as the feature maps, we have developed spatially-aware discrete distributions. 

As shown in Figure \ref{fig:444_object_detector}, given a detected bounding box, we consider a non-uniform grid with R+1 cells: $R$ cells subdivide the detected box into $r$ small subregions ($r>0$), and an additional subregion is considered for the background ($r=0$). Hence, for a given object $l$ being detected (we keep l as the index of the features, in this case object detections), we model a discrete distribution over the R+1 defined cells as $p(r | \beta_{l})$, where $r$ is a cell in the grid (which is object dependent), and $\beta_{l}$ are the parameters of the discrete distribution for the object category $l$. The distributions are then factorized for every object category and instance (in case that more than one object of a given category are detected on the same frame). By means of discrete spatial distributions that divide objects into several sub-regions, we are able to learn which parts of the object are more attractive, taking advantage of this knowledge to provide more accurate estimations of visual attention.

\section{Latent Topic Models}\label{sec:latent_topic_models}
As introduced in Section \ref{sec:machine_learning}, generative models not only make predictions on unseen data, but also offer an interesting interpretation about how this information was generated. Generative probabilistic \acsp{LTM} such as \acf{PLSA} \cite{hofmann2001unsupervised} or \acf{LDA} \cite{LDA} have, besides, an additional advantage: they can be used both in unsupervised and supervised contexts. 

This section primarily describes the \acs{LDA} graphical approach by David M. Blei et al. \cite{LDA}, which is the most frequently \acs{LTM} used, and two of its supervised extensions \cite{sLDA, DBAyang}, which are the basis of the contributions to visual attention understanding and estimation presented in this chapter. \acsp{LTM} are thought to model large collections of discrete data, such as corpus of texts, images or audio tracks. For this reason, we will begin by defining the \acf{BoW} notation, employed to organize hierarchically these entities.

\subsection{Bag-of-Words model}\label{sec:bow}
\begin{figure*}[!t]
\centering
	\includegraphics[trim=0cm 0cm 0cm 0cm, width=1\textwidth]{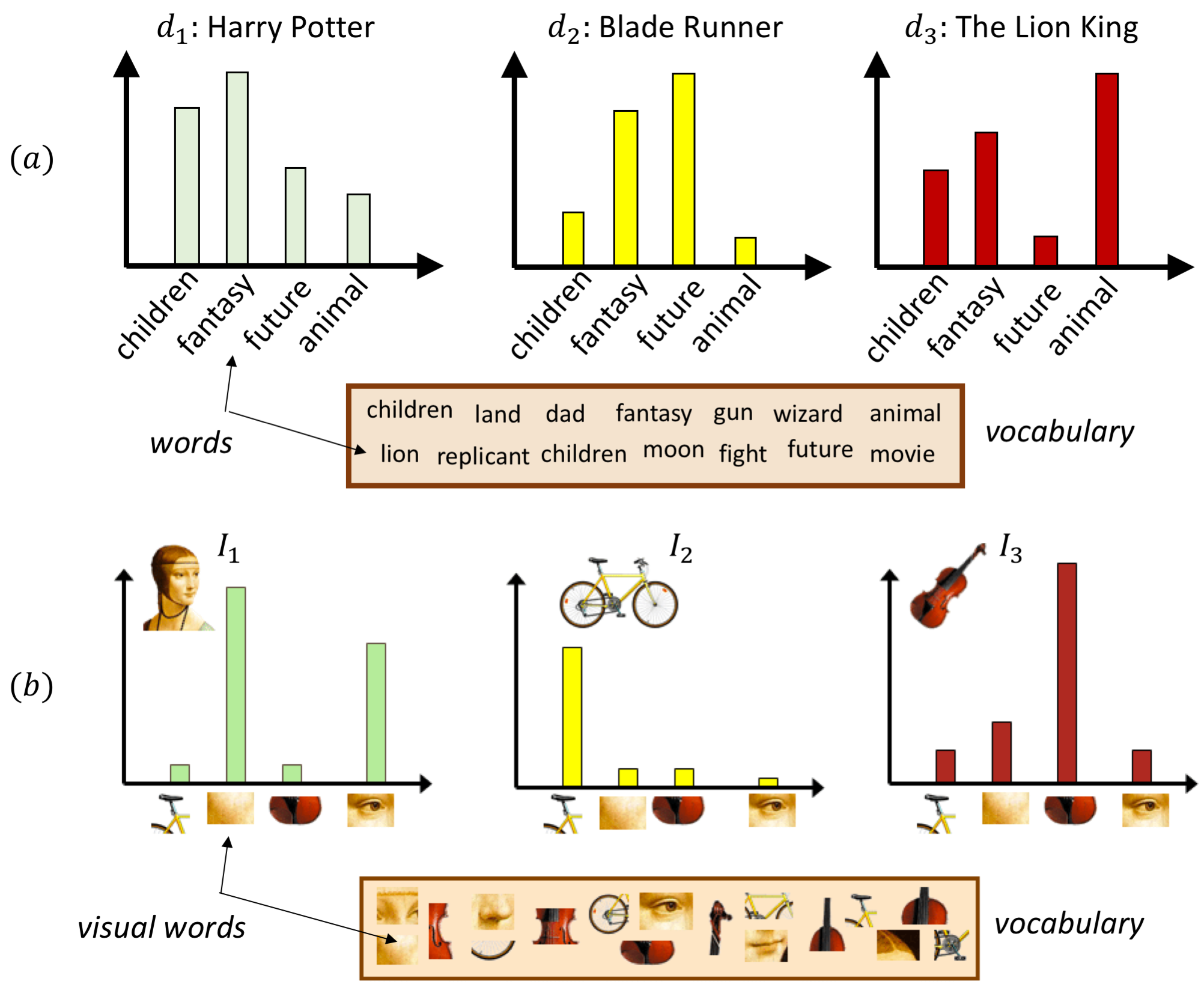}
\caption[\acf{BoW} model applied to: (a) A corpus of texts from movie reviews, (b) A corpus of images.]{\acf{BoW} model applied to: (a) A corpus of texts from movie reviews, where each review $\{\mathbf{d}_{1},\mathbf{d}_{2},\mathbf{d}_{3}\}$ is represented by means of a histogram of word occurrences; (b) A corpus of images, where each image $\{\mathbf{I}_{1},\mathbf{I}_{2},\mathbf{I}_{3}\}$ corresponds to a histogram of visual word occurrences. In both cases, words are taken from a finite vocabulary. Figure adapted from \cite{feifei09_course}.}
\label{fig:331_bow}
% \vspace{-0.4cm}
\end{figure*}
The \acf{BoW} model \cite{sivic2009efficient, LDA} is a hierarchical representation method classically used in \acf{NLP} and text categorization, which was later extended to image recognition and retrieval in the computer vision field. Given a large collection of texts, the \acs{BoW} model defines the following terms, which help to understand the intuition behind the \acs{LDA} approach:

\begin{itemize}
\item A \emph{word} $\mathbf{w}$ is an item from a finite vocabulary, and constitutes the basic unit of discrete data. 

\item A \emph{vocabulary} $\{\mathbf{w}_{1},\mathbf{w}_{2},...,\mathbf{w}_{V}\}$ is a finite collection of $V$ words. Each word $\mathbf{w}_{v}$ in the vocabulary is represented by a $V$-vector with a $1$ at the position $w^v$ of the word in the vocabulary ($w^v=1$) and $0$ everywhere else ($w^u=0$ for all $u \neq v$).

\item A \emph{document} $\mathbf{d}=(\mathbf{w}_{1},\mathbf{w}_{2},...,\mathbf{w}_{N_{d}})$ consists of a sequence of $N_{d}$ words.

\item A \emph{corpus} $\mathcal{D} = \{\mathbf{d}_{1},\mathbf{d}_{2},...,\mathbf{d}_{D}\}$ is a collection of $D$ documents. 
\end{itemize}

Figure \ref{fig:331_bow} shows two examples of application of the \acs{BoW} model in text and image corpora. Similarly to a document, an \emph{image} $\mathbf{I}$ is decomposed into a set of \emph{keypoints} represented by means of $N_{I}$ \emph{visual descriptors} $(\mathbf{w}_{1},\mathbf{w}_{2},...,\mathbf{w}_{N_{I}})$, which are associated with a finite \emph{vocabulary} of $V$ \emph{visual words}.

\subsection{Latent Dirichlet Allocation}\label{sec:lda}
\begin{figure*}[!t]
\centering
	\includegraphics[trim=0cm 0cm 0cm 0cm, width=1\textwidth]{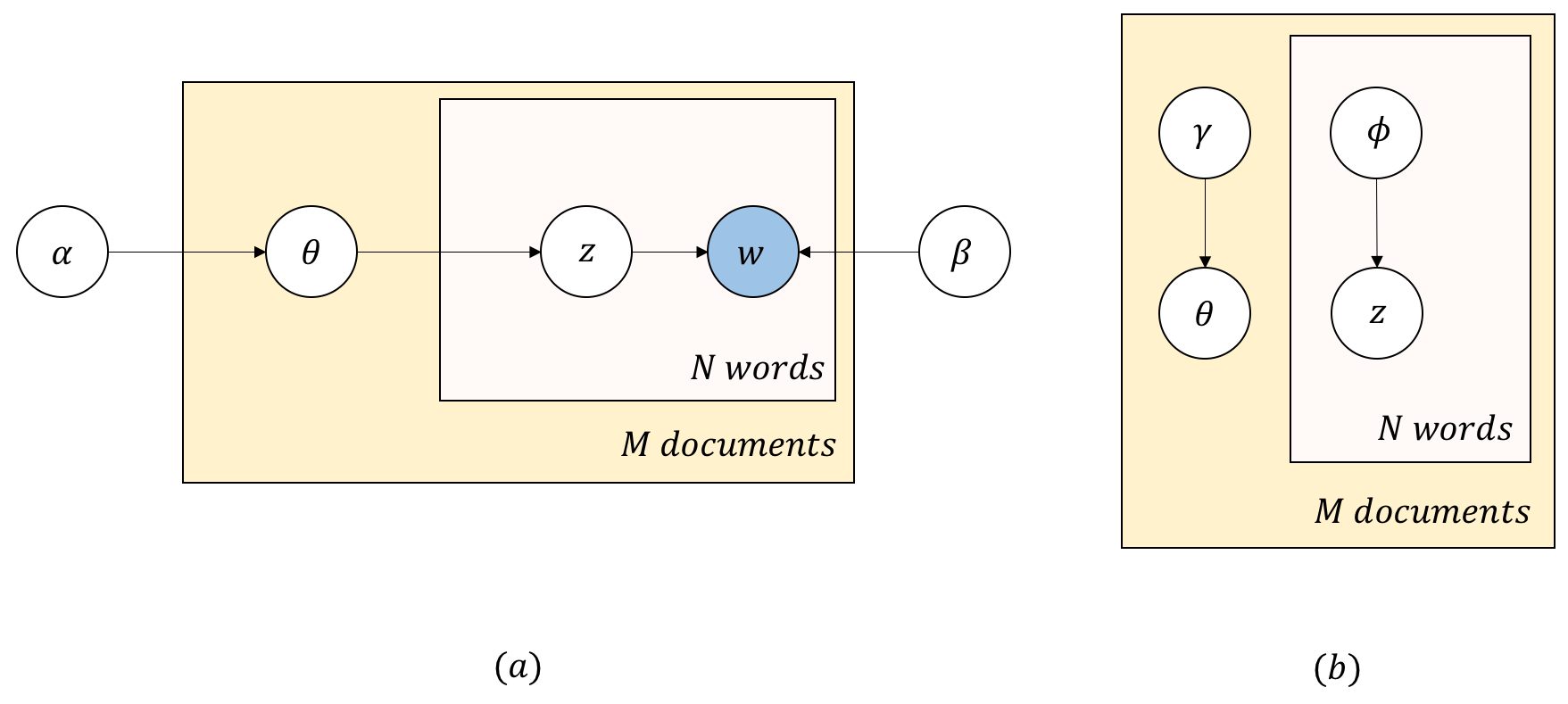}
\caption[(a) Graphical representation of \acf{LDA} \cite{LDA}. (b) Graphical representation of the variational distribution used to approximate the posterior in \acs{LDA}.]{(a) Graphical representation of \acs{LDA} \cite{LDA}. (b) Graphical representation of the variational distribution used to approximate the posterior in \acs{LDA}. Shaded nodes represent observed variables, while white nodes denote latent variables to be inferred. Boxes mean independent repetitions, and edges show the dependencies among variables.}
\label{32_lda}
% \vspace{-0.4cm}
\end{figure*}

\acf{LDA} \cite{LDA} is a hierarchical Bayesian method initially conceived to model large collections of discrete data, such as text documents. Given a \emph{corpus} of $D$ documents, the model provides an explicit representation of each \emph{document} $\mathbf{d} \in \mathcal{D}$ as a finite mixture over an underlying set of latent topics which, in turn, are modeled by a distribution over words. In fact, a document is defined as a sequence of $N_{d}$ words denoted by $\mathbf{d} = (\mathbf{w}_{1},\mathbf{w}_{2},...,\mathbf{w}_{N_{d}})$, being a \emph{word} an item from a finite vocabulary $\{\mathbf{w}_{1},\mathbf{w}_{2},...\mathbf{w}_{V}\}$.

As can be appreciated in the graphical representation of the model shown in Figure \ref{32_lda}(a), \acs{LDA} establishes a three-level representation hierarchy. The model first assumes a known and fixed number of topics $K$ in the corpus $\{\mathbf{z}_{1},\mathbf{z}_{2},...,\mathbf{z}_{K}\}$. At corpus-level, the $K$-dimensional Dirichlet variable $\alpha$ sets the global distribution of the topics or topic proportions $\{\alpha_{1},\alpha_{2},...,\alpha_{K}\}$, being $\alpha_{k}>0$, in the whole corpus. In addition, $\beta$ includes a collection of $K$ $V$-dimensional discrete or multinoulli variables with the probabilities of each word in the vocabulary. Then, at document-level, the variable $\theta_{\mathbf{d}}$ represents the particular topic proportions in each document $\mathbf{d}$. Finally, at word-level, the variable $\mathbf{z}_{\mathbf{d}n}$ stands for the topic associated with each word $\mathbf{w}_{\mathbf{d}n}$ in each document $\mathbf{d}$. ${\bf z_{\mathbf{d}n}}$ is defined as a $K$-vector with a $1$ at the position of the topic assigned and $0$ everywhere else.

Hence, for each document $\mathbf{d}$ in a corpus $\mathcal{D}$, \acs{LDA} involves the following generative process. For the sake of simplicity, let us note that we have omitted the document subindex $d$ in those document-dependent variables:

\begin{enumerate}
  \item Draw the document particular proportions $\theta$ of $K$ topics using a corpus-level Dirichlet distribution of parameter $\alpha$: $\theta | \alpha \sim Dir(\alpha)$.
  \item For each word ${\bf w}_{n} \in N_{d}$ in the document $\mathbf{d}$:
  \begin{enumerate}
    \item Draw topic assignment $p({\bf z_{n}}|\theta)$ using a multinomial distribution over the topic proportions $\theta$: ${\bf z_{n}} | \theta \sim Mult(\theta)$.
    \item Draw a word ${\bf w_{n}}$ using $p({\bf w_{n}}|{\bf z_{n}},\beta)$, which is a multinomial probability conditioned on the topic ${\bf z_{n}}$.
  \end{enumerate}
\end{enumerate}

Given a document $\mathbf{d}$ in the corpus and the corpus-level parameters $\alpha$ and $\beta$, the joint distribution of a topic mixture $\theta$, a set of $K$ topics $\mathbf{z}$ and a set of $N_{d}$ words $\mathbf{w}$ is expressed as follows:

\begin{gather}
p(\theta,\mathbf{z},\mathbf{w}|\alpha,\beta) = p(\theta|\alpha) \prod_{n=1}^{N_{d}} p(\mathbf{z}_{n}|\theta)p(\mathbf{w}_{n}|\mathbf{z}_{n},\beta).
\end{gather}    

In order to apply the \acs{LDA} method, we have to compute the posterior distribution of the latent variables $\theta, \mathbf{z}$ given a document $\mathbf{d}$: 

\begin{gather}
p(\theta,\mathbf{z}|\mathbf{w},\alpha,\beta) = \frac{p(\theta,\mathbf{z},\mathbf{w}|\alpha,\beta)}{p(\mathbf{w}|\alpha,\beta)}.
\end{gather}

This distribution, nevertheless, is intractable for exact inference due to the coupling between $\theta$ and $\beta$, so it is necessary to consider an approximate algorithm such as the convexity-based variational inference proposed in \cite{jordan1999introduction}. The idea is to make use of Jensen's inequality to arise an adjustable lower bound on the log likelihood, drawing on some variational parameters. These parameters are estimated via an optimization process that tries to find the tightest possible lower bound. 

As shown in Figure \ref{32_lda}(b), by dropping the edges between $\theta$, $\mathbf{z}$ and $\mathbf{w}$, and also the $\mathbf{w}$ nodes, we achieve the following variational distribution:

\begin{gather}
q(\theta,\mathbf{z}|\gamma,\phi) = q(\theta|\gamma) \prod_{n=1}^{N_{d}} q(\mathbf{z}_{n}|\phi_{n}),
\end{gather}

\noindent being the Dirichlet parameter $\gamma$ and the multinomial parameters $(\phi_{1},...,\phi_{N})$ the new free variational parameters.

The optimization problem to find the tightest lower bound to the posterior consists in minimizing the \acf{KL} between the variational distribution and the true posterior $p(\theta,\mathbf{z}|\mathbf{w},\alpha,\beta)$, and is defined as:

\begin{gather}\label{eq:lda_opt}
(\gamma*,\phi*) = \argmin_{(\gamma,\phi)} D(q(\theta,\mathbf{z}|\gamma,\phi) || p(\theta,\mathbf{z}|\mathbf{w},\alpha,\beta))
\end{gather}

The problem can be solved by means of a variational \acf{EM} algorithm, which involves the following iterative procedure:

\begin{enumerate}
	\item \emph{E-step}: For each document $\mathbf{d}$, find the optimum values of the variational parameters $\{\gamma^{*}_{\mathbf{d}},\phi^{*}_{\mathbf{d}}\}$.
	\item \emph{M-step}: Maximize the obtained lower bound on the log-likelihood with respect to the parameters of the model $\alpha$ and $\beta$. 
\end{enumerate}

\subsection{Supervised topic models}
\begin{figure*}[!t]
\centering
	\includegraphics[trim=0cm 0cm 0cm 0cm, width=0.75\textwidth]{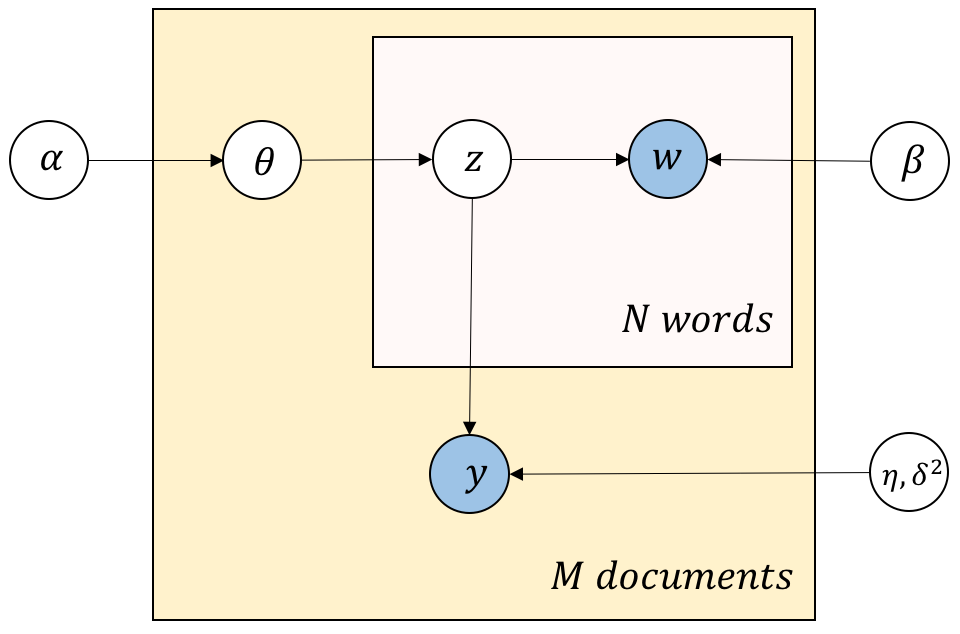}
\caption[Graphical representation of \acf{sLDA} \cite{sLDA}.]{Graphical representation of \acs{sLDA} \cite{sLDA}, which incorporates to \acs{LDA} \cite{LDA} a response variable $y$, modeled using a lineal regression model with parameters $\{\eta, \delta^2\}$. In the case of the \acs{DBA} \cite{DBAyang} extension of \acs{LDA}, the representation is similar except for the parameter of the logistic regression model, which is simply $\eta$. Shaded nodes represent observed variables, while white nodes denote latent variables to be inferred. Boxes mean independent repetitions, and edges show the dependencies among variables.}
\label{fig:32_slda}
% \vspace{-0.4cm}
\end{figure*}

So far, we have discussed about the advantages of generative models and the ability of \acsp{LTM} to represent texts as a mixture over topics, which can be inferred from a large collection of documents. However, the topics discovered by algorithms like \acs{LDA} are implicit, so that human expertise is required to arise a comprehensible interpretation of their semantics (e.g. to relate a topic with high probabilities of terms ``match'', ``players'', ``games'', ``ball'', with the semantic concept of ``sport'').

The objective of supervised extensions of \acs{LDA} described in the following paragraphs is thus to infer latent topics that not only explain the distribution of words in documents, but also serve to automatically predict a response variable. Both approaches consider response variables $y$ at document-level, as shown in the graphical representation of Figure \ref{fig:32_slda}.

\subsubsection*{Supervised Latent Dirichlet Allocation}
\acf{sLDA} \cite{sLDA} incorporates to \acs{LDA} a continuous response variable $y$ associated with each document. The documents and the responses are jointly modeled in order to find latent topics that predict the response variable of new unlabeled documents. Examples of applications of \acs{sLDA} based on \acs{NLP} include the prediction of the numerical rating of a movie review or the number of visits of a website depending on their content. 

Figure \ref{fig:32_slda}(a) shows the graphical representation of \acs{sLDA}. For each document $\mathbf{d}$ in a corpus $\mathcal{D}$, the model involves the following generative process. For the sake of simplicity, we have omitted again the document subindex $d$ in document-dependent variables:

\begin{enumerate}
  \item Draw the document particular proportions $\theta$ of $K$ topics using a corpus-level Dirichlet distribution of parameter $\alpha$: $\theta | \alpha \sim Dir(\alpha)$.
  \item For each word $\mathbf{w}_{n} \in N_{d}$ in the document $\mathbf{d}$:
  \begin{enumerate}
    \item Draw topic assignment $p({\bf z_{n}}|\theta)$ using a multinomial distribution over the topic proportions $\theta$: ${\bf z_{n}} | \theta \sim Mult(\theta)$.
    \item Draw a word ${\bf w_{n}}$ using $p({\bf w_{n}}|{\bf z_{n}},\beta)$, which is a multinomial probability conditioned on the topic ${\bf z_{n}}$.
  \end{enumerate}
  \item Draw a Gaussian response variable $y | {\bf z_{1:N}},\eta,\delta^2 \sim N(\eta^T{\bf \overline{z}}, \delta^2)$, based on a linear regression model. 
\end{enumerate}

Indeed, in addition to the original \acs{LDA}, a third step is included at document-level corresponding to the introduced response variable $y$, which is modeled using a normal lineal regression model $N(\eta^T{\bf \overline{z}}, \delta^2)$, where ${\bf \overline{z}} := (1/N)\sum_{n=1}^{N} {\bf z_{n}}$ represents the empirical frequencies of the topics in the document, $\eta$ is a $L$-length vector containing the regression coefficients of the response variable, and $\delta^{2}$ is a dispersion parameter, which provides certain flexibility when modeling the variance of $y$. 

The posterior distribution to solve \acs{sLDA} is $p(\theta,\mathbf{z}|\mathbf{w},y,\alpha,\beta,\eta,\delta^2)$, and can be again approximated by means of the \acs{KL} optimization problem defined in Eq. \ref{eq:lda_opt}. Now, at the M-step of the algorithm, we also maximize the lower bound on the log-likelihood with respect to the supervision parameters $\eta$ and $\delta^2$. 

\subsubsection*{Dirichlet-Bernoulli Alignment}
\acf{DBA} \cite{DBAyang} presents an alternative supervised extension to \acs{LDA}, with the purpose of considering multi-class, multi-label and multi-instance classification tasks, where each document from a corpus consists of multiple instances and is related to multiple classes. For instance, a \acs{NLP} application of \acs{DBA} could be, given a social network profile (\emph{document}), the classification of its publications into a set of categories (\emph{topics}). 

Hence, each document is modeled as a mixture over a set of predefined classes. Then, each word is generated independently conditioned on the sampled class, and the label of the document is generated conditioned on all the sampled labels used for generating its words.  

The graphical representation of \acs{DBA} is similar to the one presented in Figure \ref{fig:32_slda} for \acs{sLDA}. However, in contrast to \acs{sLDA}, \acs{DBA} response variable is not continuous but categorical; therefore, a multinomial logistic regression model given by a Bernoulli or softmax distribution $y | {\bf z_{1:N}},\eta \sim Be\left(\frac{\exp(\eta^T\overline{{\bf z}})}{1+exp(\eta^T\overline{{\bf z}})}\right)$ automatically aligns the topics discovered from the data to the predefined classes. 

The posterior distribution to solve \acs{DBA} is $p(\theta,\mathbf{z}|\mathbf{w},y,\alpha,\beta,\eta)$, and can be again approximated by means of the \acs{KL} optimization problem defined in Eq. \ref{eq:lda_opt}. Now, at the M-step of the algorithm, we also maximize the lower bound on the log-likelihood with respect to the supervision parameter $\eta$.  

\subsection{Applications to Computer Vision}
The description of the \acs{LTM} methods discussed in the previous sections has been focused on the analysis of corpus of texts. However, \acs{LTM} models have been also applied to other types of data \cite{blei2012probabilistic}, such as audio and music \cite{kim2009acoustic} or population genetics \cite{pritchard2000inference}, among others. 

What is more, they have been widely-used in computer vision for image retrieval \cite{fei2005bayesian}, segmentation \cite{GONZALEZDIAZ20132437} and captioning \cite{blei2003modeling}. In video processing, they have been applied for action recognition \cite{wang2009unsupervised}. In most of these applications, image features are quantized into discrete values so that they take the form of words in documents. However, this is not a straightforward step and requires computing dictionaries of visual words \cite{zhang2009descriptive,zhang2010building}.

Notwithstanding this, to our knowledge, the \acs{LTM} presented in the next section is the first approach proposed for spatio-temporal visual attention on the basis of this type of graphical models.

\section{Visual Attention Topic Model}\label{sec:sub_atom}
In this section, we describe in detail the system proposed for spatio-temporal visual attention understanding and prediction, which we have called \acf{ATOM}. 

% \clearpage
\begin{figure*}[!htb]
\centering
	\subfloat{%
       \includegraphics[trim=4cm 1cm 3.5cm 0cm, width=1\textwidth]{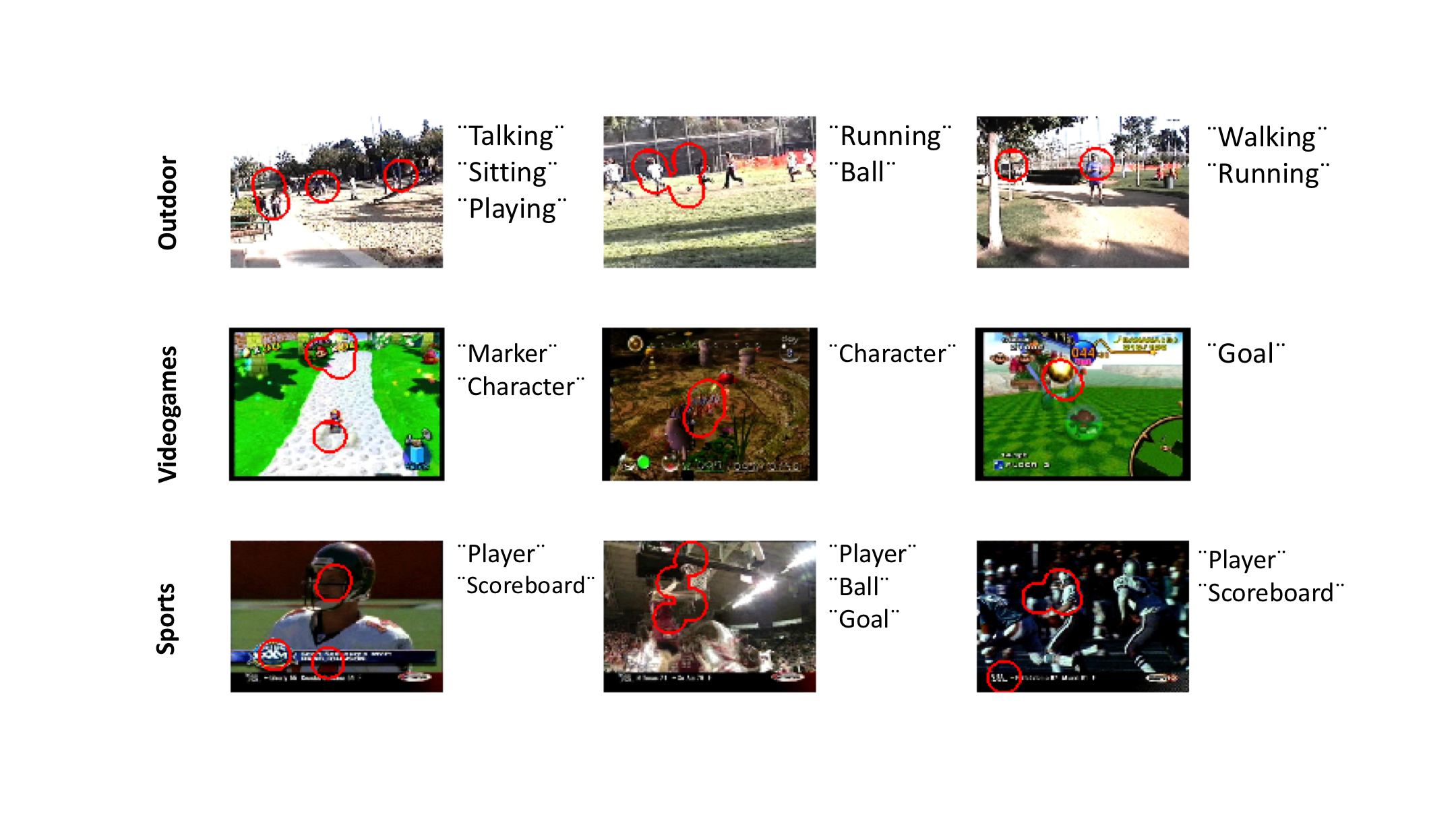}}
    \label{44_atom_examples1}
  	\subfloat{%
        \includegraphics[trim=3.5cm 1cm 3cm 3cm, width=0.8\textwidth]{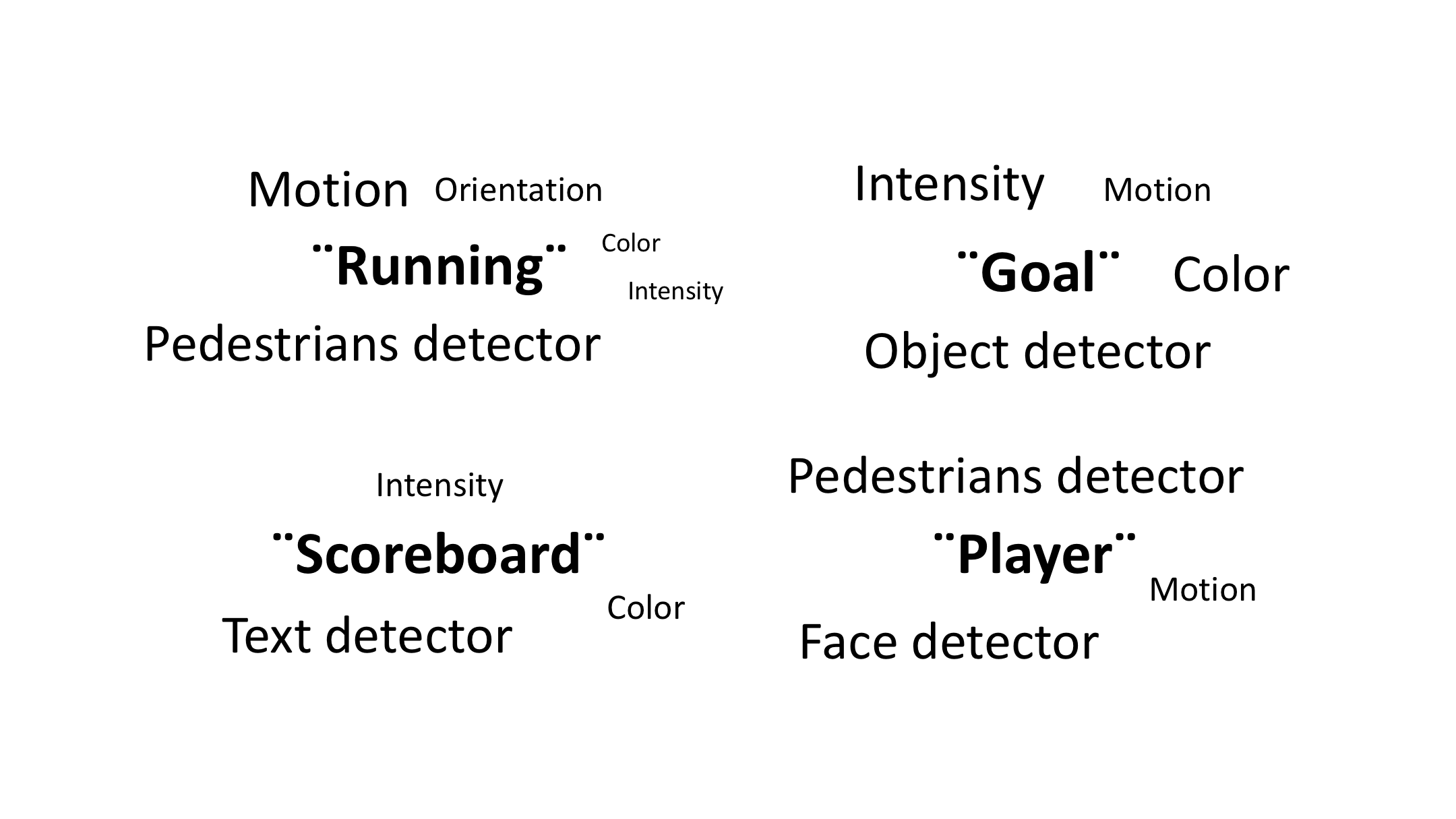}}
    \label{44_atom_examples2}
\caption[Visual attention modeled in three different scenarios taken from CRCNS-ORIG \cite{Itti_Carmi09crcns} database as a mixture of several relevant sub-tasks, associated with particular areas of special importance for observers.]{Visual attention modeled in three different scenarios taken from CRCNS-ORIG \cite{Itti_Carmi09crcns} database (\emph{Outdoor}, \emph{Videogames} and \emph{Sports}) as a mixture of several relevant sub-tasks (e.g. ``Running'', ``Goal'', ``Player'', etc.), associated with particular areas of special importance for observers, which are highlighted in the example frames on the left side. Some of them may appear similarly in different contexts, such as ``Ball'' or ``Goal''. On the bottom of the figure, word clouds show how some sub-tasks (bold central words) are represented as a combination of features (surrounding words: intensity, motion, detectors, etc.). Feature importance, represented by the font size of each word showing a feature, varies from one sub-task to another. For example, motion information and pedestrian detections are more relevant for ``Running''; in contrast, an object detector, along with intensity and color features are more advantageous to represent a ``Goal'' in a videogame.}
\label{44_atom_examples}
\end{figure*}

\subsection{Model overview}\label{sec:overview4}
\acs{ATOM} generative model is supported by the following assumption \cite{cbmi2016miguel}: 

\begin{center} 
\emph{Task- or context-driven visual attention in video can be modeled as a mixture of several sub-tasks which, in turn, can be represented as combinations of low-, mid- and high-level spatio-temporal features obtained from video frames}.
\end{center} 

The generative model thus receives as input a set of visual features, which are used to learn several related sub-tasks. These sub-tasks automatically lead the attention of the system to the most appealing areas of a scene. Depending on the scenario, visual attention may be attracted by different events. Our goal is not to detect these events of interest for a particular application, but to efficiently guide the later processing to areas of special importance in the video.

Figure \ref{44_atom_examples} illustrates our hypothesis for three different scenarios in CRCNS-ORIG \cite{Itti_Carmi09crcns} database. First, looking at the contexts given, visual attention may be attracted by different events or elements in the scene: people \emph{running} and \emph{walking} in the case of \emph{Outdoor}; \emph{game character} and \emph{goals} or \emph{items} in \emph{Videogames}; and \emph{players} and \emph{scoreboards} in \emph{Sports}. Note that some contexts may share similar attractions, like \emph{ball}, which is present both on \emph{Outdoor} and \emph{Sports} videos. Our goal is to automatically discover sub-tasks that guide later processing to the areas where those occur. In turn, these sub-tasks can be modeled as combinations of spatio-temporal features. For instance, the use of a motion feature combined with a detected face or pedestrian could be useful to represent the sub-task ``Player''. In contrast, the sub-task ``Scoreboard'' is well-defined by some intensity or color features, together with a detected text.  

Probabilistic \acfp{LTM}, which have been commonly used to extract hidden semantic structures (\emph{latent topics}) from a text corpus, can be helpful to unsupervisely understand large amounts of information, such as the human perception features that are quickly and parallely processed by the brain. Our approaches involve thus a \acs{LTM} which relies on the well-known \acs{LDA} algorithm \cite{LDA} and its supervised extension \acs{DBA} \cite{DBAyang}.

First, by understanding frames as a mixtures over topics, \acs{LDA} allows to interpret them using unsupervised statistical distributions, which associate each frame to multiple topics with different proportions. In our particular scenario, task-driven visual attention is modeled as a finite mixture over a set of $K$ topics, which represent the sub-tasks contributing to model visual attention, either by attracting or by inhibiting it. Note that both terms, topics and sub-tasks, are used interchangeably along the thesis. In parallel, for a given video frame $I_{t}$, a set of $L$ visual descriptors $\bf f$ $={\{f_{1}, f_{2}, ..., f_{L}\}}$ is computed at each spatial location $n$, so that the latent topics are in turn modeled as combinations of these features. 

The original \acs{LDA} is completely unsupervised, so that the topics are learned to maximize the likelihood of a corpus, and requires of human knowledge to align topics and semantic concepts. In our case, in contrast, we aim to learn how humans guide their attention to visual stimuli, so that the \acf{GT} fixations provided by different subjects will drive our training step. Visual attention is thus estimated by means of a logistic regression model over the topic assignments. This logistic regression is in charge of aligning the topics discovered from frames to the information gathered in \acs{GT} binary fixation maps. Hence, our final model draws on the \acs{DBA} introduced in \cite{DBAyang}. Let us note that the latent nature of the topics remains unchanged in our supervised models, as the human fixations used in the training phase are not supervising the topics but, instead, the binary response variable learned by the logistic regression. 

\begin{figure*}[!t]
\centering
	\includegraphics[trim=0cm 0cm 0cm 0cm, width=1\textwidth]{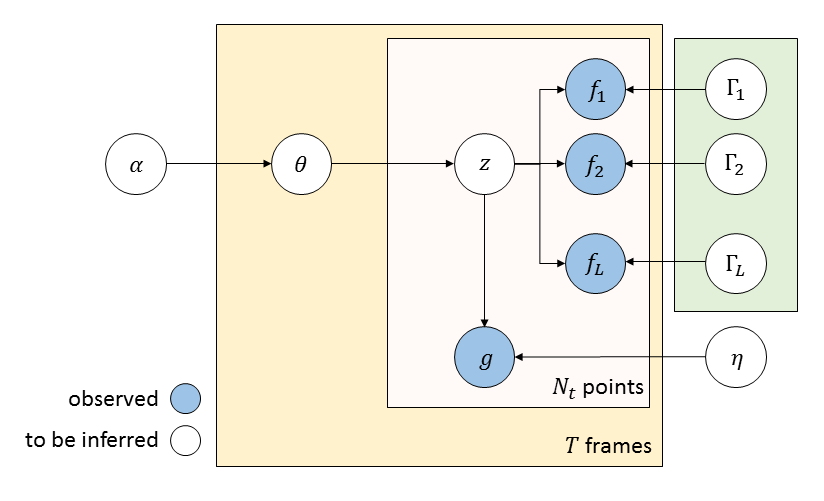}
\caption[Graphical representation of the proposed \acf{ATOM} generative model.]{Graphical representation of the proposed \acf{ATOM} generative model. Shaded nodes represent observations from frames, white nodes indicate hidden variables to be inferred, and boxes mean independent repetitions. Edges show the dependencies among variables.}
\label{44_atom}
% \vspace{-0.4cm}
\end{figure*}

The graphical representation of the model is shown in Figure \ref{44_atom}. In the same way as \acs{LDA}, \acs{ATOM} establishes a three-level representation hierarchy. The model first assumes a known and fixed number of latent topics $K$ in the video corpus $\{\mathbf{z}_{1},\mathbf{z}_{2},...,\mathbf{z}_{K}\}$, which represent the sub-tasks that contribute to model visual attention. Let us note that some of these sub-tasks may attract human attention whereas others may inhibit it. At video corpus-level, the $K$-dimensional Dirichlet variable $\alpha$ sets the global distribution of the sub-tasks or sub-task proportions $\{\alpha_{1},\alpha_{2},...,\alpha_{K}\}$, being $\alpha_{k}>0$, in the whole video corpus. High values of all components $\alpha_{k}$ of the variable $\alpha$ result in mixtures where all sub-tasks are considered to estimate visual attention in every video frame. In contrast, low values of only some $\alpha_{k}$ provide more particular mixtures of sub-tasks for each frame, being the attention determined by only few prevailing sub-tasks. Moreover, $\Gamma$ includes a collection of $K$ $L$-dimensional variables which define the distribution of each feature $l$ given the topic $k$. Depending on the nature of features, it is possible to model them using the most suitable distribution: e.g. normal, exponential, discrete, etc. Then, at frame-level, the variable $\theta$ represents the particular sub-task proportions in each frame $I_{t}$. Finally, at spatial location-level, the variable $\mathbf{z}_{n}$ stands for the sub-task associated with each spatial location $n$ in each frame $I_{t}$. ${\bf z}_n$ is an indexing $K$-dimensional vector with all zeros in except of a 1 in the position of the selected topic.

The proposed \acs{ATOM} thus involves the following generative process for each frame $I_t$ in a video corpus $\mathcal{I} = {\{I_{1}, I_{2}, ..., I_{T}\}}$. Let us note that, for simplicity, we have removed the sub-index $t$ of the frame in the notation:

\begin{enumerate}
  \item Draw the frame particular proportions $\theta$ of $K$ topics using a corpus-level Dirichlet distribution of parameter $\alpha$: $\theta | \alpha \sim Dir(\alpha)$.
  \item For each spatial location $n \in N$ in the frame $I_{t}$:
  \begin{enumerate}
    \item Draw topic assignment using a multinomial distribution over the topic proportions $\theta$: ${\bf z_{n}} | \theta \sim Mult(\theta)$.
    \item Represent the local appearance of the spatial location $n$ by drawing $L$ independent visual features $f_{ln}$ using the topic particular distributions $p(f_{ln} | {\bf z_{n}}, \Gamma)$, where $\Gamma$ includes the parameters of the distributions of the $L$ features, given the selected topic ${\bf z_n}$.
	\item Draw the binary response variable $g_n$ modeling the visual attention using a logistic regression model given by the following Bernoulli distribution: \\ $g_n | {\bf z_n}, \eta \sim Be\left(\frac{\exp{(g_n \eta^{T} {\bf z_n}})}{1+\exp{(\eta^{T} {\bf z_n}})}\right)$, where $\eta$ is the parameter vector that models attention based on the selected topic ${\bf z_n}$.
  \end{enumerate}
\end{enumerate}

Hence, for each frame $I_{t}$, we first generate a particular mixture of these topics $\theta$ based on the distribution with the global topic proportions $\alpha$. Once $\theta$ is known, we analyze the different spatial locations of the frame such that, for each $n$, we first select a sub-task by using the index-variable $\bf z_n$. Based on $\bf z_n$, we draw the local appearance of the spatial location using the particular feature-topic distribution $f_{nl} | {\bf z_{n}}, \Gamma$, where $\Gamma$ stands for the parameters of the distributions of the $L$ features considered. Sub-task is thus chosen so that its corresponding distribution parameters are the ones that maximize the likelihood of the visual features observed at this location. 

For the sake of simplicity, we assume that $p({\bf z_n}|\theta)$ is independent for all locations $n$, which makes the solution tractable, both simplifying the definition of the algorithm and, at the same time, improving the system efficiency. In contrast, other approaches such as \acsp{MRF} \cite{KATO20061103}, applied to image segmentation, are able to capture such spatial constrains. Nonetheless, it should be noted that some of the visual features that we extract for each sampled location (e.g. color, intensity, orientation, \acsp{CNN}-based) consider beforehand this spatial dependency. Moreover, we assume conditional independence among the $L$ features, so that the joint distribution of features for a particular topic can be factorized into the individual probability distributions $p(f_l|{\bf z},\Gamma)$. Finally, we also generate the attention response $g_n$ by computing the logistic regression model over the selected topics. 

\subsection{Guiding features extraction}\label{sec:guiding_features}
Motivated by the general conclusions of psychological theories about attention \cite{TreismanGelade80,wolfe1994guided}, the general hierarchical probabilistic framework presented may operate over a great number of diverse features. Depending on their nature, they may be modeled using various probability distributions: e.g. \emph{normal}, \emph{exponential}, \emph{discrete}, etc. It should be remarked that our model is not feature-dependent, so that any kind of feature can be incorporated by selecting the appropriate distribution. Furthermore, for each application scenario and based on human fixations, our model will automatically discover which particular features are more and less discriminant to model attention and correspondingly assign appropriate parameters to their distributions. Hence, one could include a broad general set of features as the model will automatically reduce or neglect the influence of those that do not guide the attention in a particular context. 

Hereunder is a list of the features extracted for our experiments in Chapter \ref{ch:atom_experiments}, which correspond to 24 feature maps, including the section where they were explained. Some of the feature maps are handcrafted and allow us to perform a meaningful interpretation of the estimated visual attention. They carry continuous values, and are modeled using a Gaussian probability density function:

\vspace{0.5cm}

\emph{Basic features} (Section \ref{sec:basic_features})
\begin{enumerate}
	\item \emph{Color} (C) 
	\item \emph{Intensity} (I) 
	\item \emph{Orientation} (O) 
\end{enumerate}

\vspace{0.5cm}

\emph{Motion-based features} (Section \ref{sec:motion_features})
\begin{enumerate}
	\setcounter{enumi}{3}
	\item \emph{Velocity or motion magnitude} (M)
	\item \emph{Acceleration} (A) 
\end{enumerate}

\vspace{0.5cm}

\emph{Novelty features} (Section \ref{sec:novelty_features})
\begin{enumerate}
	\setcounter{enumi}{5}
	\item \emph{Spatial Coherency (Luminance)} (SC (Lum.)) 
	\item \emph{Spatial Coherency (Motion)} (SC (Mot.)) 
	\item \emph{Temporal Coherency (Luminance)} (TC (Lum.)) 
	\item \emph{Temporal Coherency (Motion)} (TC (Mot.)) 
	\item \emph{Spatio-Temporal Coherency (Luminance)} (STC (Lum.)) 
	\item \emph{Spatio-Temporal Coherency (Motion)} (STC (Mot.)) 
\end{enumerate}

\vspace{0.5cm}

Then, camera motion is modeled as a multivariate Gaussian $\mathcal{CM} \sim N(\cv \odot \uv, \Sigma)$, as described in Section \ref{sec:motion_features}. Due to the diagonal nature of the covariance matrix $\Sigma$, we can decompose it into two independent univariate Gaussians (feature maps 12 and 13).

\vspace{0.5cm}

Next, we have used some object detectors in order to compute high-level spatial feature maps, which are modeled by means of discrete spatial distributions, as explained in Section \ref{sec:object_features}:

\vspace{0.5cm}

\begin{enumerate}
	\setcounter{enumi}{13}
	\item \emph{Frontal faces detector} (F)
	\item \emph{Upper bodies detector} (B)
	\item \emph{Profile faces detector} (PF)
	\item \emph{Pedestrians detector} (P)
	\item \emph{Text detector} (T)
\end{enumerate}

\vspace{0.5cm}

Finally, we decide to consider 6 feature maps (19-24) derived a \acs{CNN}, which are modeled using Gaussian distributions. As in other computer vision applications, there is no doubt about the success of \acsp{CNN} for visual attention modeling. Nevertheless, despite their capability of discovering discriminant high-level visual features, it is still necessary to clarify the relationship between the feature maps derived from  \acp{CNN} and the psychophysical stimuli that guide attention. This implies the development of complementary modules able to provide this mapping, such as our hierarchical method, which facilitates the integration with such \acs{NN} schemes. Indeed, our intermediate sub-task level can be placed straightforwardly over the top layers of a deep network.

\acs{CNN}-based features have been drawn from the Deep Contrast Network for salient object detection recently introduced by Li et al. \cite{DeepSaliencyObject}. For the sake of completeness, and due to its use in our system for modeling visual attention in the temporal domain in Chapter \ref{ch:anomaly_detection}, we also take a brief look to this architecture in Section \ref{sec:cnns_based_features6}. The reason is twofold: first, they allow modeling more general objects than those identified by previously mentioned detectors; and second, they demonstrate the ability of our model to find efficient and diverse combinations of features that help to understand how visual attention works in a given scenario. We employ the models trained by the authors on a different image dataset, and use the feature maps of the penultimate layer to obtain features modeling general objectness.

% \clearpage
\subsection{Inference process}\label{sec:inference}
This section explains the inference process of our probabilistic model. As in the original \acs{LDA} \cite{LDA} and its extension \cite{DBAyang}, exact inference is not possible due to the coupling between the variables $\theta$ and $\bf z$, which prevents from inferring the posterior distribution of the parameters given the data. Therefore, we propose to use a simplified variational distribution $q$ (that is tractable) and mean-field variational inference, so that the \acs{KL} between the variational distribution $q$ and the posterior distribution is computed. The proposed variational distribution is as follows:

\begin{equation}
\begin{split}
  q(\theta,{\bf z}|\gamma,\phi_{1:N}) = q(\theta|\gamma) \prod_{n=1}^N q({\bf z_n}|\phi_n)
\end{split}
\end{equation}
\noindent that incorporates two new variational parameters: $\phi$, which is the parameter of a multinomial distribution $q({\bf z_n}|\phi_n)$, and $\gamma$, the parameter of a Dirichlet distribution $q(\theta|\gamma)$. This optimization is equivalent to maximize the \acf{ELBO} over the log-likelihood of all the frames in the corpus. In particular, using Jensen's inequality, the \acs{ELBO} of the log-likelihood of a frame can be expressed as:

\begin{align} \label{eq:elbo}
  &log \ p(f_{1:N,1:L},g_{1:N}|\alpha,\Gamma_{1:K,1:L},\eta) \geq  E_{q}[log \ p(\theta|\alpha)] \nonumber \\
  &+ \sum_{n=1}^{N} E_{q}[log \ p({\bf z_n}|\theta)] + \sum_{n=1}^N E_{q}[log \ p(f_{n,1:L} | {\bf z_n}, \Gamma_{1:K,1:L})] \nonumber  \\
  &+ \sum_{n=1}^N E_{q}[log \ p(g_n | {\bf z_n}, \eta)] + H(q)
\end{align}
\noindent where $E_{q}[\cdot]$ and $H(\cdot)$ are, respectively, the expectation over the variational distribution $q$ and the entropy of a distribution.

The first two terms of Eq. (\ref{eq:elbo})  and the entropy of the variational distribution are identical to the corresponding terms in \acs{ELBO} for unsupervised \acs{LDA} and are described in \cite{LDA}. The third term is the expected log probability of the features given the related topic model parameters. As was mentioned in Section \ref{sec:overview4}, we assume conditional independence among features. In the following paragraphs, we particularize this expression for the considered distributions.

\begin{itemize}
\item If the feature map $f_{nl}$ is modeled with a univariate \emph{Gaussian distribution} $\Gamma_{1:K,l} \sim \{\mu_{1:K,l}, \sigma^{2}_{1:K,l}\}$, such as for basic and novelty spatio-temporal features or CNN-based features, the equation for this term is:

\begin{equation}
\begin{split}
  &E_{q}[log \ p(f_{nl} | {\bf z_n}, \Gamma_{1:K,l})] = -\sum_{k=1}^{K} \phi_{nk}\log(\sigma_{kl}\sqrt{2\pi})\\ 
  & - \sum_{k=1}^{K} \phi_{nk}\frac{(f_{nl}-\mu_{kl})^2}{2\sigma_{kl}^2}
\end{split}
\end{equation}
\noindent where $\phi_{nk}$ is the probability that the location $n$ has been drawn by the topic $k$. 

\item In the case of camera motion features, the distribution is a multivariate Gaussian $p(\xv_{n} | {\bf z_n}, \mu_k,\Sigma_k)$ with $\mu_k = \cv_k \odot \uv$, being ${\bf c_{k}}$ a parameter to be estimated and ${\bf u}=(u,v)$ the camera motion vector. However, due to the diagonal nature of the covariance matrix $\Sigma_k$ we can decompose it into two independent univariate Gaussians and apply the previous expression.

\item In contrast, if the feature is modeled as a \emph{discrete probability distribution} over cells $r$ in a grid, as happens for objects-based features, the expression is:
\begin{equation}
\begin{split}
  &E_{q}[log \ p(r_n | {\bf z_{n}}, \beta_{l{z_n}})] = \sum_{k=1}^{K} \phi_{nk} \log(\beta_{klr_n})\\ 
\end{split}
\end{equation}
\noindent where $r_n$ stands for the region in the non-uniform grid defined for the object $l$ that contains the location $n$, and $\beta_{klr_n}$ is the value of the discrete distribution in region $r_{n}$ that contains the point $n$ for the object $l$ and the topic $k$.
\end{itemize} 

The fourth term includes the visual attention response variable $g_n$ and is drawn as a logistic regression model over the topic assignment $\bf z_n$ with parameter $\eta$:
\begin{equation}
\begin{split}
  E_{q}[log \ p(g_n | {\bf z_n}, \eta)] = E_{q}\bigg[\left(g_n -\frac{1}{2}\right) \eta^{T} \bf z_n \bigg] \\
  -E_{q}\bigg[log\left(exp{\left(\frac{\eta^{T} \bf z_n}{2}\right)}+exp{\left(\frac{-\eta^{T} \bf z_n}{2}\right)}\right)\bigg]
\end{split}
\end{equation}
\noindent By taking second derivatives, it can be noticed that the second term above is a convex function in the variable $\eta^{T^{2}} {\bf z_n}^{2} = (\eta^{T} \odot \eta^{T})({\bf z_n} \odot {\bf z_n})$, so we can bound it by using the lower bound for logistic function \cite{jaakkola2000bayesian}, which is the first order Taylor expansion in the variable $\eta^{T^{2}} \bf z_n^{2}$:

\begin{equation}
\begin{split}
  &log\left(exp{\left(\frac{\eta^{T} \bf z_n}{2}\right)}+exp{\left(\frac{-\eta^{T} \bf z_n}{2}\right)}\right)\\
  &\geq -\frac{\xi_{n}}{2} -log(1+exp(-\xi_{n})) \\
  &-\frac{1}{4\xi_{n}}tanh\left(\frac{\xi_{n}}{2}\right)E_{q}\bigg[\eta^{T^{2}} {\bf z_{n}^{2}} - \xi_{n}^{2}\bigg] \\
  &\approx -\frac{\xi_{n}}{2} -log(1+exp(-\xi_{n})) \\
  &-\frac{1}{4\xi_{n}} tanh\left(\frac{\xi_{n}}{2}\right)(\eta^{T^{2}}\phi_{n} - \xi_{n}^{2}) 
\end{split}
\end{equation}

\noindent where $\phi_{n}$ is the vector of topic proportions $\phi_{nk}$ in the location $n$ and $\xi_{n}$ is an additional variational parameter associated with each point $n$. 

It should be noted that, during variational inference, we work on expected values. This means that the indexing variable ${\bf z}_{n}$ is replaced by the variational $\phi_{n}$, which now contains the expected values of the topic assignments given a location $n$. Therefore, since $\phi_{n}$ is a vector with real values (the topic proportions for that sampled location), in practice each location $n$ is in turn modeled as the mixture of sub-tasks that best explains its visual appearance.

Computing the derivatives of the \acs{KL} with respect to the parameters and setting them equal to zero allows us to obtain the update equations for the variational procedure. In particular, in the \emph{variational E-step} we must update the variational parameters:

\begin{align}
\phi_{nk} \propto & \frac{\prod_{l=1}^{L_{D}} \beta_{klr_n}}{\prod_{l=1}^{L_{C}} \sigma_{kl}} \exp \bigg[\Psi(\gamma_{k})-\Psi\left(\sum_{j=1}^{k} \gamma_{j}\right) + \nonumber \\ 
&\left(g_{n}-\frac{1}{2}\right)\eta_{k} -\frac{1}{4\xi_{k}}tanh\left(\frac{\xi_{k}}{2}\right)\eta^{2}_{k} - \nonumber \\
&\sum_{l=1}^{L_{C}} \frac{(f_{nl}-\mu_{kl})^2}{2\sigma_{kl}^2}\bigg] \\
\gamma_{k} =& \alpha_{k} + \sum_{n=1}^{N} \phi_{nk} \\
\xi_{nk} =& \eta_{k} \phi_{nk}
\end{align}

\noindent being $L_{C}$ and $L_{D}$ the number of continuous (Gaussian) and discrete features respectively, and $L = L_{C} + L_{D}$ the total number of features. Note that we have used the expression $E_{q}[log(p(\theta_{k}|\gamma)] = \Psi(\gamma_{k})-\Psi\left(\sum_{j=1}^{k} \gamma_{j}\right)$, where $\Psi(\cdot)$ is the digamma function. 

In the M-step, we maximize the corpus-level \acs{ELBO} with respect to the model parameters $\Gamma_{1:K,1:L}, \eta$, in order to compute their optimal values. 

First, parameters $\mu_{kl}$ and $\sigma^{2}_{kl}$ are computed for each Gaussian feature $l$ and topic $k$. 

\begin{align}
&\mu_{kl}  = \frac{1}{\Delta_{kl}} \sum_{t=1}^{T} \sum_{n=1}^{N_t} \phi_{tnk} f_{tnl} \\
&\sigma^{2}_{kl} = \frac{1}{\Delta_{kl}} \sum_{t=1}^{T} \sum_{n=1}^{N_t} \phi_{tnk} (f_{tnl}-\mu_{kl})^2
\end{align}
\noindent where $\Delta_{kl}=\sum_{t=1}^{T} \sum_{n=1}^{N_t} \phi_{tnk}$ is the normalization factor.

In the case of camera motion, as mentioned above, the parameter is the vector ${\bf \cv_k} = (c_{kx}, c_{ky})$ that multiplies the camera motion vector ${\bf \uv_{t}} = (u_{t}, v_{t})$ to determine the mean of the Gaussian distribution:

\begin{eqnarray}
{\bf \cv_{k}}= \frac{\sum_{t=1}^{T} \sum_{n=1}^{N_t} \phi_{tnk} {\bf u_{t}} {\bf x_{tn}}}{\sum_{t=1}^{T} \sum_{n=1}^{N_t} \phi_{tnk} {\bf u_{t}}^2}  
\end{eqnarray}
\noindent where ${\bf \xv_{tn}} = (x_{tn}, y_{tn})$ stands for the spatial coordinates vector of the location $n$ in frame $t$.

Finally, for the case of object-based discrete features, the probabilities $\beta_{klr}$ of the regions $r$ defined from the outputs of the the object-detector $l$, and for every topic $k$ are:

\begin{eqnarray}
\beta_{klr} \propto \sum_{t=1}^{T} \sum_{n=1}^{N_t} \phi_{tnk} 1[r_{nl}=r] 
\end{eqnarray}
\noindent where $1[r_{nl}=r]$ means that we have a 1 just in case that the point $n$ belongs to the region $r$ (otherwise we have a zero). It is worth noting that we have added the subindex $t$ when necessary to indicate the frame number in the corpus. 

Furthermore, during the training step, we use the \acs{GT} response value $g_{tn}$ of all points in the corpus to learn the parameter of the logistic regression model:

\begin{align}
&\eta_{k} = \frac{2\sum_{t=1}^{T} \sum_{n=1}^{N_t} \phi_{tnk} (g_{tn}-\frac{1}{2})}{\sum_{t=1}^{T} \sum_{n=1}^{N_t} \frac{\phi_{tnk}}{\xi_{nk}} tanh(\frac{\xi_{nk}}{2})}
\end{align} 

A more comprehensive development of the \acs{ELBO} and the previous formulas for parameters estimation is provided in Appendix \ref{ch:atom_formulas}.

% \clearpage
\subsection{Learning sub-tasks for spatio-temporal visual attention estimation}\label{sec:subtasks_estimation}
\begin{sidewaysfigure}
\centerline{\includegraphics[angle=0, trim=0cm 0cm 0cm 0cm, width=0.85\textwidth]{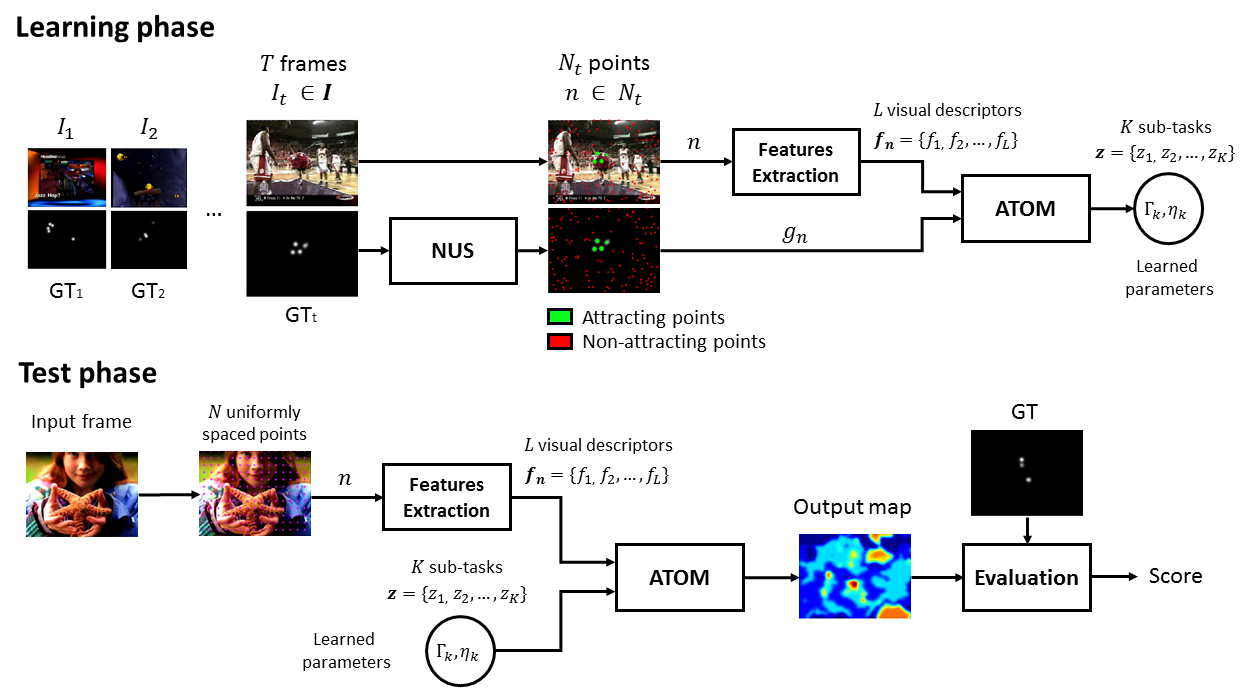}}
\caption[Processing pipelines of the generative probabilistic approach proposed for spatio-temporal visual attention modeling.]{Processing pipelines of the approach proposed. First, in the learning phase, we learn the optimal values for the parameters associated to the $K$ sub-tasks that model visual attention. A \acs{NUS} \cite{ivanNUS} strategy allows to generate training datasets that balance the number of attracting and non-attracting points. Then, in the test phase, attention is predicted for each frame at $N$ uniformly spaced locations.}
\label{44_phases_diagram}
%\vspace{-0.4cm}
\end{sidewaysfigure}

As in other supervised approaches, we can distinguish two main stages in our framework, as shown in Figure \ref{44_phases_diagram}. First, in the learning phase, optimal values for the parameters that maximize the \acs{ELBO} of the log-likelihood are learned. As we need to learn from annotated data, we first describe how we sample this data from the annotated video datasets. Since we are on a highly unbalanced scenario, in which the areas that attract visual attention are strongly less prominent than those that inhibit it, we need to prevent the later dominating the learning process, which might lead to a poor performance. For that end, we have used the \acf{NUS} strategy proposed in \cite{ivanNUS}, which allows to generate training datasets that balance the number of attracting and non-attracting points. While the first are selected based on the \acs{GT} masks computed from human fixations for a given video frame, non-attracting points are sampled from those spatial locations which have not been fixated by viewers in any frame of the same video. In addition, the sampling process also provides the ground truth binary response $g_{n}$ for each sampled spatial location ($g_{n}=1$ for attracting points, and zero otherwise).

Once models are trained, in the test phase, attention is predicted at uniformly spaced locations $n$ in frames. For that end, we remove all terms relating to the supervision (variable $g$) and estimate the visual attention maps using the expected value of the logistic regression over the topic or sub-task assignments:

\begin{equation}
E[g_{n}|f_{n,1:L},\alpha,\Gamma_{1:K},\eta] \approx \frac{\exp{(\eta^{T} \phi_{n}})}{1+\exp{(\eta^{T} \phi_{n}})}
\end{equation}

In addition, knowing that given a particular frame visual attention is usually focused on small areas of the size occupied by fixations, a histogram equalization procedure is carried out to highlight the most significant regions detected, which helps to improve the system performance.

%\begin{landscape}
%\newgeometry{right=3cm}
%\restoregeometry
%\end{landscape}

%*****************************************
%*****************************************
%*****************************************
%*****************************************
%*****************************************

%************************************************
\chapter{Experiments on context-driven visual attention understanding and prediction}\label{ch:atom_experiments} 
\chaptermark{Context-driven visual attention understanding}
%************************************************

\section{Introduction}\label{sec:introduction_5}
In this chapter, we provide an in-depth analysis of our proposal for visual attention modeling described in Chapter \ref{ch:atom}. We give a meaningful insight about the information reflected in each of the sub-tasks that decompose the visual attention. To this end, we illustrate how our approach successfully learns hierarchical guiding representations adapted to several contexts. Furthermore, we perform a comparison with quite a few methods reported in the literature of visual attention in video. 

Experiments show how our proposal successfully learns particularly adapted hierarchical explanations of visual attention in diverse video genres, outperforming several leading models in the literature.

\section*{Chapter overview}
First, the experimental design is described in Section \ref{sec:experimental_design5}, introducing the databases and the evaluation metrics used to provide the results, and also the initialization of the model. Then, the \acf{ATOM} is used for context-driven visual attention understanding in Section \ref{sec:va_understanding}. Experimental results, together with an analysis of the obtained models and a comparison with \emph{state-of-the-art} methods, are provided in Sections \ref{sec:results5} and \ref{sec:comparison_soa5}. Finally, Section \ref{sec:where_we_are5} discusses the model strengths and limitations, and Section \ref{sec:conclusions5} summarizes our conclusions and motivates and outlines future work.

\section{Experimental design}\label{sec:experimental_design5}
The purpose of our experiments is to demonstrate the ability of the proposed \acs{ATOM} to learn meaningful sub-tasks that can be used to understand what guides visual attention in different contexts, drawing conclusions on whether observers are either driven by similar generic sub-tasks or, in contrast, by certain specific tasks related to each particular scenario. For this reason, we have selected the well-known freely-accessible CRCNS-ORIG \cite{Itti_Carmi09crcns} and DIEM \cite{mital2011clustering} as benchmark datasets.

\subsection{Databases}\label{sec:databases5}
In this section we briefly describe the databases used for the experiments. Further information about the division of the database videos into categories can be found in Appendix \ref{ch:databases}.

\subsubsection*{CRCNS-ORIG database}\label{sec:orig}
CRCNS-ORIG \cite{Itti_Carmi09crcns} dataset contains eye movement recordings from eight distinct subjects freely watching 50 different video clips (over 46,000 video frames, 25 minutes total, $640\times480$). Eye traces have been obtained using a 240 Hz ISCAN RK-464 eye-tracker. As set out in Table \ref{fig:421_scenarios_stats}(a), clips include complex video stimuli that can be divided into seven categories: \emph{Outdoor, Videogames, Commercials, TV News, Sports, Talk Shows} and \emph{Others}. Eye fixations of at least 4 subjects are provided for each clip.

The dataset was delivered some years ago with this same intention pursued with our analysis, and has been employed to evaluate a lot of \emph{state-of-the-art} saliency models. However, to our knowledge, none of them had attempted so far to offer a data interpretation such as the one resulted from our approach.  

\subsubsection*{DIEM database}\label{sec:diem}
DIEM \cite{mital2011clustering} dataset contains eye movement recordings from over 250 participants freely watching 84 high-definition natural videos (over 240,000 video frames, 134 minutes total, variable dimensions). Eye traces have been obtained using a 1,000 Hz SR Research Eyelink 2000 desktop mounted eye tracker. As is summarized in Table \ref{fig:421_scenarios_stats}(b), clips have been classified into seven categories: \emph{TV Shows, Documentaries, Commercials, Talk Shows, Sports, Cooking} and \emph{TV News}. Eye fixations from approximately 50 subjects are provided for each clip.

In contrast to CRCNS-ORIG \cite{Itti_Carmi09crcns}, DIEM \cite{mital2011clustering} constitutes a greater source of video annotated with \acs{GT} fixations, which serves not only to provide a more truthful result of the method proposed but also to train deeper \acsp{CNN} such as the one presented in the next chapter for motion-based feature maps extraction. 

\subsection{Experimental setup}\label{sec:experimental_setup5}
In order to both assess the performance and gain insight into the latent information provided by the proposed probabilistic method for visual attention estimation, we will compare two different approaches for each database: a) a \acf{C-G} model trained using frames belonging to videos in all the categories; and b) 7 \acf{C-A} models trained on those videos belonging to each category or genre. 

The performance over every video in the datasets is evaluated by conducting a 4-fold cross validation procedure, in the case of CRCNS-ORIG \cite{Itti_Carmi09crcns}, and a 5-fold cross validation, in the case of DIEM \cite{mital2011clustering}, so that at each iteration some videos are picked for evaluation. For the purpose of avoiding over-fitting, all frames of a video are always grouped together in the same set (train or test).

\begin{table}[!t]
  \small
  \centering
  \caption{Categories into which (a) CRCNS-ORIG \cite{Itti_Carmi09crcns} and (b) DIEM \cite{mital2011clustering} databases are divided.}\label{fig:421_scenarios_stats}
  \captionsetup[subfloat]{position=top}
  \subfloat[CRCNS-ORIG \cite{Itti_Carmi09crcns}]{%
    \hspace{0cm}%
    \begin{tabular}{l|c|c}
    		\hline
    		{Context} & {\centering \# clips} & {Frames} \\ \hline
    		{Outdoor} & $17$ & $8,357$ \\  
    		{Videogames} & $9$ & $15,809$ \\  
    		{Commercials} & $4$ & $2,618$ \\  
    		{TV News} & $7$ & $8,071$ \\   
    		{Sports} & $5$ & $4,851$ \\ 
    		{Talk Shows} & $4$ & $4,244$ \\  
    		{Others} & $4$ & $2,539$ \\ \hline
    		{TOTAL} & $50$ & $46,489$ \\ \hline
    	\end{tabular}
    	%\label{fig:421_scenarios_stats_ORIG}
    \hspace{0cm}%
  }\hspace{0.1cm}
  \subfloat[DIEM \cite{mital2011clustering}]{%
    \hspace{0cm}%
    \begin{tabular}{l|c|c}
    		\hline
    		{Context} & {\centering \# clips} & {Frames} \\ \hline
    		{TV Shows} & $12$ & $34,271$ \\  
    		{Documentaries} & $18$ & $56,382$ \\  
    		{Commercials} & $15$ & $40,558$ \\  
    		{Talk Shows} & $5$ & $8,657$ \\   
    		{Sports} & $20$ & $54,293$ \\ 
    		{Cooking} & $7$ & $23,684$ \\  
    		{TV News} & $7$ & $22,607$ \\ \hline
    		{TOTAL} & $84$ & $240,452$ \\ \hline
    	\end{tabular}
    	%\label{fig:421_scenarios_stats_DIEM}
    \hspace{0cm}%
  }
  \captionsetup[subfloat]{position=bottom}
\end{table}

\subsection{Evaluation metrics}\label{sec:atom_evaluation_metrics}
In parallel to the proposal of computational models for saliency and visual attention, a great effort has been made to evaluate their performance. As summarized in the excellent comprehensive study by Bylinskii et al. \cite{BylinskiiMetrics}, this has resulted in a wide variety of metrics based on different assumptions: how the \acf{GT} fixation map is represented, whether center bias is considered or not or the type of normalization applied to \acs{VAM}, among others.

In this section, we define those metrics that we have used to assess the performance of the spatio-temporal visual attention methods proposed in this thesis, as well as those taken from the \emph{state-of-the-art}. 

\begin{figure}[t]
\centering
\includegraphics[trim=0cm 0cm 0cm 0cm, clip, width=1\linewidth]{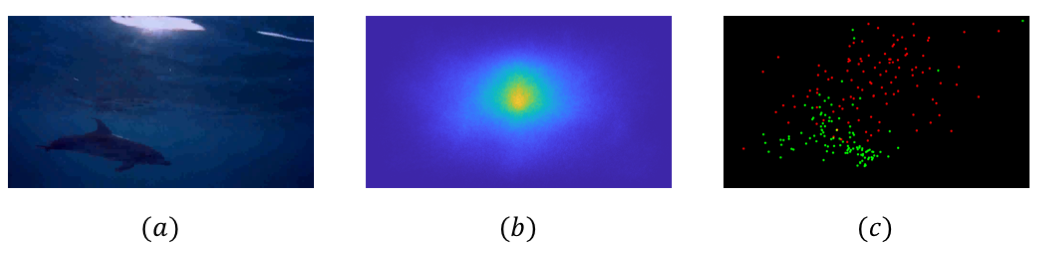}
\caption[\acsp{TP} and \acsp{FP} sampled in an example frame taken from a documentary video in DIEM \cite{mital2011clustering} database, according to a probabilistic shuffle map.]{\acsp{TP} and \acsp{FP} sampled in an example frame taken from a documentary video in DIEM \cite{mital2011clustering} database. \acsp{TP} correspond to fixations locations, while \acsp{FP} are sampled according to a probabilistic shuffle map with fixations in frames from all other videos in the dataset. (a) Example frame. (b) Shuffle map. (c) \acsp{TP} and \acsp{FP} sampled, indicated in green and red, respectively. Image has been dilated for a better visualization.}
\label{fig:422_evaluation_metrics}
\end{figure}

The following metrics have been selected according to the suggestions in \cite{BylinskiiMetrics} for models conceived for fixation prediction, which is the aim of the experiments in this chapter, and video surveillance scenarios, as those presented in Chapter \ref{ch:surveillance_experiments}:

\begin{itemize}
	\item \emph{\acf{sAUC}}: The \acf{AUC} \cite{fawcett2006introduction} is the most used metric in the literature for the evaluation of visual attention models. Given an image or a video frame and its corresponding \acs{GT} fixation map, a \acs{VAM} normalized between 0 and 1 can be understood as the soft output of a binary classifier of fixations; hence, the area under a \acf{ROC} curve, which measures the trade-off between \acfp{TP} and \acfp{FP} at different threshold values, provides a performance score. 
	Depending on how \acsp{TP} and \acsp{FP} are calculated, we can distinguish different \acs{AUC} implementations \cite{borji2013quantitative, judd2009learning}. For our experiments, we have chosen a probabilistic \acf{sAUC} metric \cite{zhang2015exploiting}. This score counteracts the effect of the commonly-observed central fixation bias in scene viewing \cite{tatler2007central}, which is advantageous for those models that consider a center prior. The \acs{sAUC} chosen to provide our results probabilistically samples \acsp{FP} from fixated locations in other images or videos, instead of uniformly at random. \\	
	Despite the effectiveness of the \acsp{AUC} scores, they are invariant to monotonic transformations of the \acs{VAM}. What is more, attention maps that place different amounts of density at fixated locations receive similar scores as long as they keep fixed the order of the locations. Therefore, it is recommended to supplement them with other metrics.
	
	\item \emph{\acf{sNSS}}: The \acf{NSS} metric, firstly introduced in \cite{peters2005components}, is given by the averaged normalized visual attention at fixated locations. Given an attention map $VAM$, it is computed as follows:
	
	\begin{gather}
		NSS = \frac{1}{N_{TP}} \sum_{i=1}^{N_{TP}} \frac{VAM_{i}-\mu_{VAM}}{\sigma_{VAM}}
	\end{gather}
	
	where $N_{TP}$ is the total number of \acs{GT} fixations (\acsp{TP}), and $\mu_{VAM}$ and $\sigma_{VAM}$ represent the mean and standard deviation of the \acs{VAM} values, respectively. \acs{NSS} is sensitive to \acsp{FP} and monotonic transformations of the map, in contrast to classical \acsp{AUC} scores, so it constitutes an interesting complement to these. For the evaluation of the methods considered in the thesis, we make use of the \acs{sNSS} version of this score, proposed by Leborán et al. \cite{GallegaSaliency}. Unlike in the original \acs{NSS}, $N_{boot}$ sets of \acsp{FP} sampled from fixated locations in different images or videos are obtained, in order to compute a mean $\mu_{VAMi} = \frac{1}{M} \sum_{m=1}^{M} VAM_{m}$ and a standard deviation $\sigma_{VAMi} = \sqrt{\frac{1}{M-1} \sum_{m=1}^{M} (VAM_{m}-\mu_{VAMi})^2}$, where $VAM_{m} \in \{TP \cup FP\}$ and $M = card(TP \cup FP)$. Then, \acs{sNSS} compensates the center bias effect, by computing the average of $N_{boot}$ scores for each frame:
	
	\begin{gather}
		sNSS = \frac{1}{N_{boot}N_{TP}} \sum_{i=1}^{N_{boot}} \sum_{j=1}^{N_{TP}} \frac{VAM_{j}-\mu_{VAMi}}{\sigma_{VAMi}}
	\end{gather}
	
	 Positive \acs{sNSS} indicates correspondence between maps above chance, while a high number of \acsp{FP} drives the overall \acs{sNSS} down.
	
%	\item \emph{\acf{IG}}: This ratio, recently proposed in \cite{InformationGain}, computes the average information gain of a given $VAM$ over a baseline map $B$. This baseline can capture, for instance, fixation biases, or might be the attention map provided by a different model. Both maps are normalized to sum to 1 and treated as probability densities. Then, \acs{IG} is computed as:
%	
%	 \begin{gather}
%	 	IG = \frac{1}{N_{TP}} \sum_{i=1}^{N_{TP}} [log_{2}(VAM_{i}+\epsilon)-log_{2}(B_{i}+\epsilon)] 
%	 \end{gather}
%	
%	 If \acs{IG} score is above zero, the $VAM$ has a better prediction for the fixated locations than $B$. 
%	 Quitar si finalmente no se utiliza en las figuras de comparación de métodos.
\end{itemize}

In order to evaluate the performance of visual attention models in a particular video, a probabilistic map that consists of fixations in frames from all other videos in the dataset is used as shuffle map for both scores. Figure \ref{fig:422_evaluation_metrics} shows how \acsp{TP} (green locations) and \acsp{FP} (red locations) are sampled in an example frame taken from DIEM \cite{mital2011clustering} database. As can be appreciated in the shuffle map, viewers have a tendency to look at the center of the image, as discussed above. Hence, more \acsp{FP} are sampled close to the center of the frame, which prevents \acs{sAUC} and \acs{sNSS} metrics from being affected by the center bias. Moreover, 95\% confidence bounds are provided for both metrics used.

Finally, for comparison purposes, we have considered the three baseline models introduced by Judd et al. in \cite{Judd_2012}: 

\begin{itemize}
	\item CHANCE: The model generates a \acs{VAM} for each frame by randomly selecting some pixels as salient, which leads to a poor performance.
	\item CENTER: The model consists in a stretched symmetric \acs{2D} Gaussian distribution centered on the frame, in such a way that closer locations to the center are more salient. This model serves as a good indicator to determine if the evaluation metrics used are affected by center bias or not.
	\item H50: For each frame, the model generates a \acs{VAM} that contains the fixations of the 50\% of subjects available. It constitutes a good realistic upper bound, which puts into perspective the efficiency of the assessed approaches.
\end{itemize}

\subsection{Model initialization}%\label{sec:applications}
Due to the stochastic nature of our approach, a correct initialization of the parameters is important to both fasten the convergence and reach an optimal model. As the goal is to learn sub-tasks that either attract or inhibit attention, we initialize basic, novelty and \acsp{CNN}-based feature distributions as follows: we initialize some topics that inhibit and other that attract visual attention, with $\mu_{kl} = 0$ and $\mu_{kl} = 1$, respectively (remember that our features are maps in the range $[0, 1]$). Then, in order to provide initial variances for the topics, we compute two separate sets of variances with respect to $\mu_{kl} = \{0, 1\}$, from non-attracting and attracting locations respectively. Then, we run a separate \emph{k-means} over the variance values and obtain the corresponding $K$ centroids, one per topic. For camera motion features, the parameters $\cv_k$ are randomly initialized with values close to $0$ whereas, as we have already mentioned, $\Sigma_{k}$ is empirically set to $\Sigma_{k} = diag(0.25)$. Finally, discrete distribution features for object detection are initialized uniformly for every region in the non-uniform grid.

Last but not least, the main parameter of the proposed model is the number $K$ of sub-tasks or topics that contribute to model visual attention. For simplicity, we have used the same number of attracting and inhibiting topics in our initialization. As indicated in the next sub-section, $K = 60$ is the number of topics used for the rest of the experiments. Initial global topic proportions $\alpha$ have been empirically set to $\alpha_{k}=0.01$.

\section{Understanding visual attention as a mixture of sub-tasks}\label{sec:va_understanding}
The most outstanding outcome of our probabilistic approach is determined by the topics inferred, which effectively help to interpret how visual attention works. Firstly, by means of the proportions in which those are blended, we can establish which sub-tasks are more prevailing for guidance. We have statistically estimated the importance of each topic by examining the value $\eta_{k}$ of the logistic regression model and the topic proportions $\phi_{nk}$ obtained for each spatial location $n$ evaluated on the test set, as both variables are linearly related to the model response which generates the visual attention map. In particular, the relevance score of each sub-task $k$ is computed as:

\begin{align}
\mathcal{S}_{k} = \eta_{k} \sum_{n=1}^{N} \phi_{nk}
\end{align}

% given that they highlight different objects of interest with respect to the background without categorizing them, which might cause more ambiguous conclusions

\noindent Scores are later normalized between $[-1, 1]$ to simplify the analysis. 

Secondly, regarding the distribution parameters learned for features considered as input, we can further study the meaning of the sub-tasks, providing useful information about the most conspicuous regions in a given scenario. For the sake of interpretability, it should be noted that we have not considered CNNs-based features in this analysis, since they constitute very high-level representations at different scales whose content is more difficult to understand. Besides, since we know in advance for all features considered in the experiments that low feature values (close to $0$) correspond to non-salient locations in frames, while regions with high feature values (close to $1$) are very salient, Gaussians' means are not learned and remain fixed in $\mu_{kl} = 0$ and $\mu_{kl} = 1$ during the whole inference process. Then, we consider topics centered in $\mu_{kl} = 0$ and $\mu_{kl} = 1$ as those topics inhibiting (\acs{IT}) or attracting attention (\acs{AT}), respectively. Furthermore, the camera motion distribution has been also removed from the analysis as it has been observed that there is not a strong influence of this feature in any of the categories, since parameters $\cv_k$ learned for the most prevailing topics have all similar values. Under this simplified scenario, we can evaluate the relevance of \emph{basic and novelty features}, using their learned standard deviation values $\sigma_{kl}$ : 

\begin{align}
\mathcal{S}^{C}_{kl} = \frac{\sigma^{F}_{l}}{\sigma_{kl}}
\end{align}

\noindent with values in the range $[0, + \infty)$. Given a sub-task $k$ and under our simplified scenario with fixed means ($\mu_{kl} = \{0, 1\}$), a feature $l$ will be representative if its standard deviation $\sigma_{kl}$ is lower compared to the deviation $\sigma^{F}_{l}$ measured on areas that correspond with the topic type $F$ (fixated areas if the topic is attracting attention, and viceversa).
%, but are displayed between $[0, 5]$ for a better visualization. 

Moreover, scores for \emph{object-based features} are calculated by computing the cumulative probability of the cells that lie inside the detected bounding box ($r>0$, excluding the background cell): 

\begin{align}
\mathcal{S}^{D}_{kl} = \sum_{r=1}^{R} \beta_{klr} 
\end{align}

\noindent with values between $[0, 1]$.

\begin{figure*}[p]
    \centering
  	\subfloat[Outdoor]{%
        \includegraphics[trim=10cm 1cm 12cm 1cm, width=0.6\textwidth]{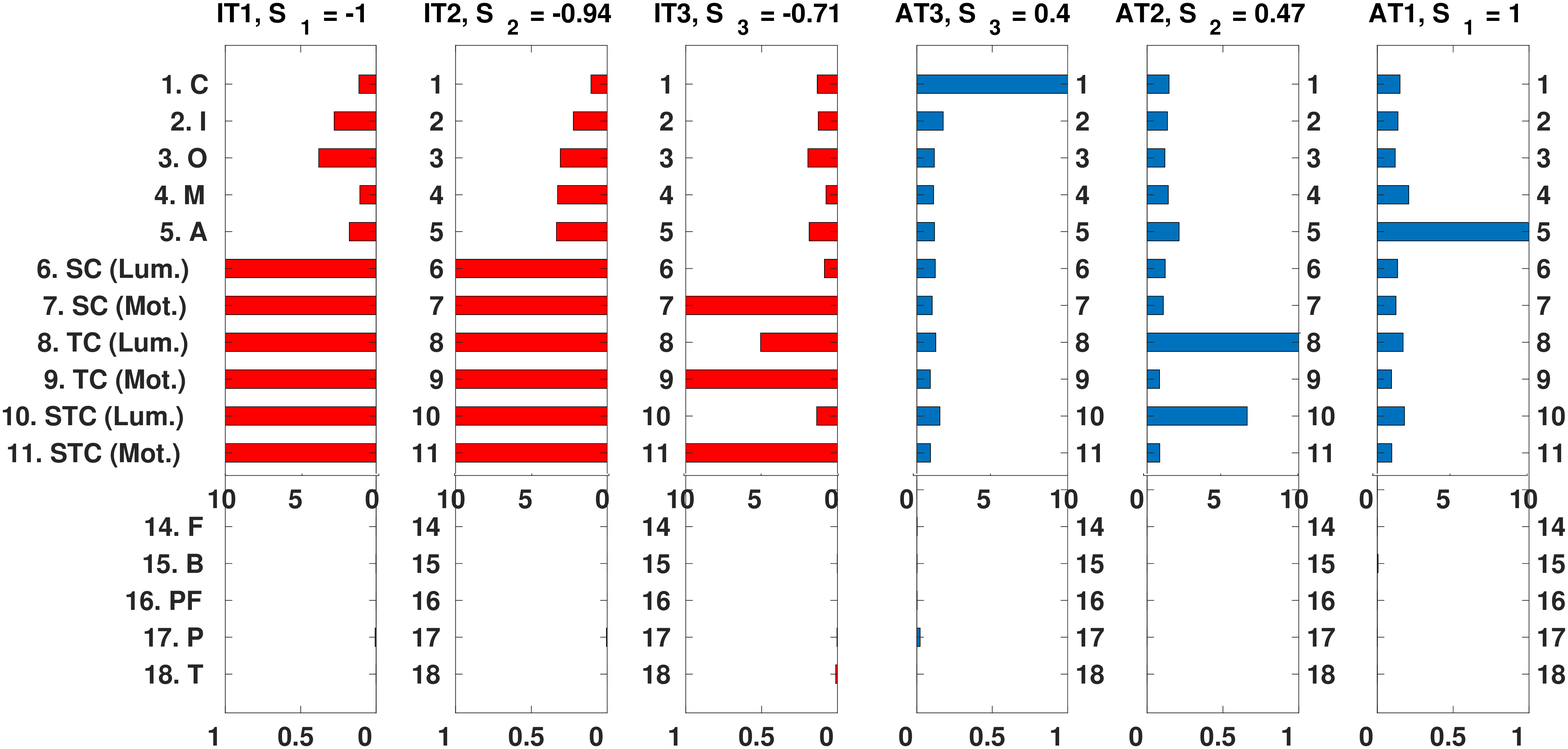}
    \label{fig:Outdoor_ORIG}}\\
    \subfloat[TV News]{%
       \includegraphics[trim=10cm 1cm 12cm 1cm, width=0.6\textwidth]{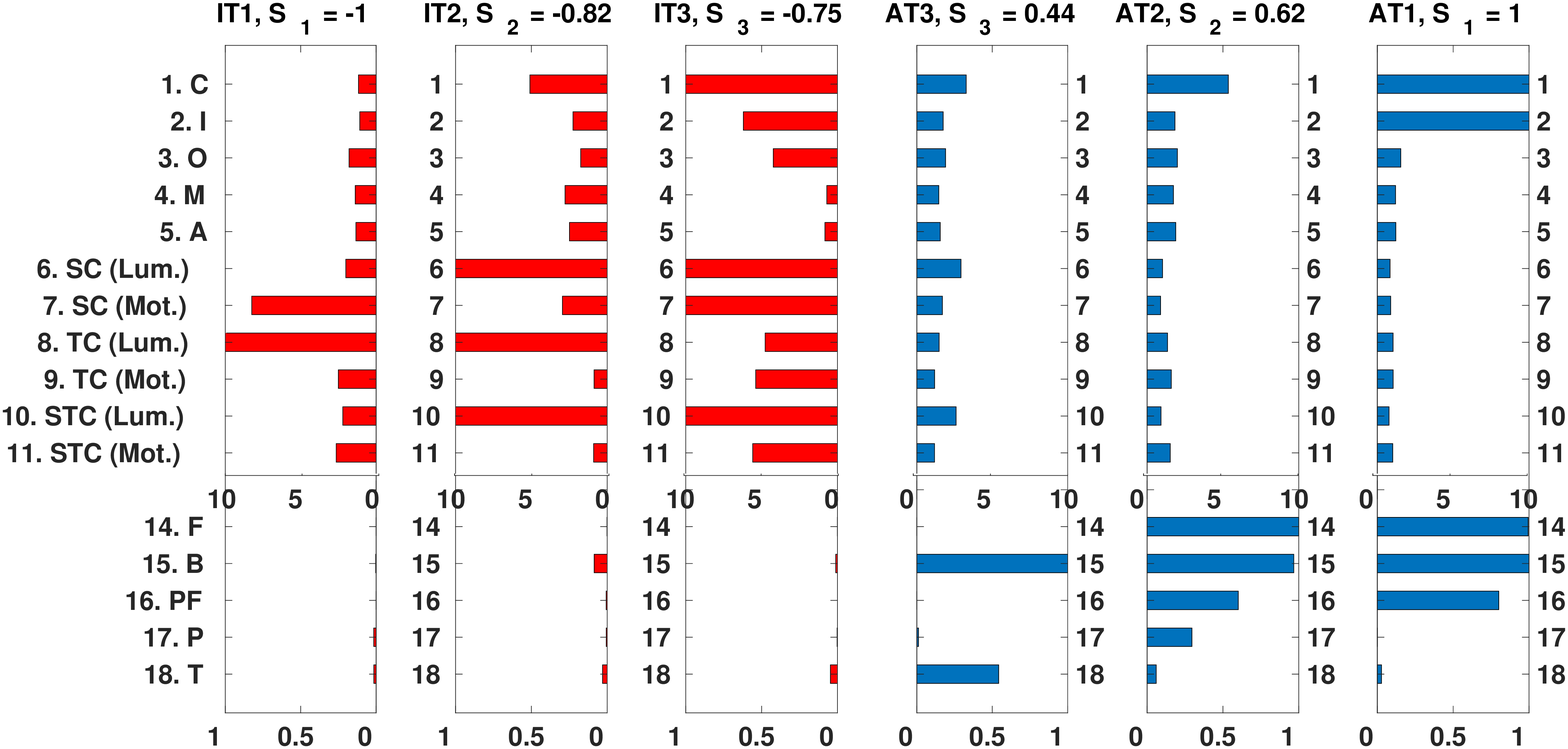}
    \label{fig:TVNews_ORIG}}\\
    \caption[Three most prominent attracting (\acs{AT}) and inhibiting (\acs{IT}) sub-tasks inferred by (a) \emph{Outdoor} and (b) \emph{TV News} \emph{context-aware} models learned based on CRCNS-ORIG \cite{Itti_Carmi09crcns} database.]{Three most prominent attracting (\acs{AT}) and inhibiting (\acs{IT}) sub-tasks inferred by (a) \emph{Outdoor} and (b) \emph{TV News} \emph{context-aware} models learned based on CRCNS-ORIG \cite{Itti_Carmi09crcns} database. For each category, \acsp{IT} ($\mu_{kl} = 0$) are shown in red on the left side of the bar graph, while \acsp{AT} ($\mu_{kl} = 1$) appear in blue on the right side. Then, the relevance score $S_{k}$ of each sub-task $k$ is indicated on top of its graph. Moreover, sub-tasks are represented as combinations of some of the features described in Section \ref{sec:guiding_features}: basic and novelty features, such as \emph{color} (C), \emph{intensity contrast} (I), \emph{orientation} (O), \emph{velocity} (M), \emph{acceleration} (A), \emph{luminance spatial coherence (SC (Lum.))}, \emph{motion spatial coherence (SC (Mot.))}, \emph{luminance temporal coherence (TC (Lum.))}, \emph{motion temporal coherence (TC (Mot.))}, \emph{luminance spatio-temporal coherence (STC (Lum.))}, \emph{motion spatio-temporal coherence (STC (Mot.))}; and object-based features, such as \emph{frontal} (F) and \emph{profile faces} (PF), \emph{upper bodies} (B), \emph{pedestrians} (P) and \emph{text} (T). Each bar is associated to a feature score $S_{kl}^{C}$, for basic and novelty features, or $S_{kl}^{D}$, for object-based features. High values of scores in \acsp{IT} correspond to inhibiting features, which reduce the attentional response. In contrast, high values of scores in \acsp{AT} highlight those features that are more attracting for each category.}
	\label{fig:53_category_subtasks_orig}
\end{figure*}

\begin{figure*}[p]
    \centering
    \subfloat[Commercials]{%
        \includegraphics[trim=10cm 1cm 12cm 1cm, width=0.6\textwidth]{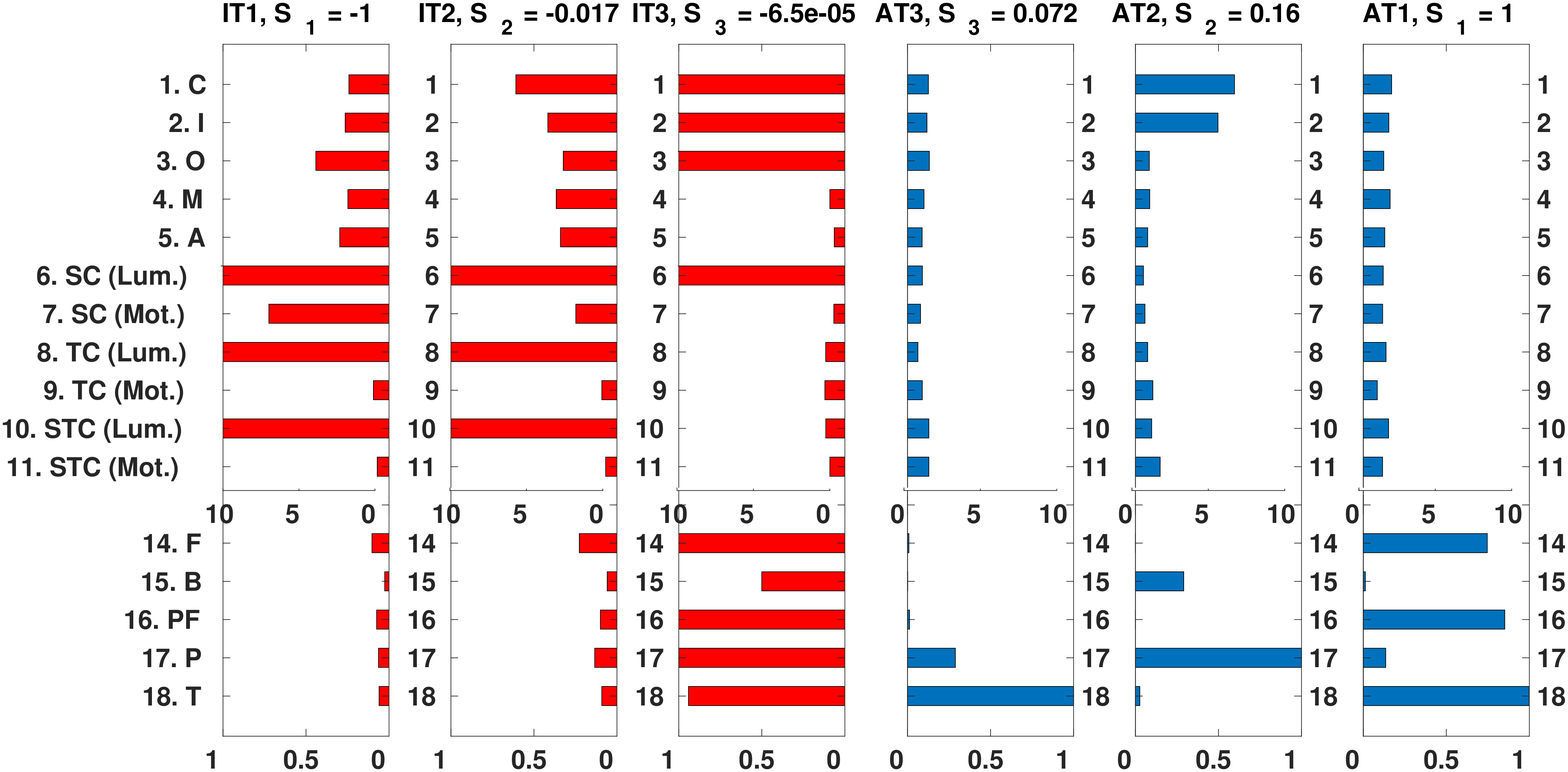}
    \label{fig:Commercials_DIEM}}\\
    \subfloat[Sports]{%
       \includegraphics[trim=10cm 1cm 12cm 1cm, width=0.6\textwidth]{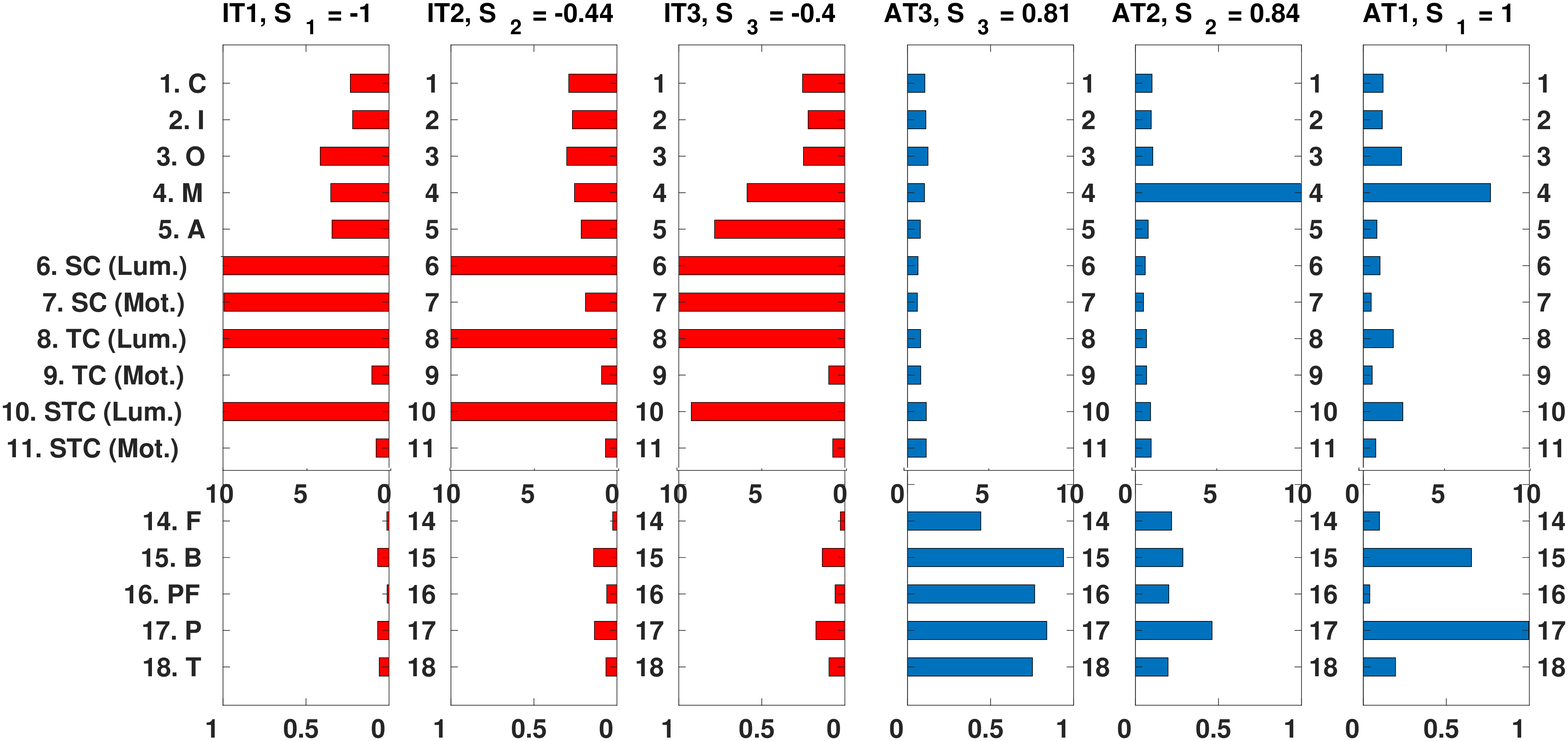}
    \label{fig:Sports_DIEM}}\\
    \caption[Three most prominent attracting (\acs{AT}) and inhibiting (\acs{IT}) sub-tasks inferred by (a) \emph{Commercials} and (b) \emph{Sports} \emph{context-aware} models learned based on DIEM \cite{mital2011clustering} database.]{Three most prominent attracting (\acs{AT}) and inhibiting (\acs{IT}) sub-tasks inferred by (a) \emph{Commercials} and (b) \emph{Sports} \emph{context-aware} models learned based on DIEM \cite{mital2011clustering} database. For each category, \acsp{IT} ($\mu_{kl} = 0$) are shown in red on the left side of the bar graph, while \acsp{AT} ($\mu_{kl} = 1$) appear in blue on the right side. Then, the relevance score $S_{k}$ of each sub-task $k$ is indicated on top of its graph. Moreover, sub-tasks are represented in the graphs as combinations of some of the features described in Section \ref{sec:guiding_features}: basic and novelty features, such as \emph{color} (C), \emph{intensity contrast} (I), \emph{orientation} (O), \emph{velocity} (M), \emph{acceleration} (A), \emph{luminance spatial coherence (SC (Lum.))}, \emph{motion spatial coherence (SC (Mot.))}, \emph{luminance temporal coherence (TC (Lum.))}, \emph{motion temporal coherence (TC (Mot.))}, \emph{luminance spatio-temporal coherence (STC (Lum.))}, \emph{motion spatio-temporal coherence (STC (Mot.))}; and object-based features, such as \emph{frontal} (F) and \emph{profile faces} (PF), \emph{upper bodies} (B), \emph{pedestrians} (P) and \emph{text} (T). Each bar is associated to a feature score $S_{kl}^{C}$, for basic and novelty features, or $S_{kl}^{D}$, for object-based features. High values of scores in \acsp{IT} correspond to inhibiting features, which reduce the attentional response. In contrast, high values of scores in \acsp{AT} highlight those features that are more attracting for each category.}
	\label{fig:53_category_subtasks_diem}
\end{figure*}

Scores obtained by the three most noteworthy attraction and inhibition sub-tasks for some video genres in CRCNS-ORIG \cite{Itti_Carmi09crcns} and DIEM \cite{mital2011clustering} databases are shown in Figures \ref{fig:53_category_subtasks_orig} and \ref{fig:53_category_subtasks_diem}. Significant sub-tasks deduced for the rest of categories included in these databases are gathered in Appendix \ref{ch:databases}. For each category, \acsp{IT} ($\mu_{kl} = 0$) are represented in red on the left side of the bar graphs, while \acsp{AT} ($\mu_{kl} = 1$) appear in blue on the right side. Then, the relevance score $S_{k}$ of each sub-task $k$ is indicated on top of its graph. Moreover, sub-tasks are represented as combinations of some of the features described in Section \ref{sec:guiding_features}: basic and novelty features, such as \emph{color} (C), \emph{intensity contrast} (I), \emph{orientation} (O), \emph{velocity} (M), \emph{acceleration} (A), \emph{luminance spatial coherence (SC (Lum.))}, \emph{motion spatial coherence (SC (Mot.))}, \emph{luminance temporal coherence (TC (Lum.))}, \emph{motion temporal coherence (TC (Mot.))}, \emph{luminance spatio-temporal coherence (STC (Lum.))}, \emph{motion spatio-temporal coherence (STC (Mot.))}; and object-based features, such as \emph{frontal} (F) and \emph{profile faces} (PF), \emph{upper bodies} (B), \emph{pedestrians} (P) and \emph{text} (T). Each bar is associated to a feature score $S_{kl}^{C}$, for basic and novelty features, or $S_{kl}^{D}$, for object-based features. High values of scores in \acsp{IT} correspond to inhibiting features, which reduce the attentional response. In contrast, high values of scores in \acsp{AT} highlight those features that are more attracting for each category.

Although the number of topics experimentally determined is quite high ($K = 60$), we have observed that only few of them are responsible of guiding attention most of the time, whereas the rest are intended to refine the estimation, specially in the less prevalent sequences.

%\subsubsection*{Analysis on CRCNS-ORIG database} 
%\label{sec:understanding_orig}

%\subsubsection*{Analysis on DIEM database}\label{sec:understanding_diem}
%COMPLETE

As can be seen, different sub-tasks are determined to model visual attention in each scenario, existing an appreciable contrast between well-separated categories such as \emph{Outdoor} or \emph{TV News}, which involve distinctive actions (see Figure \ref{fig:53_category_subtasks_orig}). While \emph{context-generic} models are adjusted to the most prominent events in the databases, which consist of faces noticeable by their color and intensity, and motion objects, \emph{context-aware} models have the ability of attaining more particular and explainable activities. Motion and acceleration features are relevant in \emph{Outdoor} (Figure \ref{fig:Outdoor_ORIG}), and \emph{Videogames} (Figure \ref{fig:B_videogames_orig}) sub-tasks, which could be related to people or characters walking or running. In contrast, faces and texts are more attractive and predominant in categories like \emph{Commercials} (Figures \ref{fig:Commercials_DIEM}, \ref{fig:B_commercials_orig}), \emph{TV News} (Figures \ref{fig:TVNews_ORIG}, \ref{fig:B_tvnews_diem}) and \emph{Talk Shows} (Figures \ref{fig:B_talkshows_orig}, \ref{fig:B_talkshows_diem}). Both motion and faces are eye-catching in \emph{Sports} (Figures \ref{fig:Sports_DIEM}, \ref{fig:B_sports_orig}) videos, which often consist of real-time edited outdoor scenes to be released on TV. Finally, low values of spatial and temporal coherency features are mostly frequent in \acs{IT}, which implies reducing the attentional response in usual and stable locations over space and time.

\begin{figure*}[!t]
\centering
		\includegraphics[trim=0cm 0cm 0cm 0cm, width=1\textwidth]{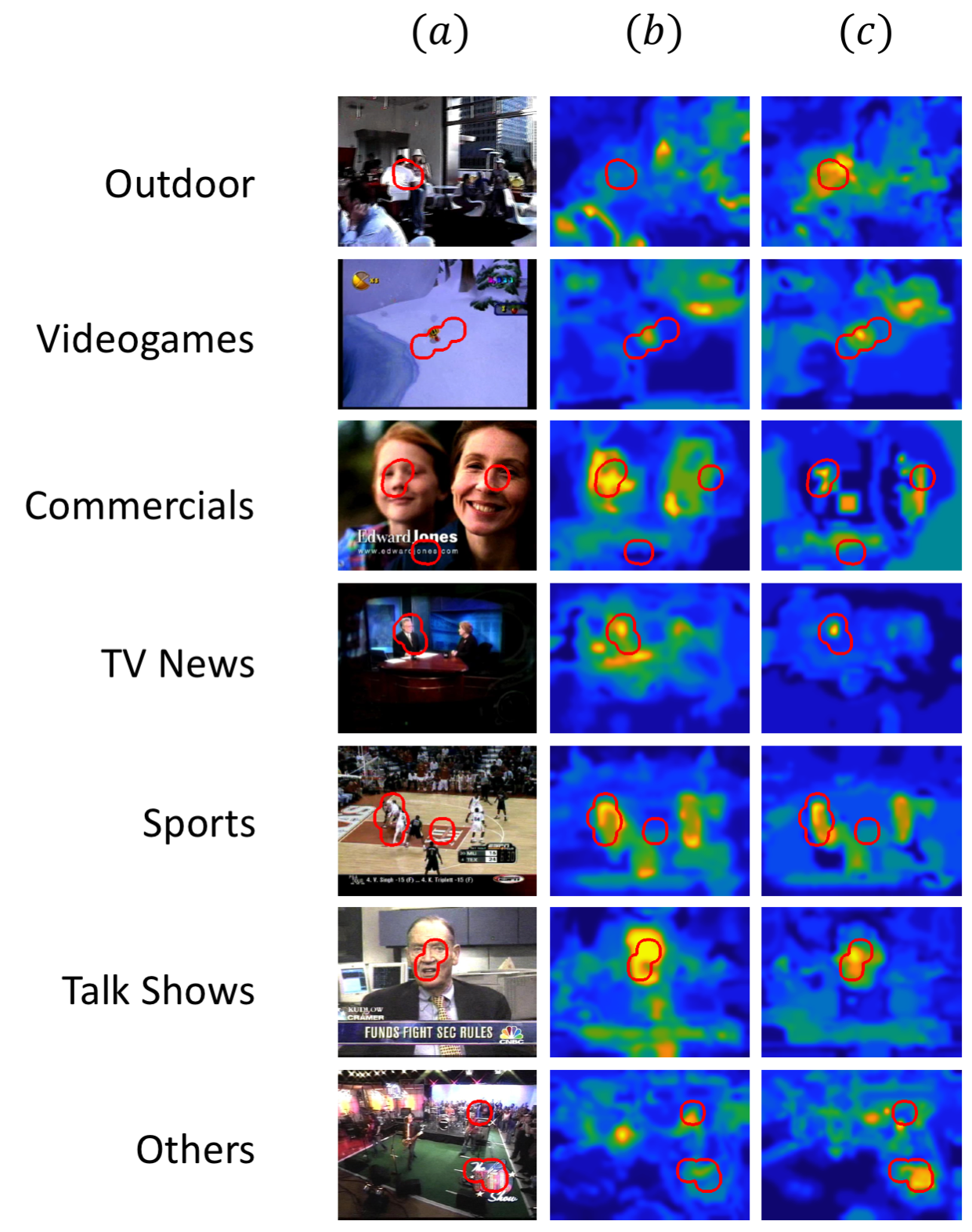}
		% \vspace{0cm}
	\caption[Visual attention maps obtained by \acs{ATOM} for some example frames from CRCNS-ORIG \cite{Itti_Carmi09crcns} database.]{Visual attention maps obtained by \acs{ATOM} for some example frames from CRCNS-ORIG \cite{Itti_Carmi09crcns} database. Red boundaries highlight high-density regions of human fixations in the \acs{GT} map. (a) Original frames. (b) Context-Generic. (c) Context-Aware.}\label{fig:54_examples_orig}
\end{figure*}

\begin{figure*}[!t]
\centering
		\includegraphics[trim=2cm 0.5cm 4cm 0cm, clip, width=1\textwidth]{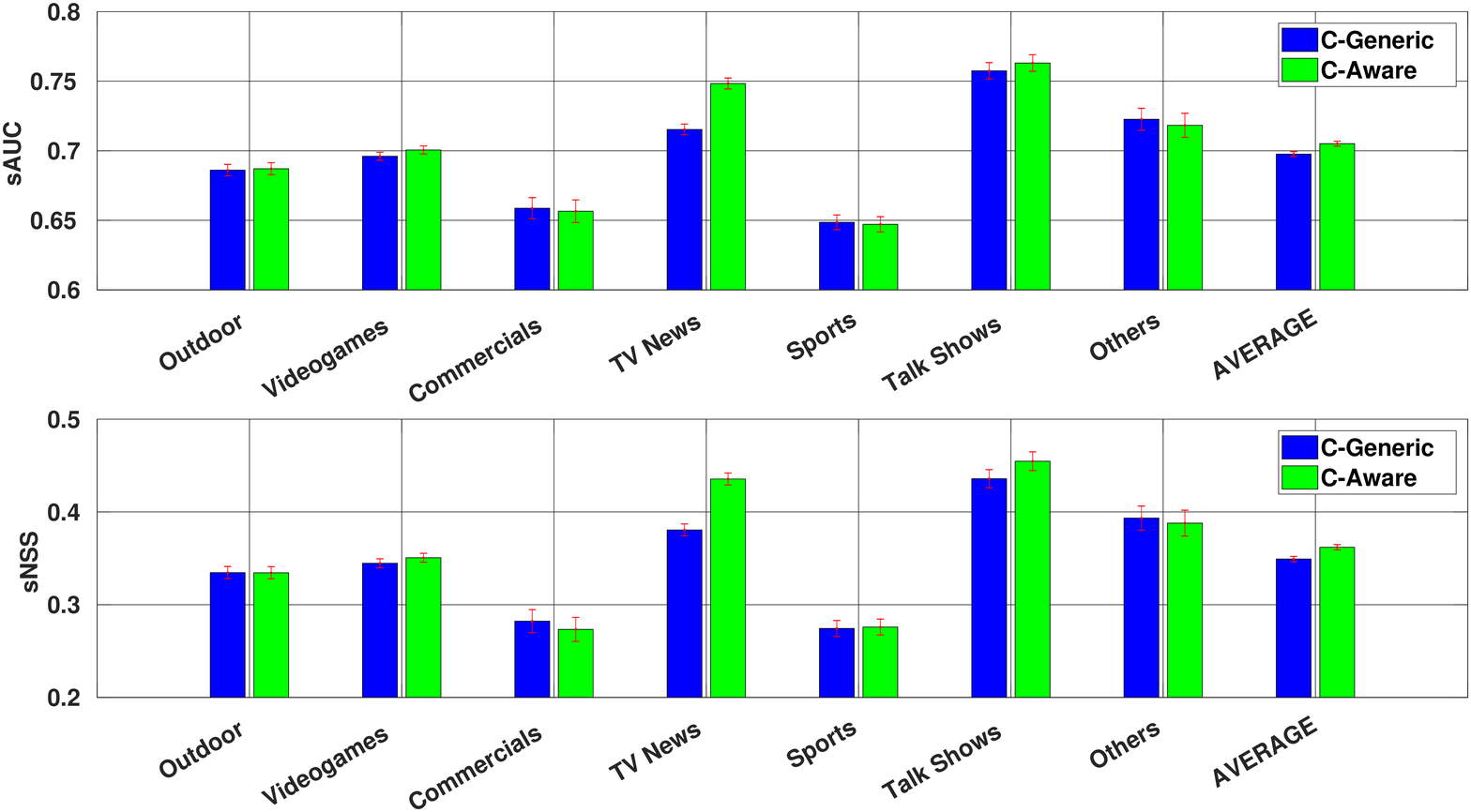}
		% \vspace{0cm}
	\caption[Results obtained by the proposed \emph{context-generic} and \emph{context-aware} ATOM models in the CRCNS-ORIG \cite{Itti_Carmi09crcns} database.]{Results obtained by the proposed \emph{context-generic} and \emph{context-aware} ATOM models in the CRCNS-ORIG \cite{Itti_Carmi09crcns} database, which consist of $K=60$ topics.}\label{fig:54_results_ORIG}
\end{figure*}

\begin{figure*}[!t]
\centering
		\includegraphics[trim=0cm 0cm 0cm 0cm, width=1\textwidth]{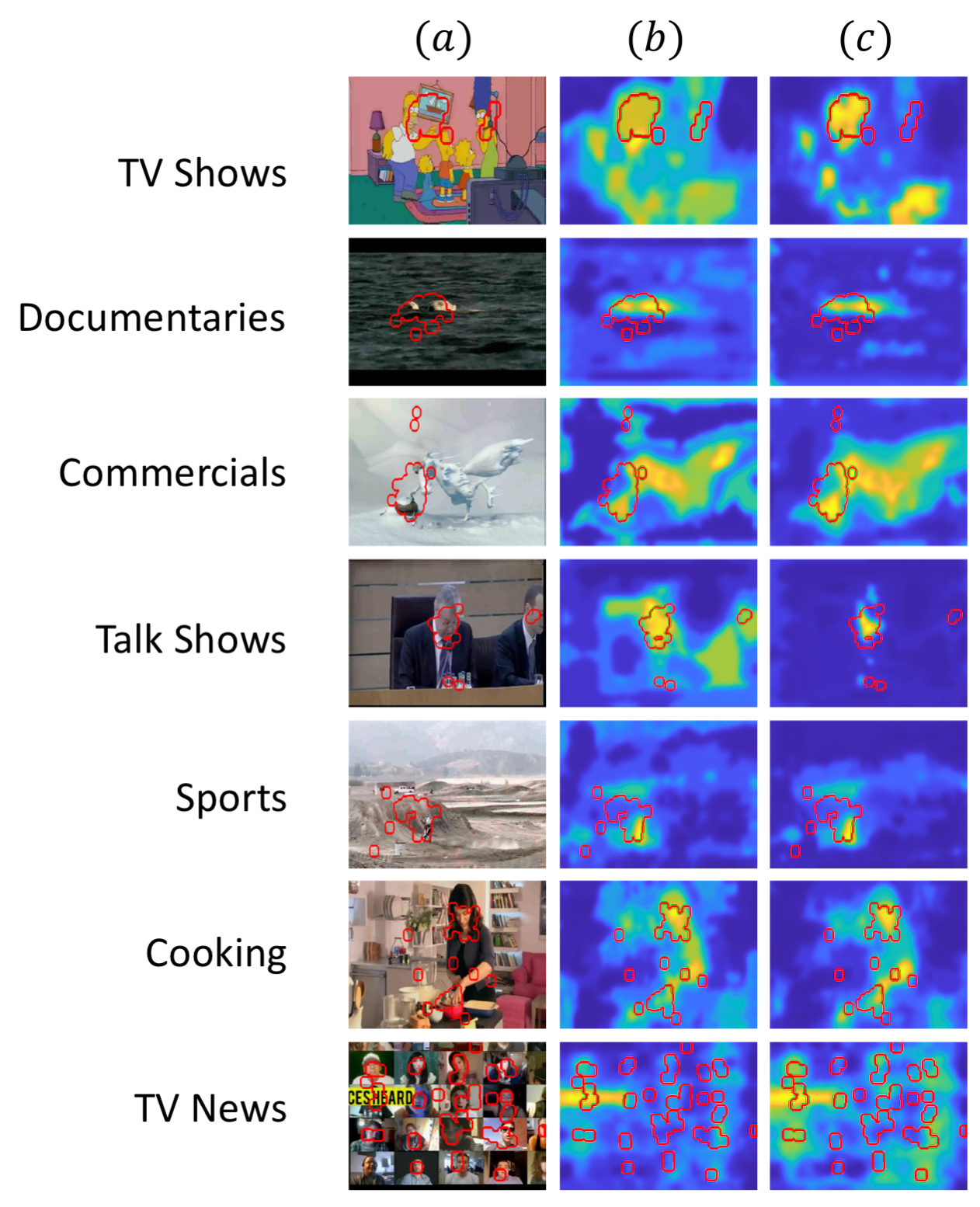}
		% \vspace{0cm}
	\caption[Visual attention maps obtained by \acs{ATOM} for some example frames from DIEM \cite{mital2011clustering} database.]{Visual attention maps obtained by \acs{ATOM} for some example frames from DIEM \cite{mital2011clustering} database. Red boundaries highlight high-density regions of human fixations in the \acs{GT} map. (a) Original frames. (b) Context-Generic. (c) Context-Aware.}\label{fig:54_examples_diem}
\end{figure*}

\begin{figure*}[!t]
\centering
		\includegraphics[trim=2cm 0cm 4cm 0cm, clip, width=1\textwidth]{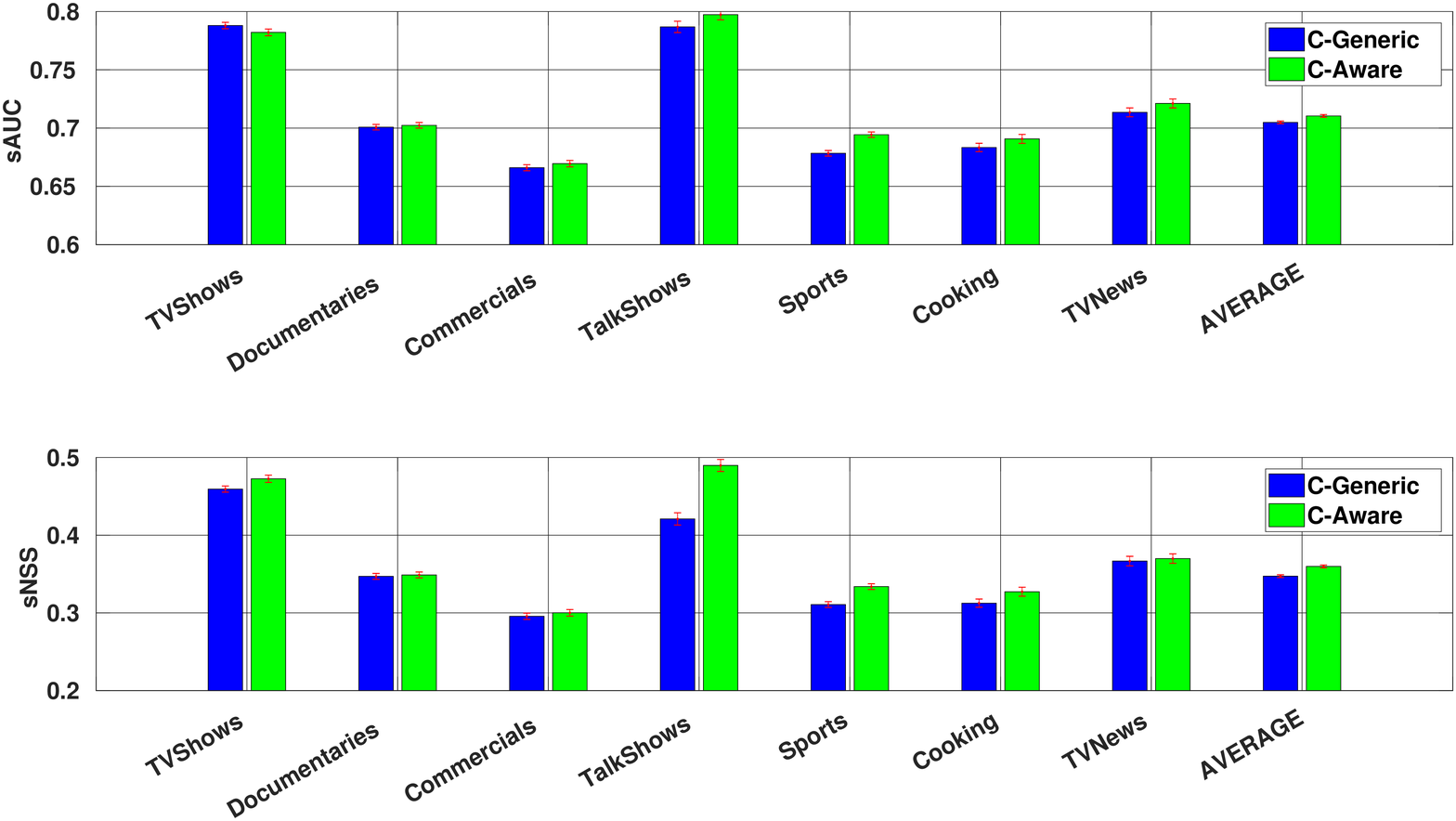}
		% \vspace{0cm}
	\caption[Results obtained by the proposed \emph{context-generic} and \emph{context-aware} ATOM models in the DIEM \cite{mital2011clustering} database.]{Results obtained by the proposed \emph{context-generic} and \emph{context-aware} ATOM models in the DIEM \cite{mital2011clustering} database, which consist of $K=60$ topics.}\label{fig:54_results_DIEM}
\end{figure*}

\section{Results on visual attention estimation}\label{sec:results5}
%\subsubsection*{Results on DIEM database}\label{sec:results_orig}
%Results obtained for the two versions of our method in each category are provided in Figure \ref{fig:54_results_DIEM}).
%\subsubsection*{Results on CRCNS-ORIG database}\label{sec:results_orig}
In this second set of experiments, \acsp{CNN}-based features are included and the \acs{ATOM} model learns unconstrained Normal distributions without fixating the means.

Results obtained for the two versions of our method in each category  and for each database are provided in Figures \ref{fig:54_results_ORIG} and \ref{fig:54_results_DIEM}, respectively. As can be seen, the \emph{context-aware} models match or outperform the \emph{generic} approach in all genres. Without considering \emph{Others} category in CRCNS-ORIG \cite{Itti_Carmi09crcns} database, which is more diverse and contains a synthetic saccade test video, best scores are obtained for \emph{TV News}, \emph{TVShows} and \emph{Talk Shows} genres, due to the high impact of object detectors (faces, pedestrians) in this genres, as shown in some of the examples provided in Figures \ref{fig:54_examples_orig} and \ref{fig:54_examples_diem}. Scores achieved for \emph{Outdoor} and \emph{Videogames} videos are also remarkable, due to the strong influence assigned to motion-related features. This reinforces the idea that, depending on the context, certain particular sub-tasks aid to guide visual attention. This can be also noticed if we look at the results obtained in categories such as \emph{Others} or \emph{Commercials}, whose associated videos cover a wide variety of contents and thus are not closely related, so consequently it has been hard to find out meaningful topics. In fact, the results for the \emph{context-generic} model in these cases are higher. From our point of view, this undesired effect might come from the fact that \acs{C-G} has been trained on a wider set of videos than \acs{C-A} approaches, and therefore has got better generalization. Therefore, it can be concluded that it is necessary to establish well-defined application scenarios where to determine these feature-based representations. In order to provide a fair comparison, we draw on the same number of topics for each of the categories in the dataset chosen, although it has been observed that the performance also depends on the complexity of the scenarios. If we compare the average performance of \emph{context-aware} models with respect to the result obtained by the \emph{context-generic} approach, there is an improvement of $4.6\%$ in terms of \acs{sNSS} and $1.1\%$ in terms of \acs{sAUC}, which is closer to the upper threshold given by H50 score. 

Thus, we can state that specific \emph{context-aware} representations of visual attention learned over particularized training sets (the training videos belonging to each category) work better than \emph{generic} models learned over larger general datasets (including all video categories). Based on these results, from now on we will use the \emph{context-aware} version of our algorithm to provide a comparison with other approaches in the state-of-the-art.

\begin{table}[t]
\caption{Comparison with state-of-the-art methods in the CRCNS-ORIG \cite{Itti_Carmi09crcns} database.}\label{tab:soa_results_orig}
\begin{center}
%\vspace{-0.25cm}
    \resizebox{11.5cm}{!} {
    \begin{tabular}{l|c|cc}
    \hline 
    \multicolumn{1}{c|}{\multirow{2}{*}{Model}} & \multirow{2}{*}{\centering Learning} &
    \multicolumn{1}{p{3cm}}{\centering $sAUC$} & \multicolumn{1}{p{3cm}}{\centering $sNSS$} \\ 
    & & \multicolumn{1}{p{3cm}}{\centering \scriptsize $mean \ (C.I.)^{Rank}$} & \multicolumn{1}{p{3cm}}{\centering \scriptsize $mean \ (C.I.)^{Rank}$} \\ \hline
    {\acs{ATOM}} & YES & $\bf 0.705 \ (0.703, 0.707)^{1}$ &  $\bf 0.362 \ (0.359, 0.365)^{1}$ \\ 
    {\acs{AWS-D} \cite{GallegaSaliency}} & NO & $0.700 \ (0.698, 0.702)^{2}$ &  $0.322 \ (0.319, 0.325)^{3}$ \\
    {DCL \cite{DeepSaliencyObject}} & YES & $0.684 \ (0.682, 0.686)^{3}$ &  $0.323 \ (0.320, 0.326)^{2}$ \\  
    {\acs{AWS} \cite{GallegaSaliency2012}} & NO & $0.675 \ (0.674, 0.677)^{4}$ &  $0.281 \ (0.278, 0.285)^{4}$ \\
    {\acs{WMAP} \cite{lopez2011scene}} & NO & $0.670 \ (0.669, 0.672)^{5}$ &  $0.236 \ (0.232, 0.239)^{12}$ \\ 
    {Hou and Zhang \cite{HouZhang}} & NO & $0.669 \ (0.667, 0.671)^{6}$ &  $0.260 \ (0.257, 0.263)^{7}$ \\  
    {DCL+ \cite{DeepSaliencyObject}} & YES & $0.666 \ (0.665, 0.668)^{7}$ &  $0.255 \ (0.251, 0.258)^{8}$ \\ 
    {\acs{ICL}-D \cite{hou2009dynamic}} & NO & $0.666 \ (0.665, 0.668)^{8}$ &  $0.217 \ (0.214, 0.220)^{14}$ \\
    {\acs{PQFT} \cite{5223506}} & NO & $0.662 \ (0.660, 0.663)^{9}$ &  $0.243 \ (0.240, 0.246)^{11}$ \\ 
    {Goferman \cite{Goferman}} & NO & $0.661 \ (0.659, 0.662)^{10}$ &  $0.263 \ (0.260, 0.266)^{6}$ \\
    {\acs{SUN} \cite{doi:10.1167/8.7.32}} & YES & $0.654 \ (0.652, 0.655)^{11}$ &  $0.251 \ (0.248, 0.254)^{9}$ \\
    {\acs{AIM} \cite{BruceT05}} & YES & $0.653 \ (0.652, 0.655)^{12}$ &  $0.270 \ (0.268, 0.273)^{5}$ \\
    {Torralba \cite{Torralba:03}} & NO & $0.648 \ (0.646, 0.650)^{13}$ &  $0.251 \ (0.248, 0.254)^{10}$ \\
    {Itti (ST) \cite{Itti:1998:MSV:297843.297870} \cite{Harel07graph-basedvisual}} & NO & $0.634 \ (0.632, 0.636)^{14}$ &  $0.217 \ (0.214, 0.220)^{15}$ \\
    {Fern\'andez-Torres \cite{cbmi2016miguel}} & YES & $0.628 \ (0.626, 0.630)^{15}$ &  $0.218 \ (0.215, 0.221)^{13}$ \\ 
    {\acs{SDSR} \cite{SeoMilanfar}} & NO & $0.627 \ (0.625, 0.628)^{16}$ &  $0.129 \ (0.126, 0.132)^{17}$ \\
    {\acs{GBVS} (ST) \cite{Harel07graph-basedvisual}} & NO & $0.621 \ (0.619, 0.623)^{17}$ &  $0.182 \ (0.179, 0.186)^{16}$ \\
    {ESA-D \cite{Rahtu2010}} & NO & $0.541 \ (0.539, 0.543)^{18}$ &  $0.075 \ (0.072, 0.078)^{18}$ \\ \hline
    {H50} & $NO$ & $0.800 \ (0.799, 0.802)$ &  $0.679 \ (0.677, 0.681)$ \\
    {CHANCE} & $NO$ & $0.500 \ (0.500, 0.500)$ &  $-0.000 \ (-0.000, 0.000)$ \\
    {CENTER} & $NO$ & $0.509 \ (0.507, 0.511)$ &  $0.057 \ (0.054, 0.060)$ \\ \hline
    \end{tabular}
    }
    \end{center}
\end{table}

\begin{table}[t]
\caption{Comparison with state-of-the-art methods in the DIEM \cite{mital2011clustering} database.}\label{tab:soa_results_diem}
\begin{center}
%\vspace{-0.25cm}
    \resizebox{11.5cm}{!} {
    \begin{tabular}{l|c|cc}
    \hline 
    \multicolumn{1}{c|}{\multirow{2}{*}{Model}} & \multirow{2}{*}{\centering Learning} &
    \multicolumn{1}{p{3cm}}{\centering $sAUC$} & \multicolumn{1}{p{3cm}}{\centering $sNSS$} \\ 
    & & \multicolumn{1}{p{3cm}}{\centering \scriptsize $mean \ (C.I.)^{Rank}$} & \multicolumn{1}{p{3cm}}{\centering \scriptsize $mean \ (C.I.)^{Rank}$} \\ \hline
    {\acs{ATOM}} & YES & $\bf 0.710 \ (0.709, 0.712)^{1}$ &  $\bf 0.360 \ (0.358, 0.362)^{1}$ \\ 
    {\acs{AWS-D} \cite{GallegaSaliency}} & NO & $0.701 \ (0.700, 0.701)^{2}$ &  $0.319 \ (0.317, 0.320)^{3}$ \\
    {DCL \cite{DeepSaliencyObject}} & YES & $0.695 \ (0.695, 0.696)^{3}$ &  $0.341 \ (0.340, 0.342)^{2}$ \\  
    {DCL+ \cite{DeepSaliencyObject}} & YES & $0.683 \ (0.682, 0.683)^{4}$ &  $0.318 \ (0.317, 0.319)^{4}$ \\ 
    {\acs{WMAP} \cite{lopez2011scene}} & NO & $0.666 \ (0.666, 0.667)^{5}$ &  $0.233 \ (0.232, 0.234)^{12}$ \\ 
    {Hou and Zhang \cite{HouZhang}} & NO & $0.663 \ (0.662, 0.664)^{6}$ &  $0.247 \ (0.246, 0.248)^{9}$ \\  
    {\acs{PQFT} \cite{5223506}} & NO & $0.662 \ (0.661, 0.663)^{7}$ &  $0.235 \ (0.233, 0.236)^{11}$ \\ 
    {Goferman \cite{Goferman}} & NO & $0.659 \ (0.658, 0.660)^{8}$ &  $0.257 \ (0.256, 0.258)^{6}$ \\
    {\acs{GBVS} (ST) \cite{Harel07graph-basedvisual}} & NO & $0.653 \ (0.652, 0.653)^{9}$ &  $0.256 \ (0.255, 0.257)^{7}$ \\
    {\acs{AWS} \cite{GallegaSaliency2012}} & NO & $0.652 \ (0.651, 0.653)^{10}$ &  $0.271 \ (0.270, 0.272)^{5}$ \\
    {Itti (ST) \cite{Itti:1998:MSV:297843.297870} \cite{Harel07graph-basedvisual}} & NO & $0.638 \ (0.637, 0.639)^{11}$ &  $0.253 \ (0.252, 0.254)^{8}$ \\
    {Torralba \cite{Torralba:03}} & NO & $0.636 \ (0.635, 0.636)^{12}$ &  $0.232 \ (0.231, 0.233)^{13}$ \\
    {\acs{SUN} \cite{doi:10.1167/8.7.32}} & YES & $0.631 \ (0.630, 0.631)^{13}$ &  $0.220 \ (0.219, 0.221)^{14}$ \\
    {Fern\'andez-Torres \cite{cbmi2016miguel}} & YES & $0.630 \ (0.629, 0.631)^{14}$ &  $0.219 \ (0.217, 0.222)^{15}$ \\
    {\acs{ICL}-D \cite{hou2009dynamic}} & NO & $0.629 \ (0.628, 0.629)^{15}$ &  $0.154 \ (0.153, 0.155)^{16}$ \\ 
    {\acs{AIM} \cite{BruceT05}} & YES & $0.618 \ (0.617, 0.618)^{16}$ &  $0.238 \ (0.237, 0.239)^{10}$ \\
    {ESA-D \cite{Rahtu2010}} & NO & $0.563 \ (0.562, 0.564)^{17}$ &  $0.150 \ (0.149, 0.151)^{17}$ \\
    {\acs{SDSR} \cite{SeoMilanfar}} & NO & $0.531 \ (0.530, 0.532)^{18}$ &  $0.022 \ (0.021, 0.023)^{18}$ \\ \hline
    {H50} & $NO$ & $0.827 \ (0.827, 0.827)$ &  $0.662 \ (0.662, 0.663)$ \\
    {CHANCE} & $NO$ & $0.500 \ (0.500, 0.500)$ &  $-0.000 \ (-0.000, 0.000)$ \\
    {CENTER} & $NO$ & $0.503 \ (0.502, 0.504)$ &  $0.054 \ (0.052, 0.055)$ \\ \hline
    \end{tabular}
    }
    \end{center}
\end{table}

\section{Comparison with state-of-the-art methods}\label{sec:comparison_soa5}
\begin{figure}[!t]
\centering
\includegraphics[trim=0cm 0cm 0cm 0cm, clip, width=1\linewidth]{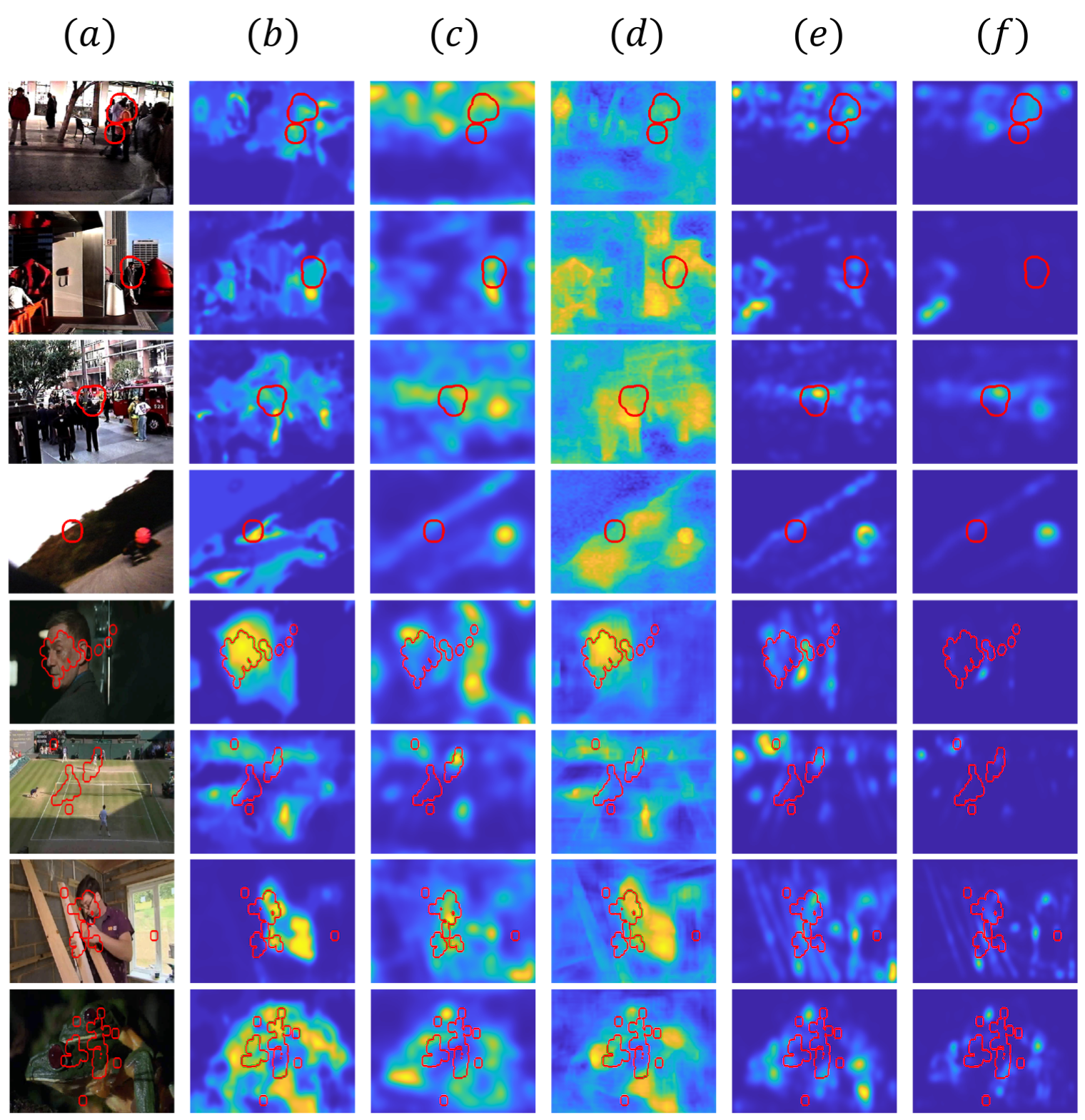}
\caption[Visual attention maps generated by some of the most outstanding methods in the \emph{state-of-the-art} for some intricate example frames taken from CRCNS-ORIG \cite{Itti_Carmi09crcns} and DIEM \cite{mital2011clustering} databases.]{Visual attention maps generated by some of the most outstanding methods in the \emph{state-of-the-art} for some intricate example frames taken from CRCNS-ORIG \cite{Itti_Carmi09crcns} and DIEM \cite{mital2011clustering} databases. Red boundaries highlight high-density regions of human fixations in the \acs{GT} map. (a) Original frames. (b) \acs{ATOM}. (c) \acs{AWS-D} \cite{GallegaSaliency}. (d) DCL \cite{DeepSaliencyObject}. (e) \acs{WMAP} \cite{lopez2011scene}. (f) \acs{ICL}-D \cite{hou2009dynamic}.}
\label{fig:55_soa_orig_diem}
\end{figure}

With the aim of assessing the performance of our approach in comparison with other methods available in the state-of-the-art, we have selected 17 static and dynamic visual attention models, which are representative of the existing diversity for visual attention prediction: we have included both \acs{BU} and \acs{TD} or learnable models, a model that uses \acsp{CNN} to predict, etc., as well as the three reference models introduced in section \ref{sec:atom_evaluation_metrics} (H50, CHANCE, CENTER). Parameters used are the ones set as default by authors. As can be verified from CENTER baseline, both metrics included in the analysis are not affected by center bias effect. 

Tables \ref{tab:soa_results_orig} and \ref{tab:soa_results_diem} contain all the results obtained for the assessed methods in CRCNS-ORIG \cite{Itti_Carmi09crcns} and DIEM \cite{mital2011clustering} databases, respectively, together with those reached by the system proposed in Chapter \ref{ch:atom} (\acs{ATOM}). We also include on the list the first approach we presented in \cite{cbmi2016miguel}, which make use of a linear regressor to estimate visual attention instead of the logistic regressor currently employed, as well as other features. Features and number of topics ($K=40$) taken for this previous configuration are those reported in \cite{cbmi2016miguel}.

The improvement achieved by our model with respect to very recent approaches such as \acs{AWS-D} \cite{GallegaSaliency}, DCL \cite{DeepSaliencyObject}, \acs{WMAP} \cite{lopez2011scene} or \acs{ICL}-D \cite{hou2009dynamic} is statistically significant. Moreover, it is also visually noticeable in some intricate cases, as those shown in Figure \ref{fig:55_soa_orig_diem}, with scenes showing crowds, multiple similar concepts that hamper visual guidance or quick actions.

Finally, we evaluate the computational time on a system with an Intel Core i7-6700K \acs{CPU} at 4.00GHz and with 32GB of \acs{RAM}. Regarding our approach, we should distinguish between the learning and the test phase. Both phases involve a feature extraction stage that takes $5.81s$ per frame, which has not been optimized, and could be highly parallelized by GPUs. Time spent in the learning phase depends on the number of topics of the model trained and the amount of input frames. For instance, training a model with $K=60$ topics and $\sim 3000$ frames would take $\sim 45min$. This time can be reduced if the number of topics is decreased to $K=40$ ($\sim 32min$) or $K=20$ ($\sim 18min$), which would slightly decrease the performance. Then, in the test phase, the average time per frame is only $0.157s$, which is competitive compared to those obtained by the two next best methods, \acs{AWS-D} \cite{GallegaSaliency} ($0.075s$) and DCL \cite{DeepSaliencyObject} ($0.2s$).

\section{Where we are: model strengths and limitations}\label{sec:where_we_are5}
Despite the improvement reached by the proposed model over the \emph{state-of-the-art} and the compelling information it provides, we are still far from reaching human capacity of almost immediately selecting the most essential elements and areas to reach a full understanding in a given scenario, or to solve a particular task, according to the H50 score reflected in Tables \ref{tab:soa_results_orig} and \ref{tab:soa_results_diem}. Nonetheless, we advocate that the inclusion of an intermediate level between features and visual attention in terms of sub-tasks is a powerful way towards comprehensible guiding representations. 

%In order to assess the influence of the topic models over the final result, we have evaluated an alternative method that uses a logistic regressor over the same set of features to directly predict visual attention. Our topic model achieves a relative improvement of $22.3\%$ in terms of \acs{sNSS} and $1.7\%$ in terms of \acs{sAUC}. This clearly demonstrates that the topic-based hierarchical modeling is useful, not only because it provides meaningful representations of top-down visual attention, but also because it successfully enhances the system performance. 

\begin{figure*} %[H]
\centering
\includegraphics[trim=0cm 0cm 0cm 0cm, clip, width=0.72\linewidth]{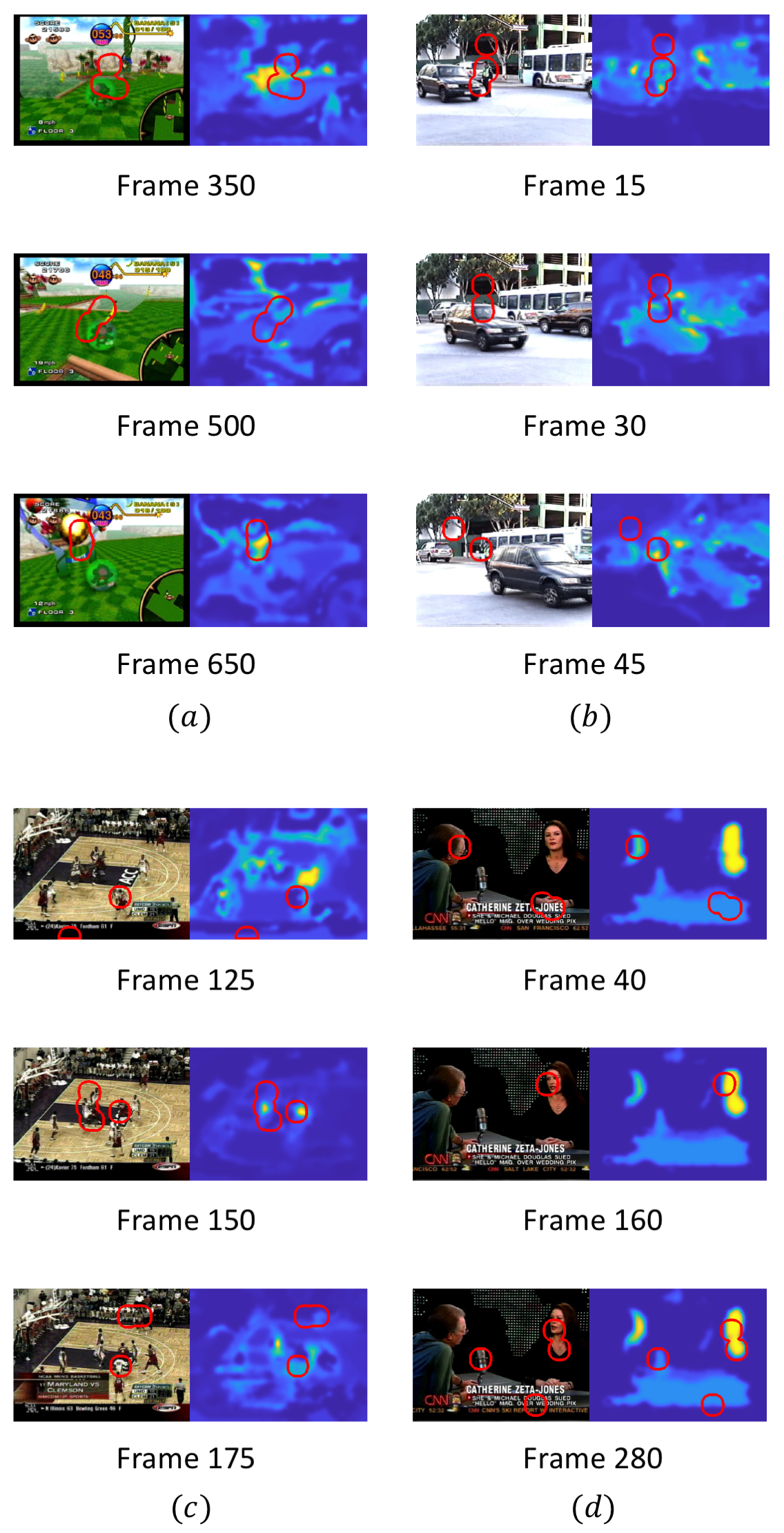}
\caption[Frame sequences taken from CRCNS-ORIG \cite{Itti_Carmi09crcns} database to analyze some \acs{ATOM} model drawbacks and define future lines of research.]{Frame sequences taken from CRCNS-ORIG \cite{Itti_Carmi09crcns} database to analyze some \acs{ATOM} model drawbacks and define future lines of research. Red boundaries highlight high-density regions of human fixations in the \acs{GT} map, both in original frames and computed visual attention maps. (a) Videogames scenario where many remarkable regions exist, making observers constantly shift their gaze. (b) Outdoor scenario where multiple salient concepts (e.g. car, policeman) overlap each other. (c) Basketball match, in which the sequence of players to follow is decisive to model visual attention. (d) TV talk show, where several quasi-static concepts appear together during a long time lapse and estimated visual attention is either distributed among all or focused in one of them.}
\label{fig:56_error_analysis}
\end{figure*}

We have demonstrated that some of the traditional basic features used (e.g. color, orientation, motion) are still useful in many cases to predict visual attention in videos. Furthermore, thanks to the object detectors introduced and the corresponding spatial discrete distributions, we are able to model simple but attractive concepts such as faces or text, putting emphasis on their most noticeable elements. The high performance achieved by these detectors in some categories leads us to reckon the integration of large-scale hierarchical networks for object recognition in future revisions of our model, such as the ones evaluated in the ImageNet Challenge \cite{ILSVRC15}. In addition, there is also a need of a deeper understanding of the scene, establishing relations between recognized concepts both in the same frame or in different frames. This would enable the system to enhance guidance in situations where many conspicuous regions exist and selecting the most significant in the task to be solved, or even an intermediate one (e.g. Figure \ref{fig:56_error_analysis}(a)); when objects are occluded during few frames (e.g. Figure \ref{fig:56_error_analysis}(b)); or to determine the sequence of objects or subjects to follow in order to interpret a scene (e.g. Figure \ref{fig:56_error_analysis}(c)), among others. In other words, we pursue the identification and modeling of sub-tasks, not only over space but also along time.

%\begin{figure}[!htb]
%\centering
%% \vspace{-0.25cm}
%\includegraphics[trim=1cm 0cm 0.75cm 0cm, width=0.95\linewidth]{images/error_analysis.png}
%\caption{Frame sequences taken from CRCNS-ORIG \cite{Itti_Carmi09crcns} database to analyze some ATOM model drawbacks and define future lines of research. Red boundaries highlight high-density regions of human fixations in the GT map, both in original frames and computed visual attention maps. (a) Videogames scenario where many remarkable regions exist, making observers constantly shift their gaze. (b) Outdoor scenario where multiple salient concepts (e.g. car, policeman) overlap each other. (c) Basketball match, in which the sequence of players to follow is decisive to model visual attention. (d) TV talk show, where several quasi-static concepts appear together during a long time lapse and estimated visual attention is either distributed among all or focused in one of them.}
%\label{fig_error_analysis}
%\vspace{-0.5cm}
%\end{figure}

Finally, the importance of \acs{GT} eye fixations has to be discussed, both in learning and evaluation stages. As it can be appreciated in some of the examples gathered throughout the chapter, not all fixations contain useful information to train a visual attention system, not only because the occlusions mentioned above, but also due to errors during the eye-tracker acquisition or to the observers' center bias present in many frames. Fixations often fall on edges, not covering completely some objects of interest, such as gaming characters or players, which are essential to infer sub-tasks. Additionally, we might take into account covert attention, which is independent of eye movements and stresses the existence of attention independent of gaze change. Hence, techniques to filter and, if necessary, to extend regions considered as \acs{GT} should be regarded in upcoming experiments. What is more, existing evaluation metrics do not seem to be appropiate in situations such as the one shown in Figure \ref{fig:56_error_analysis}(d), where many remarkable quasi-static concepts appear together during a long time lapse and estimated visual attention is either distributed among all or focused in one of them. If observers' fixations are widely dispersed and attention is switched between various locations, what should be the \acs{GT} taken for each frame in this case? Should all concepts be considered as attracting during the whole video fragment? We will seek to address these issues in future application scenarios. 

\section{Conclusions}\label{sec:conclusions5}
In Chapters \ref{ch:atom} and \ref{ch:atom_experiments}, we have presented a hierarchical probabilistic framework to estimate and understand \acs{TD} visual attention in videos. Relying on the idea of `guiding representation' supported by some of the most prevailing psychological theories about visual attention, our \acs{ATOM} model decomposes it into mixtures of several latent topics or sub-tasks, which are in turn modeled as combinations of low-, mid- and high-level spatio-temporal features obtained from video frames. For that purpose, an intermediate level between feature extraction and visual attention computation phases is introduced, aligning the latent discovered sub-tasks from frames to the information drawn from human fixations. The attention response is thus generated by computing a logistic regression model over topic proportions. It is also worth mentioning that the definition of the method is generic and independent of the input features, which enables an easy adaptation to any application scenario. 

The ability of \acs{ATOM} to successfully learn specifically adapted hierarchical representations of visual attention in diverse contexts has been demonstrated on the basis of a wide set of features. Either classical and easily interpretable feature maps, which have been effective to extract conclusions about the existing scenarios in the well-known CRCNS-ORIG \cite{Itti_Carmi09crcns} and DIEM \cite{mital2011clustering} databases, or those generated by recently adopted \acsp{CNN} structures, which allow to capture more complex concepts, have aided to significantly outperform other competent methods in the literature. Moreover, the detection of simple elements such as faces or text, and their modeling through spatial discrete distributions, has led to improve visual attention estimation in certain challenging situations. 

Experimental results show the advantage of obtaining comprehensible guiding representations to model visual attention. However, it is still necessary to deepen in some of the stages of the framework, carefully selecting the most meaningful information from fixated regions in the scene, and integrating more robust recognition and understanding techniques that enable to identify more accurate sub-tasks over space and time. To that end, future efforts will be directed towards task-driven approaches, developing video databases with human fixations to test the usefulness of the system in end-user applications. 

%*****************************************
%*****************************************
%*****************************************
%*****************************************
%*****************************************

%************************************************
\chapter{Deep Neural Networks for modeling visual attention in the temporal domain}\label{ch:anomaly_detection} 
\chaptermark{DNNs for visual attention in the temporal domain}
%************************************************

\section{Introduction}
Observers' eye movements constitute a useful source to understand how visual attention works and, consequently, what information should be selected for further processing. Given a complex and crowded scenario, if all observers fix their attention at the same location at the same time, it is very likely that something noticeable is happening.

\begin{figure*}[!t]
\centering
	\includegraphics[trim=0cm 0cm 0cm 0cm, width=1\textwidth]{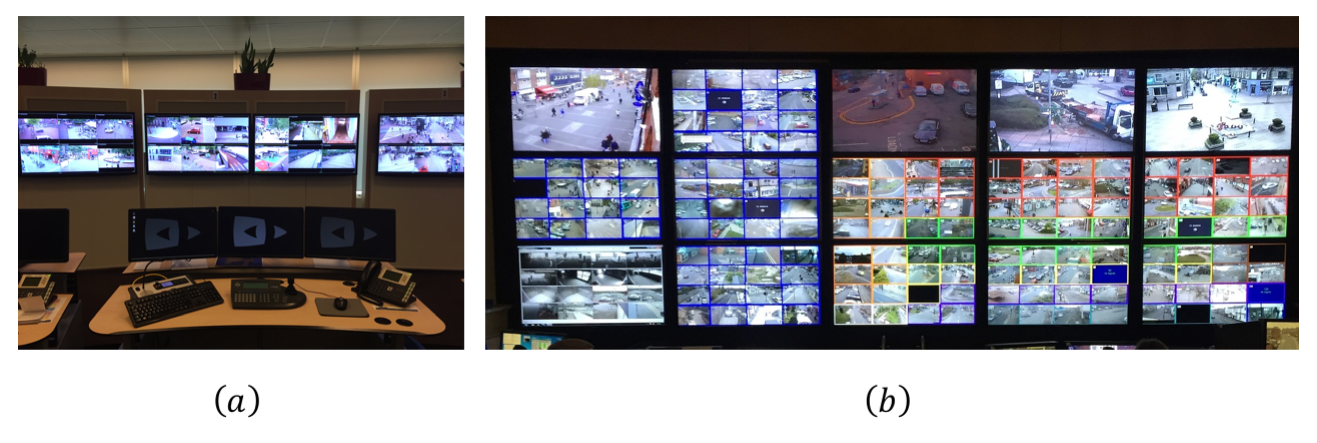}	
\caption[a) Typical video surveillance monitoring room. (b) The task of a \acs{CCTV} operator is to find a possible anomalous event amongst multiple distractors, displayed at the same time in a large array of more than 20 screens.]{(a) Typical video surveillance monitoring room. Image taken from \cite{cctv1}. (b) The task of a \acs{CCTV} operator in a video monitoring room is to find a potentially anomalous event amongst multiple distractors, displayed at the same time in a large array of more than 20 screens. Anomalous events seldom happen, which makes this task even more complex to solve. We want to meet this challenge, taking advantage of eye fixations. Image taken from \cite{cctv2}.}
\label{fig:5_cctv}
% \vspace{-0.4cm}
\end{figure*}

Photographs in Figure \ref{fig:5_cctv}(a) illustrate a typical video monitoring room. The task of \acs{CCTV} operators in this scenario is to find a potentially anomalous event (e.g. robberies, road accidents, etc.) amongst multiple distractors (e.g. crowds, similar vehicles, etc.) displayed at the same time in a large array of $20$ to more than $500$ screens \cite{stainer2013looking}, as the one shown in Figure \ref{fig:5_cctv}(b). At what screen should they look each time? Operators often have also to review many hours of surveillance recordings. Moreover, anomalous events seldom happen, which makes these tasks even more difficult to solve. How do they tackle these tasks? Is it possible to develop a system to aid experts to perform them more efficiently? We want to meet these challenges, taking advantage of eye fixations.

In this chapter, inspired by the recent success of \acfp{CNN} for learning deep hierarchical image representations and \acf{LSTM} units for time series forecasting, we propose a network architecture that goes from spatio-temporal visual attention prediction to temporal attention estimation. Visual attention in the temporal domain can be understood as a filtering mechanism, which allows to select time segments of special importance in video sequences.

Supported by the fact that eye fixation sequences of different viewers correlate well when an important or anomalous event happens, our system models visual attention over time as a fixation-based response. Hence, it could be used to prevent human errors and speed up decision making processes in real applications which require watching large amounts of visual information, at the same time and during long time periods, such as the task of video surveillance.

\section*{Chapter overview}
The chapter is organized as follows. First, in Section \ref{sec:tva_related_work}, we discuss about the importance of eye tracking to understand the behavior of experts performing real-world tasks. We review also the most relevant and recent work in visual attention estimation applying deep architectures, and present our main contributions. Then, an introduction to deep learning is carried out in the Section \ref{sec:deep_neural_networks}, describing primarily those \acsp{DNN} and techniques used for our research. Next, three feature learning architectures for attention guidance are described in Section \ref{sec:computational_cnn}, which will provide input feature maps to our system for modeling attention in the temporal domain. Afterwards, we fully describe in detail the proposed system in Section \ref{sec:statten}. For that purpose, we first introduce our assumptions about task-driven visual attention, making an overview of the complete architecture in Section \ref{sec:statten_overview}. Later, we define the process followed to generate a fixation-based temporal \acs{GT} in Section \ref{sec:fixations_gt}. Finally, we describe the complete architecture design in Section \ref{sec:architectures6}, and elaborate the two stages of our system in Sections \ref{sec:stva_autoencoder} and \ref{sec:tva_modeling}.

\section{Related work}\label{sec:tva_related_work}
Following the early experiments of Yarbus in 1967 \cite{Yarbus1967}, who concluded that visual attention is ultimately task- or goal-driven, many other investigations with still stimuli proved this statement \cite{henderson1999high, defendingYarbus}; they claim that it might be attainable to infer the attentional processes carried out by the \acs{HVS} from eye movement sequences. This has motivated researchers during the last two decades to evaluate the possibilities of eye tracking in real applications such as driving safety \cite{pradhan2005using}, aviation \cite{kilingaru2013monitoring}, production and industry control \cite{sharma2016eye}, health-care \cite{hermens2013eye} and video surveillance \cite{howard2011task}.

Visual information is constantly being updated in videos and, consequently, not all the assumptions that hold in still images can be extrapolated to videos recorded in real situations. Indeed, a recent study with almost 150 participants monitoring footage in a \acf{CCTV} crime scenario \cite{graham2018cctv} has concluded, after conducting experiments with task-oriented and non-task-oriented observers, that the complexity of dynamic environments may decrease the influence of in-advanced instructions, in contrast to what happens with static images. 

Despite the above issues, high-valuable information is still appreciated when examining fixations behavior: their typical quasi-random pattern changes just before and during a significant or suspicious event. Moreover, there is a strong correlation between fixations of different viewers when these events happen in a similar scenario \cite{howard2013suspiciousness,sharma2016eye}, both in their location and duration. Taking into consideration this fact, we propose to take a step further by developing a system able to learn this behavior from sequences of fixations, which will be used to model the temporal dimension of visual attention. 

The models introduced in this chapter goes from spatio-temporal \acs{2D} visual attention maps, used so far for fixation prediction at every frame in a video, and transforms them into temporal \acs{1D} visual attention curves. These signals highlight relevant frames in a video, which often correspond to surprising or unusual events, frequently labeled as anomalies. Therefore, our ultimate objective now is not to understand how visual attention works, as in Chapters \ref{ch:atom} and \ref{ch:atom_experiments}, but to estimate visual attention in the temporal domain. In contrast to our previous approach, which drew on probabilistic \acsp{LTM} (see Section \ref{sec:latent_topic_models}), some preliminary experiments suggested the convenience of using \acsp{CNN} for better modeling spatio-temporal visual attention, as well as \acs{LSTM}-based architectures for modeling attention in the temporal domain.    

Spatio-temporal visual attention in real scenarios is still in its infancy, as we have remarked along this thesis. The exceptional, but still lacking in analysis, performance of \acsp{CNN} has brought some new approaches to this application. They mainly attend to three attributes: a) spatial RGB-based features; b) motion, modeled either by using optical flow information at the input of a \acs{CNN} \cite{bak2018spatio} or by means of recurrent \acs{LSTM} units \cite{dhf1k}; and c) objects, located in maps generated by multi-scale \acs{CONV} architectures \cite{ledov}.

Visual attention in the temporal domain has been even less tackled in the literature up to date. First works elementarily modeled temporal attention as the mean of the saliency values predicted at each spatial location \cite{ma2005generic,peng2009keyframe}. Similarly, Ejaz et al. \cite{ejaz2013efficient} modeled the temporal saliency of a frame as an average of temporal gradients, in order to select key frames in a video summarization application. More recently, Koutras et al. \cite{koutras2018audio} defined a simple Otsu's threshold operator to transform \acsp{SM} into saliency curves. Finally, Han et al. proposed in \cite{han2014spatial} a more sophisticated probabilistic supervised method for temporal visual attention, which aims to estimate movie trailers attractiveness. This model uses the same hypothesis than our work and computes a temporal \acs{GT} for each video frame based on the fixation dispersion across several observers. 

%a) Psychological basis
%Experiments on visual attention via eye tracking
%-Yarbus on still images: Instruction given to observers
%- Multi-view scenarios
%- Fixation patterns: duration, number of...
%- Correlation between observers

%b) Spatio-temporal visual attention
%LEDOV, DHF1K, LSTM...

%c) Temporal visual attention
%Still images:
%- *Scanpath prediction on 360 degree images using saliency volumes (Marc Assens et al., ICCV)

%d) Video surveillance
%How has been treated this application in computer vision so far?
%What is our approach? Why can be useful?

\section{Deep Neural Networks}\label{sec:deep_neural_networks}
Deep learning \cite{lecun2015deep} covers a wide set of computational models with multiple processing layers. Starting from large amounts of raw information and by means of a general-purpose learning procedure, they discover multi-level abstract representations of the data which are able to effectively solve a variety of tasks. 

Indeed, during the past five years, \acfp{DNN} \cite{goodfellow2016deep} have revolutionized multiple applications using \acs{ML} algorithms, such as speech recognition \cite{hinton2012deep}, object detection \cite{krizhevsky2012imagenet} and natural language understanding \cite{collobert2011natural}, dramatically improving the existing state-of-the-art performances. At present, the deep learning field is constantly growing, so new architectures and algorithms are being tested with the objective of either understanding the behavior of \acsp{DNN} or designing more robust and less time- and computational-consuming systems.

This section makes an introduction to \acsp{DNN}, mainly providing the description of the modules that comprise \acsp{CNN} and \acsp{LSTM} architectures. \acsp{CNN} were already used for feature extraction in Chapter \ref{ch:atom}, and now they will constitute the first stages of the \acs{LSTM}-based system for modeling visual attention in the temporal domain, which is the goal of this chapter. Furthermore, several common strategies for training and optimizing \acsp{NN} are explained in the different subsections.

\subsection{Neural Networks}\label{sec:neural_networks}
\begin{figure*}[!t]
\centering
	\includegraphics[trim=0cm 0cm 0cm 0cm, width=1\textwidth]{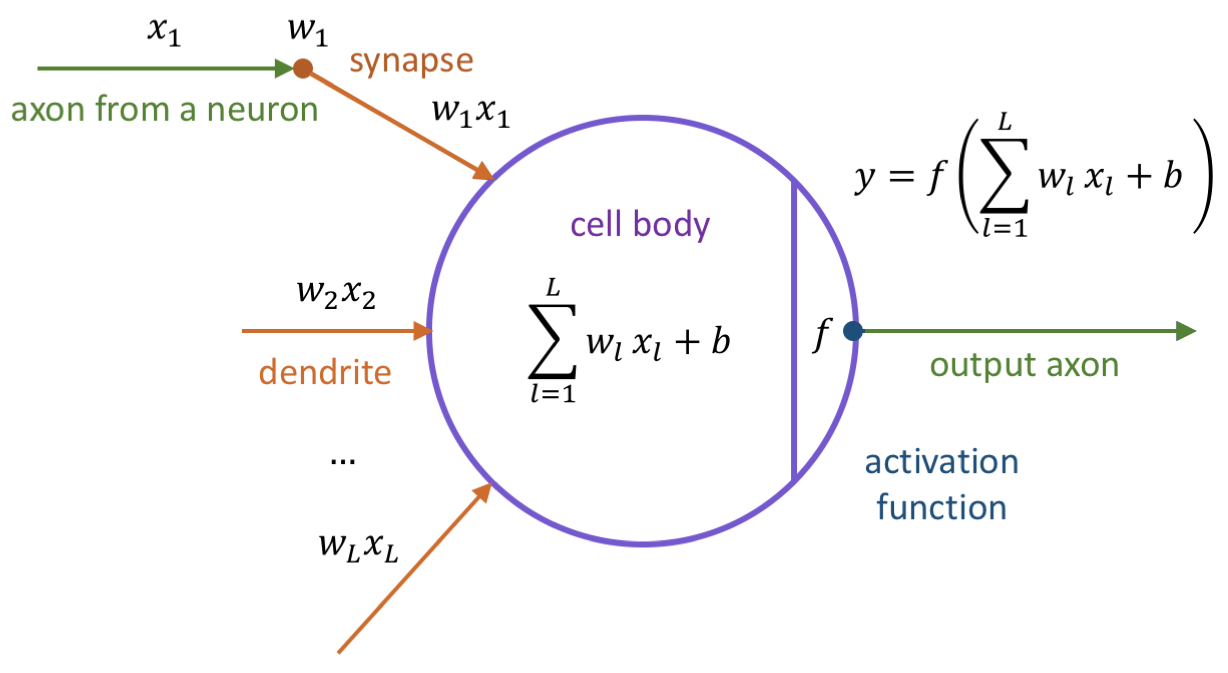}	
\caption[Mathematical model of a computational neuron.]{Mathematical model of a computational neuron with $L$ inputs ${\bf x}=\{x_{1},x_{2},...,x_{L}\}$ and one output $y$, composed by a set of weights ${\bf w}=\{w_{1},w_{2},...,w_{L}\}$ and a bias term $b$. Adapted from \cite{cs231n}.}
\label{fig:351_mathematical_neuron}
% \vspace{-0.4cm}
\end{figure*}

The basic computational cell of the brain is the neuron, as already mentioned in Section \ref{sec:brain}. Neurons receive, process and transmit information to carry out different functions, such as visual perception, learning and memory \cite{Palmer}. Inspired by this biological fact, \acfp{NN} are directed graphs of computational units, also named neurons.   

Following a similar process than the one involved by neurons in the brain, whose schematic diagram was shown in Figure \ref{fig:221_neuron}, a mathematical neuron, represented in Figure \ref{fig:351_mathematical_neuron}, computes a scalar output from a set of $L$ input signals ${\bf x}=\{x_{1},x_{2},...,x_{L}\}$ \cite{cs231n}. 

A single neuron or unit is defined as a linear classifier:

\begin{gather}\label{eq:neuron}
y = f({\bf w}^T{\bf x} + b) = f\left(\sum_{l=1}^{L}w_{l}x_{l} + b\right)
\end{gather}

\noindent Indeed, it is composed by a learnable set of weights ${\bf w}=\{w_{1},w_{2},...,w_{L}\}$ and a bias term $b$ that model synaptic strengths and control the excitatory (positive weight) or inhibitory (negative weight) influence of the neuron on subsequent neurons. Moreover, the firing rate of the biological neuron is represented by means of a non-linearity or activation function $f$. If the activation function corresponds to the sigmoid function ($y = sigm({\bf w}^T{\bf x} + b)$), the operation of a neuron corresponds to a logistic regression.

\subsubsection*{The perceptron}
The perceptron \cite{rosenblatt1958perceptron} was first introduced by Rosenblatt in 1958 and constitutes the simplest \acs{NN} model, based on a unique neuron or unit, which allows to solve a binary classification problem. For the case of a multi-class classifier with $C$ classes, a unit per class is needed. The perceptron is now defined in matrix form as follows:

\begin{gather}
{\bf y} = f({\bf W}^T{\bf x} + {\bf b}).
\end{gather}

\noindent The different variables with respect to Eq. (\ref{eq:neuron}) are ${\bf y}$, which is the vector with the $C$ output values, and the matrix ${\bf W} \in \mathbb{R}^{L \times C}$ and the vector ${\bf b} \in \mathbb{R}^{C}$, which contain the weights and biases for the $C$ units, respectively.

During the learning phase of the perceptron, its corresponding weights are iteratively updated based on samples from a training set, with the purpose of minimizing a chosen loss function. For each input sample $({\bf x}_{n}, {\bf y}_{n})$ and its predicted output ${\bf \hat{y}}_{n}$, the weight matrix ${\bf W}_{k}$ at step $k$ is updated as follows:

\begin{gather}\label{eq:perceptron_gradient_descent}
{\bf W}_{k} \longleftarrow {\bf W}_{k-1} + \epsilon({\bf \hat{y}}_{n}-{\bf y}_{n}){\bf x_{n}}^T,
\end{gather}

\noindent where $\epsilon$ denotes the learning rate, a fixed hyper-parameter to be determined that controls how quickly weights are updated.

The process is repeated until the error converges or a different criteria is met. Because all the training samples are processed independently by the algorithm at each step, this method is known as \acf{SGD}. 

\subsubsection*{Feed-forward neural networks}
\begin{figure*}[!t]
\centering
	\includegraphics[trim=0cm 0cm 0cm 0cm, width=1\textwidth]{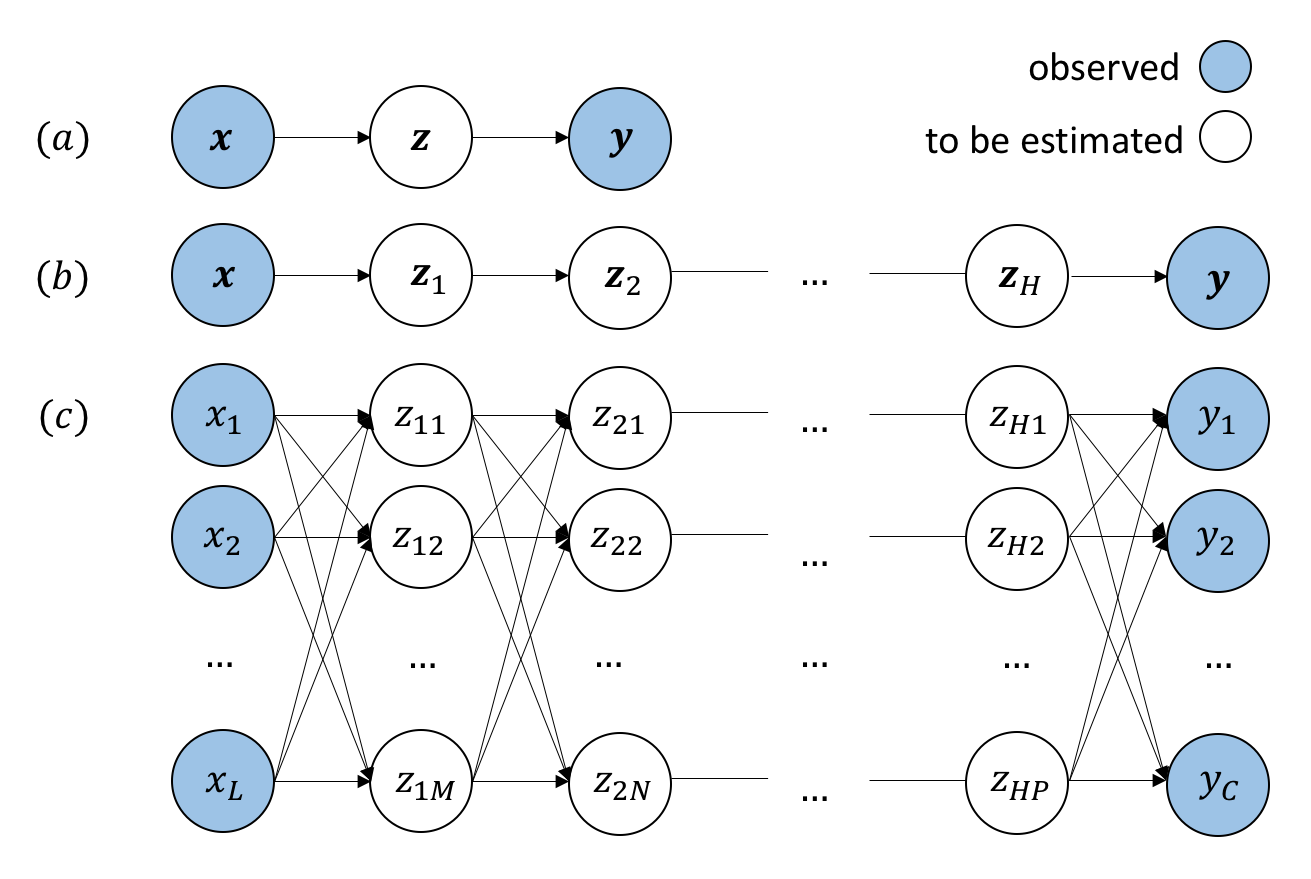}	
\caption[Graphical representations of feed-forward \acfp{NN}.]{Graphical representations of feed-forward \acsp{NN} with $L$ input features and $C$ output values. (a) Feed-forward network with a hidden layer, represented as a compact graph. Each node corresponds to a vector that contains the layer's activations. (b) Feed-forward network with $H$ hidden layers, represented as a compact graph (c) Feed-forward network with $H$ hidden layers, each of these contains a different number of units $\{M,N,...P\}$, represented as an explicit graph.}
\label{fig:351_neural_network}
% \vspace{-0.4cm}
\end{figure*}

Feed-forward neural networks or \acfp{MLP} are directed acyclic graphs of stacked groups of computational units, organized in layers, which have the ability of learning more complex functions than the ones achieved by perceptrons. A \acs{MLP} consists of an \emph{input layer} ${\bf x}$, which represent the $L$ input features; one or several \emph{hidden layers} ${\bf z}$, with the same or different number of units; and a final output layer ${\bf y}$, with as many units as values to be predicted. 

Traditional \acsp{NN} are comprised of \acf{FC} layers. While neurons between adjacent layers share fully pairwise connections, those within the same layer act in parallel and are not connected.

As can be seen in Figure \ref{fig:351_neural_network}, \acsp{MLP} can be also represented by graphs with nodes and edges, in the same way as the previously described Bayesian \acsp{LTM} (Chapter \ref{ch:atom}, Section \ref{sec:latent_topic_models}). However, it should be noted that they constitute graphical representations of functions instead of distributions.

More formally, the output of a \acs{MLP} with $H$ hidden layers can be expressed as follows:

\begin{gather}
{\bf y} = f({\bf W}^{T}{\bf z}_{H} + {\bf b}),
\end{gather} 

\noindent where ${\bf y}$ is its output layer, represented as a vector with the $C$ output values; $f$, ${\bf W}$ and ${\bf b}$ are its corresponding activation function, matrix of weights and vector of biases, respectively; and ${\bf z}_{H}$ the vector with the output values of the units in the preceding hidden layer, which is recursively defined as:

\begin{gather}
{\bf z}_{H} = f_{H}({{\bf W}_{H}}^{T}{\bf z}_{H-1} + {\bf b}_{H}).
\end{gather} 
 
\acfp{DNN} refer to \acsp{MLP} with a great number of hidden layers and units, in the same way that ``deep learning" is the field within \acs{ML} that deals with this type of models. \acfp{CNN}, described in Section \ref{sec:convolutional_neural_networks}, are \acsp{DNN} whose hidden units have local receptive fields, particularly useful to solve computer vision tasks. When feed-forward networks are extended to have feedback connections, they are called \acfp{RNN}, which will be introduced in Section \ref{sec:recurrent_neural_networks}. The following paragraphs cover the most important aspects regarding the definition and training of these architectures.

\subsubsection*{Architecture design}
The complete architecture of a feed-forward network can be summarized by its depth, which is determined by its number of layers; the width of each layer, which is the number of units these layers have; and how these units are connected to each other. Most of the existing \acsp{MLP} models follow a chain-based structure, where each layer is a function of its preceding layer.

As stated by the ``no free lunch'' theorem \cite{wolpert1997no} introduced in Chapter \ref{ch:introduction}, there is no \acs{ML} algorithm better than any other so, for the same reason, it is truly complex to define the optimal network structure for a specific application. According to the universal approximation theorem \cite{hornik1991approximation}, a large \acs{DNN} with enough capacity or hidden units is able to approximate any continuous function. Despite this, the optimization algorithm used to find the parameters corresponding to that function plays a crucial role in the learning phase and should be carefully validated. What is more, overfitting may occur in a model with higher capacity than needed, which would make it learn the noise in the data instead of the expected underlying relationships that lead to a good generalization. Regularization techniques can be helpful to prevent overfitting, as we will see in the next subsection.

\begin{figure*}[!t]
\centering
	\includegraphics[trim=0cm 0cm 0cm 0cm, width=1\textwidth]{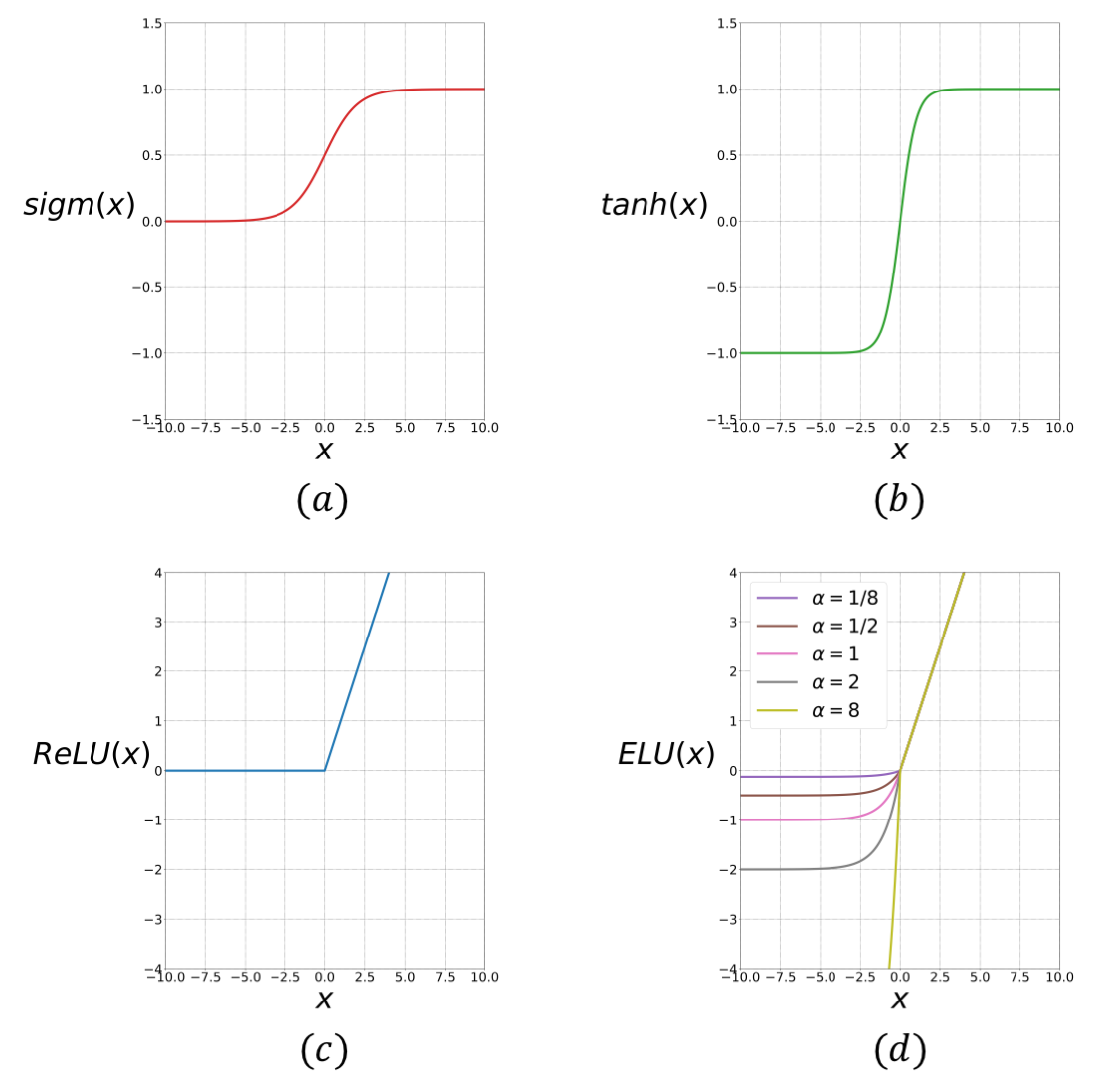}
\caption{Graphical representation of the most commonly used activation functions. (a) Sigmoid. (b) Hyperbolic tangent. (c) \acf{ReLU}. (d) \acf{ELU} represented for different values of $\alpha$.}
\label{fig:331_activation_functions}
% \vspace{-0.4cm}
\end{figure*}

On the other hand, the activation or transfer functions for both the hidden and the output units should be also taken into consideration. The most commonly used functions, which are represented in Figure \ref{fig:331_activation_functions}, are described here below:

\begin{itemize}
	\item \emph{Sigmoid}: Defined as $sigm(x)=1/(1+e^{-x})$, the sigmoid non-linearity takes a real-valued scalar and compresses it to the range $[0,1]$. Although it has been frequently used in the past, it presents two main drawbacks: first, sigmoids saturate at $0$ or $1$ when $x$ is very negative or positive, respectively, which can hinder the common gradient-based learning; besides, sigmoid outputs are non-zero centered, which may introduce troublesome zig-zagging dynamics in the gradient updates for the weights \cite{cs231n}.
	  	
	\item \emph{Hyperbolic tangent}: It compresses a real-valued scalar to the range $[-1,1]$, and can be seen as a zero-centered scaled version of the sigmoid function; consequently, it is often preferred in practice. In fact, $tanh(x)=2sigm(2x)-1$. 
	
	\item \emph{\acf{ReLU}}: The \acs{ReLU} function, defined as $f(x)=max(0,x)$, thresholds the activation at zero, notably accelerating the convergence of optimization methods with respect to the previously described sigmoidal functions, thanks to its simple, linear and non-saturating form. However, \acs{ReLU} units are sensitive to large gradients. Although this issue can be often avoided using an appropriate learning rate, generalizations from \acs{ReLU} activation such as \emph{leaky \acsp{ReLU}} and \emph{maxout} have been tested in an attempt to fix it \cite{goodfellow2016deep}, as well as \acfp{ELU} \cite{ELU}. As shown in Figure \ref{fig:331_activation_functions}(d), \acsp{ELU} have saturated negative values controlled by a hyperparameter $\alpha$, which push the mean of the activations close to zero and solve the vanishing gradient problem. If the value of $\alpha$ is set too low, \acs{ELU} and \acs{ReLU} activations become similar.
\end{itemize}
    
\acs{ReLU} non-linearities are almost always used in most hidden \acs{NN} layers. Nevertheless, it should be noted that recurrent networks, probabilistic models and autoencoders draw on sigmoidal activation functions, despite their saturation drawbacks, because of their utility when having exploding gradients \cite{goodfellow2016deep}.

Regarding the activation function chosen for the output layer of a \acs{NN}, it is completely dependent of the task to perform. While sigmoid allows to obtain class scores in a binary classification, \emph{softmax} units, which represent a discrete probability distribution with $C$ possible values, are more suitable for multi-class classifiers. In contrast, linear activations are used in regression problems to predict real values.

\subsubsection*{Gradient-based learning} 
The Goodfellow et al.'s recipe for \acs{ML} \cite{goodfellow2016deep} introduced in section \ref{sec:machine_learning} is entirely in line with the required specifications for gradient-based training a \acs{DNN}: select an optimizer, choose a cost or loss function and obtain a model. 

\begin{enumerate}
\item \emph{Select an optimizer}: First, the objective of the \emph{optimization algorithm} is to find the weights ${\bf w}$ associated with the \acs{DNN} units that minimize the error between the expected \acs{GT} values and the values predicted by the non-convex approximate function. For this purpose, iterative, gradient-based optimizers are the preferred option. 

In contrast to traditional pure optimization methods, where the cost function is not related to the measure used to evaluate the performance of the system, gradient-based optimizers have in many contexts the advantage of directly minimizing a loss function $J(f({\bf x}; {\bf w}), {\bf y})$ suited to the task to solve. 

The basic algorithm for gradient-based optimization is \acf{MBGD} \cite{bottou1998online}. At each iteration $k$, this stochastic method estimates the gradient of the approximate function as the average gradient on a small set or batch of $M$ \acs{IID} samples randomly chosen:

\begin{gather}
{\bf \hat{g}} = \nabla_{\bf w} \left(\frac{1}{M} \sum_{m=1}^M J(f({\bf x}_{m}; {\bf w}), {\bf y}_{m})\right).
\end{gather}

\noindent Then, weights are updated as follows:

\begin{gather}\label{eq:stochastic_gradient_descent}
{\bf w}_{k} \longleftarrow {\bf w}_{k-1} - \epsilon_{k} {\bf \hat{g}}.
\end{gather}

\noindent Similarly to Eq. (\ref{eq:perceptron_gradient_descent}), a critical hyper-parameter to determine is the learning rate $\epsilon_{k}$, which is now represented with the sub-index $k$ because it is common to decay its value along iterations. This allows using a higher rate at the beginning, which prevents becoming stuck at a high cost, and progressively decreasing it in order to avoid significant oscillations in the learning curve.

Another approach to increase \acs{MBGD} convergence speed is the momentum algorithm \cite{polyak1964some}, which aids to keep the direction and speed at which the parameters are updated in subsequent iterations. Formally, at each iteration $k$, it computes an \acf{EMA} of the negative past gradients ${\bf v}$ to update the weights:

\begin{gather}
{\bf v}_{k} = \alpha {\bf v}_{k-1} - \epsilon_{k} \nabla_{\bf w} \left(\frac{1}{M} \sum_{m=1}^M J(f({\bf x}_{m}; {\bf w}), {\bf y}_{m})\right),
\end{gather}

\begin{gather}
{\bf w}_{k} \longleftarrow {\bf w}_{k-1} + {\bf v}_{k}.
\end{gather} 

On the basis of \acs{MBGD}, two of the most outstanding techniques that incorporate both adaptive learning rates and momentum are \acf{RMSProp} \cite{tielemanH12} and Adam \cite{kingma2014adam}.

Furthermore, it should be noted the importance of properly setting the initial network weights. Parameters from different units need to be initialized to different values, not to converge to the same configuration; this is commonly known as ``breaking the symmetry effect''. As a convention, weights are often initialized to small values randomly chosen from a Gaussian or a uniform distribution, and biases are set to zero. While weights should be large enough to propagate information successfully, it is also desirable the use of small values to give a similar prior preference to all units. 

Among other optimization strategies it is also remarkable the use of \acf{BN} layers \cite{ioffe2015batch}, which constitute a method for adaptive reparametrization applicable to any input or hidden layer in a network. The distribution of \acs{DNN} layer's inputs changes during training. This effect, known as \emph{internal covariate shift}, makes this stage substantially difficult, but can be avoided by normalizing mini-batch samples to have mean zero and standard deviation $1$. \acf{LN} \cite{ba2016layer} is an alternative to \acs{BN}, which consists in normalizing layer activations. It may be useful when dealing with \acsp{RNN} and small mini-batches.

\item \emph{Choose a loss function}: Second, the error to minimize is defined by means of a differentiable \emph{loss function}, which is often tailored to the task at hand for a better model fit. The most widely used loss functions correspond to the classical classification and regression tasks. The function associated with a classification task with multiple categories is the cross-entropy. For a problem with $C$ classes and a set of $N$ training samples, it is computed as:

\begin{gather}
H({\bf y}, {\bf \hat{y}}) = -\sum_{n=1}^{N} \sum_{c=1}^{C} y_{n} log(\hat{y}_{nc}), 
\end{gather}

\noindent being $y_{n}$ a binary label (1 if corresponds to the samples's true class) and $\hat{y}_{nc}$ the predicted probability for the class $c$. In a regression context, either \acf{MSE} or \acf{MAE} are used, which are defined as follows:

\begin{gather}\label{eq:mse_loss}
MSE(y, \hat{y}) = \frac{1}{N} \sum_{n=1}^{N} (\hat{y}_{n} - y_{n})^2
\end{gather}

\begin{gather}
MAE(y, \hat{y}) = \frac{1}{N} \sum_{n=1}^{N} |\hat{y}_{n} - y_{n}|
\end{gather}

When a \acs{DNN} is applied to saliency or visual attention estimation, it is common to use a differentiable saliency metric, such as \acs{KL} divergence, \acf{NSS} or \acf{CC} \cite{huang2015salicon}.

With the purpose of reducing the generalization error of a model, while keeping its training error low, it is prevalent the use of regularization techniques such as \emph{weight decay} or those introduced in Chapter 7 of Goodfellow et al.'s book \cite{goodfellow2016deep}. 

\emph{Early stopping} is probably the most used regularization technique in deep learning. The method considers a validation set composed by some samples unseen during training and monitors, at each iteration, the error in this set as a reference to stop learning when the model starts overfitting. In this way, we can pre-specify a number of iterations after which the training phase finishes if the validation error has not decreased. 

Another well-known and useful strategy for regularization is \emph{Dropout} \cite{srivastava2014dropout}, which attempts to reduce overfitting in a network during training by randomly removing a certain percentage of hidden units at each iteration.

% Other common strategies are data augmentation and early stopping.
% COMPLETAR/EXTENDER EN FUNCION DE LO QUE SE UTILICE FINALMENTE EN CAPITULOS 6 Y 7

\begin{figure*}[!t]
\centering
	\includegraphics[trim=0cm 0cm 0cm 0cm, width=1\textwidth]{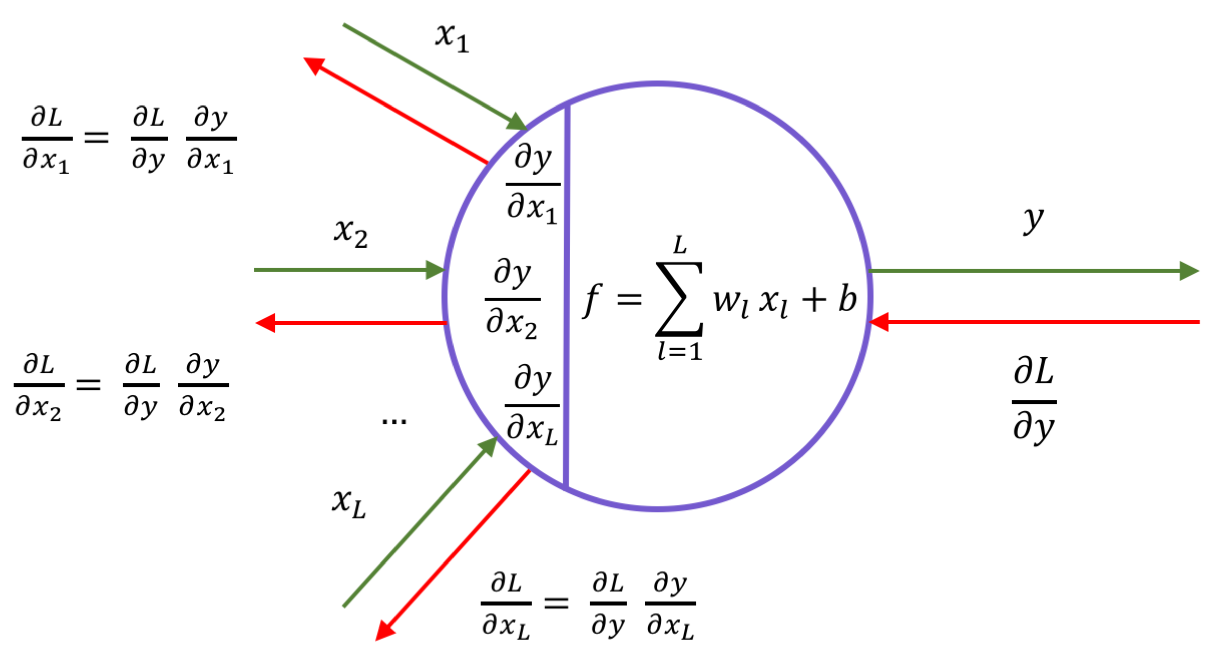}	
\caption[Computational graph of a \acf{NN} layer, where forward and back-propagation stages in gradient-based learning are represented with green and red arrows, respectively.]{Computational graph of a \acs{NN} layer with $L$ inputs $\{x_{1},x_{2},...,x_{L}\}$ and one output $y$, where forward and back-propagation stages in gradient-based learning are represented with green and red arrows, respectively. Operations involved in both stages are indicated next to the arrows. At each iteration of gradient-based learning, the predicted output value is obtained in the forward pass, while during the backward pass the gradient of the approximate function is computed, in order to update the layer weights. Adapted from \cite{cs231n}.}
\label{fig:351_backpropagation}
% \vspace{-0.4cm}
\end{figure*}

\item \emph{Obtain a model}: The \emph{back-propagation algorithm} \cite{rumelhart1986learning} allows to compute the gradient needed to iteratively update the network weights and obtain a model. During the learning phase of a \acs{DNN}, we can distinguish between two stages: forward propagation and back-propagation. In contrast, only forward propagation is performed in test.

Whereas in \emph{forward propagation} a mini-batch of $M$ input samples $({\bf x}_{1},{\bf x}_{2},...,{\bf x}_{M})$ flows through a \acs{DNN}, providing a mini-batch of outputs $({\bf \hat{y}}_{1},{\bf \hat{y}}_{2},...,{\bf \hat{y}}_{M})$, \emph{back-propagation} takes the average loss between the expected \acs{GT} values $({\bf y}_{1},{\bf y}_{2},...,{\bf y}_{M})$ and the predicted ones, which flows backwards through the network in order to obtain the gradients.

Figure \ref{fig:351_backpropagation} shows a simple computational graph for a \acs{NN} layer with $L$ inputs ${\bf x}$ and one output $y$, the same represented in Figure \ref{fig:351_mathematical_neuron}. 

\begin{itemize}
	\item If we move from left to right, the output value $y$ is computed in the forward pass, indicated by green arrows.
	\item In the backward pass, indicated by red arrows, from right to left, gradients are computed by means of the chain rule of calculus. First, we compute the gradient of the loss $J$ with respect to the output value $y$: $\frac{\partial L}{\partial y}$. Then, the gradient value for each weight $w_{l}$ is obtained as follows:
	\begin{gather}
		\frac{\partial J}{\partial w_{l}} = \frac{\partial J}{\partial y} \frac{\partial y}{\partial w_{l}}. 
	\end{gather}
	If the hidden unit participates in a \acs{DNN}, $J$ is the loss from the subsequent layer, which has to be back-propagated to every previous layer. In this case, we get the gradient value for each input $x_{l}$ as:
	\begin{gather}
		\frac{\partial J}{\partial x_{l}} = \frac{\partial J}{\partial y} \frac{\partial y}{\partial x_{l}}.
	\end{gather} 
	Back-propagation can also be expressed in vector notation. Let us consider now a vector ${\bf y}$ with $C$ output values: 
	\begin{gather}
		\nabla_{{\bf w}}J = \left(\frac{\partial {\bf y}}{\partial {\bf w}}\right)^{T}\nabla_{{\bf y}}J
	\end{gather}
	\begin{gather}
		\nabla_{{\bf x}}J = \left(\frac{\partial {\bf y}}{\partial {\bf x}}\right)^{T}\nabla_{{\bf y}}J,
	\end{gather} 
	where $\frac{\partial {\bf y}}{\partial {\bf w}}$, $\frac{\partial {\bf y}}{\partial {\bf x}}$ are $C \times L$ Jacobian matrices of gradients. The same Jacobian-gradient product is performed for each layer in the \acs{DNN}.
	It should be noted, nevertheless, that the method is usually applied to tensors of arbitrary dimensionality rather than vectors, but it is conceptually the same. Given input, output and weight tensors ${\bf \mathsf{X}}, {\bf \mathsf{Y}}, {\bf \mathsf{W}}$, respectively, we can just treat them as vectors whose indexes have multiple coordinates (e.g. three coordinates for a \acs{3D} tensor). Thus, the chain rule is applied as follows:
	\begin{gather}
		\nabla_{{\bf \mathsf{W}}}J = \sum_{j} (\nabla_{{\bf \mathsf{W}}}{\bf \mathsf{Y}}_{j})\frac{\partial J}{\partial {\bf \mathsf{Y}}_{j}}
	\end{gather}
	\begin{gather}
		\nabla_{{\bf \mathsf{X}}}J = \sum_{j} (\nabla_{{\bf \mathsf{X}}}{\bf \mathsf{Y}}_{j})\frac{\partial J}{\partial {\bf \mathsf{Y}}_{j}},
	\end{gather}
	where $j$ is an index variable to represent the complete tuple of coordinates in $\mathsf{Y}$. 
\end{itemize}

\end{enumerate}

\subsection{Convolutional Neural Networks}\label{sec:convolutional_neural_networks}
\acfp{CNN} arised inspired by the neurophysiological work of Hubel and Wiesel in 1962 \cite{hubel1962receptive}, which discovered that neurons in the early visual system are sensitive to simple patterns of light, such as oriented edges or color blotches. As mentioned in the previous section, \acsp{CNN} \cite{lecun1989backpropagation, simard2003best} are \acsp{DNN} composed of hidden units that have local receptive fields, similar to neurons in the primary visual cortex, and designed to take images or data with a grid-like topology as input. 

% COMPLETAR CON FIGURA???
Unlike traditional \acsp{NN}, a \acs{CNN} is composed of layers of neurons arranged in a \acs{3D} volume. Each layer transforms an input \acs{3D} volume into an output \acs{3D} volume. The characteristic operation of a \acs{CNN} is the discrete convolution. Furthermore, \acsp{CNN} may also involve pooling or down-sampling non-parametric operations.

The use of \acsp{CNN} has several properties and advantages \cite{goodfellow2016deep}:

\begin{enumerate}
 \item \emph{Locality of sparse interactions}: \acsp{CNN} have sparse weights, which means that they are able to detect small, meaningful features by making use of convolution kernels smaller than the input, not larger than hundreds of pixels. This reduces the memory usage, while fewer operations are required to compute outputs, which improves the efficiency of the model.
 \item \emph{Parameter sharing}: In a convolutional layer, the number of parameters is dramatically reduced, because they are shared across the image; in other words, rather than learning a set of parameters for each spatial location, the same convolutional kernel is applied on all of them. 
 \item \emph{Invariance to translation}: \acsp{CNN} are translational invariant, so they are able to identify patterns independently of their location in the input image. Conversely, they are not invariant to rotation or scale changes, and require of other techniques to handle these transformations, such as data augmentation \cite{simard2003best, hernandez2018data}. 
\end{enumerate}

\begin{figure}[!t]
\centering
	\includegraphics[trim=0cm 0cm 0cm 0cm, width=1\textwidth]{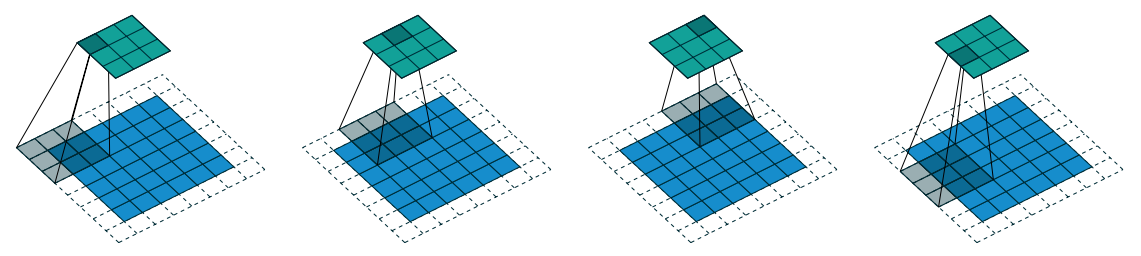}	
\caption[Application of a $k=3\times3$ convolutional kernel over a $6\times 6$ input padded with a $1\times1$ border of zeros, using a stride of $s=2$.]{Application of a $k=3\times3$ convolutional kernel over a $6\times 6$ input padded with a $1\times1$ border of zeros, using a stride of $s=2$. Figure taken from \cite{dumoulin2016guide}.}
\label{fig:352_conv2d}
% \vspace{-0.4cm}
\end{figure}

\subsubsection*{Architecture design}\label{sec:cnn_architecture_design}
Common \acsp{CNN} are built on a sequence of $A$ blocks of $B$ \acf{CONV} layers with \acs{ReLU} activations, sometimes followed by a \acf{POOL} layer, ending up in a stack of $C$ \acs{FC} layers, being the last \acs{FC} layer the one that holds the output predicted values \cite{cs231n}: \\

\begin{center}
$INPUT \rightarrow A \times$[$B \times$[\acs{CONV} $\rightarrow$ \acs{ReLU}] $\rightarrow$ \acs{POOL}?] \\ $\rightarrow C \times$[\acs{FC} $\rightarrow$ \acs{ReLU}] $\rightarrow$ \acs{FC}
\end{center}

Related to the typical layers of \acsp{CNN}, \acs{CONV} layers first involve a set of filters with learnable weights. Let width, height and depth denote the arbitrary dimensions of an input \acs{3D} volume.  During the forward pass, each filter is spatially slided along the width and the height of the channels or spatial maps stacked on the depth dimension, in order to compute the dot product at each spatial location. The connections between the input and each filter are thus local in space, but full along the depth. The \emph{filter} or \emph{kernel size} $k$ is the receptive field of the neuron, while the size of the output \acs{3D} volume depends on three hyper-parameters:

\begin{itemize}
	\item The number of filters $u$ used corresponds to the \emph{depth} of the resulting output volume.
	\item The \emph{stride} $s$ with which the filters are slided or down-sampling factor determines the spatial size of the output. Bigger strides result in smaller output volumes.
	\item \emph{Zero-padding} the input volume is another way of varying the output spatial size, increasing it spatially with a border of one or several pixels.
\end{itemize}

Given a filter $\mathsf{K}_{i,l}$ of size $k = m \times n$, which gives the connection between channel $i$ in the output volume ${\bf \mathsf{Y}}$ and channel $l$ in the input volume ${\bf \mathsf{X}}$, and is applied centered at the input location $j,k$, being $j$ and $k$ the row and column positions; assuming that both input and output have the same spatial dimensions, the value that results from a \acs{CONV} layer at the output location $j,k$ can be expressed more formally in the language of tensors:

\begin{gather}\label{eq:cnn_forward}
\mathsf{Y}_{i,j,k} = \sum_{l,m,n} \mathsf{X}_{l,(j-1) \times s_{W}+m,(k-1) \times s_{H}+n} \mathsf{K}_{i,l,m,n},
\end{gather}

\noindent where $s_{W}, s_{H}$ denote the strides of the convolution along width and height, respectively. Figure \ref{fig:352_conv2d} shows an example of application of a convolutional kernel.

\begin{figure*}[!t]
\centering
	\includegraphics[trim=0cm 0cm 0cm 0cm, width=1\textwidth]{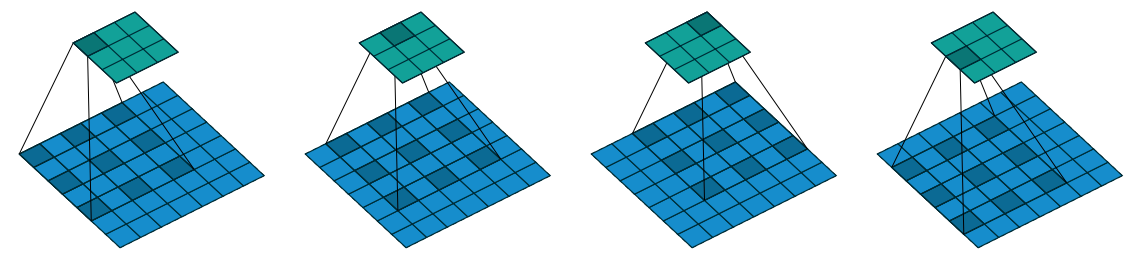}	
\caption[Application of a $k=3\times3$ dilated convolutional kernel over a $7\times7$ input, using a dilation factor of $d=2$ (1 space between kernel elements).]{Application of a $k=3\times3$ dilated convolutional kernel over a $7\times7$ input, using a dilation factor of $d=2$ (1 space between kernel elements). Figure taken from \cite{dumoulin2016guide}.}
\label{fig:352_dilatedconv2d}
% \vspace{-0.4cm}
\end{figure*}

Due to their regular use in \acsp{CNN} for the extraction of feature maps, we shall introduce a particular type of convolutions, known as \emph{dilated} or \emph{``atrous convolutions''} \cite{yu2015multi}. They differ from original convolutions in the insertion of spaces between the kernel elements depending on a dilation rate. As can be seen in the example of application in Figure \ref{fig:352_dilatedconv2d}, a dilation rate $d$ corresponds to $d-1$ spaces inserted between elements. Dilated convolutions allow to increase the receptive field of the output without increasing the size of the kernel. 

Other variants of the basic convolution function, such as transposed convolutional layers or locally connected layers, are introduced in Chapter 9 from \cite{goodfellow2016deep}, and also in the excellent guide to convolution arithmetic by Dumoulin et al. \cite{dumoulin2016guide}.

Secondly, \acs{POOL} layers are useful to make representations almost invariant to small translations of the input. Moreover, by pooling over outputs from different convolutions, the network learns to become invariant to some transformations. \acs{POOL} layers perform a function that does not have parameters. The most commonly used \acs{POOL} layer is \emph{max pooling} (MAX \acs{POOL}), which involves a maximum operation, but other functions such as the average (AVG \acs{POOL}) or the Euclidean norm are also considered. \acs{POOL} layers often entail down-sampling operations ($s>1$) along width and height, which reduce the dimensions of the feature maps.

\subsubsection*{Case studies}
\begin{figure*}[!htbp]
\centering
	\includegraphics[trim=0cm 0cm 0cm 0cm, width=1\textwidth]{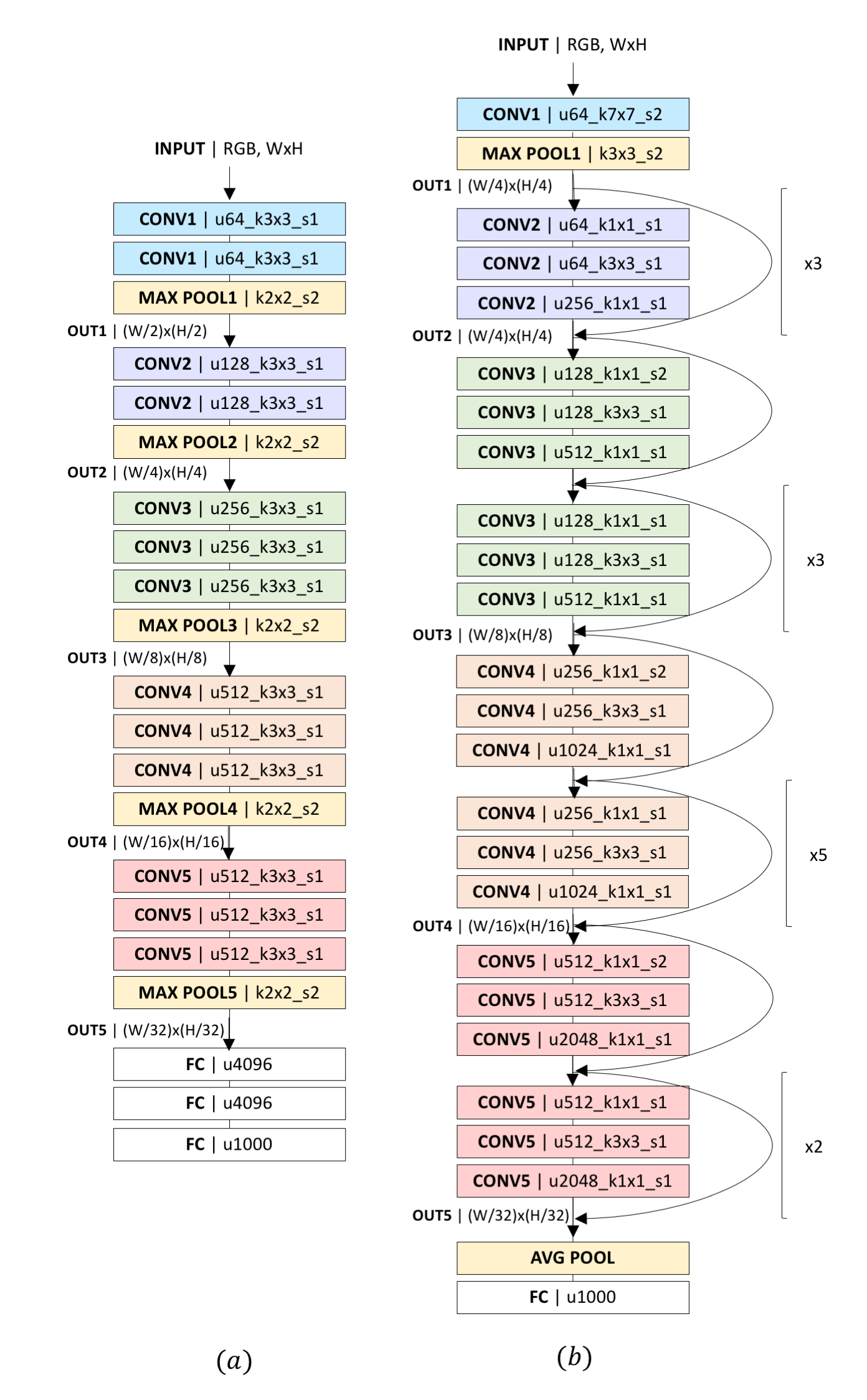}	
\caption[Architecture diagrams of the (a) \emph{VGG-16} \cite{vgg2014} and (b) \emph{ResNet-50} \cite{he2016deep} networks for image recognition.]{Architecture diagrams of the (a) \emph{VGG-16} \cite{vgg2014} and (b) \emph{ResNet-50} \cite{he2016deep} networks for image recognition. Layers are defined by their number of units $u$, kernel size $k$ and stride $s$ with which filters are slided. Given an INPUT image of dimension $WxH$, output (OUT) sizes are indicated at the end of each block. Last \acs{FC} softmax layers in both models have 1000 units, which correspond to the image classes defined in \acs{ILSVRC} \cite{ILSVRC15}.}
\label{fig:352_vgg_resnet}
% \vspace{-0.4cm}
\end{figure*}
One of the first successful applications of convolutional networks was \emph{LeNet} \cite{le1990handwritten}. Developed by Yann LeCun in 1990, it is composed of five layers and was used to read digits from zip codes. Throughout the brief history of the application of \acsp{CNN} in computer vision, three additional case studies should be highlighted, due to their successful achievements in image recognition: \emph{AlexNet} \cite{krizhevsky2012imagenet}, \emph{VGGNet} \cite{vgg2014} and \emph{ResNet} \cite{he2016deep}. Deeper (8 layers), but with a similar structure to \emph{LeNet}, \emph{AlexNet} is one of the first works that promoted the use of \acsp{CNN} in computer vision, winner of the \acf{ILSVRC} \cite{ILSVRC15} in 2012. The other two aforementioned configurations are used for visual attention feature extraction in this thesis, so they are described hereunder for the sake of completeness.

\begin{itemize}
	\item \emph{VGGNets} \cite{vgg2014} were the state-of-the-art in 2014. By increasing the depth of prior architectures up to 19 layers and using \acs{CONV} kernels with small receptive fields ($k=1\times1$, $k=3\times3$) and \acs{ReLU} non-linearities, they outperformed \emph{AlexNet} and other models proposed that year, reaffirming again that networks depth is a critical component for a better performance. The diagram in Figure \ref{fig:352_vgg_resnet}(a) shows one of the most used \emph{VGG} configurations, \emph{VGG-16}, which includes 16 weight layers, 13 $k=3\times3$ \acs{CONV} layers with stride $s=1$ and 3 \acs{FC} layers. As can be seen, some of the \acs{CONV} layers are followed by $k=2\times2$ MAX \acs{POOL} layers, with stride $s=2$.

	\item \emph{ResNets} \cite{he2016deep}, winner of \acs{ILSVRC} \cite{ILSVRC15} in 2015, arose with the aim of reducing the degradation problem caused by the saturation of performance when training deeper \acsp{NN}. They are built on a series of blocks composed of few layers, among which residual mappings are performed. These mappings consist of adding the input of each block to its output, by making use of skip connections, as can be noticed in the commonly-used 50-layer architecture in Figure \ref{fig:352_vgg_resnet}(b). The configuration contains a $k=7\times7$ \acs{CONV} layer, followed by a MAX \acs{POOL} layer, both with $s=2$, and then a stack of fully \acs{CONV} residual blocks with increasing number of units. The first \acs{CONV} layer of each block reduces the dimensionality of its input by using a stride of $s=2$.
\end{itemize}

At the end of either \emph{VGG} or \emph{ResNet} networks, the final \acs{FC} softmax layer contains 1000 channels, associated with the image classes considered in \acs{ILSVRC} \cite{ILSVRC15}. It should be noted that last \acs{FC} layers are removed in tasks such as image segmentation or visual attention estimation, which ultimately provide a pixel-wise score map instead of an image-wise score. Moreover, because these applications require more accurate spatial predictions, some \acs{CONV} layers are often substituted by dilated convolutions, in order to preserve a larger spatial resolution across layers, as we will see in the architectures used for salient feature extraction in Section \ref{sec:cnns_features6}.

\subsubsection*{Transfer learning in \acsp{CNN}}
Transfer learning \cite{pan2010survey, thrun2012learning} is a \acs{ML} method which consists in reusing the knowledge acquired while solving a particular task with the aim of addressing a second different but related task.   

In order to train a \acs{CNN} to operate in a particular scenario, it is necessary to annotate a large image database, which might become a highly arduous task. That is the reason why an entire \acs{CNN} is seldom trained from scratch \cite{cs231n}. Instead, it is quite common to pretrain the network on a big dataset, such as the well-known ImageNet \cite{deng2009imagenet}. This pretrained model is used then as a fixed feature extractor, or its (or some of its) layers are \emph{fine-tuned} on the new smaller database. Smaller learning rates are often used for fine-tuned \acsp{CNN}, assuming that already existing weights are sufficiently good, so it is better not to significantly change them.

\subsection{Encoder-Decoder Networks}\label{sec:autoencoders}
%\afterpage{
%\begin{landscape}
\begin{sidewaysfigure}
\centerline{\includegraphics[trim=0cm 0cm 0cm 0cm, width=0.85\textwidth]{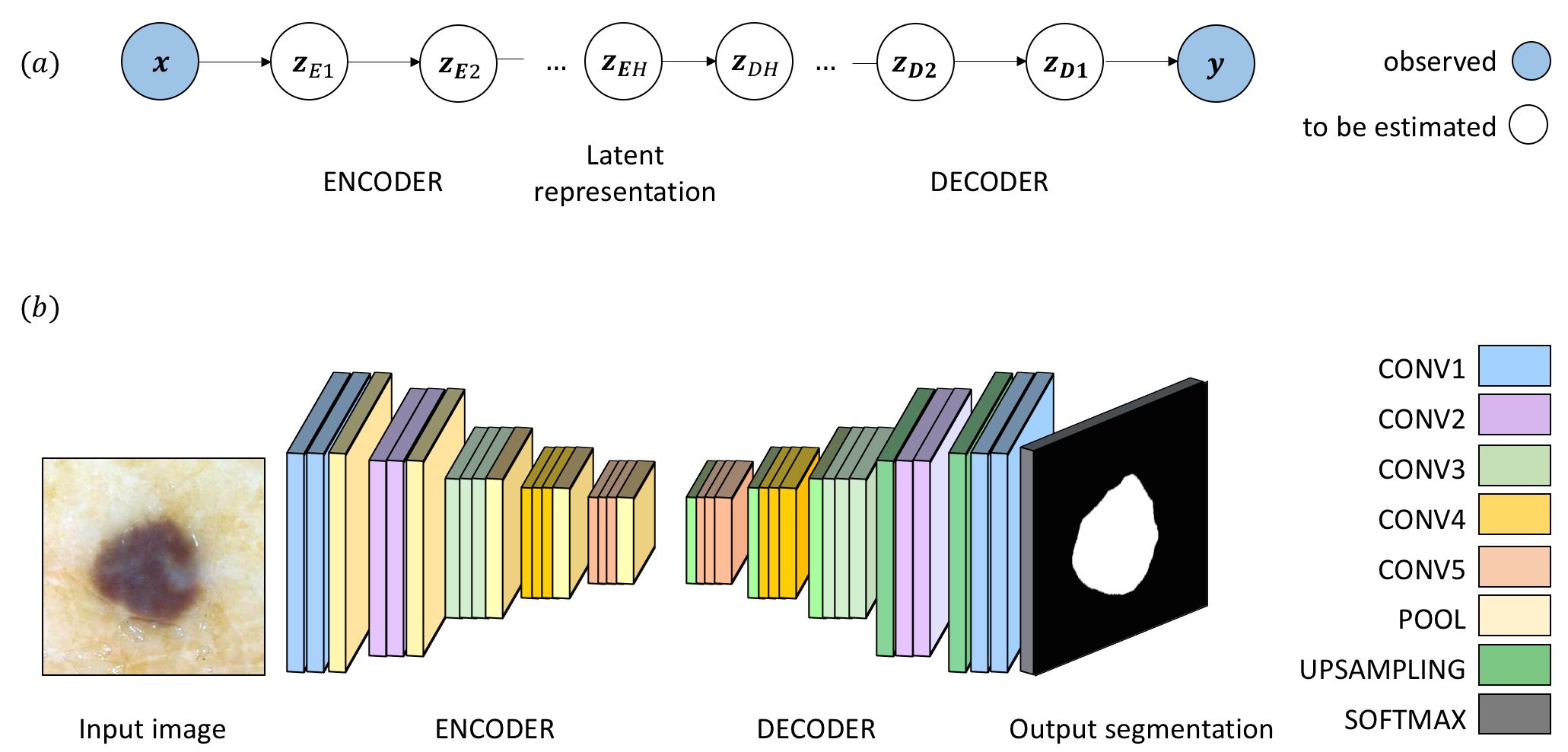}}	
\caption[(a) Graphical representation of an \acf{EDN}. (b) Example diagram of a convolutional encoder-decoder architecture for medical image segmentation.]{(a) Graphical representation of an \acs{EDN} composed by an encoder with $E_{H}$ hidden layers, which takes an input $\mathbf{x}$ and generates a latent representation $\mathbf{z_{E_{H}}}$, and a decoder with $D_{H}$ hidden layers, which produces an output $\mathbf{y}$ from the representation, tailored to the task to solve. Each node corresponds to a vector that contains the layer's activations. The \acs{EDN} has $E_{H}+D_{H}$ layers in total. (b) Example diagram of a convolutional encoder-decoder architecture for skin lesion image segmentation. Both encoder and decoder networks consist of the 13 convolutional layers in the \emph{VGG-16} \cite{vgg2014} network represented in Figure \ref{fig:352_vgg_resnet}(b).}
\label{fig:354_edn}
% \vspace{-0.4cm}
\end{sidewaysfigure}
%\end{landscape}
%}

The impressive ability of \acsp{DNN} to capture good hierarchical representations of data has also served to update the way of computing invariant features \cite{huang2007unsupervised}. These features were traditionally achieved by means of unsupervised methods for dimensionality reduction or clustering (e.g. \acs{PCA}, K-means \cite{Murphy2013Machine}), as well as through hand-crafted histograms, such as \acf{SIFT} \cite{lowe2004distinctive} or \acf{HOG} \cite{dalal2005histograms} descriptors.

\acfp{EDN} are a special case of feed-forward \acsp{NN}, explicitly designed to learn efficient feature representations. As shown in the graphical representation of Figure \ref{fig:354_edn}(a), these architectures are composed of two consecutive networks:

\begin{enumerate}
\item An \emph{encoder} network, which takes an input $\mathbf{x}$ and, after one or several hidden layers, represents it as a feature code or latent representation $\mathbf{z}_{E_{H}}=f_{E_{H-1}}(\mathbf{z}_{E_{H-1}})$, being $E_{H}$ its number of hidden layers.

\item A \emph{decoder} network with $D_{H}$ hidden layers, which reconstructs the feature code $\mathbf{z}_{E_{H}}$, in order to produce an output $\mathbf{y}$ tailored to the task to solve. For that purpose, the reconstruction error with respect to a \acs{GT} is measured (e.g. segmentation mask in an image segmentation system, see Figure \ref{fig:354_edn}(b)).
\end{enumerate}

Following this notation, \acsp{EDN} have $E_{H}+D_{H}$ layers in total. They can be trained by using the same algorithms described for feed-forward networks (see Section \ref{sec:neural_networks}). In the particular case when inputs and outputs are equal, those networks are called \emph{autoencoders} \cite{goodfellow2016deep}; and its encoder and decoder often have the same number of hidden layers ($E_{H}=D_{H}$).

We can distinguish between two types of \acsp{EDN}. The \acs{EDN} is \emph{undercomplete} when the hidden code dimension is lower than the input dimension, and attempts to extract the most useful or salient properties of the input data. In contrast, the latent representation has a dimension higher than the input in \emph{overcomplete} \acsp{EDN}.  

If the decoder is linear and \acs{MSE} is considered as loss function, an undercomplete autoencoder learns a similar subspace to the one obtained by \acs{PCA}. However, a \acs{EDN} with nonlinear encoder and decoder functions is able to learn more powerful representations than \acs{PCA}, which is restricted to linear transformations of the data. Besides, if the capacity of the network becomes too high, it might end up copying the training data instead of gathering the most prominent underlying information in the data distribution. In that case, regularization techniques can be helpful to avoid overfitting in overcomplete \acs{EDN}, providing more sparse and shift-invariant representations \cite{goodfellow2016deep}.  

\acsp{EDN} may involve either \acs{FC} \acsp{DNN}, \acsp{CNN}, \acsp{RNN} or a mixture of all three. They have been recently applied to solve tasks such as \acs{LSTM}-based sequence to sequence translation \cite{sutskever2014sequence}, image segmentation \cite{badrinarayanan2015segnet,ronneberger2015u}, or image captioning \cite{mun2017text}, combining a \acs{CNN} encoder with a \acs{LSTM} decoder. Image restoration has also been addressed by using denoising autoencoders with symmetric skip connections \cite{mao2016image}. The following is a brief explanation of convolutional \acsp{EDN}, which we will use in our system described in Section \ref{sec:statten} for spatio-temporal visual attention estimation.

\subsection*{Convolutional Encoder-Decoder Networks}
An image comprises a set of features located in different regions. Given an input image, deep \acfp{CED} compute an invariant feature vector that encodes \emph{what} features are in the image (their presence or absence), via a set of transformation parameters, which entail \emph{where} these features are found within the image (their location) \cite{huang2007unsupervised}. 

Figure \ref{fig:354_edn}(b) shows an example diagram of a \acs{CED} network for object segmentation. On the one hand, the encoder network consists of a series of \acs{CONV} and \acs{POOL} sub-sampling layers, just like \acsp{CNN}. If we look at the encoder of the example, we will notice that it is composed by the 13 \acs{CONV} layers in the \emph{VGG-16} \cite{vgg2014} represented in Figure \ref{fig:352_vgg_resnet}(b). As explained above in Section \ref{sec:convolutional_neural_networks}, \acs{POOL} layers help to manage translation invariance to small spatial shifts in the image and, at the same time, reduce its dimensionality. Therefore, they provide compacted representations which can be useful to efficiently carry out classification or regression problems. On the other hand, the decoder network is constituted by several \acs{CONV} and upsampling layers, which reconstruct the feature vector and converge into a final \acs{CONV} output layer. A pixel-wise binary classification task is performed at the output in order to achieve the desired skin lesion segmentation mask. The decoder of the example has the same number of layers than the encoder, establishing a correspondence between \acs{POOL} and upsampling layers. 

\subsection{Recurrent Neural Networks}\label{sec:recurrent_neural_networks}
\acfp{RNN} \cite{rumelhart1986learning, graves2012supervised} are \acsp{NN} with feedback connections specialized for the processing of sequential data. Widely-used for language modeling in speech recognition \cite{mikolov2010recurrent} since 2010, when they outperformed the standard n-gram models \cite{goodman2001bit}, \acsp{RNN} have more recently been used in computer vision applications such as image captioning \cite{xu2015show} or video action recognition \cite{sharma2015action}.

\acsp{RNN} operate on input sequences composed of vectors ${\bf x}_{t}$, which are denoted by the time step index $t$, and have a repetitive structure: drawing on cycles, they are able to capture the influence of a present variable ${\bf x}^{(t)}$ at a time $t$ on its future value ${\bf x}^{(t+1)}$, by learning a set of parameters ${\bf w}$ that are shared across the network states. States correspond to the hidden units ${\bf z}^{(t)}$ of the network, whose relationship is typically expressed by the following equation: 

\begin{gather}
{\bf z}^{(t)} = f({\bf z}^{(t-1)},{\bf x}^{(t)};{\bf w}).
\end{gather}

\begin{figure*}[!t]
\centering
	\includegraphics[trim=0cm 0cm 0cm 0cm, width=1\textwidth]{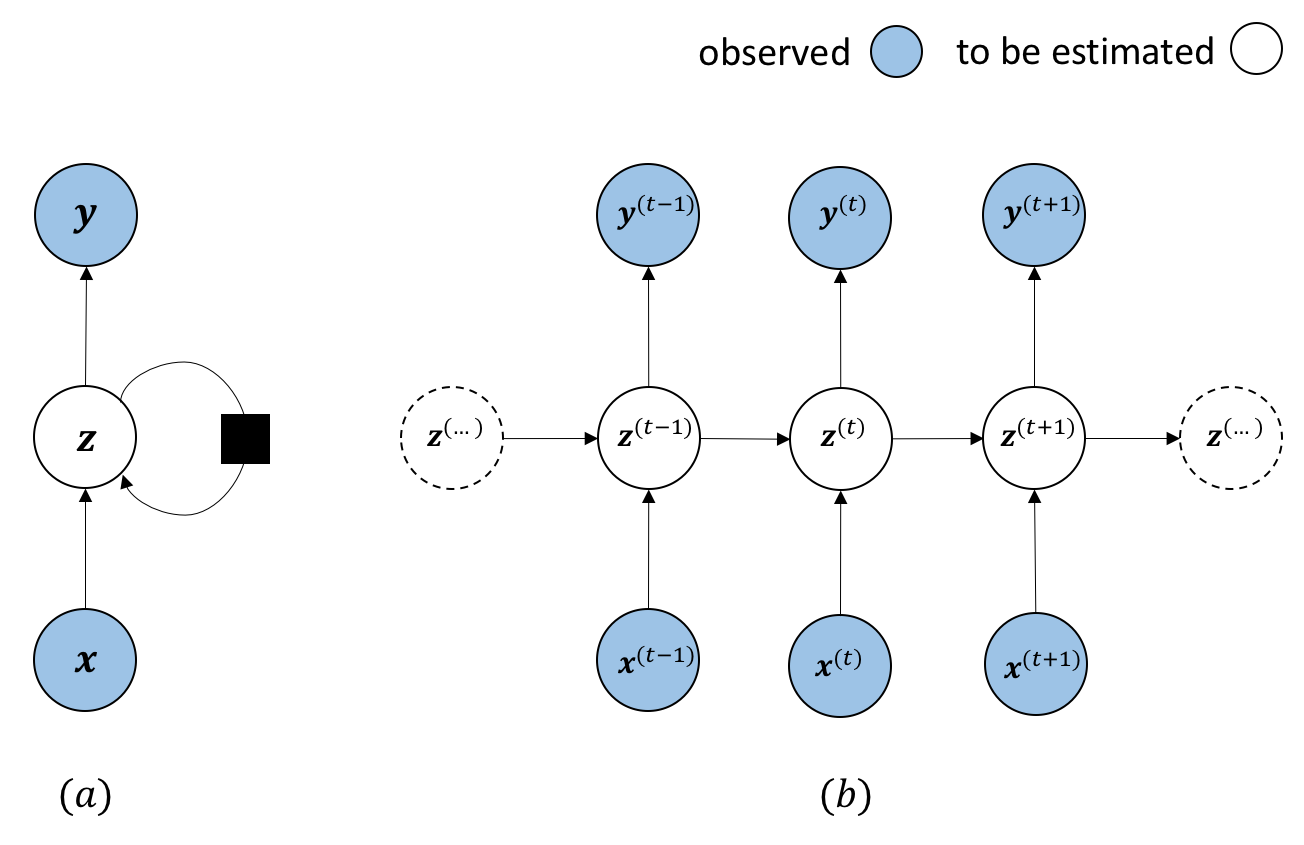}	
\caption[Graphical representations of a \acf{RNN}.]{Graphical representations of a \acs{RNN} that maps an input sequence of ${\bf x}$ values to an output sequence ${\bf y}$. (a) \acs{RNN} represented as a compact graph. Each node corresponds to a vector that contains the layer's activations. The black square means a delay of a single time step, from the state $z^{(t)}$ to $z^{(t+1)}$. (b) \acs{RNN} represented as a time-unfolded graph. Each node corresponds to a particular time instance.}
\label{fig:353_rnn}
% \vspace{-0.4cm}
\end{figure*}

\acsp{RNN} have two primary advantages \cite{goodfellow2016deep}. First, they learn a model based on transitions between states, which always has the same input size. Moreover, the model can be independent of the sequence length by making use of the same transition function $f$ with the same parameters ${\bf w}$ at every time step, instead of learning a separate model for each one. Given a long sequence, \acsp{RNN} usually work on mini-batches of shorter length $\tau$, similar to how feed-forward networks deal with samples.

It should be noted that we still have not mentioned the output layers that use the information from states to make predictions. Depending on the architecture design, \acsp{RNN} can generate either an output at each time step or a single output for a whole sequence. Besides, they can have recurrent connections between units or, in contrast, from the output at a time step to the hidden units at the subsequent time step.

Figure \ref{fig:353_rnn} shows both the compact and the time-unfolded graph of a representative \acs{RNN} example. The network maps an input sequence ${\bf x}$ of values or vectors to an output sequence of ${\bf y}$. Given a initial state ${\bf z^{(0)}}$ and a mini-batch of length $\tau$, the following update equations are applied from $t=1$ to $t=\tau$:

\begin{align}
& {\bf z}^{(t)} = g({\bf W}{\bf z}^{(t-1)} + {\bf U}{\bf x}^{(t)} + {\bf b}) \\
& {\bf y}^{(t)} = h({\bf V}{\bf z}^{(t)} + {\bf c})
\end{align}

\noindent where ${\bf U}$, ${\bf V}$ and ${\bf W}$ are weight matrices that model input-to-hidden, hidden-to-output and hidden-to-hidden connections, respectively; ${\bf b}$ and ${\bf c}$ are bias vectors; $g$ is the activation functions for the hidden units; and $h$ the activation function at the output.

\subsubsection*{Gradient-based learning in \acsp{RNN}} 
The unfolded graph in Figure \ref{fig:353_rnn}(b) illustrates how information ${\bf x}^{(t)}$ flow in the network during the forward pass from left to right, computing the output values ${\bf y}^{(t)}$. The total loss $J$ for the sequence is given by the sum of the loss $J^{(t)}$ at each time step. Then, in the backward pass, gradients are computed for each time step. 

The algorithm to compute gradients in \acsp{RNN} is called \acf{BPTT} \cite[Section 10.2.2]{goodfellow2016deep}, and applies back-propagation to the unfolded graph. For each node, the gradient is computed recursively, based on the gradients of the following nodes in the graph. In vector notation, given $\frac{\partial J}{\partial J^{(t)}} = 1$, the gradient $(\nabla_{\mathbf{y}^{(t)}}J)_{i}$ for each value $y_{i}^{(t)}$ on the output sequence at time step $t$ can be written as:

\begin{gather}
(\nabla_{\mathbf{y}^{(t)}}J)_{i} = \frac{\partial J}{\partial J^{(t)}} \frac{\partial J^{(t)}}{\partial y_{i}^{(t)}} = \frac{\partial J^{(t)}}{\partial y_{i}^{(t)}}, 
\end{gather}
	
\noindent Then, the gradient on the hidden state $\nabla_{\mathbf{z}^{(t)}}J$ is as follows:

\begin{gather}
\nabla_{\mathbf{z}^{(t)}}J = \left(\frac{\partial \mathbf{z}^{(t+1)}}{\partial \mathbf{z}^{(t)}}\right)^T \left(\nabla_{\mathbf{z}^{(t+1)}}J\right) + \left(\frac{\partial \mathbf{y}^{(t)}}{\partial \mathbf{z}^{(t)}}\right)^T \left(\nabla_{\mathbf{y}^{(t+1)}}J\right).
\end{gather}

%In the language of tensors, given $\frac{\partial J}{\partial J^{(t)}} = 1$, the gradient $\nabla_{\mathsf{Y}^{(t)}}J$ on the output at time step $t$ can be written as:
%
%\begin{gather}
%\nabla_{\mathsf{Y}^{(t)}}J = \sum_{i} \frac{\partial J}{\partial J^{(t)}} \frac{\partial J^{(t)}}{\partial \mathsf{Y}_{i}^{(t)}} = \sum_{i} \frac{\partial J^{(t)}}{\partial \mathsf{Y}_{i}^{(t)}}, 
%\end{gather}
%	
%\noindent Then, the gradient on the hidden state $\nabla_{\mathsf{Z}^{(t)}}J$ is as follows:
%
%\begin{gather}
%\nabla_{\mathsf{Z}^{(t)}}J = \sum_{i} (\nabla_{\mathsf{Z}^{(t)}} \mathsf{Y}_{i}^{(t)}) \frac{\partial J}{\partial \mathsf{Y}_{j}^{(t)}} + \sum_{j} (\nabla_{\mathsf{Z}^{(t)}} \mathsf{Z}_{j}^{(t+1)}) \frac{\partial J}{\partial Z_{j}^{(t+1)}}.
%\end{gather}
%
%\noindent In both equations, $i$ and $j$ is are index variables to represent the complete tuple of coordinates in $\mathsf{Y}^{(t)}$ and $\mathsf{Z}^{(t)}$, respectively. 

Once these gradients are computed for the computational graph associated with a mini-batch of length $\tau$ taken from an entire longer sequence, the gradients with respect to the weights matrices ${\bf U}$, ${\bf V}$, and ${\bf W}$, and bias vectors ${\bf b}$ and ${\bf c}$ can be obtained, so that they can be subsequently updated.
	
\subsubsection*{Long Short-Term Memory Units} 
The main issue when learning long-term dependencies with a \acs{RNN} is that gradients propagated over many states are very small or large in magnitude, either vanishing or exploding, dramatically hampering the optimization process. In order to reduce these effects, the scale of the initial weights has to be chosen carefully.

Although the problem of learning long-term dependencies continues being one of the main challenges in deep learning, several techniques have been proposed in order to alleviate it (e.g. \emph{gradient clipping} \cite{pascanu2013difficulty} before the weights update rule produced very large gradient magnitudes). Here we discuss a special type of sequence models, known as \emph{gated \acsp{RNN}}, whose objective is to define paths through time with non-vanishing and non-exploding derivatives. Among gated \acsp{RNN}, it is worth mentioning \acf{LSTM} units \cite{hochreiter1997long}, used in our system for modeling attention in the temporal dimension (see Section \ref{sec:statten}) and described below, and \acfp{GRU} \cite{cho2014learning}. 

\begin{figure*}[!t]
\centering
	\includegraphics[trim=0cm 0cm 0cm 0cm, width=1\textwidth]{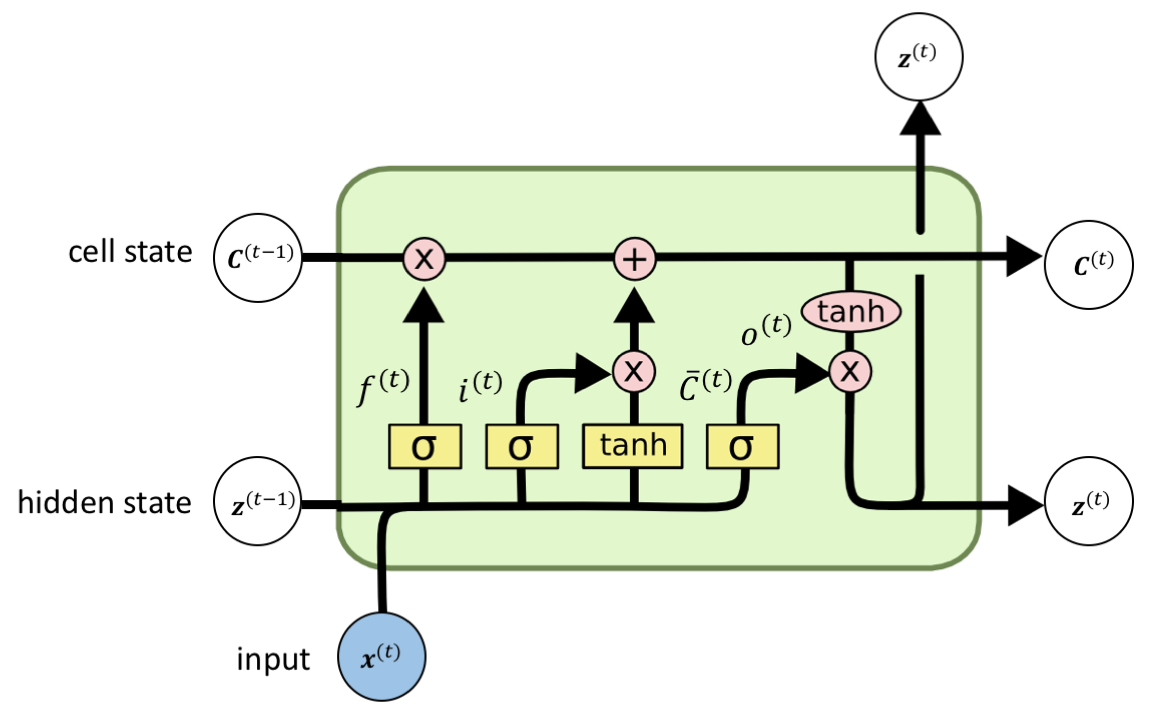}	
\caption[Diagram of a \acf{LSTM} unit.]{Diagram of a \acs{LSTM} unit. Pink circles indicate point-wise operations, while yellow boxes represent, from left to right, forget $f^{(t)}$, input $i^{(t)}$, cell state candidate values $\overline{C}^{(t)}$ and output $o^{(t)}$ gates, which control the flow of information by adding or removing information to/from the cell state $C^{(t)}$, based on the input ${\bf x}^{(t)}$, and determine the hidden state $z^{(t)}$. Adapted from \cite{olah2017august}.}
\label{fig:353_lstm}
% \vspace{-0.4cm}
\end{figure*}

Figure \ref{fig:353_lstm} shows the block diagram of a common \acs{LSTM} unit. A \acs{LSTM} model is composed by cells characterized by a state ${\bf C}^{(t)}$, represented in the diagram by the top horizontal line. Cells involve an internal recurrence or self-loop which complements the outer recurrence of traditional \acsp{RNN}. This internal recurrence serves to control the flow of information by adding or removing information to/from the cell state, and is defined by means of a system of gating units:

\begin{itemize}
	\item First, the \emph{forget gate} unit ${\bf f}^{(t)}$ decides which information to throw away from the cell state, based on the input ${\bf x}^{(t)}$ and the previous hidden units ${\bf z}^{(t-1)}$, and using a sigmoid activation function:
	
	\begin{gather}
	{\bf f}^{(t)} = sigm \left( {\bf U}_{f}{\bf x}^{(t)} + {\bf W}_{f}{\bf z}^{(t-1)} + {\bf b}_{f} \right)
	\end{gather} 
	
	\noindent where ${\bf U}_{f}$, ${\bf W}_{f}$ and ${\bf b}_{f}$ are input weights, recurrent weights and biases for the forget gate.
	
	\item Then, in order to decide what new information is going to be stored, there are two gates: the \emph{input gate}, which determines the values to be updated:
	
	\begin{gather}
	{\bf i}^{(t)} = sigm \left( {\bf U}_{i}{\bf x}^{(t)} + {\bf W}_{i}{\bf z}^{(t-1)} + {\bf b}_{i} \right)
	\end{gather}
	
	\noindent and a \emph{tanh} activation, which creates a vector of new candidate values to be added to the cell state:  
	
	\begin{gather}
	{\bf \overline{C}}^{(t)} = tanh \left( {\bf U}_{C}{\bf x}^{(t)} + {\bf W}_{C}{\bf z}^{(t-1)} + {\bf b}_{C} \right)
	\end{gather}
	
	\noindent where ${\bf U}_{i}$, ${\bf U}_{C}$ are input weights; ${\bf W}_{i}$, ${\bf W}_{C}$ recurrent weights; and ${\bf b}_{i}$, ${\bf b}_{C}$ are biases. Based on ${\bf f}^{(t)}$, ${\bf i}^{(t)}$ and ${\bf \overline{C}}^{(t)}$, the \emph{cell state} ${\bf C}^{(t)}$ is updated as follows:
	
	\begin{gather}
	{\bf C}^{(t)} = {\bf f}^{(t)}{\bf C}^{(t-1)} + {\bf i}^{(t)}{\bf \overline{C}}^{(t)}
	\end{gather}
	
	\item Finally, the output value, at the \emph{output gate} ${\bf o^{(t)}}$, and the current hidden state ${\bf z}^{(t)}$ are computed:
	
	\begin{gather}
	{\bf o}^{(t)} =  sigm \left( {\bf U}_{o}{\bf x}^{(t)} + {\bf W}_{o}{\bf z}^{(t-1)} + {\bf b}_{o} \right)
	\end{gather}
	
	\noindent where ${\bf U}_{o}$, ${\bf W}_{o}$ and ${\bf b}_{o}$ are input weights, recurrent weights and biases again; and
	
	\begin{gather}
	{\bf z}^{(t)} = {\bf o}^{(t)} tanh({\bf C}^{(t)}).
	\end{gather}
\end{itemize}

Firstly introduced for precipitation nowcasting in \cite{xingjian2015convolutional}, convolutional \acsp{LSTM} constitute an extension of \acsp{LSTM} to operate on images and \acs{2D} feature maps, and are also used as part of the spatio-temporal visual attention \acsp{EDN} presented in Section \ref{sec:stva_autoencoder}. \acs{CONV}-\acsp{LSTM} simply incoporate convolutional operators either to the input-to-hidden or the hidden-to-hidden transitions, being expressed as follows:

\begin{align}
& {\bf F}^{(t)} = sigm \left( {\bf U}_{f}*{\bf X}^{(t)} + {\bf W}_{f}*{\bf Z}^{(t-1)} + {\bf b}_{f} \right) \\
& {\bf I}^{(t)} = sigm \left( {\bf U}_{i}*{\bf X}^{(t)} + {\bf W}_{i}*{\bf Z}^{(t-1)} + {\bf b}_{i} \right) \\
& {\bf \overline{C}}^{(t)} = tanh \left( {\bf U}_{C}*{\bf X}^{(t)} + {\bf W}_{C}*{\bf Z}^{(t-1)} + {\bf b}_{C} \right) \\
& {\bf C}^{(t)} = {\bf F}^{(t)}\circ{\bf C}^{(t-1)} + {\bf I}^{(t)}\circ{\bf \overline{C}}^{(t)} \\
& {\bf O}^{(t)} =  sigm \left( {\bf U}_{o}*{\bf X}^{(t)} + {\bf W}_{o}*{\bf Z}^{(t-1)} + {\bf b}_{o} \right) \\
& {\bf Z}^{(t)} = {\bf O}^{(t)}\circ tanh({\bf C}^{(t)}),
\end{align}

\noindent where $*$ and $\circ$ denote the convolution operator and the Hadamard product, respectively.

\begin{figure*}[!t]
\centering
	\includegraphics[trim=0cm 0cm 0cm 0cm, width=1\textwidth]{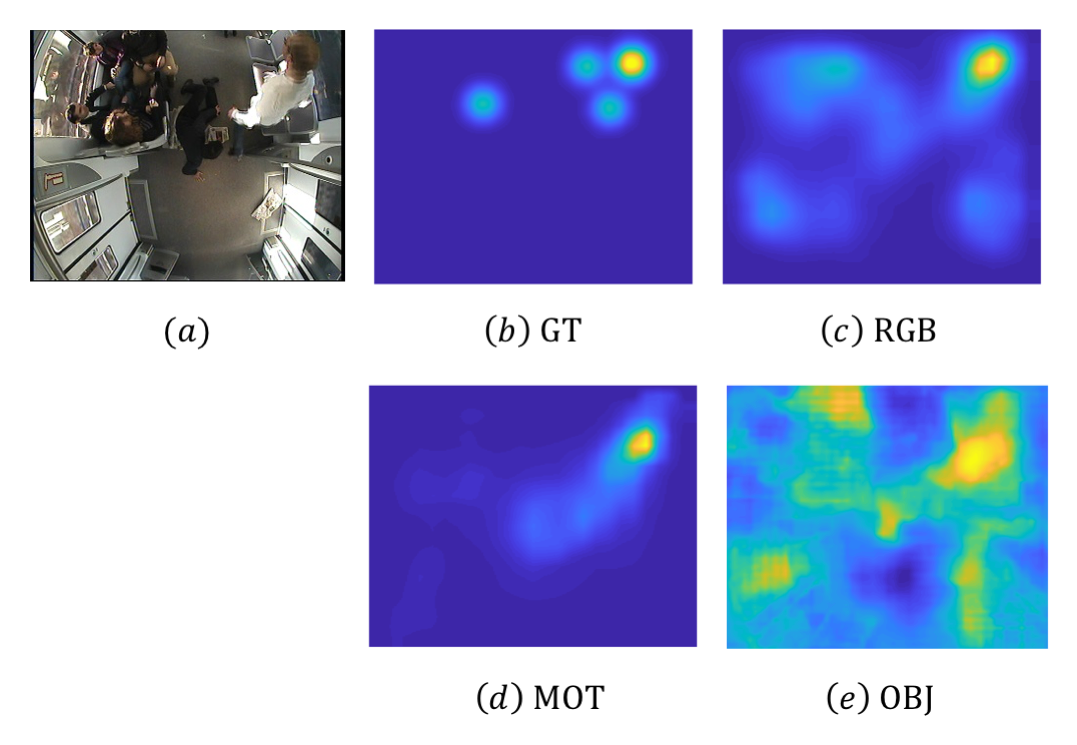}
\caption[RGB-based, motion and objectness feature maps computed for an example frame taken from BOSS \cite{BOSS} database.]{RGB-based, motion and objectness feature maps computed for an example frame taken from BOSS \cite{BOSS} database. (a) Original frame. (b) \acs{GT} fixation map. (c) RGB-based feature map. (d) Motion feature map. (e) Objectness feature map.}
\label{fig:54_feature_learning}
% \vspace{-0.4cm}
\end{figure*} 

\section{Feature learning for visual attention guidance}\label{sec:cnns_features6}
In this section, we describe three feature extraction \acsp{CNN} for visual attention guidance. Unlike in our first system, presented in Chapter \ref{ch:atom}, our main goal now is not to understand how visual attention works in diverse contexts, but to model attention in the temporal domain using spatio-temporal \acsp{VAM} as an input. 

For that purpose, we will first make use of three fundamental visual attention feature maps to estimate spatio-temporal visual attention: RGB-based spatial, optical flow-based motion, and objectness-based maps. 

While the networks used to obtain spatial and motion feature maps directly learn from \acs{GT} fixation maps, the model for objectness minimizes a loss function depending on \acs{GT} annotated object masks. These feature maps have been obtained by adapting the well-known VGG-16 \cite{vgg2014} and ResNet-50 \cite{he2016deep} networks, successfully applied first to image recognition, and more recently to saliency estimation, mainly in still images \cite{lstm2018cornia}. 

\subsection{RGB-based spatial network}\label{sec:cnn_rgb}
A spatial feature map $f_{RGB}$ is first computed by using a modified version of the ResNet-50 \cite{he2016deep} introduced in Section \ref{sec:cnn_architecture_design}, and a $224\times224$ RGB image as input. Similarly to the dilated residual convolutional block presented by Marcella Cornia et al. in \cite{lstm2018cornia} as part of the \acs{SAM} model, we first remove the \acs{FC} layers at the end of the network and then introduce dilated convolutions in either \acs{CONV}4 or \acs{CONV}5 blocks with dilation rate $d=2$ and $d=4$, respectively. Then, the output tensor of the \acs{CONV}5 block, with 2048 units, is fed into an additional \acs{CONV} block composed of two $k=3\times3$ and $k=1\times1$ layers, both with 128 units. Finally, we place a final layer with a unique unit and linear activation, which corresponds to the output spatial map, with dimension $26\times26$. Figure \ref{fig:54_feature_learning}(c) shows an example of a RGB-based feature map associated with a frame taken from BOSS \cite{BOSS} database.

\subsection{Optical flow-based motion network}\label{sec:cnn_motion}
Motion feature maps can be achieved by means of \acsp{CNN} that take as input a pair of subsequent frames or, in contrast, the previously estimated optical flow from these, similarly to our traditional feature extraction process in Section \ref{sec:motion_features} or to the networks proposed by Cagdas Bak et al. in \cite{bak2018spatio}. The network used for our experiments has been inspired by the latter work but, unlike this approach and for the sake of continuity, it builds over the modified ResNet-50 \cite{he2016deep} architecture used for RGB-based maps to extract a motion feature map $f_{MOT}$. The model now receives as input an $224\times224$  image with three channels, corresponding to the horizontal and vertical optical flow components and its associated motion magnitude map. All these channels are first re-scaled to the range $[0,255]$. Figure \ref{fig:54_feature_learning}(d) includes an example of a motion feature map for a frame taken from BOSS \cite{BOSS} database.

\subsection{Objectness-based network}\label{sec:cnns_based_features6}
Computational models for salient object detection have also demonstrated the importance of objects regardless their semantic category \cite{Rahtu2010,wang2018video}. Any object in motion is often noticeable and, if more than one object appears on a frame, attention will choose one depending on the conspicuity of their associated low-level properties.

As we anticipated in Section \ref{sec:guiding_features}, we make again use of the Deep Contrast Network for general object detection introduced by Li et al. in \cite{DeepSaliencyObject}. Unlike in our previous system, we now only consider the final fused objectness feature map, which we denote as $f_{OBJ}$ and is the output of the DCL model, prior to the \acs{CRF}-based refinement applied then. In order to achieve this map, the model makes use of two streams:

\begin{itemize}
	\item The first stream of the architecture is based on the VGG-16 \cite{vgg2014} network presented in Section \ref{sec:cnn_architecture_design}. In the same way as most deep networks for saliency estimation, it includes two additional \acs{CONV} layers after the final \acs{POOL}5 layer, which in fact replace the original \acs{FC} ones. In addition, it substitutes some of the traditional \acs{CONV} layers by dilated ones. More precisely, the three \acs{CONV}5 layers and the two extra \acs{CONV} layers incorporated after the final pooling layer have dilation rates $d=2$ and $d=4$, respectively. Moreover, with the purpose of implementing a multi-scale version of VGG-16 \cite{vgg2014}, the authors connect three more \acs{CONV} blocks to each of the first four MAX \acs{POOL} layers. 
	\item The second stream is a superpixel or segment-wise pooling network, which models both visual contrast between regions and discontinuities in object boundaries. 	
\end{itemize}

The final objectness-based feature map is the linear combination of the four outputs from the multi-scale \acs{CONV} blocks, the final output map of the first stream and the segment-wise map from the second stream. An example of an objectness-based feature map corresponding to a frame taken from BOSS \cite{BOSS} database is shown in Figure \ref{fig:54_feature_learning}(e).

\begin{figure*}[!t]
\centering
	\includegraphics[trim=0cm 0cm 0cm 0cm, width=1\textwidth]{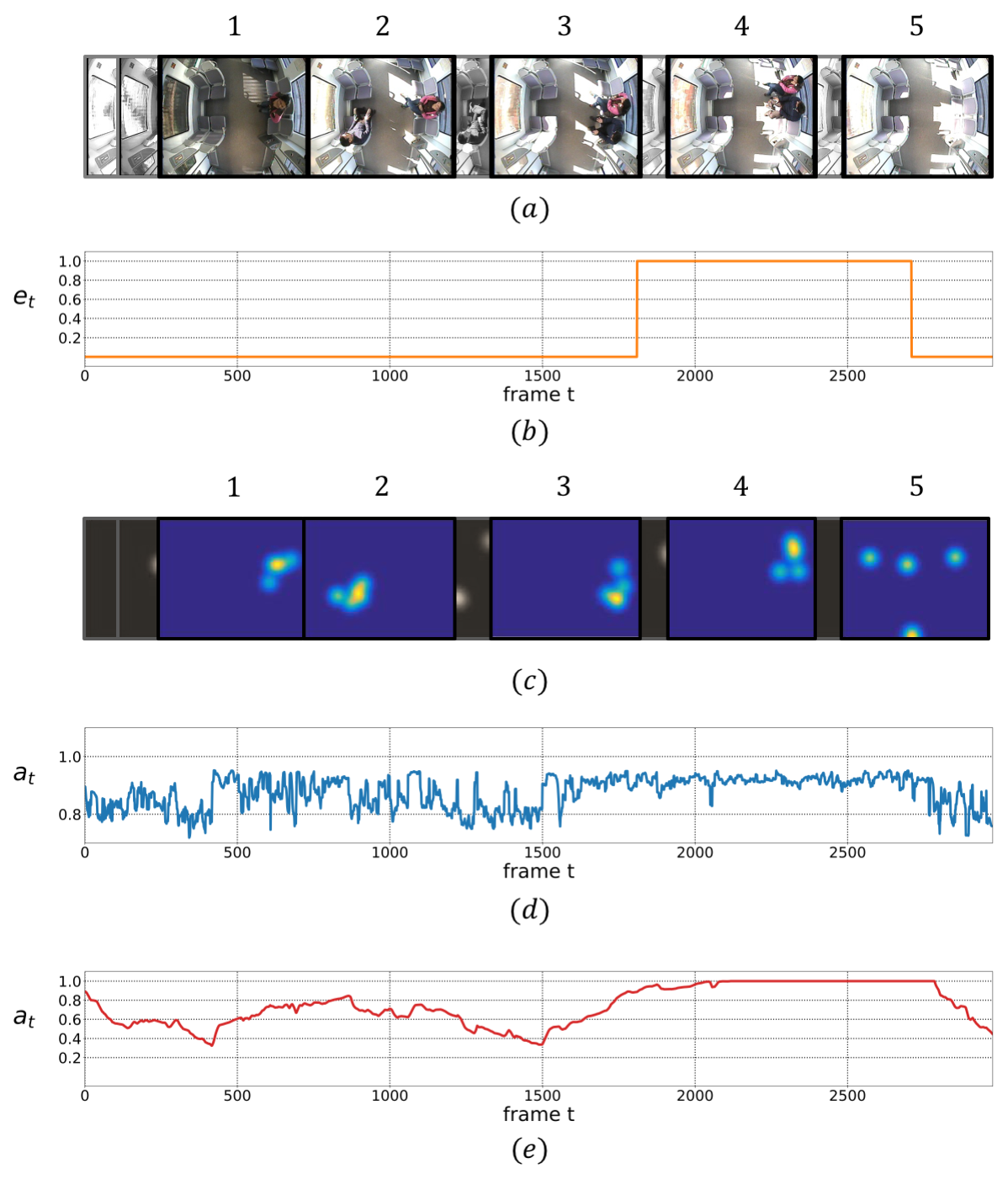}
\caption[Visual attention in the temporal domain modeled in a video-surveillance sequence taken from BOSS \cite{BOSS} database.]{Visual attention in the temporal domain modeled in a video-surveillance sequence taken from BOSS \cite{BOSS} database. The sequence shows a woman harassment scene on a train. (a) Sequence scenes. 1. Woman sits on the train. 2. Man gets on the train. 3. Man approaches the woman. 4. Man bothering the woman. 5. Woman and man leave the train. (b) Anomaly detection signal $e_{t}$, which is set to 1 when an anomaly happens. (c) Fixation maps. Fixations are gathered from 5 users watching the video. (d) Raw temporal attention response $a_{t}$. (e) Filtered temporal attention response $a_{t}$. According to the fixations-based signals provided, attention achieves its maximum value just before and at the moment of the harassment (3,4). However, it should be noted that other events also highly attract the attention of observers, such as the moment when the man appears on the scene (2). Therefore, temporal attention response $a_{t}$ should be considered an early filtering mechanism along time, that allows selecting time segments of special importance, which often match with anomalies.}
\label{fig:64_tva_gt}
% \vspace{-0.4cm}
\end{figure*}

\section{Spatio-Temporal to Temporal Visual Attention Network}\label{sec:statten}
In this section, we describe in detail our system for visual attention estimation in the temporal domain, which we have called \acf{ST-T-ATTEN}.

\subsection{Fundamental hypothesis of the model}\label{sec:statten_overview}
The \acs{ST-T-ATTEN} proposed for modeling the temporal dimension of visual attention is motivated by the following two assumptions:

\begin{enumerate}
	\item A measurement of task-driven visual attention in the temporal domain can be drawn studying the fixation dispersion across viewers performing a task in a particular context.
	
	\item Visual attention in the temporal domain can be modeled from an accurate estimation of spatio-temporal visual attention.
\end{enumerate}

First, we have introduced in Section \ref{sec:tva_related_work} that there is a significant correlation between eye movement sequences of different observers performing the same task when they perceive a surprising or anomalous event \cite{howard2013suspiciousness,sharma2016eye}. Therefore:

\begin{center} 
\emph{A measurement of task-driven visual attention in the temporal domain can be drawn studying the fixation dispersion across viewers performing a task in a particular context.}
\end{center} 

Given a crowded and complex scenario, eye movements constitute a useful source to understand how visual attention works and what information should be selected for further processing. In addition, the temporal level of attention of observers might constitute a useful clue to detect suspicious events or anomalous situations to analyze: If all observers fix their attention at the same location at the same time, it is very likely that something noticeable is happening. 

Let us consider a video surveillance scenario such as the one presented in the example in Figure \ref{fig:64_tva_gt}, taken from BOSS \cite{BOSS} database. The database, further described in Appendix \ref{ch:databases}, contains video sequences with anomalous events, which have been recorded in suburban trains. The sequence in (a) shows a man harassing a woman, while (b) includes an anomaly signal $e_{t}$, set to 1 when an anomalous situation happens (see scene 4). If we look at the fixation map in (c), we can notice that all observers are fixing their attention at the location of the anomaly. However, it should be noted that other events are also attracting the attention of viewers, such as the moment when the man gets on the train (see scene 2). This may basically happen due to two reasons: there are no more events on the scene, or there have not been significant changes in the video for long time until those new events. That is why, similarly to spatio-temporal visual attention, attention in the temporal domain measured by fixations should be always considered as an early filtering mechanism, which allows to select those time segments of special importance in a video, candidates to contain anomalies. These often correlate with anomalous situations, especially in complex videos with multiple simultaneous events. Let us emphasize that our objective is not to detect anomalies, but to develop an information filtering mechanism in order to select relevant time segments where a subsequent anomaly detection system may be more efficiently applied.

Hence, just by recording sequences of fixations from different subjects watching videos and computing a dispersion measurement of fixations across viewers, we could provide a \acs{GT} measurement of visual attention in the temporal domain.

Second, we have studied along this thesis that spatio-temporal visual attention maps aim to predict viewers fixations in videos or dynamic scenarios. Therefore:

\begin{center} 
\emph{Visual attention in the temporal domain can be modeled from an accurate estimation of spatio-temporal visual attention.}
\end{center} 

Spatio-temporal \acsp{VAM} can be understood, for each frame in a video, as a \acs{2D} probability density function which might provide a temporal attention response similar to the one measured from fixations dispersion across viewers. This motivates us to develop a system to model attention in the temporal domain by taking advantage of spatio-temporal visual attention predictions. %, such as the ones achieved by our approach in Chapter \ref{ch:atom}. The closer the estimation of visual attention to the behavior of observers, the more accurate the temporal response estimated will be. This system could simplify real applications which imply a great deal of visual information processing at the same time, reducing human errors and speeding up the decision making process to prevent anomalous events.

%Continuing from the above example, a \acs{CCTV} operator in a video surveillance control room has the task of finding a possible target (e.g. robberies, explosions, road accidents, etc.) amongst multiple distractors (e.g. people running, vehicles with similar characteristics, crowds, etc.) displayed at the same time in not one, but a large array of 20 to more than 500 screens \cite{stainer2013looking}. In common scenarios, such as cameras placed in public environments or industrial plants, anomalous events seldom happen, which makes this task even more difficult for operators standing in front of screens during long time periods. Although operators have also their own controllable spot monitor to select the most outstanding camera at each time, and have been previously trained to solve this problem, the task is still ill-suited given humans perceptual and cognitive constrains. 

%\afterpage{
%\begin{landscape}
%\newgeometry{right=3cm}
\begin{sidewaysfigure}
\centerline{\includegraphics[trim=0cm 0cm 0cm 0cm, width=0.85\textwidth]{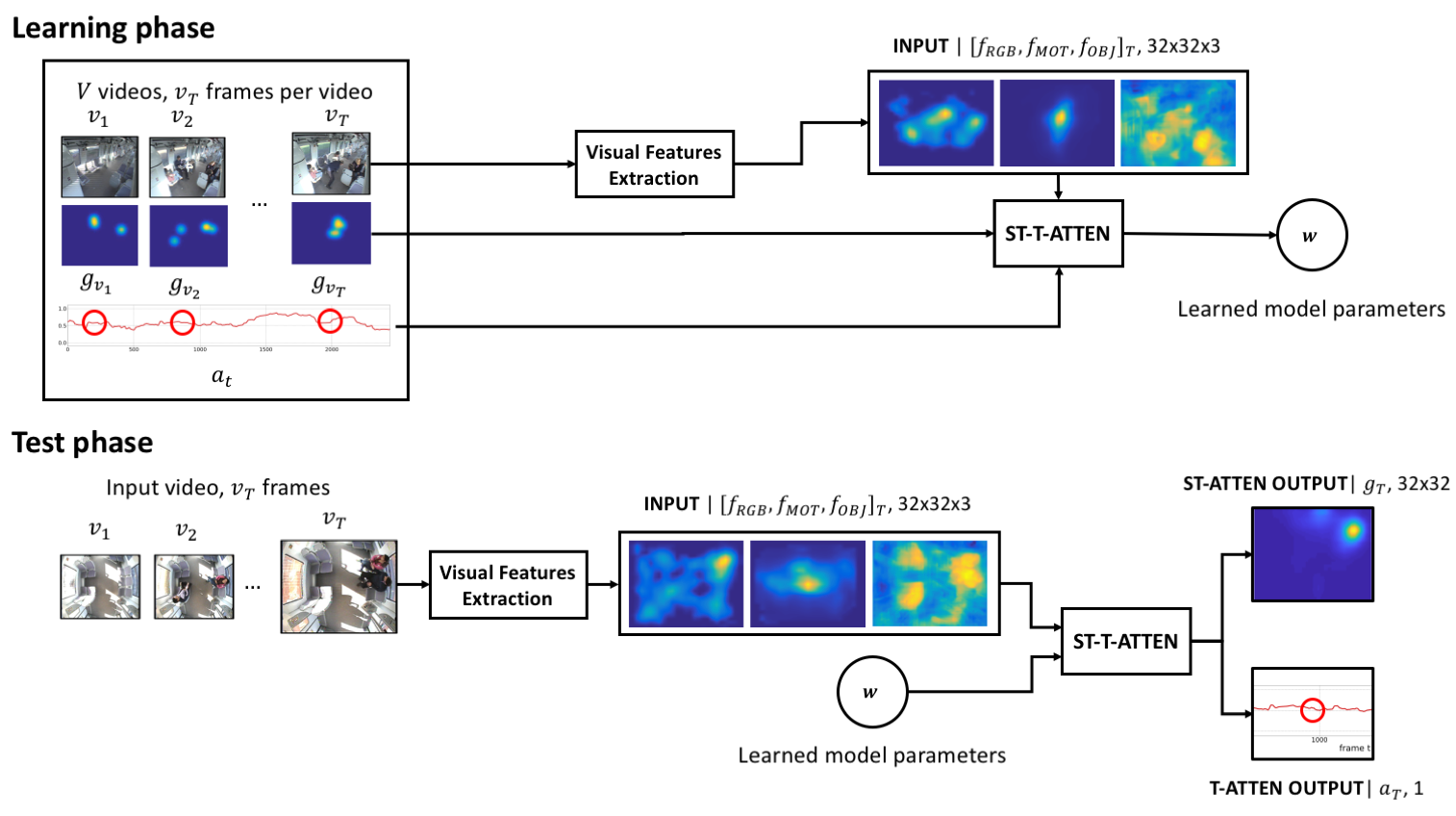}}
\caption[Processing pipelines of the \acf{ST-T-ATTEN} proposed.]{Processing pipelines of the \acs{ST-T-ATTEN} proposed. First, during the learning phase, the system receives a set of training videos and learns the optimal values for its associated parameters. Then, in the test phase, the architecture is qualified to estimate both spatio-temporal visual attention maps and temporal attention responses for new unseen videos.}
\label{fig:551_pipelines}
%\vspace{-0.4cm}
\end{sidewaysfigure}
%\restoregeometry
%\end{landscape}
%}

%, with the aim of reduce human failures and take decisions to prevent these events. 
%We have seen in section \ref{sec:deep_neural_networks} that the recent success of deep learning-based architectures has motivated researches to use them to solve a wide variety of applications \cite{krizhevsky2012imagenet,badrinarayanan2015segnet}; at the same time they attempt to understand their ability of seeing our world by means of highly effective hierarchical representations, such as the ones introduced in the previous section for visual attention guidance. These representations are nevertheless difficult to interpret at first glance most of the times, in contrast to traditional hand-crafted features (see section \ref{sec:handcrafted_features}). 

In this work, in line with the current dominant approaches in computer vision, we will make use of \acsp{CNN} for spatio-temporal visual attention estimation. In addition, we will assess the performance of \acsp{LSTM} for attention estimation in the temporal domain, which have shown impressive results for time series forecasting \cite{xingjian2015convolutional}, sequence to sequence learning \cite{sutskever2014sequence} or image captioning \cite{xu2015show}, among other applications.

Figure \ref{fig:551_pipelines} shows the processing pipelines of our supervised approach. First, during the learning phase, the system receives a set of $V$ training videos and extracts several feature maps for visual attention guidance (RGB, motion and objectness), which become high-level features representing the corresponding frames $v_{t}$. Moreover, frames in these videos are annotated with eye fixations from several subjects, which can be represented either as spatio-temporal fixations maps or their corresponding temporal responses. All these inputs are used to learn the optimal values for the parameters ${\bf w}$ of the architecture proposed, where two stages can be differentiated:

\begin{enumerate}
	\item A \acf{ST-ATTEN} consisting on a \acs{CED}, which has the important mission of providing accurate spatio-temporal visual attention maps.
	\item A \acs{LSTM}-based \acf{T-ATTEN}, which will ultimately serve to model attention in the temporal domain.
\end{enumerate}

Then, in the test phase, the \acs{ST-T-ATTEN} is qualified to estimate both spatio-temporal visual attention maps and temporal attention responses for new unseen videos.

%Different configurations for these stages are proposed in sections \ref{sec:autoencoders} and \ref{sec:tva_modeling}. Then, in the test phase, the \acs{ST-T-ATTEN} has been trained and is qualified to estimate both spatio-temporal visual attention maps and temporal attention responses for new unseen videos.

\subsection{Fixation-based temporal ground-truth}\label{sec:fixations_gt}
Given that the current objective is not to predict a spatial attention response, but a temporal one, we need to generate a frame-level temporal \acs{GT}. As stated in recent behavioral studies \cite{howard2013suspiciousness,sharma2016eye}, and as we introduced in Section \ref{sec:statten_overview}, there is a noticeable consistency between observers' eye movements in a scene. Indeed, when an anomalous or suspicious event is happening, gaze locations from different subjects are highly correlated, especially if they are experts or users trained to perform a particular task. 

Therefore, on the basis of this fact, we propose a temporal \acs{GT} $a_{t}$ for each frame $v_{t}$, which is computed attending to the dispersion at fixation spatial locations from several subjects. In order to illustrate the \acs{GT} computation process, an example sequence taken from BOSS \cite{BOSS} video surveillance database is shown in Figure \ref{fig:64_tva_gt}. So far, in Chapter \ref{ch:atom}, we have defined a binary spatial attention response $g_{tn}$ for each spatial location $n$ in a frame with $N_{t}$ pixels. This response takes the value of one if the location has been fixated by an observer, and zero otherwise. Hence, given a \acs{GT} soft spatial map $g_{t}$ that comes from convolving each gaze location with a Gaussian filter, we compute the mean $\mu_{g_{t}} = (\mu_{g_{tx}}, \mu_{g_{ty}})$ and the standard deviation $\sigma_{g_{t}} = (\sigma_{g_{tx}}, \sigma_{g_{ty}})$ of the fixation locations: 

\begin{align}
&\mu_{g_{t}} = \frac{\sum_{n=1}^{N_{t}} g_{tn} {\bf \xv_{tn}}}{\sum_{n=1}^{N_{t}} g_{tn}} \\
&\sigma_{g_{t}} = \sqrt{\frac{\sum_{n=1}^{N_{t}} g_{tn}({\bf \xv_{tn}}-\mu_{g_{tn}})^2}{\frac{M_{t}-1}{M_{t}} \sum_{n=1}^{N_{t}} g_{tn}}},
\end{align}

\noindent where ${\bf \xv_{tn}} = (x_{tn}, y_{tn})$ represents the spatial coordinates vector of each location $n$, and $M_{t}$ stands for the number of non-zero response locations. Then, the raw temporal attention response $a_{t}$ can be computed as one minus the weighted mean of the standard deviations along frame width $X$ and height $Y$:

\begin{align}
a_{t} = 1-\frac{Y\sigma_{g_{tx}} + X\sigma_{g_{ty}}}{X + Y}.
\end{align}

%It should be recalled that observers quickly change their attention, so the \acs{GT} temporal attention signal associated to a video sequence is quite noisy. This noise can be reduced in real-time by applying a first-order infinite impulse response filter. In this work, we make use of the following \acf{EMA} to compute the temporal attention response $g_{t}$:
%
%\begin{align}
%g_{t} = 
%	\begin{cases}
%		g^{raw}_{1} &\quad\text{if } t=1 \\
%		\iota g^{raw}_{t} + (1-\iota) g_{t-1}  &\quad\text{if } t>1
%	\end{cases},
%\end{align}
%
%\noindent being $\iota$ a constant smoothing coefficient empirically set %to $\iota=0.01$.
%REVISAR! La version actual es con VIDYA. 

\noindent $X$ and $Y$ are normalizers, which balance the contribution of the standard deviations considered. The response $a_{t}$ thus takes values between 0, which corresponds to uninteresting frames (maximum fixations dispersion), and 1, which stands out attractive ones (maximum correlation between fixations). 

The signal $a_{t}$, as shown in Figure \ref{fig:64_tva_gt}(d), is very noisy. The rationale behind is that observers tend to continuously scan the scene, specially when there are no changes. This noise can be reduced in real-time by applying a first-order infinite impulse response filter. In this work, we make use of an adaptive technique known as \acf{VIDYA} \cite{chande1994new}, based on an \acf{EMA}, in order to filter the temporal attention response.

In addition, the dispersion on which this \acs{GT} temporal measure is founded may depend on the camera angle and perspective with respect to the objects in the video sequence. For this reason, it is convenient to normalize $a_{t}$ by subtracting the regular mean, which centers the response around $0$, and by dividing it by three times the standard deviation, which covers the $\sim 99.7$ of the response values, for different camera views. Finally, filtered $a_{t}$ is clipped to the range $[-1,1]$ and re-scaled to be in the same interval $[0,1]$ than the initial raw response. As shown in Figure \ref{fig:64_tva_gt}(e), this final response $a_{t}$ is notably softer. 

\begin{figure*}
\centering
	\includegraphics[trim=0cm 0cm 0cm 0cm, width=1\textwidth]{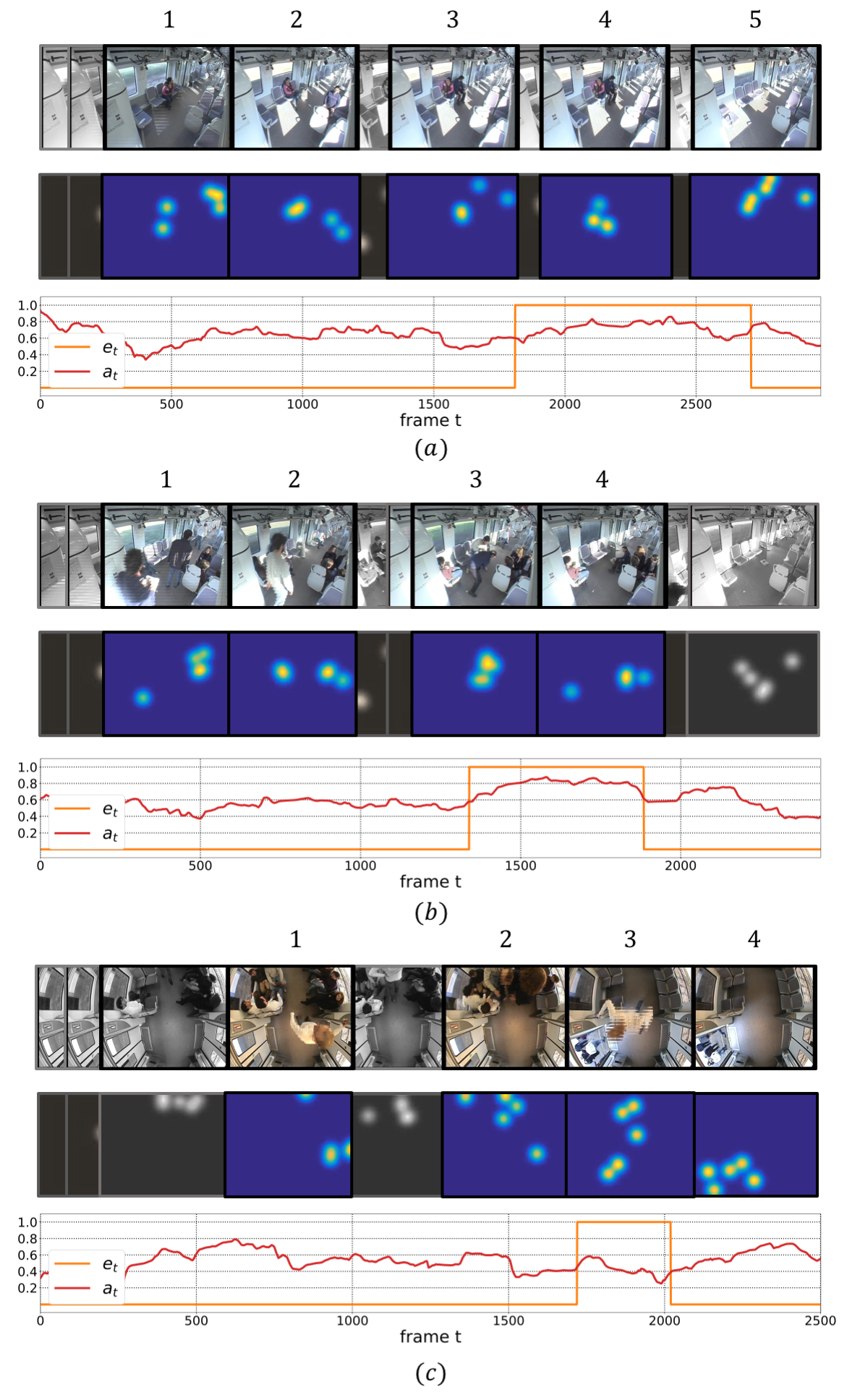}
\caption[Visual attention in the temporal domain modeled in three video-surveillance sequences taken from BOSS \cite{BOSS} database.]{Visual attention in the temporal domain $a_{t}$ modeled in three video-surveillance sequences taken from BOSS \cite{BOSS} database. Anomaly detection signal $e_{t}$ is also represented, which is set to 1 when an anomaly happens. (a) Woman harassment scene shown in Figure \ref{fig:64_tva_gt}, taken from a different camera view. (b) Passengers fighting for a newspaper. 1. A first man with a newspaper comes into the wagon. 2. A second man arrives. 3. Second man hits the first and destroys the newspaper. 4. First man lies on the ground. (c) Panic scene. 1. Passengers are warned about an accident. 2-3. Passengers running out of the wagon. 4. Nobody left on the train.}
\label{fig:73_anomaly_detection}
% \vspace{-0.4cm}
\end{figure*}

\subsection*{Hypothesis validation}
In this section we aim to validate the fundamental hypothesis of this second part of the thesis: attention in the temporal domain can be predicted using the dispersion of gaze locations recorded from several subjects.

Let us demonstrate the existing correlation between anomalies or suspicious events and the \acs{GT} temporal visual attention response $a_{t}$ proposed. For that purpose, we make use of \acs{GT} binary signals $e_{t}$, which indicate when anomalies occur in a video sequence, and consider a binary classification problem, where the objective is to classify video frames $v_{t}$ as anomalous or not, according to their associated filtered $a_{t}$. The performance of this proposed fixation-based response is thus assessed by computing the \acs{AUC} metric \cite{fawcett2006introduction} for the whole BOSS \cite{BOSS} video surveillance database. The closer is the value of this measure to 1, the higher is the existing correlation between $a_{t}$ response and the anomaly signal $e_{t}$. 

After conducting the experiment, we achieve an $AUC=0.876$, which clearly verifies the correlation between anomalies and $a_{t}$. This correlation can be also appreciated in the example in Figure \ref{fig:64_tva_gt}, where attention achieves its maximum value just before and at the moment of a woman harassment (see scenes 3 and 4). A second example is provided in Figure \ref{fig:73_anomaly_detection}(b), where two passengers fight for a newspaper. There are more people on the wagon when this situation happens (scenes 3 and 4). At that moment, fixation-based attention response becomes high, so it successfully captures the anomaly. 

Furthermore, we would also like to discuss why, although highly correlated, visual attention and anomaly are not equal variables, and therefore $AUC<1$. From a theoretical perspective, there are cases in which consistent visual attention is achieved in the absence of anomalies, such as simple scenarios with few people, as the one in Figure \ref{fig:73_anomaly_detection}(a), which is the same woman harassment scene shown in Figure \ref{fig:64_tva_gt}, taken from a different camera view. Another error case is shown in Figure \ref{fig:73_anomaly_detection}(c). The video sequence includes a panic scene, where passengers are warned about an accident (see scene 1) and run out of the wagon (see scenes 2-3). Scene involves all passengers and covers the whole image, so fixations dispersion is high at the moment of the anomaly (scene 3). 

Hence, as stated in the previous section, temporal attention can be seen as a filtering mechanism, which often correlates with anomalous situations. This correlation is particularly high in complex scenarios with multiple simultaneous events, which are those that require a greater cognitive effort to be understood. Considering that the proposed temporal attention response $a_{t}$ constitutes a filtering mechanism to be applied prior to an anomaly detection system, it is critical to obtain a low probability of non-detection ($P_{ND} \simeq 0$) with this signal, while we accept higher false-alarm probabilities ($P_{FA}>0$).    

\subsection{Model overview}\label{sec:architectures6}
\begin{figure*}
\includegraphics[trim=0cm 0cm 0cm 0cm, width=1\textwidth]{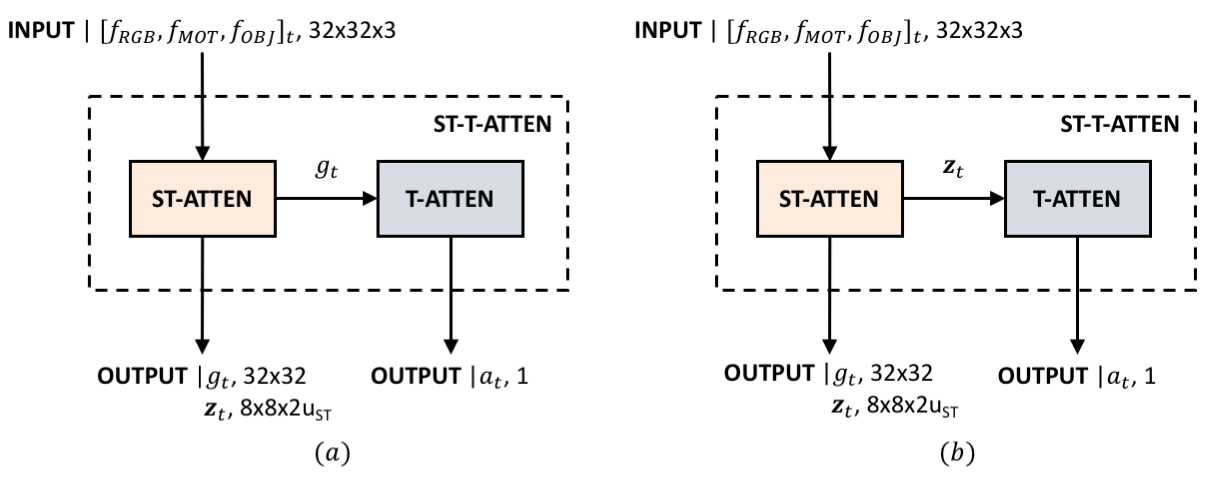}
\caption[Diagram of the \acf{ST-T-ATTEN} proposed.]{Diagram of the \acs{ST-T-ATTEN} proposed. The approach is built on the combination of two modules: 1) A \acs{ST-ATTEN} for spatio-temporal visual attention estimation; 2) A \acs{T-ATTEN} for modeling visual attention in the temporal domain. First, \acs{ST-ATTEN} receives at each timestep $t$ a frame represented by three feature maps $[f_{RGB},f_{MOT},f_{OBJ}]_{t}$. Then, \acs{T-ATTEN} receives as input (a) the spatio-temporal \acs{VAM} $g_{t}$ or (b) the latent representation ${\bf z}_{t}$ obtained at the output of \acs{ST-ATTEN} and estimates a temporal attention response $\hat{a}_{t}$.}
\label{fig:553_variable_blocks}
% \vspace{-0.4cm}
\end{figure*}

In this section, we overview the \acf{ST-T-ATTEN} proposed. The complete architecture of the system is represented in Figure \ref{fig:553_variable_blocks}. Our approach is built on the combination of two modules, which are described in the following sections: 1) A \acf{ST-ATTEN} for spatio-temporal visual attention estimation; 2) A \acf{T-ATTEN} for modeling visual attention in the temporal domain. 

First, given a video $v$, the \acs{ST-ATTEN} module consists in a \acf{CED}. At each timestep $t$, it receives an input frame $v_{t}$, which is represented by means of high-level feature maps for visual attention guidance, such as the ones described in Section \ref{sec:cnns_based_features6}: $[f_{RGB},f_{MOT},f_{OBJ}]_{v_{t}}$. The network is then able to compute, for each frame $v_{t}$, either a latent representation ${\bf z}_{t}$ of visual attention or a spatio-temporal visual attention map $\hat{g}_{t}$. 

Secondly, we propose two versions of the \acs{T-ATTEN} module, which are compared. In particular, the \acs{T-ATTEN} receives as input one of the two outputs provided by the \acs{ST-ATTEN}: either the spatio-temporal \acs{VAM} $\hat{g}_{t}$ (see Figure \ref{fig:553_variable_blocks}(a)) or the latent representation ${\bf z}_{t}$ (see Figure \ref{fig:553_variable_blocks}(b)). Then, it estimates, for each frame $v_{t}$, a temporal attention response $\hat{a}_{t}$. 

\subsection{Spatio-Temporal Visual Attention Network}\label{sec:stva_autoencoder}
\begin{figure*}
\includegraphics[trim=0cm 0cm 0cm 0cm, width=1\textwidth]{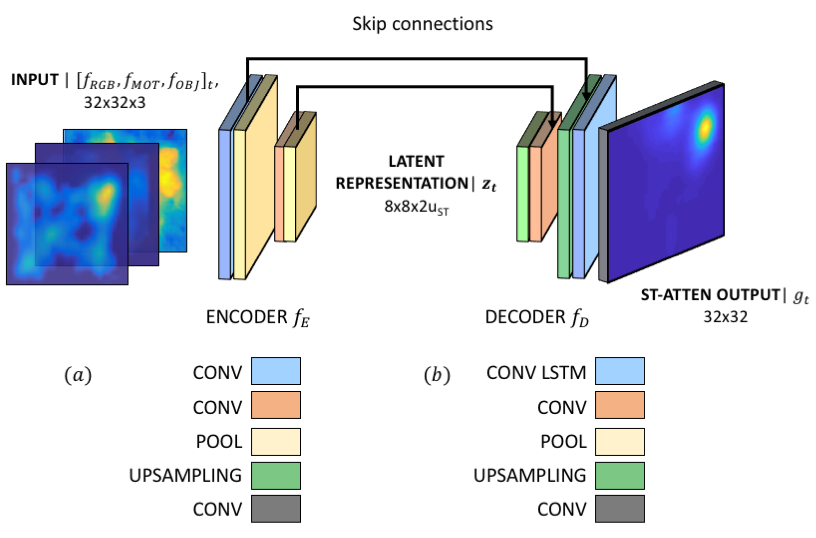}
\caption[Diagram of the \acs{ST-ATTEN} for spatio-temporal visual attention estimation, which consists of a \acs{CED} architecture.]{Diagram of the \acs{ST-ATTEN} for spatio-temporal visual attention estimation, which consists of a \acs{CED} architecture. Both encoder and decoder have two dilated \acs{CONV} layers with skip connections. Two different configurations are proposed for this stage of the system: (a) \acs{CONV}-\acs{ST-ATTEN} (b) \acs{CONV}-\acs{LSTM}-\acs{ST-ATTEN}.}
\label{fig:553_statten}
% \vspace{-0.4cm}
\end{figure*}

%\afterpage{%
%\begin{landscape}
%\centering
%\scalebox{0.7}{
\begin{sidewaystable}
\centering
  \caption[Encoder and decoder architectures for the \acs{CONV}-\acs{ST-ATTEN} and \acs{CONV}-\acs{LSTM}-\acs{ST-ATTEN} configurations of the \acf{ST-ATTEN} proposed.]{Encoder and decoder architectures for the \acs{CONV}-\acs{ST-ATTEN} and \acs{CONV} \acs{LSTM}-\acs{ST-ATTEN} configurations of the \acs{ST-ATTEN} proposed.}\label{tab:553_dced}
  \captionsetup[subfloat]{position=top}
   \subfloat[\acs{CONV}-\acs{ST-ATTEN}]{
   \resizebox{\paperwidth}{!}{
     \begin{tabular}{lllll|lllll}
    	\hline
    	%\hline 
    \multicolumn{5}{c|}{\centering Encoder $f_{E}$} &
    \multicolumn{5}{c}{\centering Decoder $f_{D}$} \\ 
    	\hline
    	\multicolumn{5}{c|}{\centering Input $[f_{RGB},f_{MOT},f_{OBJ}], 32\times32\times3$} &
    \multicolumn{5}{c}{\centering Output $g=f_{D}({\bf z},\theta_{D})$, $32\times32$, \acs{KL} loss} \\
    	\hline
    	\acs{CONV} E1 & ($k=3\times3,u_{ST},s=1,d=2$) & {MAX \acs{POOL} 1} & ($k=2\times2,s=2$), & {\acs{ELU}} & 
    	\acs{CONV} D1 & ($k=3\times3,u_{ST},s=1,d=2$) & {UPSAMPLING 1} & & {\acs{ELU}}\\
    	\hline 
    	\acs{CONV} E2 & ($k=3\times3,2u_{ST},s=1,d=2$) & {MAX \acs{POOL} 2} & ($k=2\times2,s=2$), & {\acs{ELU}} &
    	\acs{CONV} D2 & ($k=3\times3,2u_{ST},s=1,d=2$) & {UPSAMPLING 2} & & {\acs{ELU}}\\
    	\hline
    	\multicolumn{10}{c}{\centering Latent representation ${\bf z}=f_{E}([f_{RGB},f_{MOT},f_{OBJ}],\theta_{E})$, $8\times8\times2u_{ST}$} \\
    	\hline
    \end{tabular}}}
	
    \subfloat[\acs{CONV}-\acs{LSTM}-\acs{ST-ATTEN}]{
    \resizebox{\paperwidth}{!}{
     \begin{tabular}{lllll|lllll}
    	\hline
    	%\hline 
    \multicolumn{5}{c|}{\centering Encoder $f_{E}$} &
    \multicolumn{5}{c}{\centering Decoder $f_{D}$} \\ 
    	\hline
    	\multicolumn{5}{c|}{\centering Input $[f_{RGB},f_{MOT},f_{OBJ}], 32\times32\times3$} &
    \multicolumn{5}{c}{\centering Output $g=f_{D}({\bf z},\theta_{D})$, $32\times32$, \acs{KL} loss} \\
    	\hline
    	\acs{CONV}-\acs{LSTM} E1 & ($k=3\times3,u_{ST},d=2$) & {MAX \acs{POOL} 1} & ($k=2\times2,s=2$) & {\acs{ELU}} &
    	\acs{CONV}-\acs{LSTM} D1 & ($k=3\times3,u_{ST},d=2$) & {UPSAMPLING 1} & & {\acs{ELU}}\\
    	\hline 
    	\acs{CONV} E1 & ($k=3\times3,2u_{ST},d=2$) & {MAX \acs{POOL} 2} & ($k=2\times2,s=2$) & {\acs{ELU}} &
    	\acs{CONV} D1 & ($k=3\times3,2u_{ST},d=2$) & {UPSAMPLING 2} & & {\acs{ELU}}\\
    	\hline
    	\multicolumn{10}{c}{\centering Latent representation ${\bf z}=f_{E}([f_{RGB},f_{MOT},f_{OBJ}],\theta_{E})$, $8\times8\times2u_{ST}$} \\
    	\hline
    \end{tabular}}}
    \captionsetup[subfloat]{position=bottom}
\end{sidewaystable}%}
%\end{landscape}
%\clearpage
%} 

The first stage of our \acs{ST-T-ATTEN} receives as input, for each frame $v_{t}$ in a video $v$, a set of $32\times32$ feature maps for visual attention guidance, such as the ones introduced in Section \ref{sec:cnns_based_features6} ($[f_{RGB},f_{MOT},f_{OBJ}]_{v_{t}}$), and estimates spatio-temporal visual attention by means of a \acs{CED} architecture, which we have called \acf{ST-ATTEN}. 

Figure \ref{fig:553_statten}(a) illustrates the proposed \acs{ST-ATTEN}. As explained in section \ref{sec:autoencoders}, \acsp{CED} can be decomposed into two networks: an encoder and a decoder network. In our system, the first network encodes input feature maps into a latent representation ${\bf z}=f_{E}([f_{RGB},f_{MOT},f_{OBJ}]_{v_{t}},\theta_{E})$, while the second symmetric network transforms this representation into a spatio-temporal visual attention map $\hat{g}_{t}=f_{D}({\bf z}_{t},\theta_{D})$. Due to the low dimensionality of the input features, we decided to make use of dilated convolutions in an attempt to keep the number of parameters limited. Therefore, the encoder network consists of two $k=3\times3$ dilated \acs{CONV} layers with $d=2$, and $u_{ST}$ and $2u_{ST}$ filters, respectively. These layers correspond to two dilated \acs{CONV} layers in the decoder network with the same number of units. After each dilated \acs{CONV} layer in the encoder network, MAX \acs{POOL} with a $2\times2$ window and $s=2$ is performed. This operation sub-samples the \acs{CONV} layer output by a factor of 2, which allows to generate representations more robust to spatial translations. In the decoder network, these layers are replaced by upsampling operations. \acs{ELU} activations \cite{ELU} are introduced after each dilated \acs{CONV} layer. \acsp{ELU} are similar to typically used \acsp{ReLU}, but provide more robustness to noise activations with mean close to zero. Moreover, we introduce skip connections between corresponding dilated \acs{CONV} layers in the encoder and decoder, in an attempt to preserve the spatial resolution of the down-sampled input feature maps. After the decoder network, a final $3\times3$ \acs{CONV} layer with linear activation generates the output visual attention map $\hat{g}_{t}$, which has the same dimensions than the input features ($32\times32$).

We propose two configurations of \acs{ST-ATTEN}: \acs{CONV}-\acs{ST-ATTEN} (see Figure \ref{fig:553_statten}(a)), which has been described in the previous paragraph, and \acs{CONV}-\acs{LSTM}-\acs{ST-ATTEN} (see Figure \ref{fig:553_statten}(b)). The architectures of both configurations are further detailed in Table \ref{tab:553_dced}. The main difference between these two networks relies on the outer layers of the encoder and the decoder, which are \acs{CONV} in the first approach and \acs{CONV}-\acs{LSTM} in the second one. Although our system already receives dynamic spatio-temporal information from the input optical flow-based feature map $f_{MOT}$, we include \acs{CONV}-\acsp{LSTM} in an attempt to model viewers dynamic behavior during the training phase, taking advantage of the spatio-temporal information provided by fixation sequences. This information might be helpful to improve the accuracy of the predicted \acp{VAM} in complex scenarios where there are more than one object in motion, so that it is necessary to consider previous conspicuous locations for a better attention guidance.  

\begin{figure*}
\centering
	\includegraphics[trim=0cm 0cm 0cm 0cm, width=1\textwidth]{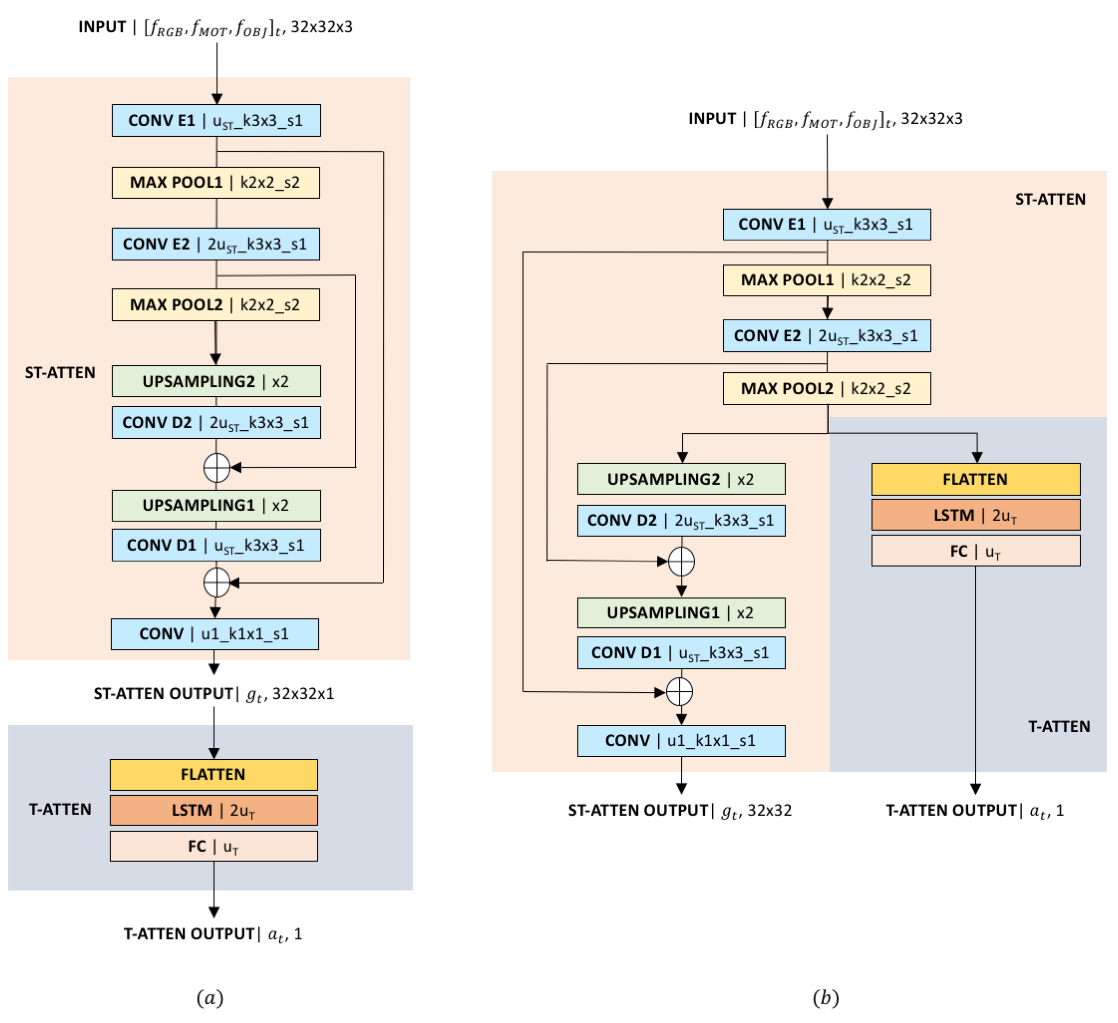}
\caption[Architecture diagrams of the \acs{ST-T-ATTEN} configurations proposed.]{Architecture diagrams of the \acs{ST-T-ATTEN} configurations proposed. The network receives as input, at each timestep $t$, a set of three feature maps $[f_{RGB},f_{MOT},f_{OBJ}]_{t}$, and generates two outputs: a spatio-temporal \acs{VAM} $\hat{g}_{t}$ and a temporal attention response $\hat{a}_{t}$. (a) In the first design, the input to the \acs{T-ATTEN} is the latent representation ${\bf z}_{t}$ computed by the \acs{ST-ATTEN} encoder. (b) In the second configuration, the input to the \acs{T-ATTEN} is the \acs{VAM} $\hat{g}_{t}$ estimated by the \acs{ST-ATTEN} decoder. Layers are defined by their number of units $u$, kernel size $k$ and stride $s$ with which the filters are slided.}
\label{fig:555_sttatten}
% \vspace{-0.4cm}
\end{figure*}

\subsection{Temporal Attention Network}\label{sec:tva_modeling}
The second stage of the system, denoted as \acf{T-ATTEN}, connects with the \acs{ST-ATTEN} explained above and estimates attention in the temporal domain. We expect to take advantage of the effective sequential representations provided by a \acs{LSTM}-based architecture in this stage. Hence, we propose an architecture composed by a layer with $u_{T}$ \acs{LSTM} units, which allows the system to learn long-term dependencies between temporal attention responses $a_{t}$ associated to frames $v_{t}$ in the same video $v$, avoiding vanishing or exploding gradients, as explained in Section \ref{sec:recurrent_neural_networks}. This layer is followed by a \acs{FC} layer, with $u_{T}/2$ units. Finally, a simple \acs{FC} layer produces the temporal attention variable $\hat{a}_{t}$.

The combination of \acs{ST-ATTEN} for spatio-temporal visual attention estimation and \acs{T-ATTEN} for modeling attention in the temporal domain gives rise to our \acf{ST-T-ATTEN}. We propose two different configurations for the architecture, as shown in Figure \ref{fig:555_sttatten}. Both approaches differ in the connection between \acs{ST-ATTEN} and \acs{T-ATTEN} modules. In the first design, the \acs{T-ATTEN} module is fed at each timestep $t$ with the estimated $32\times32$ $\hat{g}_{t}$ at the output of the \acs{ST-ATTEN} decoder. In contrast, in the second configuration, the $8\times8\times2u_{ST}$ latent representation ${\bf z}_{t}$ extracted by the encoder network is used as input to the temporal network. For each frame $v_{t}$ of a video $v$, both architectures generate at its output a linear temporal attention response $\hat{a}_{t}$.

\subsection{Implementation details}\label{sec:implementation_details6}
\subsection*{Data preprocessing}
First, we normalize feature maps at the input of the \acs{ST-T-ATTEN} by subtracting the feature mean and dividing by three times the feature standard deviation (which covers the $\sim 99.7\%$ of the data samples). Mean and standard deviation are computed over the training set. Then, we also clip feature maps values to the range $[-1,1]$. Furthermore, we normalize \acsp{VAM} at the output of the \acs{ST-ATTEN} module to sum to 1.

\subsection*{Multitask loss function}
In order to train each of the stages of the \acs{ST-T-ATTEN}, we have considered two different loss functions.

On the one hand, in order to train the \acs{ST-ATTEN} stage, we make use of the \acf{KL}, which is a distribution-based metric, frequently used as a loss function to train \acsp{CNN} for visual attention prediction due to its proven efficiency \cite{huang2015salicon}. Given a frame $v_{t}$ with $N_{t}$ spatial locations, its corresponding fixation map $g_{t}$ and a predicted visual attention map $\hat{g}_{t}$, it is defined as:

\begin{gather}\label{eq:KL}
	KL(g_{t},\hat{g}_{t}) = \sum_{n=1}^{N_{t}} g_{tn} \ log \left(\epsilon + \frac{g_{tn}}{\epsilon + \hat{g}_{tn}} \right).
\end{gather}

\noindent For each spatial location $n$, $g_{tn}$ constitutes its associated fixation-based \acs{GT}, resulting from convolving $g_{t}$ by a Gaussian filter with standard deviation equal to one degree of visual angle, in order to obtain a continuous distribution. In addition, $\hat{g}_{tn}$ represents the visual attention predicted for that location. A lower score value indicates a better approximation of the \acs{VAM} $\hat{g}_{t}$ to the fixation map $g_{t}$.

On the other hand, the \acs{T-ATTEN} stage is trained by using \acs{MSE} as loss function (see section \ref{sec:neural_networks}, Eq. \ref{eq:mse_loss}). Given a video frame $v_{t}$, the \acs{MSE} between its associated \acs{GT} temporal attention response $a_{t}$ and the attention response estimated by our \acs{T-ATTEN} $\hat{a}_{t}$ can be written as follows:

\begin{gather}\label{eq:tatten_loss}
MSE(a_{t}, \hat{a}_{t}) = (\hat{a}_{t} - a_{t})^2
\end{gather}

Finally, for each frame $v_{t}$, the multitask loss function for the overall system is expressed as follows:

\begin{gather}\label{eq:sttatten_loss}
	L_{ST-T-ATTEN}(g_{t},\hat{g}_{t},a_{t},\hat{a}_{t}) = KL(g_{t},\hat{g}_{t}) + \alpha MSE(a_{t}, \hat{a}_{t}),
\end{gather}

\noindent where $\alpha$ is a scalar that balances the contribution of the two loss functions, and has been empirically determined, as described in the next chapter. 

% multi-task learning \cite{pan2010survey}

%*****************************************
%*****************************************
%*****************************************
%*****************************************
%*****************************************

%************************************************
\chapter{Experiments on temporal visual attention estimation in a video surveillance scenario}\label{ch:surveillance_experiments} 
\chaptermark{Temporal visual attention for video surveillance}
%************************************************

\section{Introduction}\label{sec:introduction_7}
The experiments presented in this chapter have as main objective to assess various configurations for training end-to-end the \acf{ST-T-ATTEN} architecture proposed in Chapter \ref{ch:anomaly_detection}. As explained in that chapter, our system ultimately models attention in the temporal domain by aligning spatio-temporal visual attention maps estimated from video frames to frame-level fixation-based temporal attention responses.

To this end, we evaluate our system in the video surveillance scenario defined by the BOSS \cite{BOSS} database, which contains video sequences recorded in a railway transport context, showing different types of suspicious or anomalous events.

%Experiments show how training end-to-end our approach allows to estimate either spatio-temporal visual attention or attention in the temporal domain at the same time. Although we are still far from accurately estimating visual attention, we 

\section*{Chapter overview}
First, the experimental design is introduced in Section \ref{sec:experimental_design7}. Secondly, we determine the optimal architecture for the first stage of the system proposed, \acs{ST-ATTEN}, in Section \ref{sec:st_atten_validation}. Then, in Section \ref{sec:tatten_results}, the \acs{ST-ATTEN} optimal configuration is used to train end-to-end the complete \acs{ST-T-ATTEN}, and to provide results on attention estimation in the temporal domain, which allows to discuss the model strenghts and limitations. In Section \ref{sec:sttatten_anomaly_detection}, we motivate the use of our system for guiding anomaly detection in video surveillance applications. Finally, Section \ref{sec:tva_conclusions} summarizes our conclusions and motivates future work.

\section{Experimental design}\label{sec:experimental_design7}
This section explains the experimental design for the analysis of the \acs{ST-T-ATTEN}. First, the databases used to train and evaluate its different stages are introduced. Then, we describe the experimental setup and the evaluation metrics considered to assess the performance of the proposed architectures. Finally, we provide the implementation details related both to the \acs{ST-T-ATTEN} and the \acsp{CNN} previously presented in Section \ref{sec:cnns_features6}, in charge of extracting high-level visual feature maps that become the input to our model.
 
\subsection{Databases}\label{sec:databases7}
\subsubsection*{SALICON and DIEM}\label{sec:salicon}
SALICON \cite{jiang2015salicon} and DIEM \cite{mital2011clustering} databases have been considered in our experiments to train the modified ResNet-50 \cite{he2016deep} models for visual feature extraction introduced in Section \ref{sec:cnns_features6}.

On the one hand, in order to obtain RGB-based spatial feature maps, we have made use of SALICON image database to fine-tune a ResNet-50 model pre-trained on ImageNet database \cite{krizhevsky2012imagenet}. SALICON database contains a set of 10,000 context-generic training images annotated by 60 ``free-viewing´´ observers with a mouse tracking process. Although the consistency between participants' fixations is lower when compared to the information provided by an eye tracker device, mouse movements constitute a helpful approach to eye tracking in still images, which allow to efficiently annotate very large databases.  

On the other hand, DIEM video database, already introduced and analyzed in Chapter \ref{ch:atom_experiments}, has been used to train from scratch the optical flow-based network for motion feature maps extraction.

\subsubsection*{BOSS}\label{sec:boss}
Within the framework of the BOSS project \cite{BOSS}, a database with 15 video sequences recorded in RENFE suburban trains in Madrid was released with the aim of developing an efficient transmission system for video-surveillance in a railway transport context. Videos contain events such as a cell phone theft, a fight between passengers, a disease in public and several women harassment. Moreover, two additional sequences with no incidents are included. For each event, three camera views are provided.

In order to evaluate the different architectures for attention estimation in the temporal domain that were proposed in the previous chapter, we have selected the three camera views of 10 sequences from this database, and annotated them with eye fixations. In total, 30 videos (over $84,000$ video frames, $56$ minutes total, $720 \times 576$) have been used. For each video, eye traces from 5 observers have been recorded by using a 250 Hz SMI RED250mobile Eye Tracker system \cite{SMI}. The complete list of videos annotated for our experiments can be found in Appendix \ref{ch:databases}.

\subsection{Experimental setup}\label{sec:experimental_setup7}
The experiments presented in this chapter have the objective to assess which of the proposed \acs{ST-T-ATTEN} architectures models better the temporal dimension of attention. For that purpose, we will ultimately evaluate our system when estimating visual attention in the temporal domain.

We conduct our experiments by splitting the 30 videos selected from BOSS \cite{BOSS} database into three folds, each one containing 10 different sequences. In order to avoid over-fitting, the three camera views from the same sequence are grouped together in the same fold. In the following paragraphs, we will describe the process we have followed to train and optimize the different modules involved in our \acs{ST-T-ATTEN}.

First, we have determined the optimal configuration for the first stage of our system: \acs{ST-ATTEN}. To achieve this, we have evaluated the two configurations proposed for this module in Section \ref{sec:stva_autoencoder}. We selected the one that provided the best performance in terms of the evaluation metrics mentioned in the next section. Following a 3-fold cross-validation procedure, we have estimated spatio-temporal visual attention maps for each video in a fold, using the remaining two folds for training the network. 

Then, we have assessed if our \acs{ST-T-ATTEN} is able to obtain accurate estimations of temporal attention, with the ultimate goal of discussing its utility as a filtering mechanism for subsequent anomaly detection systems in video surveillance scenarios. To do this, we have trained the complete \acs{ST-T-ATTEN} architecture end-to-end. To do so, we initialized the \acs{ST-ATTEN} module using the weights learned in the previous step. As mentioned before, latent representations and \acsp{VAM} extracted by \acs{ST-ATTEN} constitute the input for the next stage of the system: \acs{T-ATTEN}. The evaluation of the complete network has been done by following the same procedure described above: we estimate either spatio-temporal visual attention or attention in the temporal domain in each fold, using the remaining two folds for training the complete architecture.

%perform a 3-fold cross validation. Each time, we use two video sets for training and validation, and evaluate in the remaining unseen one. We perform early stopping, in order to find out the optimum number of iterations for the learning phase. Then, we train the network over this number of iterations, by concatenating training and validation sets, and evaluate it in the unseen test set.

\subsection{Evaluation metrics}\label{sec:evaluation_metrics7}
\subsubsection*{Spatio-temporal visual attention estimation}
In order to evaluate the spatio-temporal \acsp{VAM} predicted by the different \acs{ST-ATTEN} architectures in the first stage, we make use of the \acs{sAUC} and \acs{sNSS} metrics described in Section \ref{sec:atom_evaluation_metrics}. Moreover, we include the \acs{KL} metric defined in Eq. \ref{eq:KL}, which is also the loss function used to train the \acs{ST-ATTEN}. Furthermore, for comparison purposes, we consider the three baseline models described in Section \ref{sec:atom_evaluation_metrics}: CHANCE, CENTER and H50. 

% sAUC, sNSS, KL? see Chapter 5

\subsubsection*{Visual attention estimation in the temporal domain} 
In addition, since our temporal attention response $a_{t}$ is a real number in the range $[0,1]$, we now introduce a rank correlation coefficient for the assessment of the temporal attention responses estimated by the \acs{T-ATTEN} proposed in Chapter \ref{ch:anomaly_detection}. The evaluation metric chosen is the well-known \acf{PCC} \cite{Benesty2009}, which here measures the linear relationship between the fixation-based temporal \acs{GT} $a_{1:T}$ and the estimated temporal attention response $\hat{a}_{1:T}$ of a given set of $T$ video frames. It can be written as follows:

\begin{gather}
PCC = \frac{\sum_{t=1}^{T} (a_{t}-\mu_{a_{1:T}})(\hat{a}_{t}-\mu_{\hat{a}_{1:T}})}{\sqrt{\sum_{t=1}^{T} (a_{t}-\mu_{a_{1:T}})^2} \sqrt{\sum_{t=1}^{T} (\hat{a}_{t}-\mu_{\hat{a}_{1:T}})^2}},
\end{gather}

\noindent where $\mu_{a_{1:T}}$ and $\mu_{\hat{a}_{1:T}}$ represent the mean values of the \acs{GT} $a_{t}$ and the estimated attention $\hat{a}_{t}$ for the considered set of frames, respectively. The coefficient lies in the range $[-1,1]$, meaning these extreme values an exact negative or positive correlation, while $PCC=0$ implies no correlation at all.

\subsection{Training and implementation details}\label{sec:implementation_details7}
Here we describe the implementation details of the complete system. To begin with, it should be mentioned that we have made use of the Keras framework \cite{chollet2015keras} with Tensorflow backend \cite{tensorflow2015-whitepaper} to build all the networks deployed. Besides, we train our models using a 12GB NVIDIA GeForce GTX TITAN Xp \acs{GPU} on a system with an Intel Core i7-6700K (4.00GHz) \acs{CPU} and 32GB of \acs{RAM}.
 
\subsection*{Feature extraction networks for visual attention guidance}
First, in order to train the ResNet-50 \cite{he2016deep} models to compute RGB-based and optical flow-based feature maps, we have considered SALICON \cite{jiang2015salicon} image and DIEM \cite{mital2011clustering} video databases, respectively. Besides, we have chosen the \acs{KL} loss function introduced in Eq. \ref{eq:KL} as loss function. In order to minimize \acs{KL}, we use \acs{SGD}, setting the learning rate to $10^{-4}$ and using a mini-batch of 10 samples.

On the other hand, we have directly used the model and weights provided by the authors of the selected objectness-based network \cite{DeepSaliencyObject}, which has been trained on MSRA-B \cite{liu2011learning} salient object database. %In contrast to our approaches, this model is built on the Caffe framework \cite{jia2014caffe}.

\subsection*{Spatio-Temporal to Temporal Visual Attention Network}
Secondly, \acs{ST-T-ATTEN} is trained in BOSS \cite{BOSS} video surveillance database to model attention in the temporal domain.

Regarding the model weights initialization, we have drawn on Glorot uniform initializer \cite{glorot2010understanding}, also known as Xavier uniform initializer, which generates random samples from a uniform distribution. 

In addition, some preliminary experiments have shown the utility of Dropout \cite{srivastava2014dropout} regularization in the \acs{ST-ATTEN} module. Therefore, we have decided to introduce dropout layers in this stage of the system, which randomly drop, at each iteration, half of the filters ($p=0.5$) of each \acs{CONV} layer, both at the encoder and the decoder. 

%\subsection*{Temporal Visual Attention Network}
With respect to the loss functions associated to each stage of the system, we have considered the multitask loss function introduced in Section \ref{sec:implementation_details6}. This loss is a linear combination of the two atomic losses: a) the \acs{KL} loss presented in Eq. \ref{eq:KL}, used to train the \acs{ST-ATTEN} stage of the system; and b) the \acs{MSE} loss defined in Eq. \ref{eq:tatten_loss}, used to train the \acs{T-ATTEN} module of the system. As defined in Eq. \ref{eq:sttatten_loss}, parameter $\alpha$ balances the contribution of each term (\acs{KL}-loss, \acs{MSE}-loss) in the multitask loss. This parameter has been empirically set to $\alpha=100$. In order to minimize the \acs{ST-T-ATTEN} loss function $L_{ST-T-ATTEN}$, we consider \acs{SGD} with a learning rate of $10^{-4}$. The network is trained over 10K iterations, using a mini-batch of 256 samples. 

%in the \acs{ST-ATTEN} stage, in order to determine its optimal configuration. Afterwards, at the time we train the whole \acs{ST-T-ATTEN} to estimate attention in the temporal domain, early stopping is performed to determine an optimum number of iterations, using one of the three video sets defined in BOSS \cite{BOSS} database for validation. In both stages, we use a mini-batch of 256 samples.

\begin{figure*}
\centering
	\includegraphics[trim=0cm 0cm 0cm 0cm, width=1\textwidth]{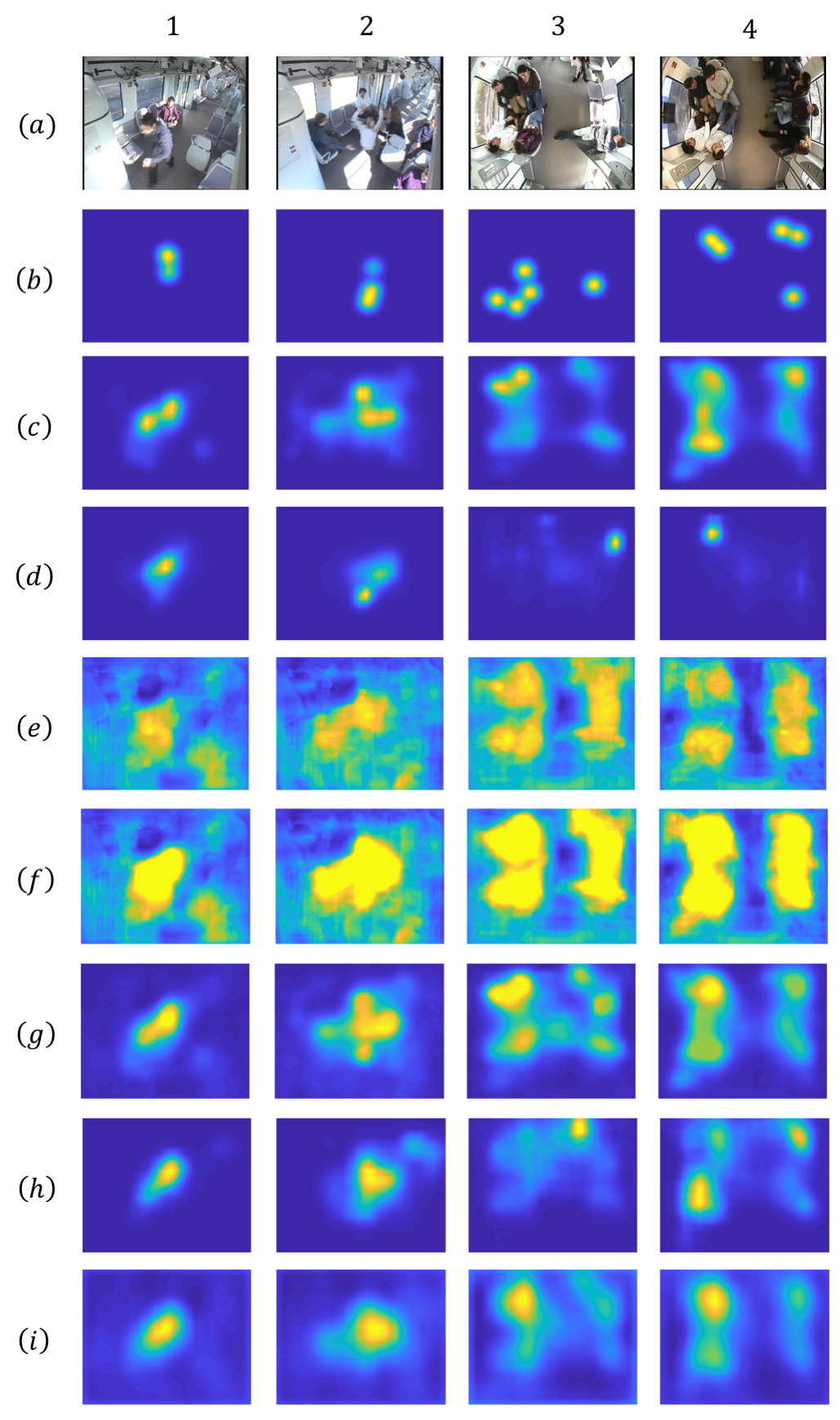}
\caption[Visual attention maps obtained by \acs{ST-ATTEN} for some example frames taken from BOSS \cite{BOSS} database.]{Visual attention maps obtained by \acs{ST-ATTEN} for some example frames taken from BOSS \cite{BOSS} database. (a) Original frames. (b) \acs{GT} fixations map. (c) RGB-based feature map. (d) Motion feature map. (e) Objectness feature map. (f) \acs{VAM} obtained from averaging the three feature maps. (g) \acs{VAM} obtained by learning a linear combination over the three feature maps. (h) \acs{CONV}-\acs{ST-ATTEN} \acs{VAM}. (i) \acs{CONV}-\acs{LSTM}-\acs{ST-ATTEN} \acs{VAM}.}
\label{fig:631_statten_examples}
% \vspace{-0.4cm}
\end{figure*}

\section{Results on spatio-temporal visual attention estimation with ST-ATTEN}\label{sec:st_atten_validation}
In this section, we aim to assess the two architectures for the \acs{ST-T-ATTEN} module proposed in Section \ref{sec:stva_autoencoder}. We evaluate both qualitatively and quantitatively the two configurations: \acs{CONV}-\acs{ST-ATTEN} and \acs{CONV}-\acs{LSTM}-\acs{ST-ATTEN}. To that end, we have first validated the parameter $u_{ST}$ associated with both networks. This parameter determines the dimension of the latent representation extracted for visual attention ($8 \times 8 \times 2u_{ST}$). After conducting several experiments with both undercomplete and overcomplete \acsp{EDN}, we set this parameter to $u_{ST}=64$, which corresponds to an overcomplete \acs{CED}, in which the representation ${\bf z}$ has a dimension greater than the network input (the three feature maps).

\begin{table}[!t]
\caption{Results obtained on the BOSS \cite{BOSS} database by the proposed \acs{ST-ATTEN} and other methods for comparison when estimating spatio-temporal visual attention.}\label{tab:statten_results}
\begin{center}
%\vspace{-0.25cm}
    \resizebox{1\textwidth}{!}{
    \begin{tabular}{ll|ccc}
    \hline 
    \multicolumn{2}{c|}{\multirow{2}{*}{Model}} &
    \multicolumn{1}{p{3cm}}{\centering $sAUC$} & \multicolumn{1}{p{3cm}}{\centering $sNSS$} & \multicolumn{1}{p{3cm}}{\centering $KL$} \\ 
    & & \multicolumn{1}{p{3cm}}{\centering \scriptsize $mean \ (C.I.)$} & \multicolumn{1}{p{3cm}}{\centering \scriptsize $mean \ (C.I.)$} & \multicolumn{1}{p{3cm}}{\centering \scriptsize $mean \ (C.I.)$} \\ \hline
    \multicolumn{2}{l|}{RGB} & $0.714 \ (0.710, 0.718)$ & $0.374 \ (0.367, 0.381)$ &  $ 1.938 \ (1.922, 1.954)$ \\
    \multicolumn{2}{l|}{MOT} & $0.588 \ (0.583, 0.592)$ & $0.127 \ (0.119, 0.134)$ &  $ 2.755 \ (2.730, 2.781)$ \\ 
    \multicolumn{2}{l|}{OBJ} & $0.667 \ (0.663, 0.671)$ & $0.303 \ (0.296, 0.310)$ &  $2.257 \ (2.249, 2.264)$ \\ \hline
    \multicolumn{2}{l|}{AVERAGE} & $0.702 \ (0.698, 0.710)$ & $0.362 \ (0.354, 0.369)$ &  $2.039 \ (2.031, 2.048)$ \\
    \multicolumn{2}{l|}{LIN. COMB.} & $0.702 \ (0.698, 0.706)$ & $0.363 \ (0.356, 0.370)$ &  $1.894 \ (1.883, 1.905)$ \\ \hline
    %{\acs{CONV}} & $u_{ST}=8$ & $0.727 \ (0.723, 0.732)$ & $0.390 \ (0.382, 0.398)$ & $1.620 \ (1.605, 1.635)$ \\
    %{\acs{CONV}} & $u_{ST}=24$ & $0.738 \ (0.734, 0.742)$ & $0.406 \ (0.398, 0.414)$ & $1.588 \ (1.572, 1.604)$ \\ 
    %{\centering \acs{ST-ATTEN}} & \multicolumn{1}{l|}{} & \multicolumn{3}{l}{} \\
    {\acs{CONV}} & & ${\bf 0.750 \ (0.746, 0.755)}$ &	${\bf 0.424 \ (0.416, 0.431)}$ &	${\bf 1.563 \ (1.547, 1.578)}$ \\  
    %{\acs{CONV}} & $u_{ST}=256$ & $? \ (?, ?)$ & $? \ (?, ?)$ &  $ ? \ (?, ?)$ \\ \hline
    %{\acs{CONV} \acs{LSTM}} & $u_{ST}=8$ & $? \ (?, ?)$ & $? \ (?, ?)$ &  $ ? \ (?, ?)$ \\
    %{\acs{CONV} \acs{LSTM}} & $u_{ST}=24$ & $? \ (?, ?)$ & $? \ (?, ?)$ &  $ ? \ (?, ?)$ \\
    {\acs{CONV}-\acs{LSTM}} & & $0.748 \ (0.744, 0.751)$ & $0.410 \ (0.403, 0.416)$ &  $1.610 \ (1.601, 1.627)$ \\ \hline
    %{\centering \acs{ST-T-ATTEN}} & \multicolumn{1}{l|}{} & \multicolumn{3}{l}{} \\
    %{\acs{CONV}} &  & $?$ &	$?$ &	$?$ \\ \hline
    %{\acs{CONV} \acs{LSTM}} & $u_{ST}=256$ & $? \ (?, ?)$ & $? \ (?, ?)$ &  $ ? \ (?, ?)$ \\  \hline
    \multicolumn{2}{l|}{H50} & $0.826 \ (0.826, 0.826)$ & $0.692 \ (0.692, 0.693)$ &  $2.137 \ (2.135, 2.139)$ \\
    \multicolumn{2}{l|}{CHANCE} & $0.500 \ (0.500, 0.500)$ & $-0.000 \ (-0.001, 0.000)$ &  $ 4.423 \ (4.335, 4.338)$ \\ 
    \multicolumn{2}{l|}{CENTER} & $0.500 \ (0.499, 0.502)$ & $0.014 \ (0.012, 0.017)$ &  $4.337 \ (4.335, 4.338)$ \\ \hline
    \end{tabular}
    }
    \end{center}
\end{table}

Then, we perform a quantitative evaluation of the two configurations proposed in terms of the \acs{sAUC}, \acs{sNSS} and \acs{KL} metrics. Table \ref{tab:statten_results} summarizes the results obtained by the proposed \acs{ST-ATTEN} on the BOSS \cite{BOSS} database. For the sake of comparison, we include the results achieved by the three considered feature maps: RGB-based (RGB), motion (MOT) and objectness (OBJ). Moreover, we provide the results offered by two simple fusion approaches: a map computed by averaging the three features (AVERAGE) and a map obtained by learning a linear combination over them (LIN. COMB). Finally, the three reference models introduced in Section \ref{sec:atom_evaluation_metrics} (H50, CHANCE, CENTER) are also included for comparison. As can be verified from CENTER baseline, both \acs{sAUC} and \acs{sNSS} metrics are not affected by center bias.

According to the results in the table, our \acs{CED} configurations for spatio-temporal visual attention estimation successfully learn non-linear fusion schemes for the three feature maps considered and notably outperform the two baseline fusion approaches (AVERAGE, LIN. COMB.). Furthermore, it can be concluded that the first configuration proposed, which only makes use of \acs{CONV} layers, is the one that offers a slightly better performance. However, whereas the performance achieved by \acs{ST-ATTEN} is close to H50 score in terms of \acs{sAUC}, we are still far from reaching highly accurate estimations of spatio-temporal visual attention according to \acs{sNSS} metric. This is probably due to some of the aspects discussed by Bylinskii et al. in \cite{DBLP:conf/eccv/BylinskiiRBOTD16}, mainly related to the modeling of high-level concepts, such as objects of action or gaze, which are not explicitly modeled by our approach. It is also worth pointing out that H50 model performs worse than our approaches in terms of distribution-based \acs{KL} metric. From our point of view, this is not a surprise, as predicting visual attention from fixations of $50\%$ of subjects available is hard when fixation dispersion across observers is high, due to the absence of a clear and conspicuous event on the scene. This happens in the majority of frames of the database, which do not contain anomalous events.

Finally, we have also assessed qualitatively the \acsp{VAM} obtained by the different methods. Figure \ref{fig:631_statten_examples} shows the output maps of the two configurations proposed for some example frames taken from BOSS \cite{BOSS} database. \acs{GT} eye fixation density maps are also displayed (b). For the sake of comparison, we include the three feature maps used as input to \acs{ST-ATTEN} (c,d,e). Moreover, we provide the output \acsp{VAM} obtained by the two baseline fusion approaches: AVERAGE (f) and LIN. COMB. (g). As can be seen, our architectures (h,i) provide the most accurate estimations of visual attention in the shown cases, providing better attention representations than baseline fusion approaches and individual features. In addition, they are able to learn more sophisticated representations of visual attention in complex situations, as the one shown in the fourth example provided, in contrast to maps obtained from a linear combination, which gives a higher weight to the feature that best model visual attention amongst the three maps considered (RGB-based, according to the results in Table \ref{tab:statten_results}). However, it can be appreciated that our model fails in estimating attention in crowded scenes with several people, as in the example gathered on the third column.

If we compare the maps estimated by the two proposed \acs{CED} configurations, it can be noticed that those obtained by a \acs{CONV}-\acs{LSTM} architecture (i) are in general smoother than the ones computed by a fully \acs{CONV} one (h), probably due to the influence of information from previous frames stored in the \acs{LSTM} state cells. This information may help to reduce noise when feature maps associated to the current frame do not constitute good estimations of the visual attention. For instance, in the third case, estimated motion is concentrated on a very small location and, therefore, it is less attracting than passengers on the train. While \acs{CONV}-\acs{ST-ATTEN} map is strongly affected by this motion, \acs{CONV}-\acs{LSTM}-\acs{ST-ATTEN} uses information stored from previous frames to keep the attention on passengers. 

However, given the fact that quantitative results achieved by both configurations are quite similar, and that the use of \acs{CONV}-\acsp{LSTM} is more computationally demanding, from now on we will make use of the \acs{CONV}-\acs{ST-ATTEN} configuration, with $u_{ST}=64$ filters, as input to the \acs{T-ATTEN} stage of the system.  

%OPTION 2- Hence, given than the quantitative results achieved by both configurations are quite similar but, qualitatively, \acs{CONV}-\acs{LSTM}-\acs{ST-ATTEN} offers some extra advantages with respect to \acs{CONV}-\acs{ST-ATTEN}, from now on we will make use of this configuration, with $u_{ST}=64$ filters, as input to the next stage of the system.  

\section{Results on attention estimation in the temporal domain with ST-T-ATTEN}\label{sec:tatten_results}
\begin{table}[!t]
\caption{Results obtained on the BOSS \cite{BOSS} database by the proposed \acs{ST-T-ATTEN} when modeling visual attention in the temporal domain.}\label{tab:tatten_results}
\begin{center}
%\vspace{-0.25cm}
    \resizebox{1\textwidth}{!}{
    \begin{tabular}{l|c|c}\hline
\multirow{2}{*}{\diagbox{Input}{Architecture}}
&\multicolumn{1}{p{4cm}|}{\centering \acs{FC}}&\multicolumn{1}{p{3.5cm}}{\centering \acs{LSTM} + \acs{FC}}\\
&\multicolumn{1}{p{4cm}|}{\centering \acs{PCC}}&\multicolumn{1}{p{3.5cm}}{\centering \acs{PCC}}\\\hline
\acs{VAM} & $0.166$ & $0.321$\\
LR & $0.106$ & ${\bf 0.323}$\\\hline %,0.068$ - %,0.287,0.299,0.274$\\\hline 
\acs{GT} & $0.278$ & $0.467$\\\hline
%\acs{GT} Latent representation & $?$ & $?$\\\hline
\end{tabular}
    }
    \end{center}
\end{table}

Once we selected the optimal configuration for the \acs{ST-ATTEN} module, we have trained end-to-end the whole \acs{ST-T-ATTEN} architecture proposed, as described in Section \ref{sec:experimental_design7}. Results obtained on the BOSS \cite{BOSS} database are summarized here below.

First, it is worth mentioning that training the whole architecture barely changes the performance achieved by the \acs{ST-ATTEN} module; consequently, for the sake of simplicity, we have omitted its analysis in this section, focusing our discussion in the ultimate goal of our system, which is to estimate attention in the temporal domain, by means of the \acs{T-ATTEN} module. The parameter $u_{T}$ of this module has been empirically set to $u_{T}=256$, which determines the number of units of its corresponding \acs{LSTM} ($u_{T}$) and \acs{FC} ($u_{T}/2$) layers. 

As described in Section \ref{sec:tva_modeling}, we have evaluated two configurations of the \acs{T-ATTEN} architecture: 1) one that takes the \acs{VAM} estimated by the \acs{ST-ATTEN} decoder as input (\acs{VAM}), and 2) one that works with the latent representation (LR) extracted by the encoder network. Besides, for comparison purposes, we have also considered a baseline \acs{T-ATTEN} architecture composed of a unique \acs{FC} layer with $u_{T}$ units. Finally, we have also trained two reference \acsp{T-ATTEN} (\acs{GT}), that work with \acs{GT} \acsp{VAM} $g_{t}$ computed using \acs{GT} fixations recorded from subjects. These two approaches provide a theoretical upper bound of the performance of the system, simulating scenarios in which \acsp{VAM} are optimal.

Table \ref{tab:tatten_results} presents the results obtained on the BOSS \cite{BOSS} database by our proposed \acs{ST-T-ATTEN} for visual attention estimation in the temporal domain, as well as by the models considered for comparison. Results are presented in terms of the \acs{PCC} (see Section \ref{sec:evaluation_metrics7}). As can be appreciated, \acs{LSTM}-based \acs{T-ATTEN} architectures notably outperform \acs{FC} ones, which confirms the benefits of using \acs{LSTM} units to model short and long-term temporal relationships between video frames. Regarding the performance of our approaches, both of them provide a similar \acs{PCC} score, in spite of the advantages that latent representations have shown in several multi-task learning paradigms \cite{zhang2017survey}. Therefore, we can conclude that, according to the experiments taken on the BOSS \cite{BOSS} database, the optimal configuration for the \acs{ST-T-ATTEN} proposed consists in a \acs{CONV}-\acs{ST-ATTEN}, and a \acs{T-ATTEN} with \acs{LSTM} and \acs{FC} layers. The latter can be fed either with the \acsp{VAM} computed at the output of the \acs{ST-ATTEN} decoder or the latent representations generated by the encoder network. 

\begin{figure*}[!t]
\centering
	\includegraphics[trim=0cm 0cm 0cm 0cm, width=1\textwidth]{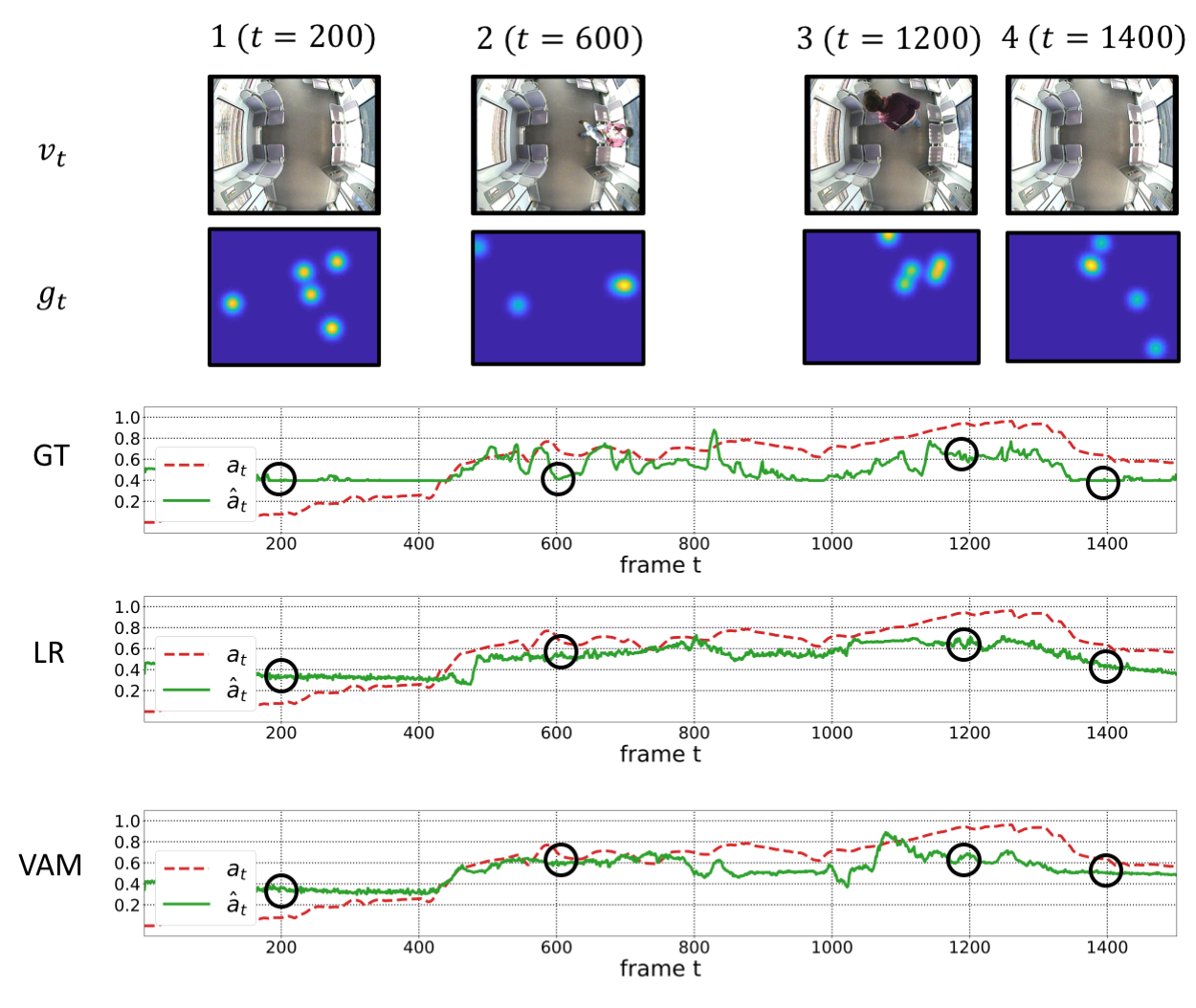}
\caption[Visual attention in the temporal domain $\hat{a}_{t}$ estimated by \acs{ST-T-ATTEN} in a video-surveillance sequence taken from BOSS \cite{BOSS} database.]{Visual attention in the temporal domain $\hat{a}_{t}$ estimated by \acs{ST-T-ATTEN} in a video-surveillance sequence taken from BOSS \cite{BOSS} database. The sequence shows the theft of a mobile phone. 1. Empty train wagon. 2. Woman sits on the train. 3. Somebody stole woman's mobile phone. 4. Woman has gone to report the incident. On the first two rows, some video frames over time $v_{t}$ are shown, together with their associated \acs{GT} fixations map $g_{t}$. Then, the temporal attention response $\hat{a}_{t}$ estimated by each of the three \acs{LSTM}-based \acs{T-ATTEN} evaluated (\acs{GT}, LR and \acs{VAM}) is displayed. For the sake of comparison, the \acs{GT} temporal attention response $a_{t}$ is also plotted.}
\label{fig:64_sttatten_results1}
% \vspace{-0.4cm}
\end{figure*}

\begin{figure*}
\centering
	\includegraphics[trim=0cm 0cm 0cm 0cm, width=0.9\textwidth]{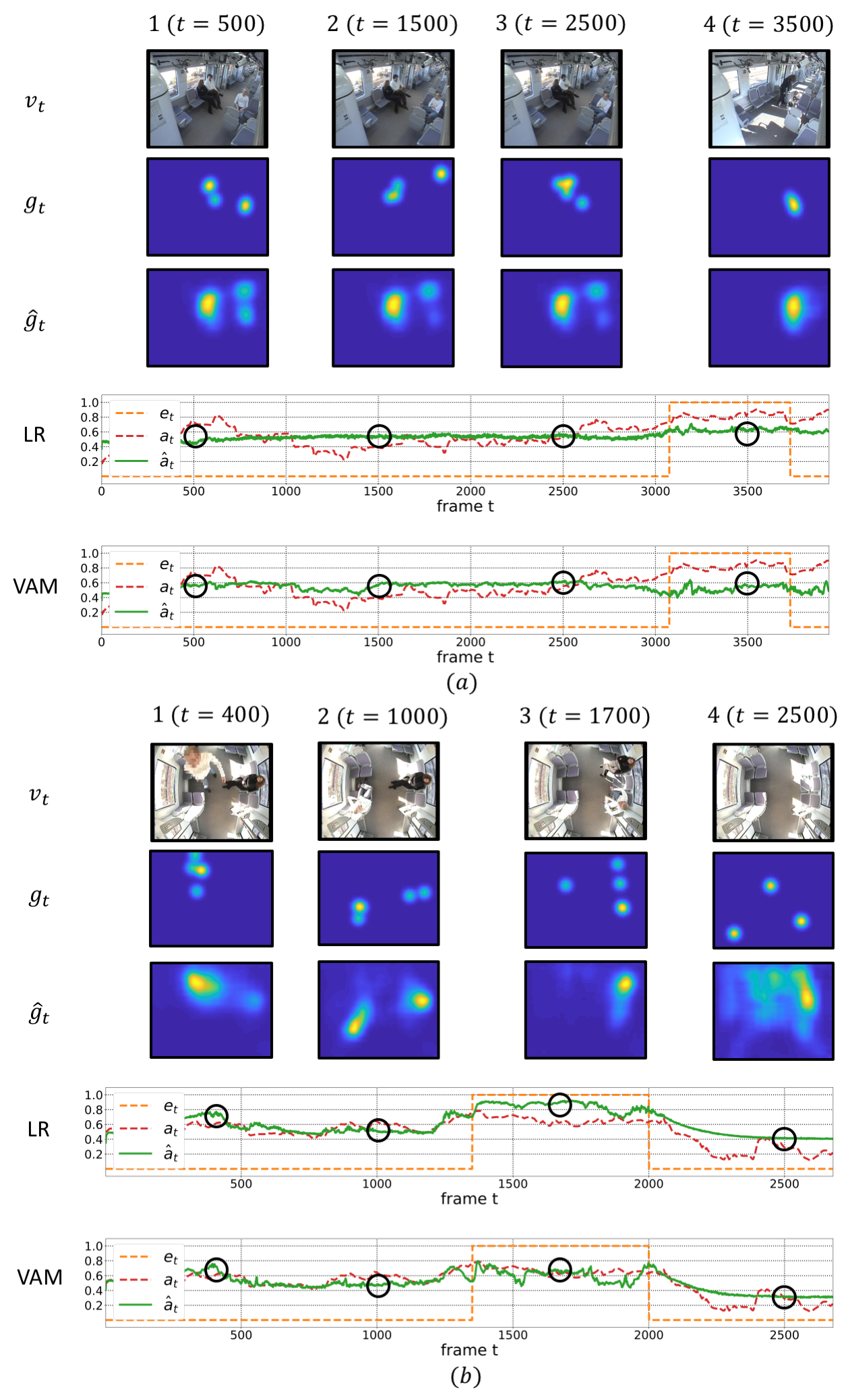}
\caption[Visual attention in the temporal domain $\hat{a}_{t}$ estimated by \acs{ST-T-ATTEN} in two video-surveillance sequences taken from BOSS \cite{BOSS} database.]{Visual attention in the temporal domain $\hat{a}_{t}$ estimated by \acs{ST-T-ATTEN} in two video-surveillance sequences taken from BOSS \cite{BOSS} database. On the first two rows, some video frames over time $v_{t}$ are shown, together with their associated \acs{GT} fixations map $g_{t}$. Then, the spatio-temporal \acsp{VAM} $\hat{g}_{t}$ estimated by the \acs{ST-ATTEN} module of the system are displayed. Finally, the temporal attention response $\hat{a}_{t}$ estimated by the two \acs{LSTM}-based \acs{T-ATTEN} proposed (LR and \acs{VAM}) is represented. For the sake of comparison, the \acs{GT} temporal attention response $a_{t}$ and the anomaly detection signal $e_{t}$ are also plotted. (a) The sequence shows two men fighting. 1,2. Three passengers are sitting on the train. 3. Two passengers start arguing. 4. Two men fight, and a woman tries to stop them. (b) The sequence shows a woman harassment scene. 1. Man sits on the train. 2. Man starts talking to the woman. 3. Man approaches the woman. 4. Woman and man left the train.}
\label{fig:64_sttatten_results2}
% \vspace{-0.4cm}
\end{figure*}

Although we are still far from achieving an exact correlation with the \acs{GT} temporal attention response $a_{t}$ that we aim to estimate, it should be noted that \acsp{PCC} obtained by our models are not far from those provided by the theoretical bounds. As can be seen in the example sequence in Figure \ref{fig:64_sttatten_results2}, the temporal attention responses $\hat{a}_{t}$ estimated by our \acs{LSTM}-based approach have a lower dynamic range compared to the expected \acs{GT} fixation-based responses $a_{t}$. This may be due to several reasons: 

\begin{itemize}
	\item Features for visual attention guidance considered might not be accurate enough for modeling spatio-temporal visual attention. We should improve them, or even incorporate additional ones, in order to handle crowded and complex scenes, such as the ones shown on the third and fourth columns of Figure \ref{fig:631_statten_examples}. 
	\item Besides, it has been observed that when the scene is static or there are not obvious conspicuous locations on it, the \acs{ST-ATTEN} is not always able to model the pseudo-random nature of eye fixation sequences, which approximately corresponds to \acsp{VAM} that distribute visual attention equally in all spatial locations. Instead, it estimates the same \acs{VAM} for static frames with the same content, which may result in a flat temporal response.
	\item We have trained \acs{ST-T-ATTEN} with a database that contains few and similar anomalous events, which might not be sufficient to demonstrate our second assumption in Section \ref{sec:statten_overview}. To address this issue, we aim to make use of large-scale video surveillance databases such as VIRAT \cite{oh2011large} or UCF-Crime \cite{sultani2018real} for a more complete analysis of the system.
\end{itemize}

Moreover, given that the \acs{T-ATTEN} has not been able to accurately estimate visual attention in the temporal domain even when the \acsp{VAM} are optimal, we come to the following conclusion, which serves to lead further work: our system, which makes use of traditional \acsp{CNN} layers and the widely-used \acs{MSE} loss for regression, does not to compute adequately the function that maps \acs{GT} fixations maps $g_{t}$ with temporal attention responses $a_{t}$. Therefore, future efforts will be made towards designing \acsp{CNN} layers and a loss function tailored to the problem we want to solve with \acs{T-ATTEN} (estimation of attention in the temporal domain), in order to improve our system to efficiently guide anomaly detection in video surveillance scenarios. An alternative to \acs{MSE} that would perhaps be worth testing is the \acf{MPSE}, which is a pairwise ranking loss that measures the differences between all possible pairs of corresponding temporal attention responses estimated $(\hat{a}_{i},\hat{a}_{j})$ and \acs{GT} fixation-based responses $(a_{i},a_{j})$ in each mini-batch.	

\section{Where we are: towards guiding anomaly detection}
\label{sec:sttatten_anomaly_detection}
As it was introduced in Chapter \ref{ch:anomaly_detection}, visual attention in the temporal domain can be understood as an information filtering mechanism which allows to select candidate time segments to contain anomalous events. This constitutes a very interesting application of visual attention, which would substantially decrease the cognitive effort made by \acs{CCTV} operators in video monitoring.

\begin{table}[!t]
\caption{Results obtained by the proposed \acs{ST-T-ATTEN} and other comparison methods considered as filtering mechanisms for guiding anomaly detection in the video surveillance scenario defined by the BOSS \cite{BOSS} database.}\label{tab:sttatten_anomaly_results}
\begin{center}
%\vspace{-0.25cm}
    \resizebox{0.7\textwidth}{!}{
    \begin{tabular}{l|c}\hline
\multicolumn{1}{p{4cm}|}{\centering Model}&\multicolumn{1}{p{3.5cm}}{\centering \acs{AUC}}\\ \hline
$\hat{a}_{t}$ & ${\bf 0.703}$ \\
AGGREGATION MAX & $0.613$\\
AGGREGATION SUM & $0.543$\\\hline
$a_{t}$ & $0.876$\\\hline
%\acs{GT} Latent representation & $?$ & $?$\\\hline
\end{tabular}
    }
    \end{center}
\end{table}

As a first approximation to the use of the \acs{ST-T-ATTEN} proposed for guiding and reducing the computational cost of an anomaly detection task, we can consider the same binary classification problem posed in Section \ref{sec:fixations_gt} for the context defined by the BOSS \cite{BOSS} database, but now measuring the existing correlation between anomalies $e_{t}$ and the estimated temporal attention response $\hat{a}_{t}$ by the best \acs{ST-T-ATTEN} configuration determined in the previous section. Moreover, for the sake of comparison, we provide the results obtained by two simple baseline methods: AGGREGATION SUM and AGGREGATION MAX. They also estimate attention in the temporal domain from \acsp{VAM} extracted by \acs{ST-ATTEN}, but aggregating the spatial dimension using a sum (AGGREGATION SUM) or a maximum (AGGREGATION MAX) operator. Table \ref{tab:sttatten_anomaly_results} summarizes the results obtained by \acs{ST-T-ATTEN} and the methods considered for comparison. After conducting the experiments, we achieve an $AUC = 0.703$, which notably outperforms the two baseline approaches. Besides, despite the $AUC$ obtained by \acs{ST-T-ATTEN} is lower to the one achieved by the \acs{GT} temporal attention response $a_{t}$ ($AUC = 0.876$), it constitutes a very promising result, given the complexity of the task to address. Furthermore, it is worth recalling that there is still room for improvement in \acs{ST-T-ATTEN}, as discussed in the previous section. 

In order to illustrate two potential applications of the \acs{ST-T-ATTEN} architecture in a video surveillance scenario, we provide two more sequences taken from BOSS \cite{BOSS} database in Figure \ref{fig:64_sttatten_results2}. Either spatio-temporal \acsp{VAM} computed by \acs{ST-ATTEN} module or temporal attention responses estimated by \acs{T-ATTEN} are shown. Moreover, in order to bring visual attention closer to the context of anomaly detection, we plot again the anomaly detection signals $e_{t}$, as we did in Figures \ref{fig:64_tva_gt} and \ref{fig:73_anomaly_detection} to validate the fundamental hypothesis of our approach.  

On the one hand, let us imagine a situation where the two video sequences in Figure \ref{fig:64_sttatten_results2} are being shown in real-time in two screens belonging to a large array of camera views. In such a situation, the attention estimated in the temporal domain could be applied to select or highlight the most outstanding screen from the monitoring array at every time, thus driving operator's attention to scenes that potentially show anomalies or suspicious events. 

Furthermore, the estimated temporal attention response could be also applied in off-line tasks which imply reviewing many hours of surveillance recordings, e.g. a surveillance operator inspecting hours of video footage searching for a particular event or anomaly. In this case, our system might reduce the amount of information to be processed by the operator. As a first demonstration of this application, let us note that identifying a unique frame for each anomaly would be sufficient for providing a \acs{CCTV} operator with temporal indicators to review footage faster and more efficiently. Considering this fact and given the temporal attention responses estimated by \acs{ST-T-ATTEN} on the BOSS \cite{BOSS} database, it would be necessary only to retrieve approximately the $16\%$ of the frames of the complete database in order to locate over time the $95\%$ of the existing anomalous events.  

Therefore, we can conclude that, with some adjustments, the \acs{ST-T-ATTEN} proposed might be able to estimate, given one or multiple camera views, spatio-temporal visual attention maps and temporal attention responses at the same time. Our system would thus provide \acs{CCTV} operators a complete experience of visual attention by highlighting the most conspicuous locations in a given scene and, besides, the most relevant time segments, according not only to previous events in the scene, but also to events happening in different camera views at the same time.

% due to we are still far from accurate estimating it. However, it should be noted that certain correlation exists anpproximatelyd... COMPLETE 

\section{Conclusions}
\label{sec:tva_conclusions}
In Chapters \ref{ch:anomaly_detection} and \ref{ch:surveillance_experiments}, we have presented a deep network architecture for visual attention modeling, which goes from \acs{CNN}-based spatio-temporal visual attention prediction to \acs{LSTM}-based attention estimation in the temporal domain. Our model is fundamentally supported by the assumption that a measurement of task-driven visual attention in the temporal domain can be drawn from the dispersion of gaze locations recorded from several subjects. Indeed, the temporal level of attention of observers constitutes an important clue to detect suspicious events or anomalous situations in crowded and complex scenarios. However, it should be borne in mind that, similarly to spatio-temporal visual attention, attention in the temporal domain has to be considered as an early filtering mechanism, which allows to select time segments candidate to contain events of special importance, and therefore reduce the complexity of subsequent anomaly detection systems or to drive the attention of human operators to particular cameras in complex multi-camera \acs{CCTV} systems.

Experimental results have determined the optimal configuration for the \acs{ST-T-ATTEN} proposed. First, it is composed by a \acs{CONV} \acs{ST-ATTEN} stage, which successfully fuses the information provided by three different visual feature maps at its input: RGB-based, motion and objectness. Then, the \acs{ST-ATTEN} module connects with a \acs{LSTM}-based \acs{T-ATTEN} architecture, which models attention in the temporal domain. After evaluating the system on the BOSS \cite{BOSS} database, either the \acs{T-ATTEN} fed with \acsp{VAM} obtained at the output of the \acs{ST-ATTEN} decoder or the one that receives as input the latent representations extracted by the encoder network have resulted in similar performances in terms of the \acs{PCC} score. However, there is still room for the analysis and the improvement of our approach. To that end, future work will address the annotation of a large-scale video surveillance dataset with eye fixations to draw some better conclusions about the behavior of the system, with the ultimate aim of demonstrating its usefulness for guiding anomaly detection in a video surveillance application.

%*****************************************
%*****************************************
%*****************************************
%*****************************************
%*****************************************

%************************************************
\chapter{Conclusions and future lines of research}\label{ch:conclusions} 
%************************************************

\section{Conclusions}\label{sec:conclusions}
In this thesis we have proposed two hierarchical frameworks for visual attention modeling in video sequences. Visual attention can be modeled in two different domains, spatial and temporal, which leads to three types of computational models: spatial, spatio-temporal and temporal. First, spatial models highlight locations of particular interest in a frame by frame basis. Second, modeling attention in the temporal domain allows either to update spatial attention based on previously selected locations (spatio-temporal) or to select time segments of special importance in a video (temporal).

In Chapter \ref{ch:atom}, we have presented our first approach, which is called \acf{ATOM} \cite{csvtmiguel}. Our proposal involves a hierarchical generative probabilistic model for spatio-temporal visual attention prediction and understanding. The definition of the system proposed is generic and independent of the application scenario. Moreover, it is founded on the most outstanding psychological studies about attention \cite{TreismanGelade80,wolfe1994guided}, which hold that attention guidance is not based directly on the information provided by early visual processes but on a contextual representation arisen from them. 

Relying on the well-known \acf{LDA} \cite{LDA} and its supervised extensions \cite{sLDA,DBAyang}, \acs{ATOM} defines task- or context-driven visual attention in video as a mixture of several sub-tasks which, in turn, can be represented as a combination of low-, mid- and high-level spatio-temporal features obtained from video frames. Therefore, given a video frame, the algorithm receives a set of visual feature maps (color, intensity, motion, object-based, etc.) as input. Then, an intermediate level of latent sub-tasks between feature extraction and visual attention modeling is introduced. Finally, latent sub-tasks are aligned to the information drawn from human fixations by means of a categorical variable response, which is generated by a logistic regression model over the sub-task proportions.

In Chapter \ref{ch:atom_experiments}, we have demonstrated the ability of \acs{ATOM} to successfully learn hierarchical representations of visual attention specifically adapted to diverse contexts (outdoors, video games, sports, TV news, etc.), on the basis of a wide set of features. For that purpose, we have made use of the well-known large-scale CRCNS-ORIG \cite{Itti_Carmi09crcns} and DIEM \cite{mital2011clustering} databases. Experiments have shown the advantage of our comprehensive guiding representations based on handcrafted features to understand how visual attention works in different scenarios. In addition, modeling simple eye-catching elements, such as faces or text, through spatial discrete distributions, as well as considering object-based representations learned by recently adopted \acsp{CNN}, our proposal significantly outperforms quite a few competent methods in the literature when estimating visual attention.

In Chapter \ref{ch:anomaly_detection}, we have introduced our second proposal, which is named \acf{ST-T-ATTEN}. This second approach takes a step further and goes from spatio-temporal visual attention estimation to attention estimation in the temporal domain. The model is fundamentally supported by the assumption that a measurement of task-driven visual attention in the temporal domain can be drawn from the dispersion of fixation locations recorded from several observers. First, to demonstrate this hypothesis, we have measured the existing correlation between eye fixation sequences of different viewers when an important or anomalous event happens on the BOSS \cite{BOSS} database. Although this temporal level of attention constitutes a useful clue to detect important events in crowded and complex scenarios, attention in the temporal domain should always be considered as an early filtering mechanism, which selects candidate time segments to contain suspicious events, and therefore reduces the later processing devoted to the anomaly detection. Based on this hypothesis, we have developed \acs{ST-T-ATTEN}, which attempts to model attention in the temporal domain from estimations of spatio-temporal visual attention.  

Inspired by the recent success of \acfp{CNN} for learning deep hierarchical representations and \acs{LSTM} units for time series forecasting, the proposed \acs{ST-T-ATTEN} is composed of two stages. The first stage, which is denoted as \acf{ST-ATTEN}, consists of a \acf{CED} network that receives at its input three high-level feature maps for visual attention guidance (RGB-based, motion and objectness), all of them computed by deep \acsp{CNN}. Then, through an encoding-decoding architecture, the network concurrently estimates spatio-temporal \acsp{VAM} and extracts latent representations of visual attention. We have proposed two configurations for this module of the system. They differ in the outer layers of the encoder and the decoder, which are \acs{CONV} in the first approach and \acs{CONV}-\acs{LSTM} in the second one. 

The second stage of \acs{ST-T-ATTEN}, which is called \acf{T-ATTEN}, involves a \acs{LSTM}-based architecture that estimates, for each frame in a video sequence, a temporal attention response. We have also distinguished between two versions of \acs{T-ATTEN}, depending on the input variable: either the \acs{VAM} at the output of the \acs{ST-ATTEN} or the latent representations generated by the encoder. 

In Chapter \ref{ch:surveillance_experiments}, the proposed \acs{ST-T-ATTEN} architecture has been evaluated in a video surveillance scenario defined by the BOSS \cite{BOSS} database, which contains video sequences recorded in a railway transport context, with different types of suspicious or anomalous events (several women harassment, a cell phone theft, a passengers fight, etc.). The main purpose of our experiments has been to assess various architectures of our proposal. Experiments have concluded that the best performing architecture is composed by a \acs{CONV} \acs{ST-ATTEN} stage, which successfully fuses the information provided by the three input feature maps. Then, either the \acs{T-ATTEN} fed with the \acsp{VAM} obtained at the output of the \acs{ST-ATTEN} decoder or the latent representations extracted by its associated encoder have resulted in similar performances in terms of the \acs{PCC} score. However, there is still room for the improvement of our system, as discussed in the experimental section. 

Finally, we have also discussed two potential end-user applications for our proposal. On the one hand, given a video surveillance scenario, the temporal attention response could be applied to select in real-time the most outstanding screens from the monitoring array, thus driving operator's attention to scenes that potentially show anomalies or suspicious events. On the other hand, the estimated response could be also applied in off-line tasks which imply reviewing many hours of surveillance recordings, reducing the information to be processed by the operator. With some adjustments, our system might be able to provide \acs{CCTV} operators a complete experience of visual attention, not only highlighting the most conspicuous locations in a scene, but also selecting the most relevant time segments, according to both previous events in the scene and events happening in different camera views at the same time.

% \clearpage
\section{Future lines of research}\label{sec:future_lines}
Lastly, we conclude this thesis by identifying and discussing potential future lines of research related to our contributions.

At this point, there is no doubt about the great benefits of visual attention modeling in the framework of \acf{AI}, nor about the infinite possibilities that such an abstract concept opens for the processing and understanding of this big data world. Despite the wide variety of computational models of visual attention existing in the literature, much remains to be done, not only to meet a system that automatically addresses this cognitive function, but also to understand how \acs{HVS} carries out this optimization process. 

Turning to the two popular representation learning paradigms introduced at the beginning of this thesis, \acf{DL} and \acf{PGM}, our contributions have shown the importance of either the task of \emph{seeing}, performed by \acs{DL} representations, or the ability of \emph{thinking}, characteristic of \acs{PGM}, for visual attention modeling and understanding. 

First, it is important to achieve good representations of the world that surrounds us for attention guidance, and it is here where \acsp{DNN} architectures and, in particular, \acsp{CNN}, play an essential role in machine perception. In addition, given that visual attention involves not only one, but several complex tasks, it is paramount to understand how computational visual attention deals with the hierarchical representations provided by \acsp{DNN}, through probabilistic methods that explain relationships between the observed variables. This direction, recently set by \acf{BDL} \cite{wang2016towards}, is the one that we plan to follow in our future research, paying special attention to \acs{BDL} for topic models, which constitutes a revision to probabilistic \acfp{LTM} \cite{LDA,sLDA,DBAyang}, on the basis of which our hierarchical \acs{ATOM} framework for visual attention understanding has been built. Discovering sub-tasks, not only over space but also along time, will allow establishing relationships between recognized concepts in one or multiple video sequences, both in the same scene or in different ones.

Secondly, in the latter part of the thesis, we have demonstrated the major advantages of modeling visual attention in the temporal domain, selecting video segments of special importance, which subsequently help to reduce the computational burden of subsequent end-user applications. Visual attention has been barely tackled from this perspective in the literature up to date, in spite of its usefulness for the processing and analysis of vast amounts of visual information in applications such as anomaly detection. 

One interesting research line we have not covered in this thesis is the interpretation of eye movement sequences, establishing relationships between the content of fixated locations. This would allow to develop more comprehensible and valuable systems for estimating the variation of visual attention over time. Reinforcement learning methods seem a promising way of addressing this challenge \cite{mnih2014recurrent}.

Finally, we are highly motivated to model spatio-temporal visual attention, as well as attention in the temporal domain, given multiple video sequences played at the same time, with the aim of assisting experts in crowded and complex scenarios. For that purpose, we will soon proceed to annotate large-scale video surveillance databases, such as VIRAT \cite{oh2011large} or UCF-Crime \cite{sultani2018real}, with human fixations, which will serve for a further analysis and the improvement of the deep \acs{ST-T-ATTEN} architecture proposed.

%*****************************************
%*****************************************
%*****************************************
%*****************************************
%*****************************************

%\include{multiToC} % <--- just debug stuff, ignore for your documents
% ********************************************************************
% Backmatter
%*******************************************************
\appendix
\cleardoublepage
%\part{Appendix}
%********************************************************************
% Appendix
%*******************************************************
% If problems with the headers: get headings in appendix etc. right
%\markboth{\spacedlowsmallcaps{Appendix}}{\spacedlowsmallcaps{Appendix}}
\chapter{Derivation of the formulas for the ATOM}\label{ch:atom_formulas}

In this appendix, we provide the derivation of the formulas for the \acf{ATOM} presented in Chapter \ref{ch:atom}.

\section{Expansion of the lower bound}
As introduced in section \ref{sec:inference}, the optimization of the probabilistic model for visual attention understanding and prediction proposed in the thesis implies maximizing the \acf{ELBO} over the log-likelihood of all the frames in a corpus of videos. In particular, using Jensen's inequality, the \acs{ELBO} of the log-likelihood of a frame with $N$ spatial locations can be expressed as:

\begin{align} \label{eq:app_elbo}
  &log \ p(f_{1:N,1:L},g_{1:N}|\alpha,\Gamma_{1:K,1:L},\eta) \geq  E_{q}[log \ p(\theta|\alpha)] \nonumber \\
  &+ \sum_{n=1}^{N} E_{q}[log \ p({\bf z_n}|\theta)] + \sum_{n=1}^N E_{q}[log \ p(f_{n,1:L} | {\bf z_n}, \Gamma_{1:K,1:L})] \nonumber  \\
  &+ \sum_{n=1}^N E_{q}[log \ p(g_n | {\bf z_n}, \eta)] + H(q)
\end{align}

\noindent where $L$ is the number of visual descriptors computed as input for the models, $K$ is the number of sub-tasks or topics inferred,  $E_{q}[\cdot]$ is the expectation over the variational distribution $q$ and $H(\cdot)$ is the entropy of the variational distribution. \\

The first two terms in the \acs{ELBO} and the entropy of the variational distribution are identical to the corresponding terms in the \acs{ELBO} for unsupervised \acs{LDA} \cite{LDA}: 

\begin{align}
&E_{q}[log \ p(\theta|\alpha)] & = & \ log \ \Gamma \left(\sum_{k=1}^{K} \alpha_{k} \right) - \sum_{k=1}^{K} log \ \Gamma \left( \alpha_{k} \right) \nonumber \\
& & + & \sum_{k=1}^{K} \left( \alpha_{k}-1 \right) E_{q}[log \ \theta_{k}] \\
&E_{q}[log \ p({\bf z_n}|\theta)] & = & \sum_{k=1}^{K} \phi_{nk} E_{q}[log \ \theta_{k}] \\
&H(q) & = & -\sum_{n=1}^{N} \sum_{k=1}^{K} \phi_{nk} log \ \phi_{nk} - log \ \Gamma \left(\sum_{k=1}^{K} \gamma_{k} \right) \nonumber \\
& & + & \sum_{k=1}^{K} log \ \Gamma(\gamma_{k}) - \sum_{k=1}^{K} (\gamma_{k}-1)E_{q}[log \ \theta_{k}],
\end{align}

\noindent where the expectation of the log of the multinomial random variable $\theta_{k}$ is:

\begin{gather}\label{eq:theta_expectation}
E_{q}[log \ \theta_{k}] = \Psi(\gamma_{k})-\Psi \left( \sum_{j=1}^{K} \gamma_{j} \right),
\end{gather}

\noindent being $\Psi(\cdot)$ the digamma function. \\

The third and fourth terms are derived in the following subsections.

\subsection{Lower bound of the local appearance model}\label{sec:bound_appearance}
The third term is the expected log probability of the data given the related topic model parameters. We assume conditional independence among features. In the following paragraphs, we derive this expression for the different distributions considered. \\

If the feature map $f_{nl}$ is modeled with a univariate \emph{Gaussian distribution} $\Gamma_{1:K,l} \sim \{\mu_{1:K,l}, \sigma^{2}_{1:K,l}\}$, such as for basic and novelty spatio-temporal features or CNN-based features, the equation for this term is:

\begin{align}
  E_{q}[log \ p(f_{nl} | {\bf z_n}, \Gamma_{1:K,l})] = -\sum_{k=1}^{K} \phi_{nk}\log(\sigma_{kl}\sqrt{2\pi}) \nonumber \\ 
  - \sum_{k=1}^{K} \phi_{nk}\frac{(f_{nl}-\mu_{kl})^2}{2\sigma_{kl}^2}
\end{align}
\noindent where $\phi_{nk}$ is the probability that the location $n$ has been drawn by the topic $k$. \\

In the case of camera motion features, the distribution is a multivariate Gaussian $p(\xv_{n} | {\bf z_n}, \mu_k,\Sigma_k)$ with $\mu_k = \cv_k \odot \uv$, being ${\bf c_{k}}$ a parameter to be estimated and ${\bf \uv}=(u,v)$ the camera motion vector. However, due to the diagonal nature of the covariance matrix $\Sigma_k$ we can decompose it into two independent univariate Gaussian distributions and apply the previous expression: 

\begin{align}
  E_{q}[log \ p({\bf \xv_n} | {\bf z_n}, \cv_{k})] = \nonumber \\ 
  -\sum_{k=1}^{K} \phi_{nk}\log(\sqrt{2\pi\Sigma_{k}}) 
  - \sum_{k=1}^{K} \phi_{nk}\frac{(\xv_{n}-\cv_{k}\uv)^T(\xv_{n}-\cv_{k}\uv)}{2\Sigma_{k}}. 
\end{align}

\noindent where ${\bf \xv_{n}} = (x_{n}, y_{n})$ is a vector with the spatial coordinates of the location $n$. As explained in section \ref{sec:motion_features}, $\Sigma_{k}$, which controls the spatial extent of the Gaussian distribution, has been empirically set to $\Sigma_{k}=diag(0.25)$ in order to cover a sufficiently wide area in the scene. \\

In contrast, if the feature is modeled as a \emph{discrete probability distribution} over cells $r$ in a grid, as happens for objects-based features, the expression is:
\begin{equation}
\begin{split}
  &E_{q}[log \ p(r_n | {\bf z_{n}}, \beta_{l{z_n}})] = \sum_{k=1}^{K} \phi_{nk} \log(\beta_{klr_n}), \\ 
\end{split}
\end{equation}
\noindent where $r_n$ stands for the region in the non-uniform grid defined for the object $l$ that contains the location $n$, and $\beta_{klr_n}$ is the value of the of the discrete distribution in that region for the object $l$ and the topic $k$. 

\subsection{Lower bound of the visual attention response}\label{sec:bound_va}
The fourth term includes the visual attention response variable $g_n$, which is generated from a Bernoulli distribution, i.e., 
	
	\begin{align}
	p(g_n|\pi_n) = (\pi_n)^{g_n}(1-\pi_n)^{(1-g_n)},
	\end{align}
	
	\noindent where $\pi$ is a logistic regression model based on a weighted empirical average of the Dirichlet realization $\eta^T{\bf z_n}$, being $\eta$ the parameter vector that models attention based on the selected topic ${\bf z_{n}}$:
	
	\begin{align}\label{eq:bernoulli}
	p(g_n|{\bf z_n},\eta) = \frac{\exp{(g_n\eta^T{\bf z_n})}}{1+\exp{(\eta^T{\bf z_n})}}.
	\end{align}
	
	Thus, the Bernoulli distribution is as follows:
	\begin{align}
	p(g_n|{\bf z_n},\eta) \sim Be \left( \frac{\exp{(g_n\eta^T{\bf z_n})}}{1+\exp{(\eta^T{\bf z_n})}} \right).
	\end{align}
	
	According to \cite{jaakkola2000bayesian}, the logistic function in Eq. \ref{eq:bernoulli} can be symmetrized as follows:
	
	\begin{align}
	p(g_n|{\bf z_n},\eta) = \frac{\exp{((g_n-\frac{1}{2})\eta^T{\bf z_n})}}{\exp{\frac{(\eta^T{\bf z_n})}{2}}+\exp{\frac{(-\eta^T{\bf z_n})}{2}}}.
	\end{align}
	
	Then, the expected log probability of the response variable given the topic assignments id expressed as:
\begin{equation}
\begin{split}
  E_{q}[log \ p(g_n | {\bf z_n}, \eta)] = E_{q}\bigg[\left(g_n -\frac{1}{2}\right) \eta^{T} \bf z_n \bigg] \\
  -E_{q}\bigg[log\left(exp{\left(\frac{\eta^{T} \bf z_n}{2}\right)}+exp{\left(\frac{-\eta^{T} \bf z_n}{2}\right)}\right)\bigg]
\end{split}
\end{equation}
\noindent By taking second derivatives, it can be noticed that the second term above, which can be denoted as $E_{q}[f(\eta^{T} \bf z_n)]$, is a convex function in the variable $\eta^{T^{2}} {\bf z_n}^{2} = (\eta^{T} \odot \eta^{T})({\bf z_n} \odot {\bf z_n})$, so we can bound it by using the lower bound for the logistic function

\begin{align}
f(\eta^{T} {\bf z_n}) \geq f(\xi_{n}) + \frac{\partial f(\xi_{n})}{\partial \xi_{n}^{2}}(\eta^{T^{2}} {\bf z_n}^{2}-\xi_{n}^{2}),
\end{align}

\noindent which is the first order Taylor expansion in the variable $\eta^{T^{2}} \bf z_n^{2}$:

\begin{equation}\label{eq:xi_lower_bound}
\begin{split}
  &log\left(exp{\left(\frac{\eta^{T} \bf z_n}{2}\right)}+exp{\left(\frac{-\eta^{T} \bf z_n}{2}\right)}\right)\\
  &\geq -\frac{\xi_{n}}{2} -log(1+exp(-\xi_{n})) \\
  &-\frac{1}{4\xi_{n}}tanh\left(\frac{\xi_{n}}{2}\right)E_{q}\bigg[\eta^{T^{2}} {\bf z_{n}^{2}} - \xi_{n}^{2}\bigg] \\
  &\approx -\frac{\xi_{n}}{2} -log(1+exp(-\xi_{n})) \\
  &-\frac{1}{4\xi_{n}} tanh\left(\frac{\xi_{n}}{2}\right)(\eta^{T^{2}}\phi_{n} - \xi_{n}^{2}), 
\end{split}
\end{equation}

\noindent where $\phi_{n}$ is the vector of topic proportions $\phi_{nk}$ in the location $n$ and $\xi_{n}$ is an additional variational parameter associated to each point $n$. 

It should be noted that, during variational inference, we work on expected values. This means that the indexing variable ${\bf z}_{n}$ is replaced by the variational $\phi_{n}$, which now contains the expected values of the topic assignments given a location $n$. Therefore, since $\phi_{n}$ is a vector with real values (the topic proportions for that sampled location), in practice each location $n$ is in turn modeled as the mixture of sub-tasks that best explains its visual appearance.

\section{Derivation of the formulas for the variational parameters}
This section includes the complete derivation of the update equation of the variational multinomial $\phi$, which is computed in the E-step of the inference process. \\

First, we begin with the lower bound that depends on $\phi$, incorporating a Lagrange parameter $\lambda$ to ensure that $\sum_{k=1}^{K} \phi_{nk}=1$:

\begin{align}
	&ELBO_{\phi} = \sum_{n=1}^{N} \sum_{k=1}^{K} \phi_{nk} E_{q}[log \ \theta_{k}] + \sum_{n=1}^N E_{q}[log \ p(f_{n,1:L} | {\bf z_n}, \Gamma_{1:K,1:L})] \nonumber \\
  &+ \sum_{n=1}^N E_{q}[log \ p(g_n | {\bf z_n}, \eta)] + H(q) \nonumber \\
  &= \sum_{n=1}^{N} \sum_{k=1}^{K} \phi_{nk} \left(\Psi(\gamma_{k})-\Psi \left( \sum_{j=1}^{K} \gamma_{j} \right) \right) \nonumber \\
	&- \sum_{n=1}^{N} \sum_{k=1}^{K} \sum_{l=1}^{L_{C}} \phi_{nk}\left(\log(\sigma_{kl}\sqrt{2\pi}) + \frac{(f_{nl}-\mu_{kl})^2}{2\sigma_{kl}^2}\right) \nonumber \\	
	&+ \sum_{n=1}^{N} \sum_{k=1}^{K} \sum_{l=1}^{L_{D}} \phi_{nk} \log(\beta_{klr_n}) \nonumber \\
	&+ \sum_{n=1}^{N} \sum_{k=1}^{K} \left(\left(g_n -\frac{1}{2}\right) \eta_{k} \phi_{nk} -\frac{1}{4\xi_{nk}} tanh\left(\frac{\xi_{nk}}{2}\right)(\eta_{k}^{2}\phi_{nk} - \xi_{nk}^{2})\right) \nonumber \\
	&- \sum_{n=1}^{N} \sum_{k=1}^{K} \phi_{nk} log \ \phi_{nk} - log \ \Gamma \left(\sum_{k=1}^{K} \gamma_{k} \right) \nonumber \\
 &+ \sum_{k=1}^{K} log \ \Gamma(\gamma_{k}) - \sum_{k=1}^{K} (\gamma_{k}-1)E_{q}[log \ \theta_{k}] \nonumber \\ &+ \sum_{n=1}^{N} \lambda_{n} \left(\sum_{k=1}^{K} \phi_{nk} -1 \right),
\end{align}

\noindent being $L_{C}$ and $L_{D}$ the number of continuous (Gaussian) and discrete features, respectively, and $L=L_{C}+L_{D}$ the total number of features. 

\noindent If we take the derivative of the \acs{ELBO} with respect to $\phi_{nk}$:
	\begin{align}
	&\frac{\partial ELBO_{\phi_{nk}}}{\partial \phi_{nk}} = \Psi(\gamma_{k})-\Psi \left( \sum_{j=1}^{K} \gamma_{j} \right) \nonumber \\
	&-\sum_{l=1}^{L_{C}} \left(\log(\sigma_{kl}\sqrt{2\pi}) + \frac{(f_{nl}-\mu_{kl})^2}{2\sigma_{kl}^2}\right) + \sum_{l=1}^{L_{D}} \log(\beta_{klr_n}) \nonumber \\
	&+\left(g_n -\frac{1}{2}\right) \eta_{k} -\frac{1}{4\xi_{nk}} tanh\left(\frac{\xi_{nk}}{2}\right)\eta_{k}^{2} -log \phi_{nk} -1 + \lambda_{n}
	\end{align}
	
\noindent and set it to zero, we obtain the equation for updating the multinomial parameter:
	\begin{align}
	\phi_{nk} \propto & \frac{\prod_{l=1}^{L_{D}} \beta_{klr_n}}{\prod_{l=1}^{L_{C}} \sigma_{kl}} \exp \bigg[\Psi(\gamma_{k})-\Psi\left(\sum_{j=1}^{k} \gamma_{j}\right) + \nonumber \\ 
&\left(g_{n}-\frac{1}{2}\right)\eta_{k} -\frac{1}{4\xi_{k}}tanh\left(\frac{\xi_{k}}{2}\right)\eta^{2}_{k} - \nonumber \\
&\sum_{l=1}^{L_{C}} \frac{(f_{nl}-\mu_{kl})^2}{2\sigma_{kl}^2}\bigg].
	\end{align}
 
In addition, it should be noted that the equations corresponding to the variational Dirichlet $\gamma$ and the Dirichlet parameters $\alpha$ are not included here, because they are identical to those in the original \acs{LDA} \cite{LDA}.
 
\section{Derivation of the formulas for the model parameters}
This section includes the complete derivation of the update equations for the model parameters computed in the M-step of the inference process, given a corpus of $T$ video frames, each one with $N_{t}$ spatial locations.

First, parameters $\mu_{kl}$ and $\sigma^{2}_{kl}$ are computed for each Gaussian feature $l$ and topic $k$. 

\begin{itemize}
	\item The \acs{ELBO} that depends on $\mu_{kl}$ is:
	\begin{align}
	&ELBO_{\mu_{kl}} = -\sum_{t=1}^{T} \sum_{n=1}^{N_t} \sum_{k=1}^{K} \phi_{nk}\frac{(f_{tnl}-\mu_{kl})^2}{2\sigma_{kl}^2}.
	\end{align}
	
	Computing its derivative with respect to $\mu_{kl}$ gives:
	\begin{align}
	&\frac{\partial ELBO_{\mu_{kl}}}{\partial \mu_{kl}} = \sum_{t=1}^{T} \sum_{n=1}^{N_t} \phi_{tnk} \frac{(f_{tnl}-\mu_{kl})}{\sigma_{kl}^2}
	\end{align}
	
	Setting it to zero, we obtain the update equation:
	\begin{align}
	&\mu_{kl}  = \frac{1}{\Delta_{kl}} \sum_{t=1}^{T} \sum_{n=1}^{N_t} 				\phi_{tnk} f_{tnl}
	\end{align}
	
	\item The \acs{ELBO} that depends on $\sigma^{2}_{kl}$ is:
	\begin{align}
	&ELBO_{\sigma^{2}_{kl}} = -\sum_{t=1}^{T} \sum_{n=1}^{N_t} \sum_{k=1}^{K} \phi_{nk}\log(\sigma_{kl}\sqrt{2\pi}) \nonumber \\ 
&- \sum_{t=1}^{T} \sum_{n=1}^{N_t} \sum_{k=1}^{K} \phi_{nk}\frac{(f_{tnl}-\mu_{kl})^2}{2\sigma_{kl}^2}.
	\end{align}
	
	Computing its derivative with respect to $\sigma^{2}_{kl}$ gives:
	\begin{align}
	&\frac{\partial ELBO_{\sigma^{2}_{kl}}}{\partial \sigma^{2}_{kl}} = -\sum_{t=1}^{T} \sum_{n=1}^{N_t} \frac{\phi_{nk}}{\sigma_{kl}} \nonumber \\ &+ \sum_{t=1}^{T} \sum_{n=1}^{N_t} \frac{\phi_{nk}}{\sigma_{kl}} \frac{(f_{tnl}-\mu_{kl})^2}{\sigma_{kl}^2} 
	\end{align}
	
	Setting it to zero, we obtain the update equation:
	\begin{align}
	&\sigma^{2}_{kl} = \frac{1}{\Delta_{kl}} \sum_{t=1}^{T} \sum_{n=1}^{N_t} 	\phi_{tnk} (f_{tnl}-\mu_{kl})^2
	\end{align}
\end{itemize}

\noindent For both $\mu_{kl}$ and $\sigma^{2}_{kl}$, $\Delta_{kl}=\sum_{t=1}^{T} \sum_{n=1}^{N_t} \phi_{tnk}$ is the normalization factor. \\

In the case of camera motion, as mentioned above, the parameter is the vector ${\bf \cv_k} = (c_{kx}, c_{ky})$ that multiplies the camera motion vector ${\bf \uv_{t}} = (u_{t}, v_{t})$ to determine the mean of the Gaussian distribution. The \acs{ELBO} that depends on this parameter is:

\begin{align}
	&ELBO_{\bf \cv_k} = -\sum_{t=1}^{T} \sum_{n=1}^{N_t} \sum_{k=1}^{K} \phi_{nk}\frac{(\xv_{tn}-\cv_{k}\uv_{t})^T(\xv_{tn}-\cv_{k}\uv_{t})}{2\Sigma_{k}},
\end{align}
	
\noindent where ${\bf \xv_{tn}} = (x_{tn}, y_{tn})$ stands for the spatial coordinates vector of the location $n$ in frame $t$. By computing its derivative with respect to ${\bf \cv_k}$:

\begin{align}
	&\frac{\partial ELBO_{\bf \cv_k}}{\partial {\bf \cv_k}} = \sum_{t=1}^{T} \sum_{n=1}^{N_t} \phi_{nk}\frac{(\uv_{t}\xv_{tn}-\uv_{t}^{2}\cv_{k})}{\Sigma_{k}}
\end{align}

\noindent and setting it to zero, we obtain the following update equation:

\begin{eqnarray}
{\bf \cv_{k}}= \frac{\sum_{t=1}^{T} \sum_{n=1}^{N_t} \phi_{tnk} {\bf u_{t}} {\bf x_{tn}}}{\sum_{t=1}^{T} \sum_{n=1}^{N_t} \phi_{tnk} {\bf u_{t}}^2}.  
\end{eqnarray}

Finally, for the case of object-based discrete features, the \acs{ELBO} that depends on the probabilities $\beta_{klr}$ of the regions $r$ defined on the object-detector $l$ and for every topic $k$ is:

\begin{align}
	&ELBO_{\beta_{klr}} = \sum_{t=1}^{T} \sum_{n=1}^{N_t} \sum_{k=1}^{K} \phi_{nk} \log(\beta_{klr_n}) + \sum_{k=1}^{K} \lambda_{kl} \left(\sum_{r=1}^{R} \beta_{klr_n}-1 \right).
\end{align}

\noindent where we have added the Lagrange multipliers $\lambda_{kl}$. By computing its derivative with respect to $\beta_{klr}$:

\begin{align}
	&\frac{\partial ELBO_{\beta_{klr}}}{\partial \beta_{klr}} = \sum_{t=1}^{T} \sum_{n=1}^{N_t} \frac{\phi_{nk}}{\beta_{klr}} + \lambda_{kl}.
\end{align}

\noindent Setting this derivative to zero gives the following update equation:

\begin{eqnarray}
\beta_{klr} \propto \sum_{t=1}^{T} \sum_{n=1}^{N_t} \phi_{tnk} 1[r_{nl}=r] 
\end{eqnarray}
\noindent where $1[r_{nl}=r]$ means that we have a 1 just in case the region of the point $n$ for the detector $l$ is $r$ (otherwise we have a zero).

\section[Derivation of the formulas for the logistic regression model]{Derivation of the formulas for the parameters of the logistic regression model}
This section includes the complete derivation of the update equations for the parameters of the logistic regression model proposed to estimate visual attention over the underlying topics obtained, either in the E-step or in the M-step of the inference process.

\begin{itemize}
	\item In the E-step, the variational parameter $\xi$ has to be updated. The \acs{ELBO} that depends on $\xi_{nk}$ is:
\begin{align}
&ELBO_{\xi_{nk}} = -\sum_{t=1}^{T} \sum_{n=1}^{N_{t}} \sum_{k=1}^{K} \left(\frac{\xi_{tnk}}{2} +log(1+exp(-\xi_{tnk}))\right) \nonumber \\
  &+\sum_{t=1}^{T} \sum_{n=1}^{N_{t}} \sum_{k=1}^{K} \frac{1}{4\xi_{tnk}} tanh \left(\frac{\xi_{tnk}}{2}\right)(\eta^{2}_{k}\phi_{tnk} - \xi_{tnk}^{2}),
\end{align}

which corresponds to the lower bound in Eq. \ref{eq:xi_lower_bound}. This lower bound is exact if $\xi_{nk}^{2}=\eta^{2}_{k}\phi_{nk}$. Consequently, the update equation for this parameter is:
\begin{align}
\xi_{nk} = \eta_{k} \phi_{nk}.
\end{align} 
		
	\item In the M-step, we update the parameter $\eta$, attending to the lower bound that depends on it:
\begin{align}
&ELBO_{\eta_{k}} = \sum_{t=1}^{T} \sum_{n=1}^{N_{t}} \sum_{k=1}^{K} \left(g_{tn} -\frac{1}{2}\right) \eta_{k} \phi_{tnk} \nonumber \\ &-\sum_{t=1}^{T} \sum_{n=1}^{N_{t}} \sum_{k=1}^{K} \frac{1}{4\xi_{tnk}} tanh\left(\frac{\xi_{tnk}}{2}\right)(\eta_{k}^{2}\phi_{tnk} - \xi_{tnk}^{2}).
\end{align}

Computing the derivative with respect to $\eta_{k}$:
\begin{align}
	&\frac{\partial ELBO_{\eta_{k}}}{\partial \eta_{k}} = \sum_{t=1}^{T}  \sum_{n=1}^{N_{t}}  \left(g_{tn} -\frac{1}{2}\right) \phi_{tnk} \nonumber \\
	&-\sum_{t=1}^{T}  \sum_{n=1}^{N_{t}} \frac{1}{2\xi_{nk}} tanh\left(\frac{\xi_{nk}}{2}\right)\eta_{k}\phi_{nk}.
\end{align}

Setting it to zero, we obtain the update equation:
\begin{align}
&\eta_{k} = \frac{2\sum_{t=1}^{T} \sum_{n=1}^{N_t} \phi_{tnk} 			(g_{tn}-\frac{1}{2})}{\sum_{t=1}^{T} \sum_{n=1}^{N_t} 					\frac{\phi_{tnk}}{\xi_{nk}} tanh(\frac{\xi_{nk}}{2})}.
\end{align} 
\end{itemize}

%********************************************************************
% Appendix
%*******************************************************
% If problems with the headers: get headings in appendix etc. right
%\markboth{\spacedlowsmallcaps{Appendix}}{\spacedlowsmallcaps{Appendix}}
\chapter{Eye-tracking databases used in the thesis}\label{ch:databases}
This appendix summarizes the databases used in Chapters \ref{ch:atom_experiments} and \ref{ch:surveillance_experiments} from this thesis, with the aim of providing the list of videos included in each of the categories established for our experiments.

It should be noted that SALICON \cite{jiang2015salicon} database has also been considered for RGB-based feature maps extraction in Chapter \ref{ch:surveillance_experiments}, where it has been described. This database is not covered in this appendix because it was not necessary neither to divide its images into categories nor to annotate it with more observers' fixations, so it has been used as in other related works in the state-of-the-art.

Last but not least, the three most significant attraction (\acs{AT}) and inhibition (\acs{IT}) sub-tasks determined by the spatio-temporal model for visual attention understanding presented in Chapter \ref{ch:atom} are provided for all the video genres in CRCNS-ORIG \cite{Itti_Carmi09crcns} and DIEM \cite{mital2011clustering} databases. For further comprehension of the diagrams provided, the reader is referred to section \ref{sec:va_understanding}.

\section{CRCNS-ORIG database}
\subsection{Description}
CRCNS-ORIG \cite{Itti_Carmi09crcns} dataset contains eye movement recordings from eight distinct subjects freely watching 50 different video clips (over 46,000 video frames, 25 minutes total, $640\times480$). Eye traces have been obtained using a 240 Hz ISCAN RK-464 eye-tracker. Eye fixations of at least 4 subjects are provided for each clip.

\begin{table}[H]
  \small
  \centering
  \caption{Categories in the CRCNS-ORIG \cite{Itti_Carmi09crcns} database. Clips included in each category are enumerated, together with their number of frames.}\label{fig:B_scenarios_ORIG}
  \subfloat[Outdoor, 17 clips]{%
    \hspace{0cm}%
    \begin{tabular}[b]{l|c}
    		\hline
    		{Clip name} & {Frames}\\ \hline
    		{beverly01} & $490$ \\  
    		{beverly03} & $481$ \\  
    		{beverly05} & $546$ \\  
    		{beverly06} & $521$ \\   
    		{beverly07} & $357$ \\ 
    		{beverly08} & $237$ \\ 
    		{monica03} & $1,526$ \\  
    		{monica04} & $640$ \\  
    		{monica05} & $611$ \\  
    		{monica06} & $164$ \\   
    		{standard01} & $254$ \\ 
    		{standard02} & $515$ \\ 
    		{standard03} & $309$ \\ 
    		{standard04} & $612$ \\ 
    		{standard05} & $483$ \\ 
    		{standard06} & $434$ \\  
    		{standard07} & $177$ \\ \hline
    		{TOTAL} & $8,357$ \\ \hline
    	\end{tabular}
    \hspace{0cm}%
  }\hspace{0.1cm}
  \subfloat[Videogames, 9 clips]{%
    \hspace{0cm}%
    \begin{tabular}[b]{l|c}
    		\hline
    		{Clip name} & {Frames}\\ \hline
    		{gamecube02} & $1,819$ \\  
    		{gamecube04} & $2,083$ \\  
    		{gamecube05} & $213$ \\  
    		{gamecube06} & $2,440$ \\  
    		{gamecube13} & $898$ \\  
    		{gamecube16} & $2,814$ \\  
    		{gamecube17} & $2,114$ \\  
    		{gamecube18} & $1,999$ \\  
    		{gamecube23} & $1,429$ \\ \hline
    		{TOTAL} & $15,809$ \\ \hline
    	\end{tabular}
    \hspace{0cm}%
  }\hspace{0.1cm}
  \subfloat[Commercials, 4 clips]{%
    \hspace{0cm}%
    \begin{tabular}[b]{l|c}
    		\hline
    		{Clip name} & {Frames}\\ \hline
    		{tv-ads01} & $1,077$ \\  
    		{tv-ads02} & $387$ \\  
    		{tv-ads03} & $841$ \\  
    		{tv-ads04} & $313$ \\ \hline
    		{TOTAL} & $2,618$ \\ \hline
    	\end{tabular}
    \hspace{0cm}%
  }\hspace{0.1cm}
  \subfloat[TV News, 7 clips]{%
    \hspace{0cm}%
    \begin{tabular}[b]{l|c}
    		\hline
    		{Clip name} & {Frames}\\ \hline
    		{tv-news01} & $918$ \\  
    		{tv-news02} & $1,058$ \\  
    		{tv-news03} & $1,444$ \\  
    		{tv-news04} & $491$ \\   
    		{tv-news05} & $1,341$ \\ 
    		{tv-news06} & $1,643$ \\  
    		{tv-news09} & $1,176$ \\ \hline
    		{TOTAL} & $8,071$ \\ \hline
    	\end{tabular}
    \hspace{0cm}%
  }\hspace{0.1cm}
  \subfloat[Sports, 5 clips]{%
    \hspace{0cm}%
    \begin{tabular}[b]{l|c}
    		\hline
    		{Clip name} & {Frames}\\ \hline
    		{tv-sports01} & $579$ \\  
    		{tv-sports02} & $444$ \\  
    		{tv-sports03} & $1,460$ \\  
    		{tv-sports04} & $982$ \\  
    		{tv-sports05} & $1,386$ \\ \hline
    		{TOTAL} & $4,851$ \\ \hline
    	\end{tabular}
    \hspace{0cm}%
  }\hspace{0.1cm}
  \subfloat[Talk Shows, 4 clips]{%
    \hspace{0cm}%
    \begin{tabular}[b]{l|c}
    		\hline
    		{Clip name} & {Frames}\\ \hline
    		{tv-talk01} & $1,651$ \\  
    		{tv-talk03} & $783$ \\  
    		{tv-talk04} & $1,258$ \\  
    		{tv-talk05} & $552$ \\ \hline
    		{TOTAL} & $4,244$ \\ \hline
    	\end{tabular}
    \hspace{0cm}%
  }\hspace{0.1cm}
  \subfloat[Others, 4 clips]{%
    \hspace{0cm}%
    \begin{tabular}[b]{l|c}
    		\hline
    		{Clip name} & {Frames}\\ \hline
    		{saccadetest} & $516$ \\  
    		{tv-action01} & $567$ \\  
    		{tv-announce01} & $434$ \\   
    		{tv-music01} & $1,022$ \\ \hline
    		{TOTAL} & $2,539$ \\ \hline
    	\end{tabular}
    \hspace{0cm}%
  }
\end{table}

\subsection{Video categories}
For our experiments on context-driven visual attention understanding and prediction presented in Chapter \ref{ch:atom_experiments}, we have divided the dataset into seven categories: \emph{Outdoor, Videogames, Commercials, TV News, Sports, Talk Shows} and \emph{Others}. Videos included in each category are enumerated in Table \ref{fig:B_scenarios_ORIG}.

\subsection{Context-aware visual attention understanding}
\begin{figure}[H]
    \centering
        \includegraphics[trim=10cm 1cm 12cm 0.5cm, width=0.6\textwidth]{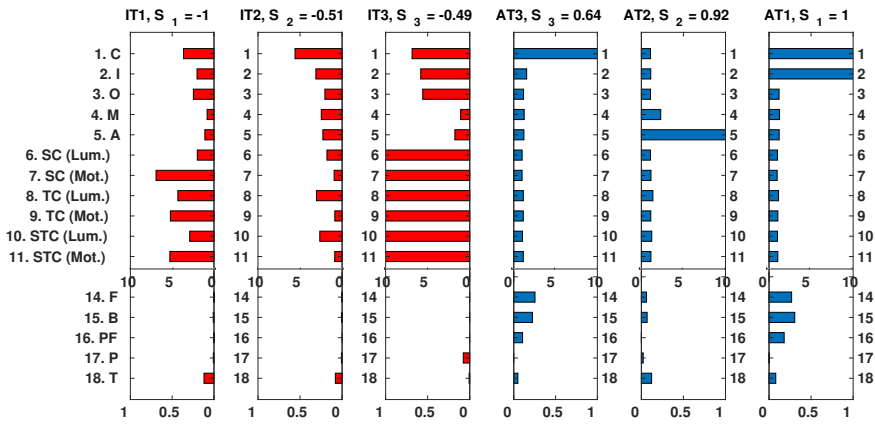}
    \caption{CRCNS-ORIG \cite{Itti_Carmi09crcns} database: \emph{Context-Generic}}
	\label{fig:B_generic_orig}
\end{figure}

\begin{figure}[H]
    \centering
        \includegraphics[trim=10cm 1cm 12cm 0.5cm, width=0.6\textwidth]{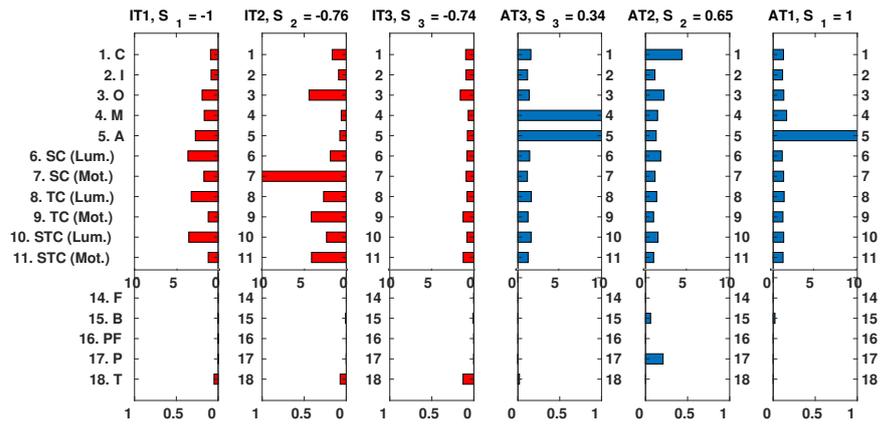}
    \caption{CRCNS-ORIG \cite{Itti_Carmi09crcns} database: \emph{Videogames}}
	\label{fig:B_videogames_orig}
\end{figure}

\begin{figure}[H]
    \centering
        \includegraphics[trim=10cm 1cm 12cm 0.5cm, width=0.6\textwidth]{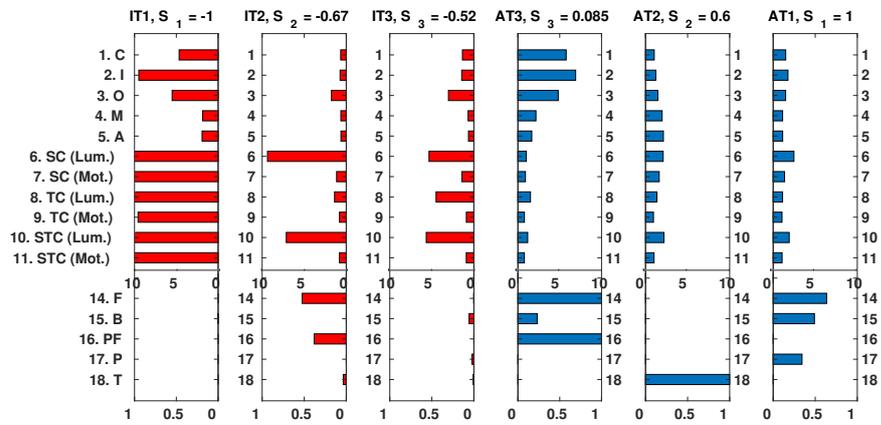}
    \caption{CRCNS-ORIG \cite{Itti_Carmi09crcns} database: \emph{Commercials}}
	\label{fig:B_commercials_orig}
\end{figure}

\begin{figure}[H]
    \centering
        \includegraphics[trim=10cm 1cm 12cm 0.5cm, width=0.6\textwidth]{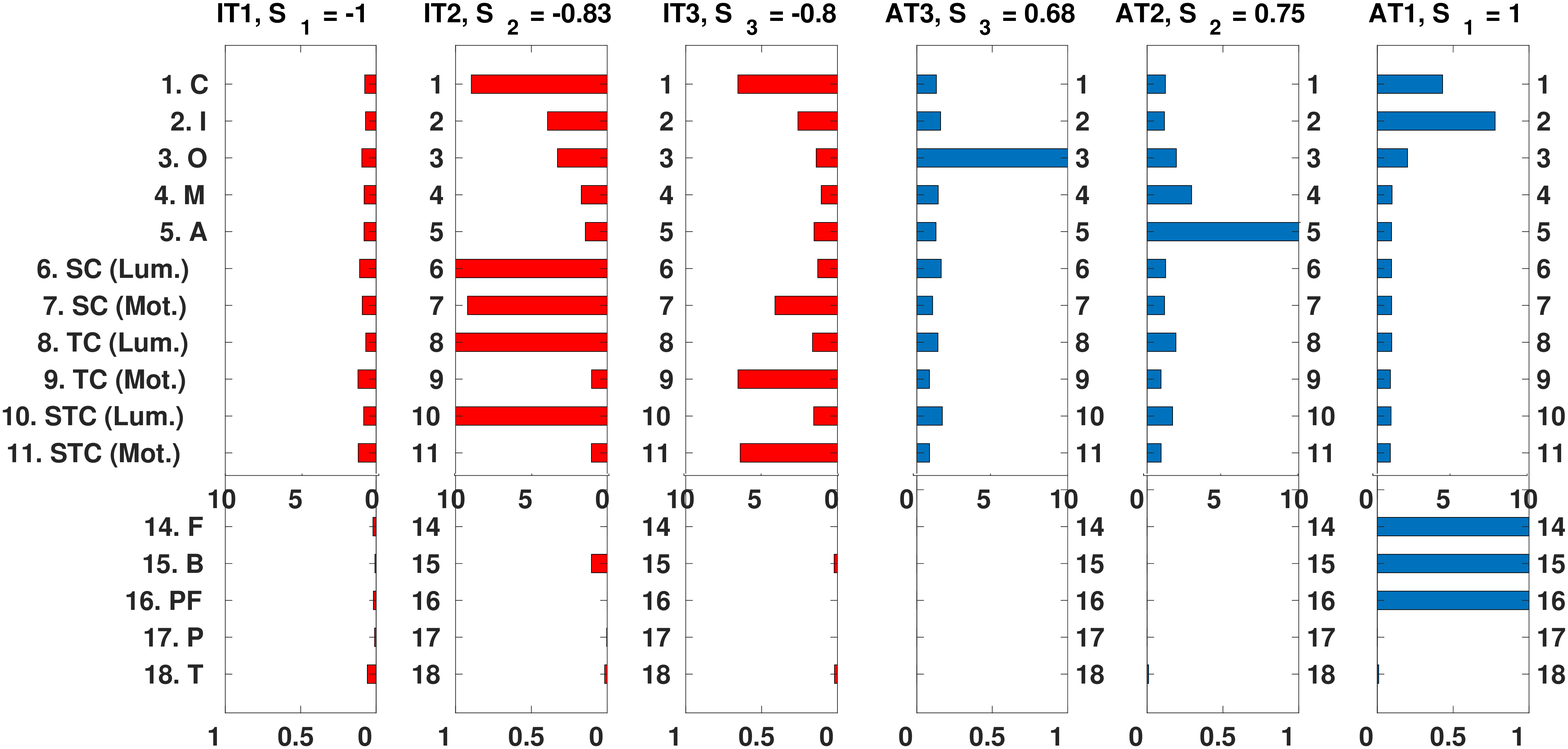}
    \caption{CRCNS-ORIG \cite{Itti_Carmi09crcns} database: \emph{Sports}}
	\label{fig:B_sports_orig}
\end{figure}

\begin{figure}[H]
    \centering
        \includegraphics[trim=10cm 1cm 12cm 0.5cm, width=0.6\textwidth]{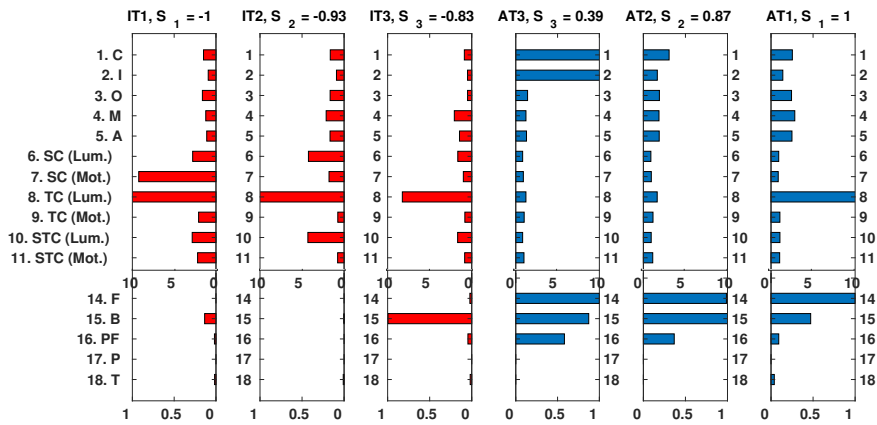}
    \caption{CRCNS-ORIG \cite{Itti_Carmi09crcns} database: \emph{Talk Shows}}
	\label{fig:B_talkshows_orig}
\end{figure}

\begin{figure}[H]
    \centering
        \includegraphics[trim=10cm 1cm 12cm 0.5cm, width=0.6\textwidth]{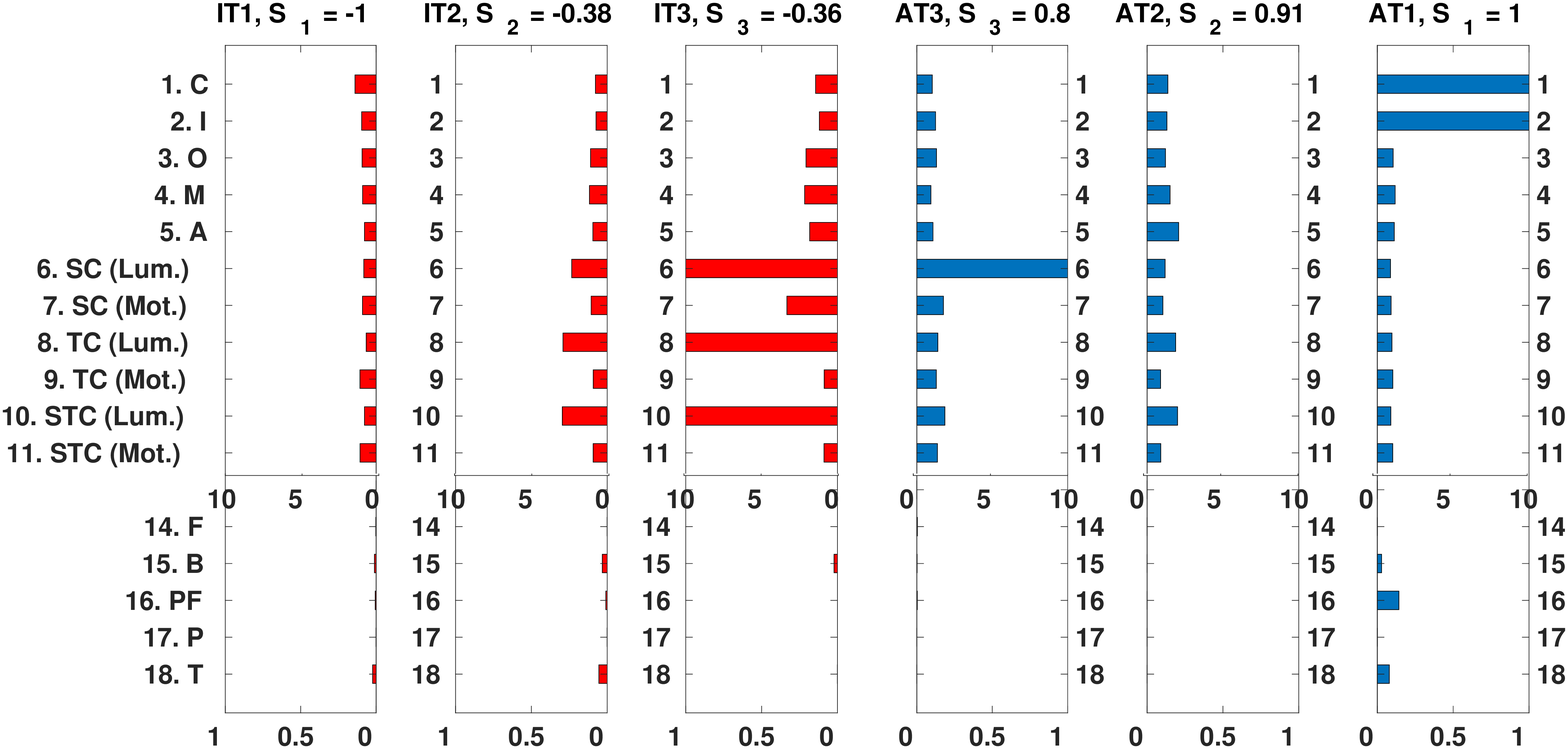}
    \caption{CRCNS-ORIG \cite{Itti_Carmi09crcns} database: \emph{Others}}
	\label{fig:B_others_orig}
\end{figure}

Figures \ref{fig:B_outdoor_orig}-\ref{fig:B_others_orig} illustrate the three most prominent attraction (\acs{AT}) and inhibition (\acs{IT}) sub-tasks determined by the above-mentioned approach for modeling visual attention in all database contexts. Moreover, for the sake of comparison, significant sub-tasks that define a \emph{context-generic} model trained on frames from the whole database are also provided in Figure \ref{fig:B_generic_orig}.

\begin{figure}[H]
    \centering
        \includegraphics[trim=10cm 1cm 12cm 0.5cm, width=0.6\textwidth]{gfx/chapter5/53_Outdoor.eps}
    \caption{CRCNS-ORIG \cite{Itti_Carmi09crcns} database: \emph{Outdoor}}
	\label{fig:B_outdoor_orig}
\end{figure} 

\begin{figure}[H]
    \centering
        \includegraphics[trim=10cm 1cm 12cm 0.5cm, width=0.6\textwidth]{gfx/chapter5/53_TVNews.eps}
    \caption{CRCNS-ORIG \cite{Itti_Carmi09crcns} database: \emph{TV News}}
	\label{fig:B_tvnews_orig}
\end{figure} 

\section{DIEM database}
\subsection{Description}
DIEM \cite{mital2011clustering} dataset contains eye movement recordings from over 250 participants freely watching 84 high-definition natural videos (over 240,000 video frames, 134 minutes total, variable dimensions). Eye traces have been obtained using a 1,000 Hz SR Research Eyelink 2000 desktop mounted eye tracker. Eye fixations from approximately 50 subjects are provided for each clip. 

\subsection{Video categories}
For the experiments on context-driven visual attention understanding and prediction presented in Chapter \ref{ch:atom_experiments}, database clips have been classified into seven categories: \emph{TV Shows, Documentaries, Commercials, Talk Shows, Sports, Cooking} and \emph{TV News}, as enumerated in Table \ref{fig:B_scenarios_DIEM}.

\begin{table}[H]
  \small
  \centering
  \caption{Categories in the DIEM \cite{mital2011clustering} database. Clips included in each category are enumerated, together with their number of frames.}\label{fig:B_scenarios_DIEM}
  \subfloat[TV Shows, 12 clips]{%
    \vspace{3cm}%
    \begin{tabular}{l|c}
    		\hline
    		{Clip name} & {Frames}\\ \hline
    		{50{\_}people{\_}brooklyn{\_}1280x720} & $3,669$ \\  
    		{50{\_}people{\_}brooklyn{\_}no{\_}voices{\_}1280x720} & $3,669$ \\  
    		{50{\_}people{\_}london{\_}1280x720} & $3,840$ \\  
    		{50{\_}people{\_}london{\_}no{\_}voices{\_}1280x720} & $3,840$ \\   
    		{DIY{\_}SOS{\_}1280x712} & $1,200$ \\ 
    		{home{\_}movie{\_}Charlie{\_}bit{\_}my{\_}finger{\_}again{\_}960x720} & $1,661$ \\ 
    		{one{\_}show{\_}1280x712} & $1,430$ \\  
    		{stewart{\_}lee{\_}1280x712} & $2,412$ \\  
    		{tv{\_}graduates{\_}1280x720} & $4,045$ \\  
    		{tv{\_}ketch2{\_}672x544} & $2,286$ \\   
    		{tv{\_}the{\_}simpsons{\_}860x528} & $3,642$ \\  
    		{tv{\_}uni{\_}challenge{\_}final{\_}1280x712} & $2,577$ \\ \hline
    		{TOTAL} & $34,271$ \\ \hline
    	\end{tabular}
    \hspace{0cm}%
  }\vspace{3cm}
  \end{table}
  \begin{table}\ContinuedFloat
  \centering
  \subfloat[Documentaries, 18 clips]{%
    \hspace{0cm}%
    \begin{tabular}{l|c}
    		\hline
    		{Clip name} & {Frames}\\ \hline
    		{Antarctica{\_}landscape{\_}1246x720} & $2,135$ \\  
    		{BBC{\_}life{\_}in{\_}cold{\_}blood{\_}1278x710} & $3,401$ \\  
    		{BBC{\_}wildlife{\_}eagle{\_}930x720} & $3,960$ \\  
    		{BBC{\_}wildlife{\_}serpent{\_}1280x704} & $1,038$ \\  
    		{BBC{\_}wildlife{\_}special{\_}tiget{\_}1276x720} & $4,320$ \\  
    		{artic{\_}bears{\_}1066x710} & $2,786$ \\  
    		{documentary{\_}adrenaline{\_}rush{\_}1280x720} & $3,282$ \\  
    		{documentary{\_}coral{\_}reef{\_}adventure{\_}1280x720} & $2,969$ \\  
    		{documentary{\_}discoverers{\_}1280x720} & $4,560$ \\
    		{documentary{\_}dolphins{\_}1280x720} & $3,181$ \\  
    		{documentary{\_}mystery{\_}nile{\_}1280x720} & $2,604$ \\  
    		{documentary{\_}planet{\_}earth{\_}1280x704} & $5,082$ \\
    		{hummingbirds{\_}closeups{\_}960x720} & $4,217$ \\  
    		{hummingbirds{\_}narrator{\_}960x720} & $1,162$ \\  
    		{nightlife{\_}in{\_}mozambique{\_}1280x580} & $1,421$ \\
    		{planet{\_}earth{\_}jungles{\_}frogs{\_}1280x704} & $4,371$ \\  
    		{planet{\_}earth{\_}jungles{\_}monkeys{\_}1280x704} & $4,475$ \\  
    		{university{\_}forum{\_}construction{\_}ionic{\_}1280x720} & $1,418$ \\ \hline
    		{TOTAL} & $56,382$ \\ \hline
    	\end{tabular}
    \hspace{0cm}%
  }\hspace{0.1cm}
  \subfloat[Commercials, 15 clips]{%
    \hspace{0cm}%
    \begin{tabular}{l|c}
    		\hline
    		{Clip name} & {Frames}\\ \hline 
    		{advert{\_}bbc4{\_}bees{\_}1024x576} & $1,217$ \\  
    		{advert{\_}bbc4{\_}library{\_}1024x576} & $1,202$ \\  
    		{advert{\_}bravia{\_}paint{\_}1280x720} & $2,167$ \\
    		{advert{\_}iphone{\_}1272x720} & $900$ \\  
    		{game{\_}trailer{\_}bullet{\_}witch{\_}1280x720} & $3,720$ \\  
    		{game{\_}trailer{\_}ghostbusters{\_}1280x720} & $3,103$ \\  
    		{game{\_}trailer{\_}lego{\_}indiana{\_}jones{\_}1280x720} & $3,314$ \\
    		{game{\_}trailer{\_}wrath{\_}lich{\_}king{\_}shortened{\_}subtitles{\_}1280x548} & $5,420$ \\  
    		{harry{\_}potter{\_}6{\_}trailer{\_}1280x544} & $2,928$ \\  
    		{movie{\_}trailer{\_}alice{\_}in{\_}wonderland{\_}1280x682} & $2,538$ \\  
    		{movie{\_}trailer{\_}ice{\_}age{\_}3{\_}1280x690} & $3,283$ \\
    		{movie{\_}trailer{\_}quantum{\_}of{\_}solace{\_}1280x688} & $2,998$ \\  
    		{music{\_}gummybear{\_}880x720} & $888$ \\  
    		{music{\_}red{\_}hot{\_}chili{\_}peppers{\_}shortened{\_}1024x576} & $5,597$ \\  
    		{music{\_}trailer{\_}nine{\_}inch{\_}nails{\_}1280x720} & $1,283$ \\ \hline
    		{TOTAL} & $40,558$ \\ \hline
    	\end{tabular}
    \hspace{0cm}%
  }\hspace{0.1cm}
  \end{table}
  \begin{table}\ContinuedFloat
  \centering
  \subfloat[Talk Shows, 5 clips]{%
    \hspace{0cm}%
    \begin{tabular}{l|c}
    		\hline
    		{Clip name} & {Frames}\\ \hline
    		{ami{\_}ib4010{\_}closeup{\_}720x576} & $1,080$ \\  
    		{ami{\_}ib4010{\_}left{\_}720x576} & $1,067$ \\  
    		{ami{\_}is1000a{\_}closeup{\_}720x576} & $1,262$ \\  
    		{ami{\_}is1000a{\_}left{\_}720x576} & $1,270$ \\  
    		{scottish{\_}pariliament{\_}1152x864} & $3,978$ \\ \hline
    		{TOTAL} & $8,657$ \\ \hline
    	\end{tabular}
    \hspace{0cm}%
  }\hspace{0.1cm}
  \subfloat[Sports, 20 clips]{%
    \hspace{0cm}%
    \begin{tabular}{l|c}
    		\hline
    		{Clip name} & {Frames}\\ \hline
    		{basketball{\_}of{\_}sorts{\_}960x720} & $3,476$ \\  
    		{one{\_}show{\_}1280x712} & $900$ \\  
    		{pingpong{\_}angle{\_}shot{\_}960x720} & $1,170$ \\  
    		{pingpong{\_}closeup{\_}rallys{\_}960x720} & $3,300$ \\  
    		{pingpong{\_}long{\_}shot{\_}960x720} & $3,772$ \\ 
    		{pingpong{\_}miscues{\_}1080x720} & $1,371$ \\
    		{pingpong{\_}no{\_}bodies{\_}960x720} & $4,371$ \\  
    		{sport{\_}F1{\_}slick{\_}tyres{\_}1280x720} & $2,259$ \\  
    		{sport{\_}barcelona{\_}extreme{\_}1280x720} & $1,721$ \\  
    		{sport{\_}cricket{\_}ashes{\_}2007{\_}1252x720} & $2,574$ \\  
    		{sport{\_}football{\_}best{\_}goals{\_}976x720} & $2,478$ \\ 
    		{sport{\_}golf{\_}fade{\_}a{\_}driver{\_}1280x720} & $2,410$ \\  
    		{sport{\_}poker{\_}1280x640} & $3,480$ \\  
    		{sport{\_}scramblers{\_}1280x720} & $1,525$ \\  
    		{sport{\_}slam{\_}dunk{\_}1280x720} & $5,747$ \\  
    		{sport{\_}surfing{\_}in{\_}thurso{\_}900x720} & $2,357$ \\ 
    		{sport{\_}wimbledon{\_}baltacha{\_}1280x704} & $5,818$ \\  
    		{sport{\_}wimbledon{\_}federer{\_}final{\_}1280x704} & $2,772$ \\  
    		{sport{\_}wimbledon{\_}magic{\_}wand{\_}1280x704} & $1,768$ \\  
    		{sport{\_}wimbledon{\_}murray{\_}1280x704} & $2,627$ \\  
    		{sports{\_}kendo{\_}1280x710} & $2,768$ \\ \hline
    		{TOTAL} & $54,293$ \\ \hline
    	\end{tabular}
    \hspace{0cm}%
  }\hspace{0.1cm}
  \end{table}
  \begin{table}[H]\ContinuedFloat
  \centering
  \subfloat[Cooking, 7 clips]{%
    \hspace{0cm}%
    \begin{tabular}{l|c}
    		\hline
    		{Clip name} & {Frames}\\ \hline
    		{chilli{\_}plasters{\_}1280x712} & $3,697$ \\  
    		{growing{\_}sweetcorn{\_}1280x712} & $2,223$ \\  
    		{hairy{\_}bikers{\_}cabbafe{\_}1280x712} & $3,121$ \\  
    		{hydraulics{\_}1280x712} & $3,611$ \\  
    		{nigella{\_}chocolate{\_}pears{\_}1280x712} & $5,393$ \\  
    		{scottish{\_}starters{\_}1280x712} & $3,123$ \\  
    		{spotty{\_}trifle{\_}1280x712} & $2,516$ \\ \hline
    		{TOTAL} & $23,684$ \\ \hline
    	\end{tabular}
    \hspace{0cm}%
  }\hspace{0.1cm}
  \subfloat[TV News, 7 clips]{%
    \hspace{0cm}%
    \begin{tabular}{l|c}
    		\hline
    		{Clip name} & {Frames}\\ \hline
    		%{news{\_}bee_parasites{\_}768x576} & $3,072$ \\  
    		{news{\_}newsnight{\_}othello{\_}720x416} & $2,295$ \\  
    		{news{\_}sherry{\_}drinking{\_}mice{\_}768x576} & $1,999$ \\   
    		{news{\_}tony{\_}blair{\_}resignation{\_}720x540} & $1,413$ \\
    		{news{\_}us{\_}election{\_}debate{\_}1080x600} & $2,572$ \\
    		{news{\_}video{\_}republic{\_}960x720} & $6,276$ \\
    		{news{\_}wimbledon{\_}macenroe{\_}shortened{\_}1024x576} & $4,980$ \\ \hline
    		{TOTAL} & $22,607$ \\ \hline
    	\end{tabular}
    \hspace{0cm}%
  }
\end{table}

\subsection{Context-aware visual attention understanding}
\begin{figure}[H]
    \centering
        \includegraphics[trim=10cm 1cm 12cm 0.5cm, width=0.6\textwidth]{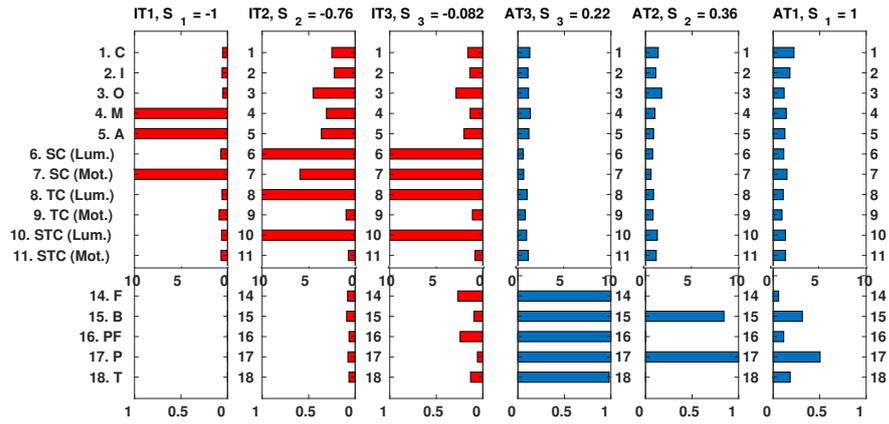}
    \caption{DIEM \cite{mital2011clustering} database: \emph{Context-Generic}}
	\label{fig:B_generic_diem}
\end{figure}

\begin{figure}[H]
    \centering
        \includegraphics[trim=10cm 1cm 12cm 0.5cm, width=0.6\textwidth]{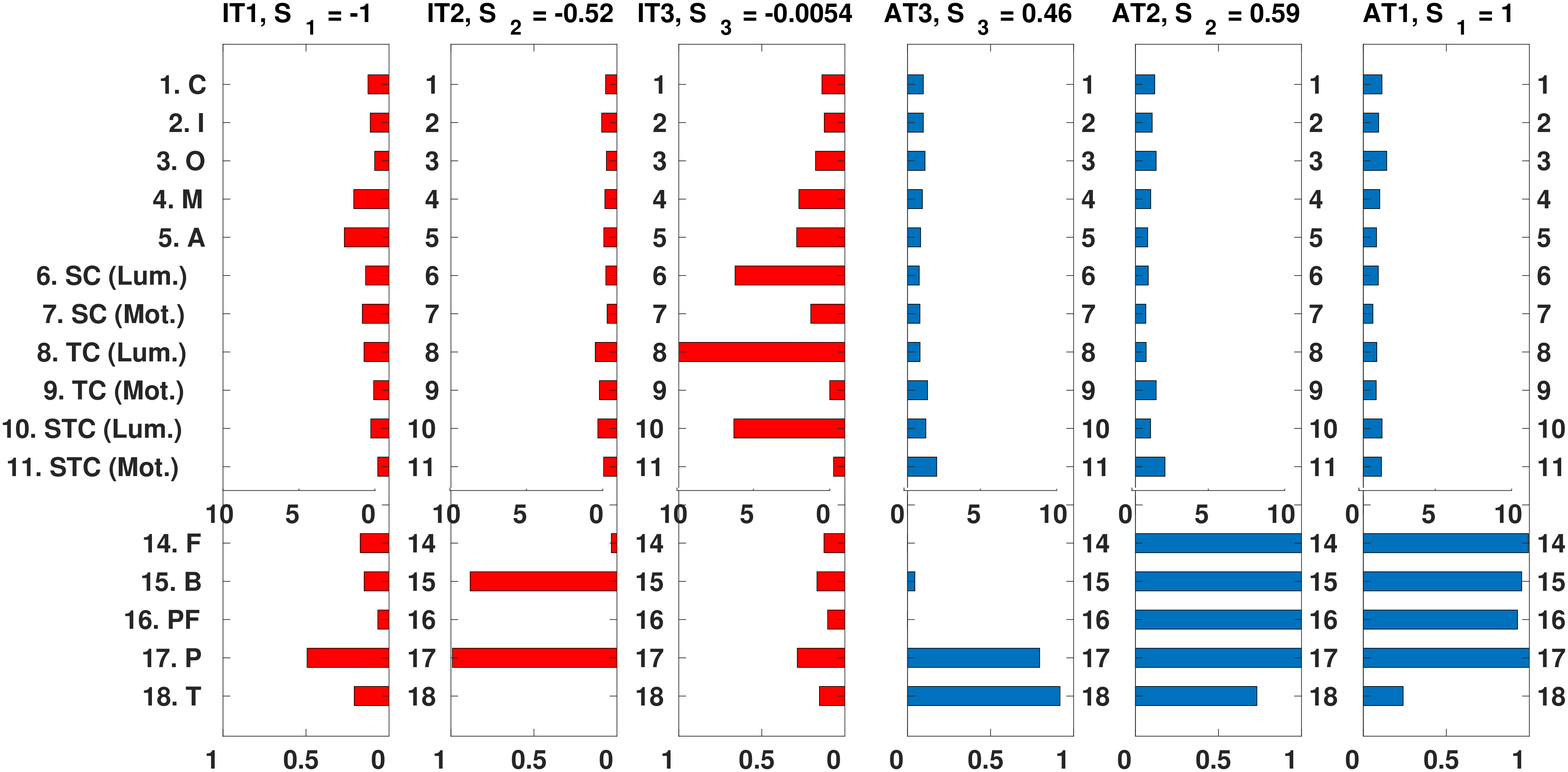}
    \caption{DIEM \cite{mital2011clustering} database: \emph{TV Shows}}
	\label{fig:B_tvshows_diem}
\end{figure}

\begin{figure}[H]
    \centering
        \includegraphics[trim=10cm 1cm 12cm 0.5cm, width=0.6\textwidth]{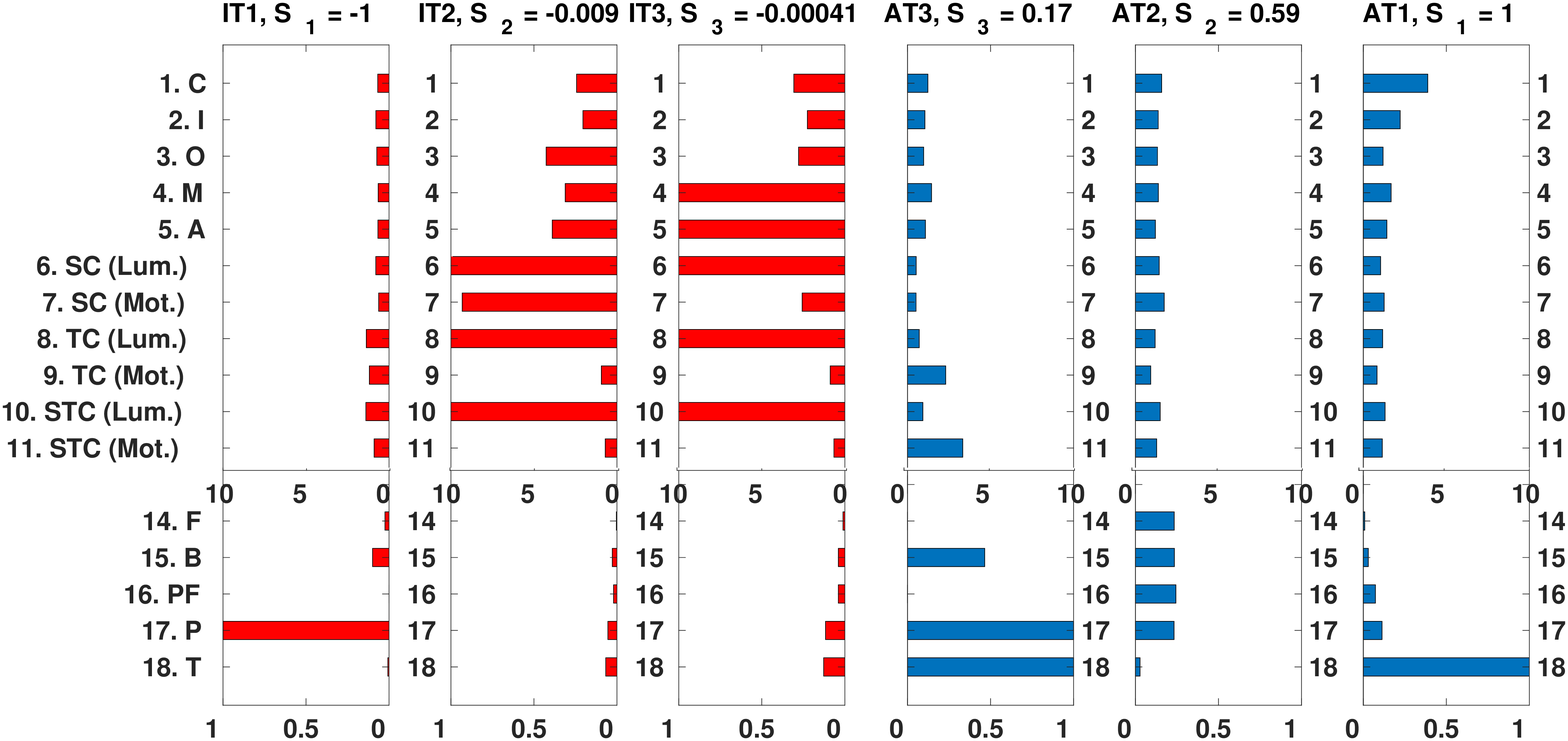}
    \caption{DIEM \cite{mital2011clustering} database: \emph{Documentaries}}
	\label{fig:B_documentaries_diem}
\end{figure}

\begin{figure}[H]
    \centering
        \includegraphics[trim=10cm 1cm 12cm 0.5cm, width=0.6\textwidth]{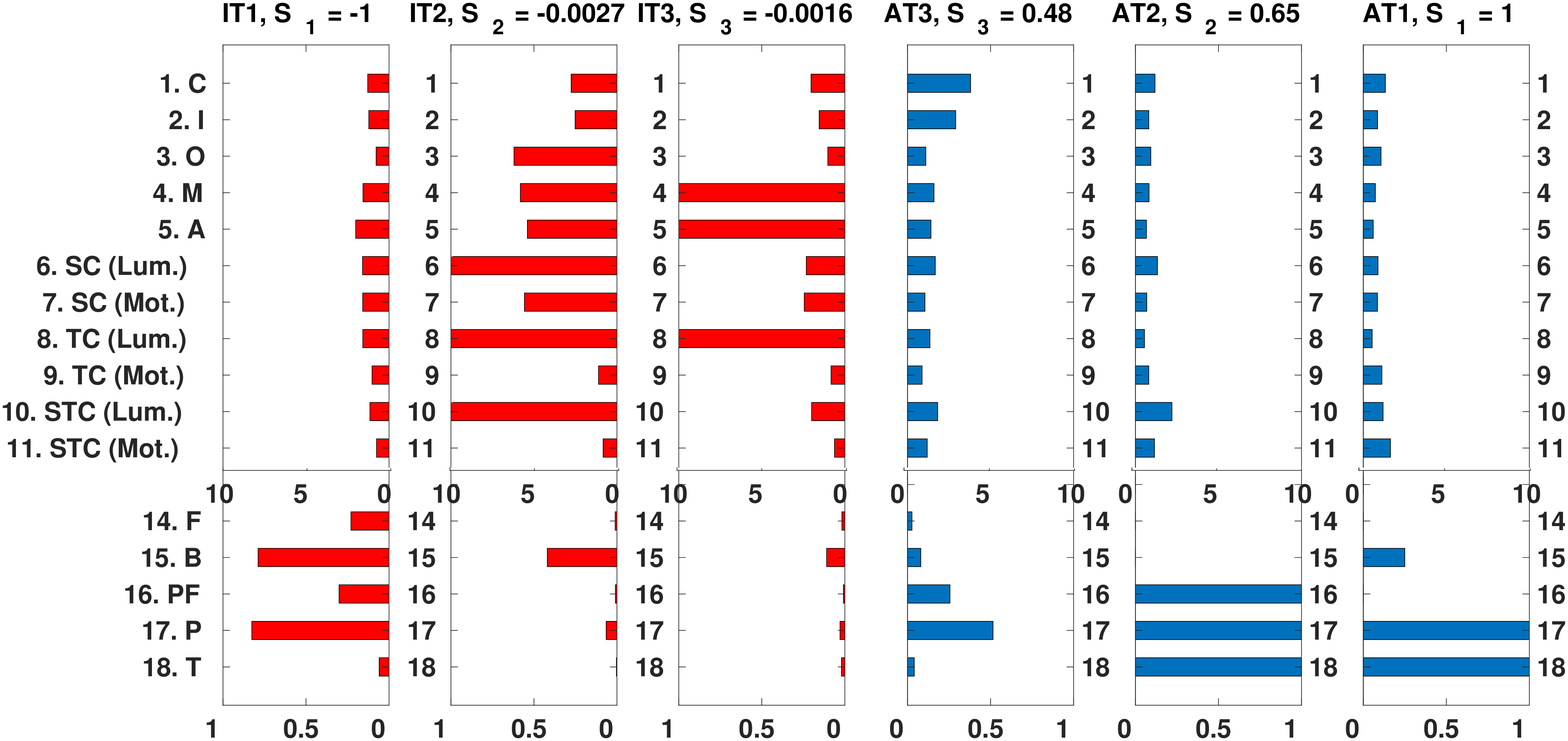}
    \caption{DIEM \cite{mital2011clustering} database: \emph{Talk Shows}}
	\label{fig:B_talkshows_diem}
\end{figure}

\begin{figure}[H]
    \centering
        \includegraphics[trim=10cm 1cm 12cm 0.5cm, width=0.6\textwidth]{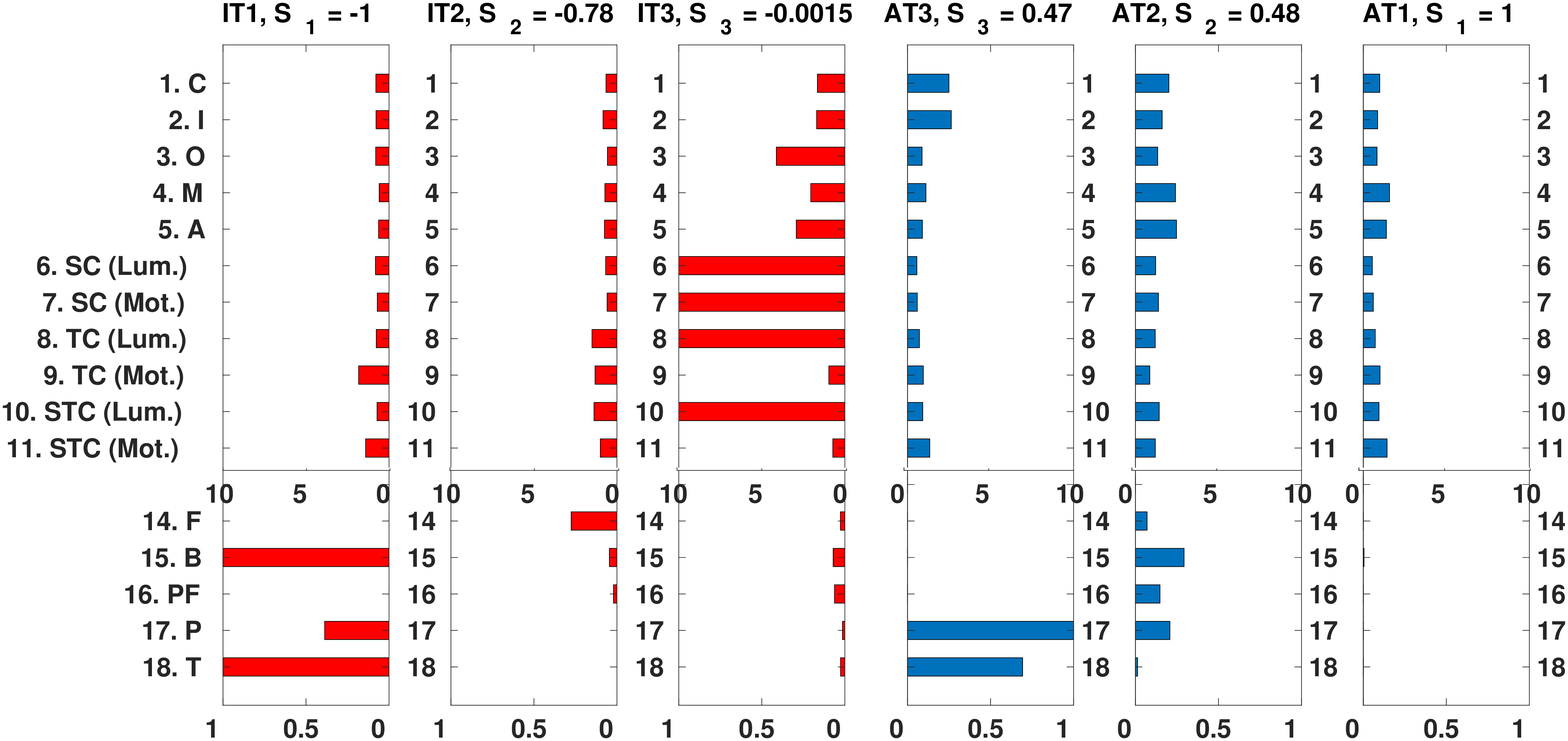}
    \caption{DIEM \cite{mital2011clustering} database: \emph{Cooking}}
	\label{fig:B_cooking_diem}
\end{figure}

\begin{figure}[H]
    \centering
        \includegraphics[trim=10cm 1cm 12cm 0.5cm, width=0.6\textwidth]{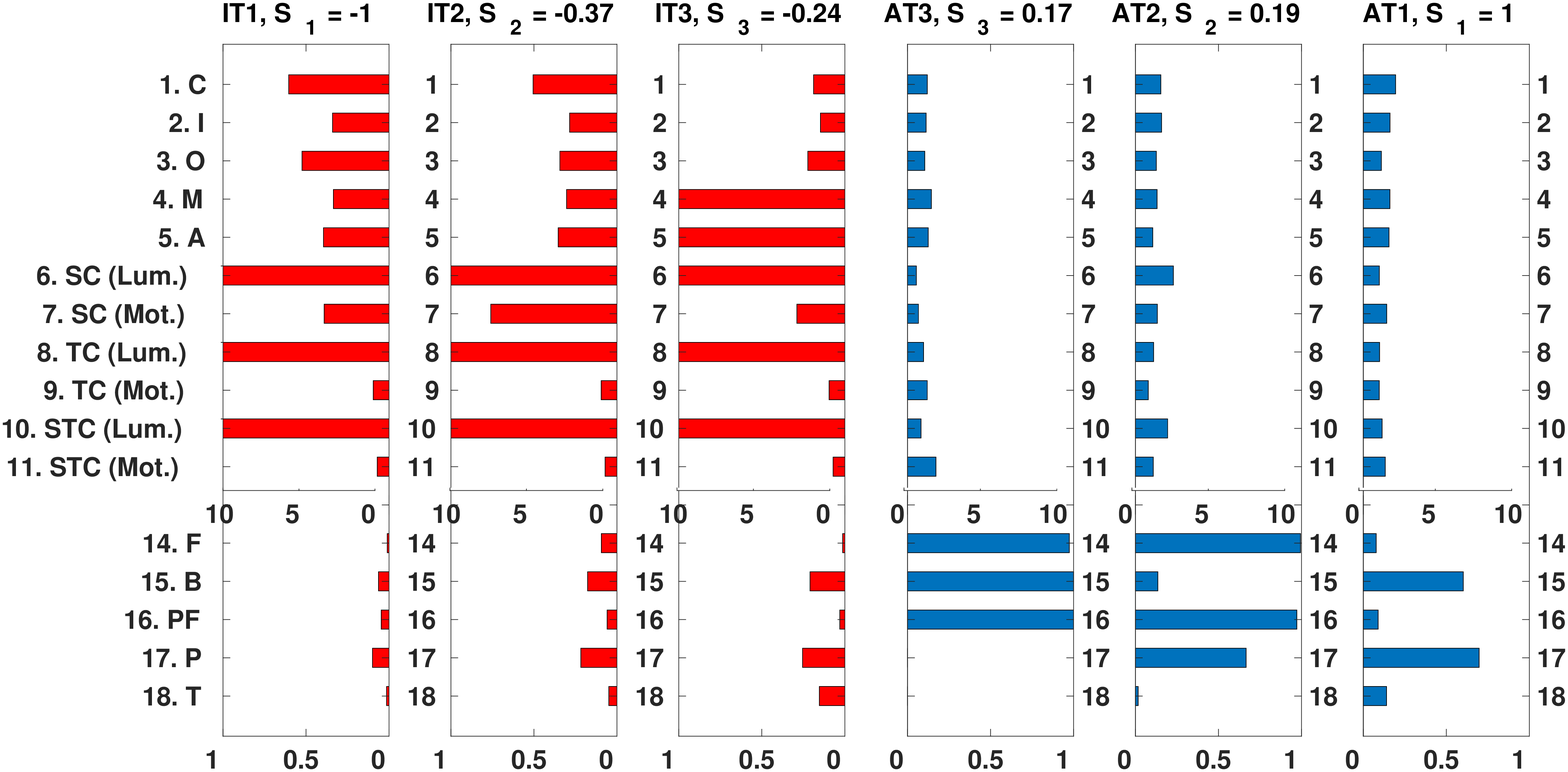}
    \caption{DIEM \cite{mital2011clustering} database: \emph{TV News}}
	\label{fig:B_tvnews_diem}
\end{figure}

Figures \ref{fig:B_tvshows_diem}-\ref{fig:B_tvnews_diem} illustrate the most noteworthy attraction (\acs{AT}) and inhibition (\acs{IT}) sub-tasks determined by the above-mentioned approach for modeling visual attention in all database contexts. Moreover, for the sake of comparison, significant sub-tasks that define a \emph{context-generic} model trained on frames from the whole database are also provided in Figure \ref{fig:B_generic_diem}.

\begin{figure}[H]
    \centering
        \includegraphics[trim=10cm 1cm 12cm 0.5cm, width=0.6\textwidth]{gfx/chapter5/53_CommercialsD.eps}
    \caption{DIEM \cite{mital2011clustering} database: \emph{Commercials}}
	\label{fig:B_commercials_diem}
\end{figure} 

\begin{figure}[H]
    \centering
        \includegraphics[trim=10cm 1cm 12cm 0.5cm, width=0.6\textwidth]{gfx/chapter5/53_SportsD.eps}
    \caption{DIEM \cite{mital2011clustering} database: \emph{Sports}}
	\label{fig:B_sports_diem}
\end{figure} 

\section{BOSS database}
\subsection{Description}
Within the framework of the BOSS project \cite{BOSS}, a database with 15 video sequences recorded in RENFE suburban trains from Madrid was released, with the aim of developing an efficient transmission system for video-surveillance in a railway transport context. Videos contain events such as a cell phone theft, a passengers fight, a disease in public and several women harassment. Moreover, two additional sequences with no incidents are included. For each event, three camera views are provided.

\subsection{Video sequences}
In order to evaluate the architectures for visual attention modeling in the temporal domain proposed in Chapter \ref{ch:anomaly_detection}, we have selected the three camera views of 10 sequences from this database to be annotated with eye fixations. In total, 30 videos (over $84,000$ video frames, $56$ minutes total, $720 \times 576$) have been used, which are enumerated in Table \ref{fig:B_videos_BOSS}. For each video, eye traces from 5 observers have been recorded by using a 250 Hz SMI RED250mobile Eye Tracker system \cite{SMI}. 

\begin{table}[H]
\renewcommand\thetable{B.3}
  \small
  \centering
  \caption{Videos from the BOSS \cite{BOSS} database for the experiments in Chapter \ref{ch:surveillance_experiments}. Clips are enumerated together with their number of frames.}\label{fig:B_videos_BOSS}
  \vspace{0.5cm}%
    \begin{tabular}{l|c}
    		\hline
    		{Clip name} & {Frames}\\ \hline
    		{Cell{\_}phone{\_}Spanish.Cam1} & $1,501$ \\
    		{Cell{\_}phone{\_}Spanish.Cam2} & $1,501$ \\
    		{Cell{\_}phone{\_}Spanish.Cam3} & $1,501$ \\
    		{Checkout{\_}French.Cam1} & $3,941$ \\
    		{Checkout{\_}French.Cam2} & $2,810$ \\
    		{Checkout{\_}French.Cam3} & $2,843$ \\
    		{Disease{\_}Public.Cam1} & $3,082$ \\
    		{Disease{\_}Public.Cam2} & $3,086$ \\
    		{Disease{\_}Public.Cam3} & $3,088$ \\
    		{Harass{\_}French.Cam1} & $2,674$ \\
    		{Harass{\_}French.Cam2} & $2,679$ \\
    		{Harass{\_}French.Cam3} & $2,679$ \\
    		{Harass2{\_}French.Cam1} & $2,976$ \\
    		{Harass2{\_}French.Cam2} & $2,976$ \\
    		{Harass2{\_}French.Cam3} & $2,976$ \\ 
    		{Harass{\_}Spanish.Cam1} & $2,976$ \\
    		{Harass{\_}Spanish.Cam2} & $2,976$ \\
    		{Harass{\_}Spanish.Cam3} & $2,976$ \\
    		{Newspaper{\_}Spanish.Cam1} & $2,438$ \\
    		{Newspaper{\_}Spanish.Cam2} & $2,438$ \\
    		{Newspaper{\_}Spanish.Cam3} & $2,438$ \\
    		{No{\_}Event.Cam1} & $2,630$ \\
    		{No{\_}Event.Cam2} & $2,635$ \\
    		{No{\_}Event.Cam3} & $2,636$ \\
    		{No{\_}Event2.Cam1} & $4,001$ \\
    		{No{\_}Event2.Cam2} & $4,001$ \\
    		{No{\_}Event2.Cam3} & $4,001$ \\
    		{Panic.Cam1} & $2,501$ \\
    		{Panic.Cam2} & $2,501$ \\
    		{Panic.Cam3} & $2,501$ \\ \hline
    		{TOTAL} & $83,962$ \\ \hline
    \end{tabular}
    \hspace{0cm}%
\end{table}
%\include{Chapters/Chapter0C}
% ********************************************************************
% DEMO CHAPTERS
%*******************************************************
%\include{Chapters/Chapter01demo}
%\include{Chapters/Chapter02demo}
%\include{Chapters/Chapter03demo}
%\include{Chapters/Chapter0Ademo}
%********************************************************************
% Other Stuff in the Back
%*******************************************************
\cleardoublepage%********************************************************************
% Bibliography
%*******************************************************
% work-around to have small caps also here in the headline
% https://tex.stackexchange.com/questions/188126/wrong-header-in-bibliography-classicthesis
% Thanks to Enrico Gregorio
\defbibheading{bibintoc}[\bibname]{%
  \phantomsection
  \manualmark
  \markboth{\spacedlowsmallcaps{#1}}{\spacedlowsmallcaps{#1}}%
  \addtocontents{toc}{\protect\vspace{\beforebibskip}}%
  \addcontentsline{toc}{chapter}{\tocEntry{#1}}%
  \chapter*{#1}%
}
\printbibliography[heading=bibintoc]

% Old version, will be removed later
% work-around to have small caps also here in the headline
%\manualmark
%\markboth{\spacedlowsmallcaps{\bibname}}{\spacedlowsmallcaps{\bibname}} % work-around to have small caps also
%\phantomsection
%\refstepcounter{dummy}
%\addtocontents{toc}{\protect\vspace{\beforebibskip}} % to have the bib a bit from the rest in the toc
%\addcontentsline{toc}{chapter}{\tocEntry{\bibname}}
%\label{app:bibliography}
%\printbibliography

%\cleardoublepage\include{FrontBackmatter/Declaration}
\cleardoublepage\pagestyle{empty}

\hfill

\vfill

\pdfbookmark[0]{Colophon}{colophon}
\section*{Colophon}
This document was typeset using the typographical look-and-feel \texttt{classicthesis} developed by Andr\'e Miede and Ivo Pletikosić.
The style was inspired by Robert Bringhurst's seminal book on typography ``\emph{The Elements of Typographic Style}''.
\texttt{classicthesis} is available for both \LaTeX\ and \mLyX:
\begin{center}
\url{https://bitbucket.org/amiede/classicthesis/}
\end{center}
%Happy users of \texttt{classicthesis} usually send a real postcard to the author, a collection of postcards received so far is featured here:
%\begin{center}
%\url{http://postcards.miede.de/}
%\end{center}
%Thank you very much for your feedback and contribution.

\bigskip

\noindent We gratefully acknowledge the support of NVIDIA Corporation with the donation of the NVIDIA GeForce GTX TITAN Xp GPU used for part of the research covered in this thesis.

\bigskip

\noindent\finalVersionString

\end{document}